\documentclass[journal]{IEEEtran}

\usepackage[numbers, compress]{natbib}
\usepackage{moreverb,url}
\usepackage{setspace}
\usepackage{hyperref}
\usepackage[flushleft]{threeparttable}
\usepackage{tikz}

\newcommand\BibTeX{{\rmfamily B\kern-.05em \textsc{i\kern-.025em b}\kern-.08em
     T\kern-.1667em\lower.7ex\hbox{E}\kern-.125emX}}

\IEEEoverridecommandlockouts
\makeatletter
\renewcommand*{\@opargbegintheorem}[3]{\trivlist
  \item[\hskip \labelsep{\it\quad  #1\ #2:}] {\it(#3)}\ }
\makeatother

\newtheorem{thm}{Theorem}[section]
\newtheorem{cor}[thm]{Corollary}
\newtheorem{lem}[thm]{Lemma}
\newtheorem{asmp}[thm]{Assumption}
\newtheorem{prop}[thm]{Proposition}
\newtheorem{problem}{Problem}
\newtheorem{exmp}{Example}
\newtheorem{defn}[thm]{Definition}
\newtheorem{rem}[thm]{Remark}
\newcounter{listcounter}

\newenvironment{noindlist}
 {\begin{list}{(\alph{listcounter})~~}{\usecounter{listcounter} \labelsep=0em \labelwidth=0em \leftmargin=0em \itemindent=0em}}
 {\end{list}}

\newcommand{\ltlx}{ {LTL}$^\chi$ }
\newcommand{\ltlz}{ {LTL}$^0$ }

\newcommand{\clause}[1]{\mathsf{cls}(#1)}
\newcommand{\cp}[2]{\ccalC_{#1}^{#2}}

\newcommand{\auto}[1]{\ccalA_{\textup{#1}}}
\newcommand{\autop}{\ccalA_{\phi}}
\newcommand{\vertex}[1]{v_{\textup{#1}}}
\newcommand{\ag}[2]{\langle#1,#2\rangle}
\newcommand{\simplies}{\DOTSB\Longrightarrow}

\usepackage{stmaryrd}
\usepackage{graphicx} 
\usepackage{bbm}
\usepackage{bm}
\usepackage{amsmath}
\usepackage{amssymb}
\usepackage{epstopdf}

\usepackage[noend,ruled,linesnumbered,resetcount]{algorithm2e}
\SetKwComment{Comment}{$\triangleright$\ } {}
\usepackage{multirow}
\usepackage{rotating}
\usepackage{subfigure}
\usepackage{color}
\usepackage{mysymbol}
\usepackage{url}
\usepackage{ltlfonts}
\usepackage{tabularx,ragged2e,booktabs}
\usepackage[normalem]{ulem}
\usepackage[utf8x]{inputenc}
\renewcommand{\ap}[3]{\mathcal{\pi}_{{#1},{#2}}^{#3}}
\newcommand{\aap}[4]{\mathcal{\pi}_{{#1},{#2}}^{#3,#4}}

\newcommand{\RNum}[1]{\uppercase\expandafter{\romannumeral #1\relax}}

\usepackage[hang,flushmargin]{footmisc}
\makeatletter
\newcommand{\algorithmendnote}[2][\footnotesize]{%
  \let\old@algocf@finish\@algocf@finish
  \def\@algocf@finish{\old@algocf@finish
    \leavevmode\rlap{\begin{minipage}{\linewidth}
    #1#2
    \end{minipage}}%
  }%
}
\makeatother

\makeatletter
\newcommand{\pushright}[1]{\ifmeasuring@#1\else\omit\hfill$\displaystyle#1$\fi\ignorespaces}
\newcommand{\pushleft}[1]{\ifmeasuring@#1\else\omit$\displaystyle#1$\hfill\fi\ignorespaces}
\makeatother

\usepackage{amsmath,scalerel}
\newcommand\doverline[1]{\ThisStyle{%
  \setbox0=\hbox{$\SavedStyle\overline{#1}$}%
  \ht0=\dimexpr\ht0-.15ex\relax
  \overline{\copy0}%
}}

\makeatletter
  \setbox0\hbox{$\xdef\scriptratio{\strip@pt\dimexpr
    \numexpr(\sf@size*65536)/\f@size sp}$}
\newcommand{\scriptveryshortarrow}[1][3pt]{{%
    \hbox{\rule[\scriptratio\dimexpr\fontdimen22\textfont2-.2pt\relax]
               {\scriptratio\dimexpr#1\relax}{\scriptratio\dimexpr.4pt\relax}}%
   \mkern-4mu\hbox{\let\f@size\sf@size\usefont{U}{lasy}{m}{n}\symbol{41}}}}

\makeatother

\newenvironment{cexmp}[2]
{\addtocounter{exmp}{-1}\begin{exmp}{\textit{continued} (#2)}}
  {\end{exmp}}

  \usepackage{etoolbox}

\usepackage{graphicx,calc}
\newlength\myheight
\newlength\mydepth
\settototalheight\myheight{Xygp}
\newcounter{para}[subsubsection]
\renewcommand{\thepara}{(\arabic{para})}

\newenvironment{mysubparagraph}[2]%
{
\refstepcounter{para}
{\normalsize\itshape \thepara \space #1:\space}  {#2}
}%

\newcounter{phase} 
\newcounter{subphase}[para] 

\renewcommand{\thesubphase}{(\arabic{para}\alph{subphase})}

\newenvironment{subphase}[2]%
{
\refstepcounter{subphase}
{\normalsize\itshape \thesubphase \space #1: \space}  {#2}
}%

\newcounter{parag}[subsubsection]

\renewcommand{\theparagraph}{(\alph{paragraph})}

\renewenvironment{paragraph}[2]%
{
\refstepcounter{paragraph}
{\normalsize\itshape \theparagraph \space #1:\space}  {#2}
}%

\begin{document}

\title{Temporal Logic Task Allocation in Heterogeneous Multi-Robot Systems}

\author{Xusheng Luo and Michael~M.~Zavlanos,~\IEEEmembership{Senior Member,~IEEE} \thanks{Xusheng Luo and Michael M. Zavlanos are with the Department of Mechanical Engineering and Materials Science, Duke University, Durham, NC 27708, USA. $\left\{\text{xusheng.luo, michael.zavlanos}\right\}$@duke.edu. This work is supported in part by ONR under agreement $\#$N00014-18-1-2374 and by AFOSR under the award $\#$FA9550-19-1-0169.}}

\maketitle

\begin{abstract}
  In this paper, we consider the problem of optimally allocating tasks, expressed as global Linear Temporal Logic (LTL) specifications,   to teams of heterogeneous mobile robots. The robots are classified in different types that capture their different capabilities, and each task may require robots of multiple types. The specific robots assigned to each task are immaterial, as long as they are of the desired type. Given a discrete workspace, our goal is to design paths, i.e., sequences of discrete states, for the robots so that the LTL specification is satisfied.  To obtain a scalable solution to this complex temporal logic task allocation problem, we propose a hierarchical approach that first allocates specific robots to tasks using the information about the tasks contained in  the Nondeterministic B$\ddot{\text{u}}$chi Automaton (NBA) that captures the LTL specification, and then designs low-level executable plans for the robots that respect the high-level assignment. Specifically, we first prune and relax the NBA by removing all negative atomic propositions. This step is motivated by ``lazy collision checking" methods in robotics and allows to simplify the planning problem by checking constraint satisfaction only when needed. Then, we extract sequences of subtasks from the relaxed NBA along with their temporal orders, and formulate  a Mixed Integer Linear Program (MILP) to allocate these subtasks to the robots.  Finally, we define generalized multi-robot path planning problems to obtain low-level executable robot plans that satisfy both the high-level task allocation and the temporal constraints captured by the negative atomic propositions in the original NBA. We show that our method is complete for a subclass of LTL that covers a broad range of tasks and present numerical simulations demonstrating that it can generate paths with lower cost, considerably faster than existing methods.

\end{abstract}

 \section{Introduction}

  Robot motion planning traditionally consists of generating robot trajectories between a start and a goal region,  while avoiding obstacles \citep{lavalle2006planning}. More recently, new planning methods  have been proposed that can handle a richer class of tasks than standard  point-to-point navigation  that also include   temporal goals subject to  time constraints. Such tasks can be captured using formal
 languages, such as Linear Temporal Logic (LTL)~\citep{baier2008principles}, and include sequencing or coverage \citep{fainekos2005temporal}, data gathering \citep{guo2017distributed}, intermittent communication \citep{kantaros2018distributed}, and persistent surveillance \citep{leahy2016persistent}, to name a few. {A survey on formal specifications and synthesis techniques for robotic systems can be found in~\cite{luckcuck2019formal}.}

 In this paper, we consider LTL tasks that require robots of different types to collaborate to satisfy the specification. The different robot types capture the different robot  capabilities, and each task may require robots of multiple types to accomplish. The specific robots assigned to each task are immaterial, as long as they are of the desired type. An example of such an LTL task is: {\em At most five robots of type 1 pick up the mail by visiting houses in a given order. Next, visit a delivery site. Never leave the delivery site until one ground robot of type 2 is present to pick up the mail (a robot of type 2 can carry mail from at most 5 robots of type 1). Repeat this process infinitely often.}  In this task, several robots are required to work cooperatively and meet simultaneously at the same place. Note that the specific robots to participate in this task are not important and are not specified by the LTL formula. Instead, it is only required that no more than five robots of type 1 and exactly one robot of type 2 collaborate to accomplish this task. Therefore, there are multiple ways that this LTL task can be satisfied, which grow combinatorially with the number of robots, robot types, and the complexity of the LTL task.
 We refer to this problem as the Multi-Robot Task Allocation (MRTA) problem for LTL tasks, in short, LTL-MRTA. {Existing   control synthesis methods under temporal logic specifications, such as the ones proposed in~\cite{smith2010optimal,ulusoy2013optimality,guo2015multi,kantaros15asilomar},} build a large product automaton composed of the Nondeterministic B$\ddot{\text{u}}$chi Automaton (NBA) that captures the LTL specification and the discrete transition systems describing the motion of each one of the robots in the world. Then, these methods employ graph search techniques on this product graph to find the optimal plan that satisfies the LTL specification. However, as the number of robots, the size of the environment, and the complexity of the LTL task grows, the size of this product graph grows exponentially large and, therefore, graph search methods become intractable. This is more so the case for LTL-MRTA problems as the number of possible assignments of robots to tasks increases the complexity of the LTL specification dramatically.

 To mitigate the computational complexity of the LTL-MRTA problem, we propose a novel hierarchical approach that first allocates specific robots to tasks using the information about tasks provided by the Nondeterministic B$\ddot{\text{u}}$chi Automaton (NBA) that captures the LTL specification, and then designs low-level executable plans for the robots that respect the high-level assignment. Specifically, we first prune and relax the NBA by removing all negative atomic propositions. This step is motivated by "lazy collision checking" methods in robotics~\citep{sanchez2003single,hauser2015lazy} and allows to simplify the planning problem by checking constraint satisfaction only when needed. Then, we extract sequences of subtasks from the relaxed NBA along with their temporal orders, and formulate a Mixed Integer Linear Program (MILP), inspired by the vehicle routing problem~\citep{bredstrom2008combined}, to allocate these subtasks to the robots, while respecting the temporal order between subtasks. The solution to this MILP  generates a time-stamped task allocation plan for each robot, which is a sequence of essential waypoints that the robot needs to visit. Finally, given this time-stamped task allocation plan for each robot, we formulate a sequence of generalized multi-robot path planning (GMRPP) problems, one for each subtask, to obtain executable paths that also respect the negative atomic propositions that were relaxed from the original NBA. We show through  extensive simulations  that our method can handle LTL-MRTA problems with up to $10^{90}$ states in the product graph, considerably outperforming existing methods. Moreover, we provide theoretical guarantees on the completeness and soundness of our proposed framework, under mild assumptions on the structure of the NBA that were satisfied by  all meaningful LTL specifications we considered in practice, no matter their complexity. While not theoretically optimal, our method is still able to improve on the cost of the returned plans, unlike existing methods in the literature that only focus on feasibility.

 \subsection{Related work}
 In existing literature on optimal control synthesis methods from LTL specifications, LTL tasks are either assigned locally to the robots in a multi-robot team, as in~\cite{guo2015multi,tumova2016multi} or a global LTL specification is assigned to the team that captures the collective behavior of all robots. In the latter case, the global LTL specification can explicitly assign tasks to the individual robots, as in~\cite{loizou2004automatic,smith2011optimal,saha2014automated,kantaros2015intermittent,kantaros2017sampling,kantaros2018distributedOpt,kantaros2018sampling,kantaros2018temporal,kantaros2020stylus,xluo_CDC19,luo2019abstraction}, or it may not explicitly assign tasks to the robots as in~\cite{kloetzer2011multi,shoukry2017linear,moarref2017decentralized,lacerda2019petri}, and our current work in this paper.

 Global temporal logic specifications that do not explicitly allocate tasks to robots typically  need to be decomposed in order to obtain the required allocation. For example,~\cite{tumova2015decomposition,kantaros2016distributed} decompose a global specification directly into local specifications and assign them to individual robots. Similarly,~\cite{camacho2017non,xluo_CDC19,camacho2019ltl,schillinger2019hierarchical} decompose a global specification into multiple subtasks by exploiting the structure of the finite automata. Particularly,~\cite{camacho2017non} convert temporal planning problems to standard planning problems by defining actions based on  transitions in the NBA, while~\cite{xluo_CDC19} define subtasks associated with transitions in the NBA and synthesize plans for these subtasks which they store in a library so that they can be reused to efficiently synthesize plans  for new LTL formulas.~\cite{schillinger2019hierarchical} also define subtasks associated with transitions in the automaton, but use reinforcement learning to learn plans that execute these subtasks  under uncertainty.~\cite{camacho2019ltl} also use reinforcement learning but with the purpose of converting formal languages to reward machines that capture the structure of the task. Similar to these works, here too we define subtasks associated with transitions in the NBA. However, we do not assume that these subtasks are preassigned to the robots.

 Temporal logic control synthesis without an explicit assignment of robots to tasks has been considered in~\cite{karaman2011linear} that combine the vehicle routing problem with metric temporal logic specifications and leverage MILP to solve this problem for heterogeneous robots. However, this approach can only handle  finite horizon tasks and does not  design the low-level executable paths as we do here. An alternative approach is proposed in~\cite{chen2011formal,leahy2015distributed} that decomposes a global automaton into individual automata that are assigned to the heterogeneous robots  and then builds a synchronous product of these automata to synthesize  parallel plans. However, the size of the synchronous product automaton   grows exponentially large  with the number of robots. Also,  the requirement that parallel plans exist does not allow application of this method to tasks that lack such parallel executions. Furthermore, this method also focuses only  on finite robot trajectories. 
 In relevant literature, teams  of homogeneous robots have also been  modeled using Petri Nets as in~\cite{lacerda2019petri,kloetzer2020path}. Specifically,~\cite{lacerda2019petri} propose a job shop problem  under safe temporal logic specifications, but  do not consider the ``eventually'' operator so that liveness in terms of good future outcomes can not be guaranteed. Additionally, this approach only focuses  on robot coordination at the task level without considering execution. To the contrary, \cite{kloetzer2020path}  select multiple shortest accepting runs in the NBA and for each accepting run, determine whether an executable plan exists. Finally,~\cite{schillinger2018decomposition,schillinger2018simultaneous,faruq2018simultaneous,banks2020multi} automatically decompose the automaton representation of the LTL formula into independent subtasks that can be fulfilled by different robots. However, they only consider LTL formulas that can be satisfied by finite robot  trajectories, limiting the applicability of the proposed method to tasks such as recurrent sequencing and persistent monitoring. Also, subtasks subject to precedence relations can only be executed by a single robot.

 Common in the above approaches  is that they do not consider cooperative tasks where robots of the same or different types need to meet at a common location  to complete a task,  Such tasks require strong synchronization between robots. In our recent work~\citep{kantaros2018sampling,kantaros2020stylus}, we have proposed  a sampling-based planning method named STyLuS$^*$  that incrementally builds trees to approximate the product of the NBA and the  model of the team. Using the powerful biased sampling method proposed in~\cite{kantaros2020stylus}, STyLuS$^*$ can synthesize plans for product automata with up to $10^{800}$ states without considering collision avoidance. 
 However, STyLuS$^*$  requires global LTL specifications that explicitly assign tasks to robots. Although a subset of specifications  we consider here can be converted into explicit LTL formulas by enumerating all possible task assignments  and connecting them with ``OR'' operators, this would result in exponentially long LTL formulas. Furthermore, the biased sampling strategy in STyLuS$^*$ needs a fixed assignment of robots to tasks and biases search towards finding a plan for this fixed assignment. If the assignment is not given, biased STyLuS$^*$ will  need to be run combinatorially many times, one for each possible assignment. With unbiased sampling,~\cite{kantaros2018sampling} show that STyLuS$^*$ can only solve problems with product automata that have $10^{10}$ states. Instead, our proposed method can synthesize plans for problems with $10^{90}$ states while considering collision avoidance. On the other hand, model-checkers like NuSMV~\citep{cimatti2002nusmv}, focus on finding feasible paths and are incapable of  optimizing cost. As stated in~\cite{kantaros2020stylus}, NuSMV can only  handle problems with $10^{30}$ states,  and can not easily process  exponentially long LTL formulas generated by explicitly expressing task assignments.

 Among other methods that focus on  cooperative tasks,~\cite{moarref2017decentralized} focus on specifications capturing behaviors of homogeneous robotic swarms at the swarm and individual levels, but they can only impose universal or existential constraints, that is, all robots or some robots visit a certain region. As a result, these specifications are incapable of imposing restrictions on the number of robots that should be present at one place at the same time. This limitation is addressed in~\cite{sahin2017provably,sahin2017synchronous,sahin2019multirobot} that relies on  counting linear temporal logic (cLTL+/cLTL)  to capture constraints on the number of robots that must be present in different regions. Specifically, the authors formulate an  Integer Linear Program (ILP)  inspired by Bounded Model Checking techniques~\citep{biere2006linear}, but can only guarantee  feasibility of the resulting paths.  Instead our hierarchical method also takes into consideration the quality of the solution at each level. A sequential planning approach is proposed in 
 \cite{JoLeVaSaSeTrBe-ISRR-2019} that augments the LTL specification by introducing time and, unlike our proposed approach, plans low-level plans for the robots, one at a time, while treating the other robots as obstacles. Common in the methods in~\cite{sahin2017provably,sahin2017synchronous,sahin2019multirobot,JoLeVaSaSeTrBe-ISRR-2019} is that the size of the workspace has a significant effect on the computation time. To mitigate the complexity due to the size of the workspace,~\cite{sahin2019multi} propose a hierarchical framework that abstracts the workspace by aggregating states with the same observations. As we show in Section~\ref{sec:sim}, our proposed method  scales better than the method in \cite{sahin2019multi}, and provides lower cost solutions with less runtime. Also, unlike our  method, the completeness of solutions is not guaranteed in~\cite{sahin2019multi}.
\subsection{Contributions}

 The contributions of this paper  can be summarized as follows: We propose a new hierarchical approach to the LTL-MRTA problem that first assigns robots to tasks and then plans robot paths that satisfy the high level assignment. Our approach differs from common methods that rely on the product automaton~\cite{smith2010optimal,ulusoy2013optimality,guo2015multi} or on the Bounded Model Checking~\citep{biere2006linear} in that it directly operates on the NBA. Under mild assumptions on the NBA that are satisfied by a subclass of LTL formulas that cover a broad class of tasks in practice, we showed that our method is complete and sound. While not theoretically optimal, our method still incorporates optimization steps  to improve on the cost of the returned plans. To the best of our knowledge, this is the first LTL-MRTA method that is both complete for a subclass of LTL and includes operations to optimize  the synthesized plans. The unique aspect of our approach is a clever pruning and relaxation of the NBA that removes  all negative atomic propositions, and is motivated by ``lazy collision checking'' methods in robotics. This step significantly simplifies the planning problem by allowing to check  constraint satisfaction only when needed and, as a result, contributes to significantly increasing scalability of our method. To the best of our knowledge, this is the first time that ``lazy collision checking'' methods that are common in point-to-point navigation are used for high-level robot planning. Another unique aspect of our method is to infer the temporal order of tasks from the automaton, which can capture the parallel execution of subtasks. Compared to existing methods, our approach returns lower cost plans in significantly less time.

 The rest of the paper is organized as follows. 
 In Sections~\ref{sec:preliminaries} and~\ref{sec:problem} we present preliminaries and define the problem under consideration, respectively. We describe the high-level task assignment component of our method in  Sections~\ref{sec:app} and \ref{sec:solution}. Specifically, in Section~\ref{sec:app} we prune and relax the NBA, identify subtasks from the NBA and infer temporal orders between them. Then, in Section~\ref{sec:solution} we formulate a MILP to obtain the high-level plans. In Section~\ref{sec:correctness}, we examine the  completeness and soundness of these plans, while in Section~\ref{sec:sim} we present  simulation results. Finally, Section~\ref{sec:conclusion} concludes the paper.  For completeness, the low-level component of our method  to obtain executable paths, which is based on existing multi-robot path planning techniques, is presented in Appendix~\ref{sec:solution2mrta}.

 \section{Preliminaries}\label{sec:preliminaries}
 \subsection{Linear temporal logic}\label{sec:ltl}
 Linear Temporal Logic (LTL) is  composed of a set of atomic propositions $\mathcal{AP}$, the boolean operators, conjunction $\wedge$ and negation $\neg$, and temporal operators, next $\bigcirc$ and until $\mathcal{U}$~\citep{baier2008principles}. LTL formulas over $\mathcal{AP}$ follow the grammar $$\phi:=\top~|~\pi~|~\phi_1\wedge\phi_2~|~\neg\phi~|~\bigcirc\phi~|~\phi_1~\mathcal{U}~\phi_2,$$ where $\top$ is unconditionally true and $\pi$ is the boolean-valued atomic proposition. Other temporal operators can be derived from $\mathcal{U}$. For instance,  $\Diamond \phi$ means $\phi$ will be eventually satisfied sometime in the future and $\square \phi$ means $\phi$ is always satisfied from now on. 

 An infinite \textit{word} $w$ over the alphabet $2^{\mathcal{AP}}$, the power set of the set of atomic propositions, is defined as an infinite sequence  $w=\sigma_0\sigma_1\ldots\in (2^{\mathcal{AP}})^{\omega}$, where $\omega$ denotes an infinite repetition and $\sigma_k\in2^{\mathcal{AP}}$, $\forall k\in\mathbb{N}$. The language $\texttt{Words}(\phi)=\left\{w|w\models\phi\right\}$ is defined as the set of words that satisfy the LTL formula $\phi$, where $\models\subseteq (2^{\mathcal{AP}})^{\omega}\times\phi$ is the satisfaction relation. An LTL $\phi$ can be translated into a Nondeterministic B$\ddot{\text{u}}$chi Automaton (NBA)  defined as follows \citep{vardi1986automata}:
 \begin{defn}[NBA]\label{def:nba}
   A \textit{Nondeterministic B$\ddot{\text{u}}$chi Automaton} $B$ is  a tuple $B=\left(\ccalQ, \ccalQ_0,\Sigma,\rightarrow_B,\mathcal{Q}_F\right)$, where $\ccalQ$ is the set of states; $\ccalQ_0\subseteq\ccalQ$ is a set of initial states; $\Sigma=2^{\mathcal{AP}}$ is an alphabet;  $\rightarrow_{B}\subseteq\ccalQ\times \Sigma\times\ccalQ$ is the transition relation;
 and $\ccalQ_F\subseteq\ccalQ$ is a set of accepting states.
 \end{defn}

 An \textit{infinite run} $\rho_B$ of $B$ over an infinite word $w=\sigma_0\sigma_1\sigma_2\dots$, $\sigma_k\in\Sigma$, $\forall k\in\mathbb{N}$, is a sequence $\rho_B=q_0q_1q_2\dots$ such that $q_0\in\ccalQ_0$ and $(q_{k},\sigma_k,q_{k+1})\in\rightarrow_{B}$, $\forall k\in\mathbb{N}$.
 An infinite run $\rho_B$ is called \textit{accepting} if $\texttt{Inf}(\rho_B)\cap\ccalQ_F\neq\varnothing$, where $\texttt{Inf}(\rho_B)$ represents the set of states that appear in $\rho_B$ infinitely often. If an LTL formula is satisfiable, then there exists an accepting run that  can be written in the prefix-suffix structure such that the prefix part, connecting an initial state to an  accepting state, is traversed only once and the suffix part, a cycle around the accepting state, is traversed infinitely often.
 The words $\sigma$ that induce an accepting run of $B$ constitute the accepted language of $B$, denoted by $\ccalL_B$. It is shown in~\cite{baier2008principles} that for any given LTL formula $\phi$ over a set of atomic propositions $\ccalA\ccalP$, there  exists a NBA $B_{\phi}$ over alphabet $\Sigma = 2^{\ccalA\ccalP}$ such that $\ccalL_{B_{\phi}}=\texttt{Words}(\phi)$, where $\texttt{Words}(\phi)$ is the set of  words accepted by $\phi$.

 \subsection{Partially ordered set}\label{sec:partial}
 A finite partially ordered set or poset $P = (X, <_P )$  is a pair consisting of a finite base set $X$ and a binary relation $<_P \subseteq X \times X$ that is reflexive, antisymmetric, and transitive. Let  $x, y \in X$ be two distinct elements. We write $x <_P y$ if $(x,y) \in <_P$, and $x \|_P y$ if  $x$ and $y$ are incomparable. Moreover, we say $x$ is covered by $y$ or $y$ covers $x$, denoted by $x \prec_P y$, if $x<_P  y$ and there is no distinct $z \in X$ such that $x  <_P  z <_P y$. An antichain is a subset of a poset  in which any two distinct elements   are incomparable. The width of a poset is the cardinality of a maximal antichain. Similarly, the height of a poset is defined as the  cardinality of a chain. Finally, a chain is a subset of a poset  in which any two distinct elements are comparable. The height of a poset is the cardinality of a maximal chain.

 A linear order $L_X=(X, <_L)$ is a poset such that $x <_L y$, $x = y$ or $y <_L x$ holds for any pair of  $x, y \in X$. A linear extension  $L_P = (X, <_L)$  of a poset $P$ is a linear order such that $x <_L y$ if $x <_P y$, i.e., a linear order that preserves the partial order.
 We define $\ccalL_P$ as the set of all linear extensions of a poset $P$. Note that a poset and its linear extensions share the same base set $X_P$. Given a collection of linear orders $\Xi$, the poset cover problem focuses on reconstructing a single poset $P$ or a set of posets $ \{P_1,\ldots,P_k\}$ such that $\Xi_P = \Xi$ or $\cup_{i=1}^k \Xi_{P_i} = \Xi$. As shown in~\cite{heath2013poset}, the poset cover problem is NP-complete. Moreover, the partial cover problem focuses on finding a single poset $P$ such that $\Xi_P$ contains the maximum number of linear orders in $\Xi$, i.e., $\Xi_P \subseteq \Xi$ and   $\nexists P'$ s.t. $\Xi_{P'} \subseteq \Xi$ and $|\Xi_{P'}| > |\Xi_{P}|$. It is shown in~\cite{heath2013poset} that the partial cover problem can be solved in polynomial time.

 \section{Problem Definition}\label{sec:problem}
 \subsection{Transition system}\label{sec:ts}
 Consider a discrete workspace containing $l\in \mathbb{N}^+$ labeled regions of interest, so that each such region can span multiple cells in the workspace, and denote by $\mathcal{L}=\{\ell_k\}_{k\in[l]}$ the set of these regions, where $[l]$ is the shorthand notation for $\{1, \ldots, l\}$. We call free cells in the workspace that do not belong to any region {\em region-free}, and paths connecting two different regions that only pass through region-free cells {\em label-free}. We also assume that the workspace contains obstacles that can span multiple cells and do not overlap with the regions of interest.  We represent the workspace by a graph $E = (S, \to_E)$ where $S$ is the finite set of vertices corresponding to free cells and $\to_{E} \subseteq S \times S$ captures the adjacency relation.

  Given the workspace $E$, we consider a team of $n$ heterogeneous robots. We assume that these robots are of $m$ different types  and every robot belongs to exactly one type. Let $\mathcal{K}_j, j\in[m]$,  denote the set that  collects all robots of type $j$, so that $\sum_{j\in [m]} |\ccalK_j| = n$ and $\ccalK_{j} \cap \ccalK_{j'} = \emptyset$ if $j \not= j'$, where $|\cdot|$ is the cardinality of a set. We collect all $n$ robots in the set $\ccalR$, i.e., $\ccalR = \{\ccalK_j\}_{j\in [m]}$. Finally, we use $[r,j]$ to represent robot $r$ of type $j$, where $r\in \ccalK_j, j\in [m]$. To model the motion  of robot $[r,j]$ in the workspace, we define a transition system (TS) for this robot as follows.

 \begin{defn}[TS]\label{def:ts}
   A transition system for robot $[r,j]$ is a tuple $\textup{TS}_{r,j} = (S, s_{r,j}^0, \to_{r,j}, \Pi_{r,j}, L_{r,j})$ where: (a) $S$ is the set of free cells; (b) $s_{r,j}^0$ is the initial location of robot $[r,j]$; (c) $\to_{r,j} \subseteq \to_{E}  \bigcup \cup_{s_{r,j}\in S} \{(s_{r,j},s_{r,j})\} $ is the transition relation that allows the robots to remain idle or move between cells; (d) $\Pi_{r,j} = \cup_{k\in [l]}\{p_{r,j}^k\} \cup \{\epsilon\}$ where the atomic proposition $p_{r,j}^{k}$ is true if robot $[r,j]$ is at region $\ell_k$ and $\epsilon$ denotes the  empty label; and (e) $L_{r,j}: S \to {\Pi_{r,j}} $ is the labeling function that returns the atomic proposition  satisfied at location $s_{r,j}^t$.
 \end{defn}

 Given the transition systems of all robots $[r,j]$ we can define the  product transition system (PTS), which captures all possible combinations of robot behaviors.
 \begin{defn}[PTS]\label{def:pts}
    Given $n$ transition systems \textup{TS}$_{r,j} = (S, s_{r,j}^0, \to_{r,j}, \Pi_{r,j}, L_{r,j})$, the product transition system  is a tuple $\textup{PTS} = (S^n, s^0, \to, \Pi, L)$ where: (a) $S^n = S\times \cdots \times S$ is the finite set of collective robot locations
; (b) $s^0$ are the initial locations of the robots; (c) $\to \subseteq S^n \times S^n$ is the transition relation so that  $(s, s') \in \to$ if $s_{r,j}\rightarrow_{r,j} s'_{r,j}$ for all $r\in \ccalK_j, \forall \,j \in [m]$;
   (d) $\Pi = \cup_{i\in[|\ccalK_j|], j\in [m], k\in [l]}\{\pi_{i,j}^k\} \cup \{\epsilon\}$, where the atomic proposition $\ap{i}{j}{k}$ is true if  there exist at least $i$ robots of type $j$, denoted by $\ag{i}{j}$,  at region $\ell_k$ at the same time, i.e., $\ap{i}{j}{k} \Leftrightarrow |\{r\in \ccalK_j: L_{r,j}(s_{r,j}^t) = p_{r,j}^k \}| \geq i$; (e) and $L: S^n \to 2^{\Pi}$ is the labeling function that returns the set of atomic propositions satisfied by all robots at time $t$.
 \end{defn}
 \subsection{Task specification}
 In this paper, we consider MRTA problems where the tasks are globally described by LTL formulas. 
 Furthermore, we consider tasks in which the same fleet of robots of a certain type may need  to visit different regions in sequence, e.g., to deliver objects between different regions. To capture such tasks, we  define {\it induced atomic propositions} over the set $\Pi$ defined in Definition~\ref{def:pts}  as follows.

 \begin{defn}[Induced atomic propositions]
    For each basic atomic proposition $\ap{i}{j}{k} \in \Pi$, we define an infinite set of  {induced} atomic propositions $\{\ap{i}{j}{k,\chi}\}_{\chi\in \mathbb{N}}$, where $\chi$ is a connector that binds the truth of  atomic propositions with identical $i, j$ and $\chi$. Specifically, when $\chi=0$, $\ap{i}{j}{k,\chi}$ is equivalent to $\ap{i}{j}{k}$ whose truth is state-dependent. When $\chi \neq 0$, the truth of $\ap{i}{j}{k,\chi}$ is state-and-path-dependent, meaning that it additionally depends on other induced atomic propositions that share the same $i,j$ and $\chi$. That is, both $\ap{i}{j}{k,\chi}$ and $\ap{i}{j}{k',\chi}$  with $\chi\not=0$ are true if  the \textup{same} $i$ robots of type $j$ visit regions $\ell_k$ and $\ell_{k'}$. Furthermore, the negative atomic proposition  $\neg \ap{i}{j}{k,\chi}$ is equivalent to its basic counterpart $\neg \ap{i}{j}{k}$, i.e., less than $\ag{i}{j}$ robots  are at region $\ell_k$.
 \end{defn}
 Let $\ccalA\ccalP$ collect all basic and induced atomic propositions and  denote by $\Sigma=2^{\ccalA\ccalP}$, its power set.  In what follows, we omit the superscript $\chi$ when $\chi=0$. We denote by {LTL}${^\chi}$, the set of formulas defined over the set of basic and  induced atomic propositions and by {LTL}$^0$, the set of formulas defined only over basic atomic propositions, respectively. Clearly, \ltlx$\supset$ {LTL}$^0$, which means that {LTL}${^\chi}$ is able to capture a broader class of tasks. While there exists literature on feasible control synthesis over \ltlz~\citep{sahin2017synchronous,sahin2019multirobot}, to the best of our knowledge there is no work on optimal control synthesis over~\ltlx formulas. Next, we introduce the notion of {\it valid} temporal logic tasks.
  \begin{defn}[Valid temporal logic task]\label{defn:valid}
 A temporal logic task specified by a\ltlx formula defined over $\ccalA\ccalP$ is valid if atomic propositions with the same nonzero connector $\chi$ involve the same number of robots of the same type.
  \end{defn}

\begin{figure}[!t]
\centering
\includegraphics[width=0.45\linewidth]{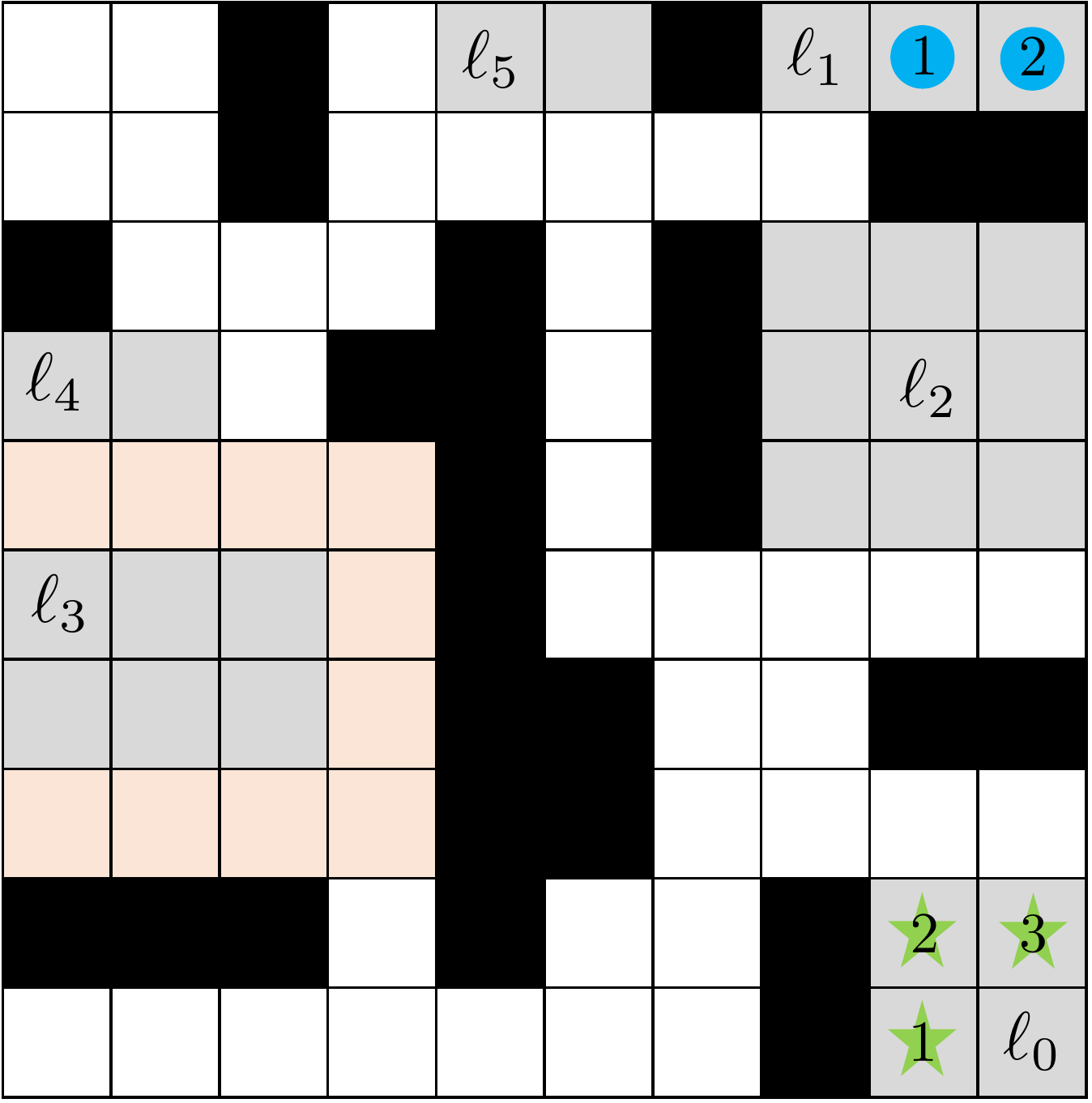}
\caption{Illustration of the workspace for Example~\ref{exmp:1}.}\label{fig:workspace}
\end{figure}
 \begin{exmp}[Valid temporal logic tasks]\label{exmp:1}
   {Consider a mail delivery task amidst the COVID-19 pandemic (shown in Fig.~\ref{fig:workspace}) where three robots of type 1 (green stars) and two robots of type 2 (blue circles) are located at region $\ell_0$ and $\ell_1$, respectively, $\ell_2$ is an office building that the robots visit to pick up the mail, $\ell_3$ and $\ell_5$ are two delivery sites, and $\ell_4$ is a control room from where other robots are driven to  the orange area between $\ell_3$ and $\ell_4$ to get disinfected and then drop  off the mail at the delivery site $\ell_3$.  We consider two delivery tasks: {(i)} \label{task:i} Two robots of type 1 visit building $\ell_2$ to collaboratively pick up  the mail  and deliver it to the delivery site $\ell_3$, and one robot of type 2 must visit the control room $\ell_4$ to disinfect robots of type 1  before they get to  the delivery site $\ell_3$. {\it (ii)} \label{task:ii} One robot of type 1 travels between  building $\ell_2$ and the delivery site $\ell_3$ back and forth to transport  equipment, assuming that the disinfection area operates automatically after task \hyperref[task:i]{(i)}. These tasks are more complex than typical  task allocation problems due to the temporal operators   like ``before'' and ``back and forth''.
  Observe that in Fig.~\ref{fig:workspace}, the  atomic propositions satisfied by initial robot locations  are $\ap{3}{1}{0}$ and $\ap{2}{2}{1}$.}   Moreover,  tasks \hyperref[task:i]{(i)} and \hyperref[task:ii]{(ii)} can be captured by the valid formulas  $\phi_1 = \lozenge \left(\left(\ap{2}{1}{2,1}\wedge \neg \ap{2}{1}{3}\right) \wedge  \lozenge \ap{2}{1}{3,1}\right)  \wedge \lozenge \ap{1}{2}{4} \wedge \neg \ap{2}{1}{3} \,\mathcal{U}\, \ap{1}{2}{4}$ and $ \phi_2 = \square \lozenge \left(\ap{1}{1}{2,1} \wedge \lozenge \ap{1}{1}{3,1}\right)$, respectively. Note that when binding the truth of atomic propositions, the value of $\chi$ is immaterial as long as it is the same non-zero number. Therefore, $\phi_2$ can also be written as $\square \lozenge \left(\ap{1}{1}{2,2} \wedge \lozenge \ap{1}{1}{3,2}\right)$.
  However, formulas $\lozenge \left(\ap{1}{1}{2,1} \wedge \lozenge \ap{2}{1}{3,1}\right)$ and $\lozenge \left(\ap{2}{2}{2,1} \wedge \lozenge \ap{2}{1}{3,1}\right)$ are two invalid formulas as they connect different numbers of robots $i$ and robot types $j$, respectively.
  \end{exmp}
 Let $s^t$ be the collective state at time $t$. A path of length $h$ is defined as   $\tau = s^0 \ldots s^h$ and it captures the collective behavior of the team  such that $s^{t-1}\to s^{t}, \forall t\in[h]$. Given a valid LTL$^\chi$ formula $\phi$, a  path $\tau=\tau^{\text{pre}}[\tau^{\text{suf}}]^\omega$ in a prefix-suffix structure that satisfies $\phi$ exists since there exists an accepting run in prefix-suffix form, where the prefix part  $\tau^{\text{pre}}=s^0 \dots s^{h_1}$ is executed once followed by the indefinite execution of the suffix part $\tau^{\text{suf}}=s^{h_1} \dots s^{h_1+h_2} s^{h_1+h_2+1}$, where $s^{h_1+h_2+1}=s^{h_1}$ \citep{baier2008principles}. We say that a path $\tau$ satisfies $\phi$ if (a) the trace, defined as $\texttt{trace}(\tau):=L(s^0)\dots L(s^{h_1})[L(s^{h_1})\dots L(s^{h_1+h_2+1})]^{\omega}$,  belongs to $\texttt{Words}(\phi^0)$, where $\phi^0$ is obtained by replacing all induced atomic propositions in $\phi$ by  their counterparts with the zero connector and (b) it is the same $\ag{i}{j}$ that satisfy the induced atomic propositions $\ap{i}{j}{k,\chi}$ in $\phi$ sharing the same nonzero connector $\chi$. In other words, condition (a) restricts  the label of the path, while condition (b) restricts the robots that participate in the satisfaction of induced atomic propositions. If $\phi \in \textup{LTL}^0$, the satisfaction conditions only include (a).

 \subsection{Problem definition}
 Given a  path $\tau_{r,j} = s_{r,j}^0, s_{r,j}^1, \ldots, s_{r,j}^h$ of length $h$ for robot $[r,j]$, we define the cost of $\tau_{r,j}$  as $J(\tau_{r,j}) = \sum_{t=0}^{h-1} d(s_{r,j}^{t}, s_{r,j}^{t+1})$, where $d: S\times S \to \mathbb{R}^+\cup\{0\}$ is a cost function that maps a pair of free cells to a non-negative value, for instance, travel distance or time. The cost of path $\tau$ that combines all robot paths $\tau_{r,j}$ of length $h$ is given by
 \begin{align}
   J(\tau) = \sum_{{r \in \ccalK_j, j \in [m]}} J(\tau_{r,j}).
 \end{align}
 For  plans written in prefix-suffix form, we get
 \begin{align}\label{eq:cost}
 J(\tau) = \beta J(\tau^{\text{pre}}) + (1-\beta) J(\tau^{\text{suf}}),
 \end{align}
 where $\beta\in [0,1]$ is a user-specified parameter. Then,  the problem addressed in this paper can be  formulated as follows.
 \begin{problem}\label{prob:1}
   Consider  a discrete workspace with labeled regions and obstacles, a team of $n$ robots of $m$ types, and a valid formula $\phi\in \textit{LTL}^\chi$. Plan a path for each robot such that the specification $\phi$ is satisfied and the cost in~\eqref{eq:cost} is minimized.
 \end{problem}

 We refer to Problem~\ref{prob:1} as the Multi-Robot Task Allocation problem  under LTL specifications or LTL-MRTA. This is a single-task robot and multi-robot task (ST-MR) problem, where a robot is capable of one task and a task may require multiple robots. Since the ST-MR problem is NP-hard~\citep{korsah2013comprehensive,nunes2017taxonomy}, so is the LTL-MRTA problem. Consequently, existing approaches to this problem become intractable  for large-scale applications \citep{sahin2017provably,sahin2017synchronous}. In this work, we propose a new hierarchical  framework  to solve LTL-MRTA problems efficiently.

 \subsection{Assumptions}\label{sec:asmp}
 In this section, we  discuss   assumptions on the workspace and the  NBA translated from the LTL specifications   that are necessary to ensure completeness of our propose hierarchical framework. As we discuss later in Section~\ref{sec:sim}, these assumptions are mild and were satisfied by all tasks we tested our method on, regardless of their complexity.
 \subsubsection{Workspace}
  The following assumption ensures that regions in the workspace are  well-defined and mutually exclusive.
  \begin{asmp}[Workspace]\label{asmp:env}
  Regions are disjoint, and each region  spans consecutive cells. There exists a label-free path between any two regions, between any two label-free cells, and between any label-free cells and any regions.
  \end{asmp}
  If regions are partially overlapping or span multiple clusters of cells, we can define additional atomic propositions to satisfy Assumption~\ref{asmp:env}.  Assumption~\ref{asmp:env} implies that there are no ``holes'' inside regions that generate different labels, label-free cells are connected, and each region is adjacent to a label-free cell.

 \subsubsection{Nondeterministic B$\ddot{\text{u}}$chi Automaton (NBA)}\label{sec:nba}
\begin{figure}[!t]
    \centering
    \subfigure[NBA $\autop$ for the task {(i)}]{
      \label{fig:nba_i}
      \includegraphics[width=0.8\linewidth]{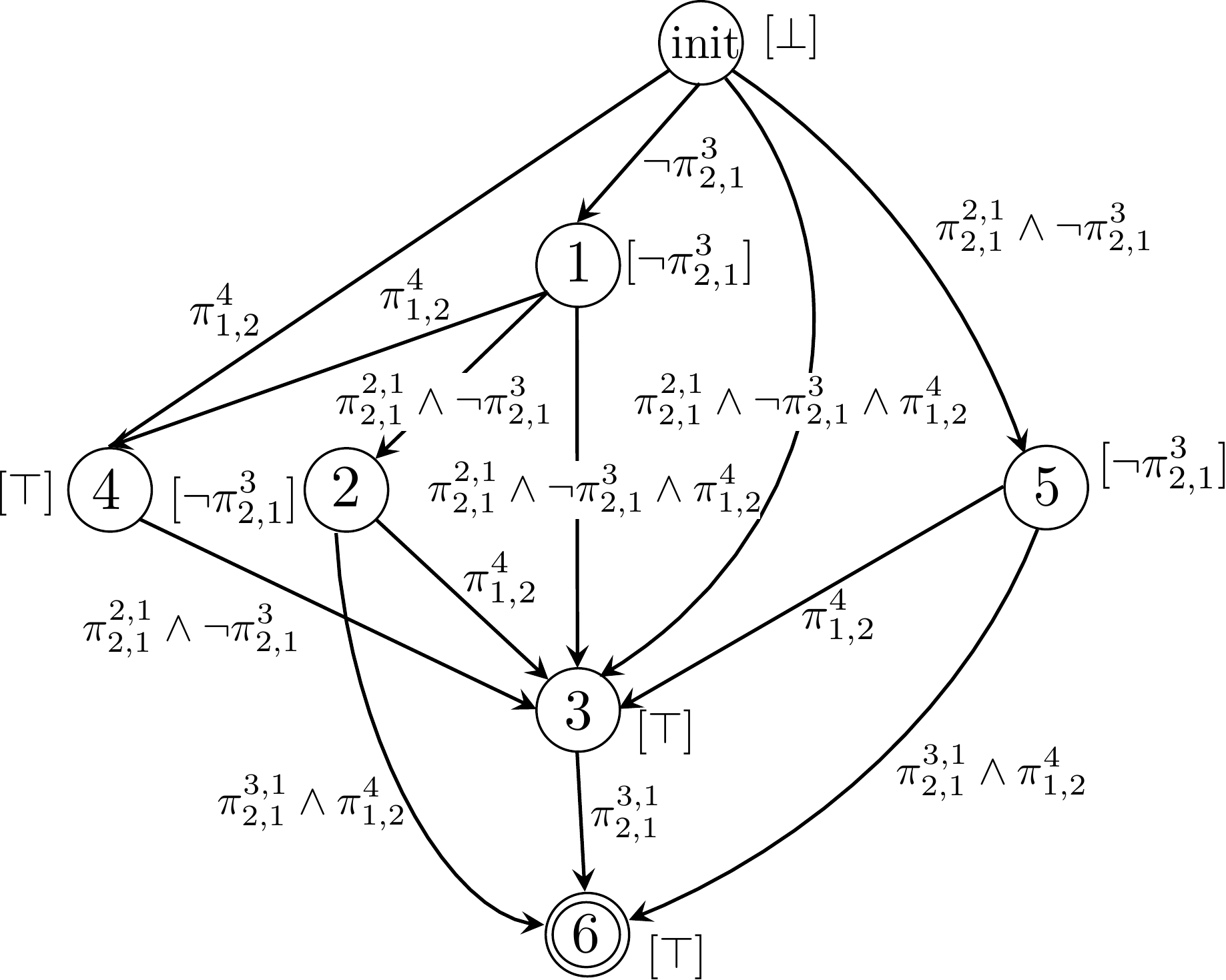}}
    \subfigure[NBA $\autop$ for the task {\it (ii)}]{
      \label{fig:nba_ii}
      \includegraphics[width=0.6\linewidth]{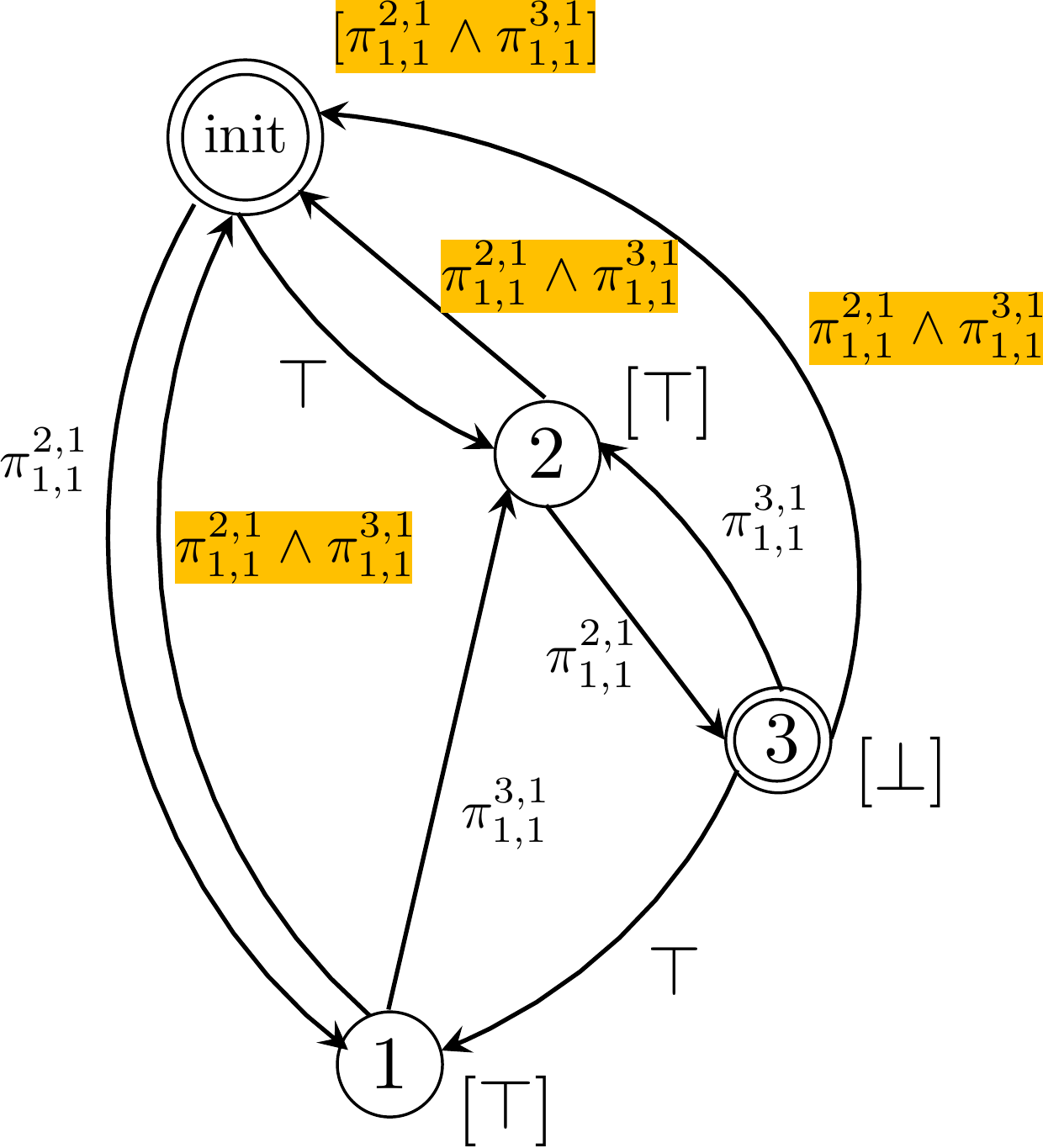}}
    \caption{NBA $\autop$ for tasks \hyperref[task:i]{(i)} and \hyperref[task:ii]{(ii)}, where self-loops are omitted and the corresponding vertex labels are placed in square brackets.}
    \label{fig:nba_iii}
 \end{figure}

 Given a team of $n$ robots and an \ltlx formula $\phi$, we can find a path $\tau$ that satisfies $\phi$ by operating  on the corresponding NBA $\autop = (\ccalV, \ccalE)$, which can be constructed using existing  tools, such as LTL2BA developed by~\cite{gastin2001fast}; see also Fig.~\ref{fig:nba_iii} for the NBA of tasks~\hyperref[task:i]{(i)} and~\hyperref[task:ii]{(ii)}. Note that the NBA in Definition~\ref{def:nba} is essentially a graph. Thus, in the remainder of this paper, we refer to the NBA by the graph $\autop$ for notational convenience. Before we discuss our assumptions on the structure of the NBA $\ccalA_\phi$, we describe a list of pre-processing steps to obtain an "equivalent" NBA that does not lose any feasible paths that satisfy the specification $\phi$. The goal is to remove infeasible and redundant transitions in the NBA to reduce its size.

 Specifically, let the propositional formula $\gamma\in\Sigma$ associated with every transition $v_1 \xrightarrow{\gamma} v_2$ in the NBA $\autop$ be in {\it disjunctive normal form} (DNF), i.e,
 $ \gamma = \bigvee_{p\in \ccalP} \bigwedge_{q\in \ccalQ_p} (\neg)\ap{i}{j}{k,\chi}$, where the negation operator can only precede the atomic propositions and $\ccalP$ and $\ccalQ_p$ are proper index sets. Note that any propositional formula has an equivalent formula in DNF~\citep{baier2008principles}. We call  $\ccalC_p^{\gamma}=\bigwedge_{q\in \ccalQ_p}(\neg) \ap{i}{j}{k,\chi}$ the $p$-th {\it clause} of $\gamma$ that includes a set $\ccalQ_p$ of positive and negative {\it literals} and each positive literal is an atomic proposition $\ap{i}{j}{k,\chi}\in \ccalA\ccalP$. Let $\mathsf{cls}(\gamma)$ denote the set of clauses $\ccalC_p^{\gamma}$ in $\gamma$. And let $\mathsf{lits}^+(\ccalC_p^{\gamma})$ and $\mathsf{lits}^-(\ccalC_p^{\gamma})$ be the {\it positive subformula} and {\it negative  subformula}, consisting of all positive literals and all negative literals in the clause $\ccalC_p^\gamma$. Those subformulas are $\top$ (constant true) if the corresponding literals do not exist. In what follows, we do not consider self-loops when we refer to edges in $\autop$, since self-loops can be captured by vertices. We call the propositional formula $\gamma$ a {\it vertex label} if $v_1=v_2$, otherwise, an {\it edge label}.  With a slight abuse of notation, let {$\gamma: \ccalV \to \Sigma $} and {$\gamma: \ccalV \times \ccalV \to \Sigma$} be the functions that map a vertex and edge in the NBA to its vertex label and edge label, respectively. Given an edge $(v_1, v_2)$, we call labels $\gamma(v_1)$ and $\gamma(v_2)$ the {\it starting} and {\it end} vertex labels, respectively. Next, we  pre-process the NBA $\autop$ by removing infeasible clauses and merging redundant literals.  In particular, given a vertex  or edge label $\gamma$ in $\autop$ we perform the following operations:

 \mysubparagraph{Absorption in $\mathsf{lits}^+(\cp{p}{\gamma})$}{For each clause $\cp{p}{\gamma} \in \clause{\gamma}$, we delete the positive literal $\ap{i}{j}{k} \in \mathsf{lits}^+(\cp{p}{\gamma})$, replacing it with $\top$, if another $\ap{i'}{j}{k,\chi'} \in \mathsf{lits}^+(\cp{p}{\gamma})$ exists such that $i \leq i'$.  This is because  if $\ag{i'}{j}$ are at region $\ell_k$, i.e., $\ap{i'}{j}{k,\chi'}$ is true, so is  $\ap{i}{j}{k}$. Similarly, we replace $\ap{i}{j}{k}$ by $\ap{i-i'}{j}{k}$ if $i > i'$, {since $i-i'$ additional robots are needed to make $\ap{i}{j}{k}$ true if  $\ap{i'}{j}{k,\chi'}$ is true.}}\label{prune:absorption1}

 \mysubparagraph{Absorption in $\mathsf{lits}^-(\cp{p}{\gamma})$} {We delete the negative literal $\neg \ap{i}{j}{k} \in \mathsf{lits}^-(\cp{p}{\gamma})$, if another $\neg \ap{i'}{j}{k} \in \mathsf{lits}^-(\cp{p}{\gamma})$ exists such that $i' < i$. This is because if $\neg \ap{i'}{j}{k}$ is true, so is $\neg \ap{i}{j}{k}$.} \label{prune:absorption2}

 \mysubparagraph{Mutual exclusion in $\mathsf{lits}^+(\cp{p}{\gamma})$}{We delete the clause $\cp{p}{\gamma} \in \mathsf{cls}(\gamma)$, replacing it with constant false $\bot$, if there exist two positive literals $\ap{i}{j}{k,\chi}, \ap{i}{j}{k',\chi} \in \mathsf{lits}^+(\cp{p}{\gamma})$ such that $k\not= k'$ and $\chi\not=0$. The reason is that  the same $i$ robots of type $j$  cannot be at different regions at the same time.} \label{prune:exclusion1}

 \mysubparagraph{Mutual exclusion in $\mathsf{lits}^+(\cp{p}{\gamma})$  and $\mathsf{lits}^-(\cp{p}{\gamma})$} { We delete the clause $\cp{p}{\gamma} \in \mathsf{cls}(\gamma)$  if there exists a positive literal $\ap{i}{j}{k,\chi} \in \mathsf{lits}^+(\cp{p}{\gamma})$ and a negative literal $\neg \ap{i'}{j}{k} \in \mathsf{lits}^-(\cp{p}{\gamma})$ such that $i' \leq i$. This is because these literals   are mutually  exclusive.} \label{prune:exclusion2}

 \mysubparagraph{Violation of team size} {For each clause $\cp{p}{\gamma} \in \clause{\gamma}$, let $\mathsf{lits}^+(j')$ denote literals in $\mathsf{lits}^+(\cp{p}{\gamma})$ that involve robots of type $j'$, i.e., $\mathsf{lits}^+(j') = \{\ap{i}{j}{k,\chi} \in \mathsf{lits}^+(\ccalC_p^\gamma)| j = j'\}$. We delete the clause $\cp{p}{\gamma}$   if the total required number of robots of type $j$ exceeds the size $|\ccalK_j|$, i.e., if there exists $j\in[m]$ such that $ \sum_{\ap{i}{j}{k,\chi}\in\mathsf{lits}^+(j)}   i > |\ccalK_{j}|$.} \label{prune:violation1}

  Note that these pre-processing steps  merely  remove infeasible clauses and merge redundant literals in the NBA $\ccalA_\phi$, and they do  compromise any accepting words in $\ccalL(\autop)$ that can be generated by a feasible path. Therefore, with a slight abuse of notation, we continue to use $\autop$ to refer to the NBA associated with formula  $\phi$ that is obtained  after these pre-processing steps.

 Consider now an edge $e = (v_1, v_2) $  and its starting vertex $v_1$ in the NBA $\autop$ and assume that the current state of $\ccalA_\phi$ is vertex $v_1$. For the NBA to transition to vertex $v_2$, certain robots need to simultaneously reach certain regions  or avoid certain regions in order to make $\gamma(v_1, v_2)$ true, while maintaining  $\gamma(v_1)$ true en route. We assume that the transition to $v_2$ occurs immediately once $\gamma(v_1, v_2)$ becomes true. Therefore, we can define by a {\em subtask} the set of actions that need to be taken by a group of robots in order to activate a transition in the NBA. Formally, we have the following definition.
 \begin{defn}[Subtask]\label{defn:subtask}
   Given an edge $(v_1, v_2)$ in the NBA $\ccalA_\phi$, a subtask is defined by the associated  edge label $\gamma(v_1, v_2)$ and  starting vertex label $\gamma(v_1)$.
 \end{defn}

 Subtasks can be viewed as generalized reach-avoid tasks where specific types of robots should visit or avoid certain regions (the ``reach'' part of the tasks) while satisfying the starting vertex labels along the way (the ``avoid'' part of the tasks, which here is defined in a more general way compared to the conventional definition that requires robots to stay away from given regions in space).

 Note that every accepting run defined in Section~\ref{sec:ltl} consists of a sequence of subtasks, as they are defined in Definition~\ref{defn:subtask}. However, not all sequences of subtasks associated with an accepting run make progress towards accomplishing the task. In what follows, we restrict the accepting runs in an NBA to those that make progress towards accomplishing the task. But first, we provide some intuition using the following example.

  \begin{cexmp}{exmp:1}{Subtask progress in the pre-processed NBA $\autop$}\label{exmp:observation}
    The pre-processed NBAs corresponding to tasks \hyperref[task:i]{(i)} and \hyperref[task:ii]{(ii)} are shown in Fig.~\ref{fig:nba_iii} where the vertex labels are placed in square brackets next to each vertex. After pre-processing, the NBA $\autop$ for task~\hyperref[task:i]{(i)} does not change whereas some labels in the NBA $\autop$ for task \hyperref[task:ii]{(ii)} become $\bot$ due to step~\hyperref[prune:exclusion1]{\it (3)}. These labels are highlighted in orange.

    In Fig.~\ref{fig:nba_i}, $\vertex{init}$ is the initial vertex and $v_6$ is the accepting vertex. Observe that all vertices have self-loops except for the initial vertex $\vertex{init}$. In each accepting run, e.g., $\vertex{init}, v_1, v_2, v_3, v_6, v_6^\omega$, the satisfaction of an edge label leads to the satisfaction of its end vertex label, assuming this end vertex label is not  $\bot$. For instance,  label $\ap{2}{1}{2,1} \wedge \neg\ap{2}{1}{3}$ of  edge $(v_1, v_2)$ implies  label $\neg\ap{2}{1}{3}$ of vertex $v_2$, and  label $\neg\ap{2}{1}{3}$ of edge  $(\vertex{init}, v_1)$ implies  label $\neg\ap{2}{1}{3}$ of  its end vertex $v_1$. Intuitively, the completion of a subtask indicated by the satisfaction of its edge label, automatically activates the subtasks that immediately follow it indicated by the satisfaction of their starting vertex labels. This is because once the edge is enabled, its end vertex label should be satisfied at the next time instant; otherwise, progress  in the NBA $\autop$ will  get stuck. The same observation also applies to the NBA in Fig.~\ref{fig:nba_ii} where the vertex $\vertex{init}$ is both an initial and accepting vertex and $v_3$ is another accepting vertex.  The accepting run $\vertex{init}, v_2, v_3, (v_1, v_2, v_3)^\omega$ includes one pair of initial and accepting vertices, $\vertex{init}$ and  $v_3$, and the accepting run $\vertex{init}, v_2, v_1, \vertex{init}^\omega$ (although infeasible) includes one pair of initial and accepting vertices, $\vertex{init}$ and $\vertex{init}$. Note that we view the two $\vertex{init}$ vertices differently, one as the initial vertex and the other as the accepting vertex. Furthermore, label $\ap{1}{1}{3,1}$ of edge $(v_1, v_2)$  implies  label $\top$ of its end vertex $v_2$; the same holds for the edge $(v_2, \vertex{init})$ and its end vertex $\vertex{init}$ (although infeasible). It is noteworthy that even though the accepting vertex $v_3$ does not have a self-loop, the satisfaction of the label $\ap{1}{1}{2,1}$ of its incoming edge $(v_2, v_3)$ leads to the satisfaction of the label $\top$ of its outgoing edge $(v_3, v_1)$. If the satisfaction of the incoming edge label does not imply satisfaction of the outgoing edge label, then progress in the NBA  will get stuck at $v_3$ since the label $\ap{1}{1}{2,1} \wedge \ap{1}{1}{3,1}$ of edge $(v_3, \vertex{init})$ is infeasible and the transition between regions $\ell_2$ and $\ell_3$ requires more than one time steps; see Fig.~\ref{fig:workspace}, which makes the label $\ap{1}{1}{3,1}$ of edge $(v_3, v_2)$ unsatisfiable at the next time instant.
  \end{cexmp}

  Motivated by the observations in Example~\ref{exmp:1}, we introduce the notions of {\it implication} and {\it strong implication} between two propositional formulas. Then, we define a {\it restricted accepting run} in the NBA $\autop$   in a prefix-suffix structure. The completeness of our method relies on the assumption that the set of restricted accepting runs in the  NBA $\autop$ is nonempty.

  \begin{defn}[Implication and strong implication]\label{defn:implication}
    Given two propositional formulas $\gamma$ and $\gamma'$ over $\ccalA\ccalP$, we say that formula  $\gamma$ implies $\gamma'$, denoted by $\gamma \simplies \gamma'$, if for each clause $\ccalC_{p}^{\gamma} \in \clause{\gamma}$, there exists a clause $\ccalC_{p'}^{\gamma'} \in \clause{\gamma'}$ such that  $\ccalC_{p'}^{\gamma'}$ is a subformula of $\ccalC_{p}^{\gamma}$, i.e., all literals in $\ccalC_{p'}^{\gamma'}$ also appear in $\ccalC_{p}^{\gamma}$. By default, $\top$ is a subformula of any clause. In addition, formula $\gamma$ strongly implies $\gamma'$, denoted by $\gamma \simplies_s  \gamma'$, if $\gamma \simplies \gamma'$, and for each clause $\ccalC_{p'}^{\gamma'} \in \clause{\gamma'}$, there exists a clause $\ccalC_{p}^{\gamma} \in \clause{\gamma}$ such that  $\ccalC_{p'}^{\gamma'}$ is a subformula of $\ccalC_{p}^{\gamma}$.
  \end{defn}

  Intuitively, if $\gamma \simplies \gamma'$ or $\gamma \simplies_s \gamma'$, robot locations that satisfy $\gamma$ also  satisfy  $\gamma'$.

  \begin{figure}[t]
   \centering
   \includegraphics[width=0.7\linewidth]{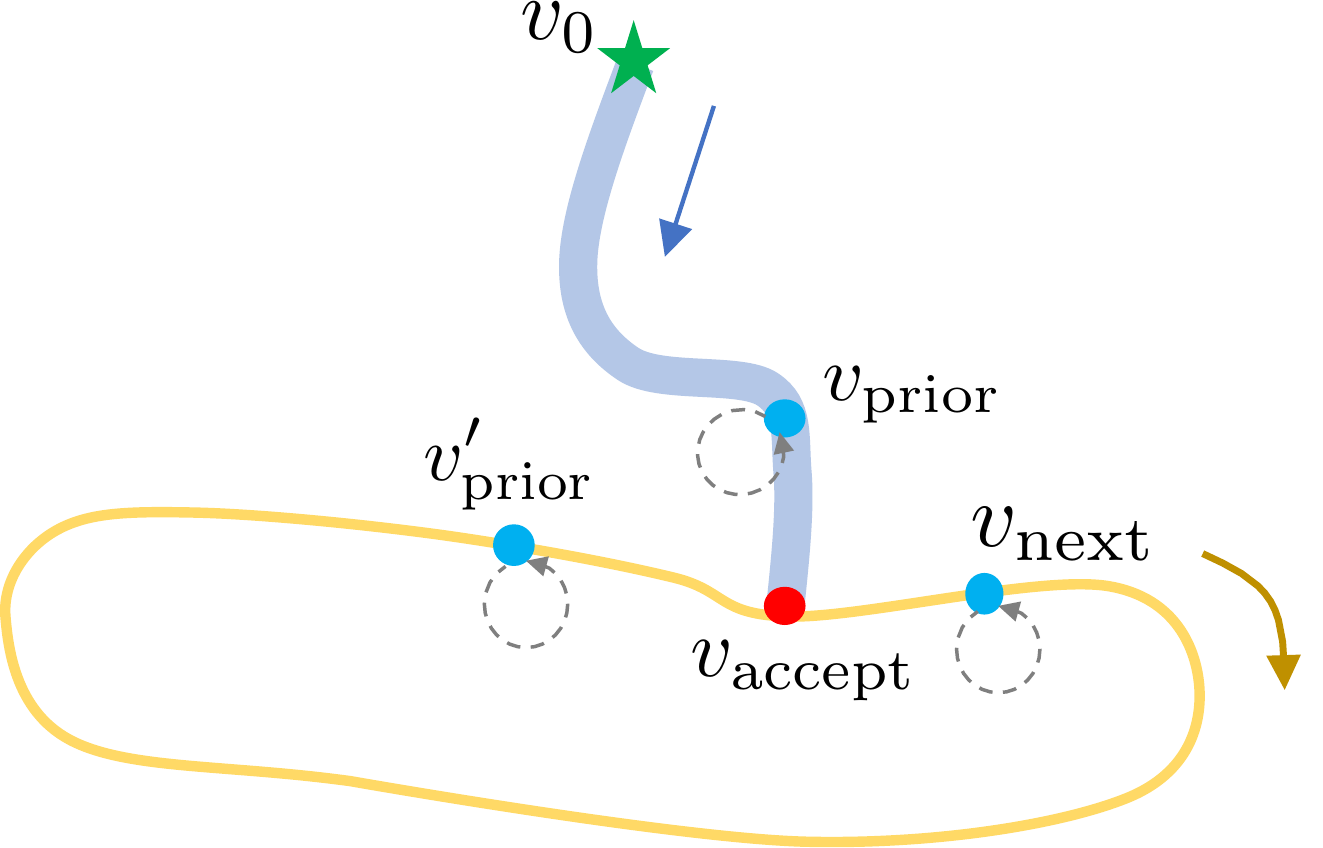}
   \caption{Graphical depiction of the accepting run in the prefix-suffix structure when $\vertex{accept}$ does not have a self-loop, which resembles a lasso. The shaded blue line and the orange loop represent the prefix and suffix part, respectively. The arrow indicates the progression direction and the gray circles indicate the self-loops.}
   \label{fig:lasso}
 \end{figure}

 \begin{defn}[Restricted accepting run]\label{defn:run}
   Given the NBA $\autop$ (after pre-processing) corresponding to an \ltlx formula, we call any accepting run in a
   prefix-suffix structure $\rho = \rho^{\textup{pre}} [\rho^{\textup{suf}}]^\omega= v_0, \ldots, \vertex{prior}, \vertex{accept} [\vertex{next}, \ldots, \vertex{prior}', \vertex{accept}]^\omega$ (see Fig.~\ref{fig:lasso}), a restricted accepting run,  if it  satisfies the following conditions:
   \begin{noindlist}
     \setlength\itemsep{0em}
   \item \label{cond:a} If a vertex is both an initial vertex $v_0 $ and an accepting vertex $\vertex{accept}$, we treat it as two different vertices, namely  an initial vertex and an accepting vertex. The accepting vertex $\vertex{accept}$ appears only once at the end in both the prefix and suffix parts. In the prefix part  $v_0, \ldots, \vertex{prior}, \vertex{accept}$, if a vertex appears multiple times,  all repetitive occurrences are consecutive.  The same holds for  the suffix part  $\vertex{next}, \ldots, \vertex{prior}', \vertex{accept}$;
   \item \label{cond:b} There only exist one initial vertex $v_0$ and one  accepting vertex $\vertex{accept}$ in the accepting run (they can appear multiple times in a row). Different  accepting runs can have different pairs of initial and accepting vertices;
   \item \label{cond:c} In the prefix part, only initial and accepting vertices, $v_0$ and $\vertex{accept}$, are allowed not to have  self-loops, i.e., their vertex labels  can be $\bot$. In the suffix part,  only the accepting vertex $\vertex{accept}$ is allowed not to have a self-loop;
   \item  \label{cond:d} For any two consecutive vertices $v_1$, $v_2$ in the accepting run $\rho$, if $v_1 \neq v_2$, $v_2\neq \vertex{accept}$ and $v_2$ has a self-loop, then the edge label $\gamma(v_1, v_2)$  strongly implies  the end vertex label $\gamma(v_2)$, i.e., $\gamma(v_1, v_2) \simplies_s \gamma(v_2)$;
    \item \label{cond:e} In the suffix part $\rho^{\textup{suf}}$, if $\vertex{accept} = \vertex{next}$ (this happens when $\vertex{accept}$ has a self-loop), then $\rho^{\textup{suf}}$ only contains the vertex   $\vertex{accept}$. Meanwhile, the label of the edge $(\vertex{prior}, \vertex{accept})$  implies the label of the vertex $\vertex{accept}$, i.e., $\gamma(\vertex{prior}, \vertex{accept}) \simplies  \gamma(\vertex{accept})$;
  \item \label{cond:f} In the suffix part, if $\vertex{accept} \neq \vertex{next}$ (this can happen when $\vertex{accept}$  does not have a self-loop), then the label of the edge $(\vertex{prior}, \vertex{accept})$ implies the label of the edge $(\vertex{accept}, \vertex{next})$, i.e., $\gamma(\vertex{prior}, \vertex{accept}) \simplies  \gamma(\vertex{accept}, \vertex{next})$. Also, the label $\gamma(\vertex{prior}, \vertex{accept})$ implies the label of the  edge $(\vertex{prior}', \vertex{accept})$, i.e., $\gamma(\vertex{prior}, \vertex{accept}) \simplies  \gamma(\vertex{prior}', \vertex{accept})$. Note that $\vertex{prior}$ and $\vertex{prior}'$ can be different.
   \end{noindlist}
  \end{defn}

 In what follows, we discuss the conditions in Definition~\ref{defn:run} in more detail. {Specifically, conditions~\ref{cond:a} and~\ref{cond:b} require that a restricted accepting run is  ``simple''.  Specifically, condition~\ref{cond:a} states that vertices  $v_0$ and $\vertex{accept}$ can be treated differently since they mark different progress towards accomplishing a task. The prefix and suffix parts of a restricted accepting run  end once  $\vertex{accept}$ is reached, as in~\cite{smith2010optimal}. By aggregating consecutive identical vertices in the prefix part of  a restricted accepting run into one single vertex,  there are no identical vertices in the ``compressed'' prefix part. That is,  it contains no cycles. The presence of a cycle is redundant since it implies negative progress towards accomplishing the task.  The same applies to the suffix part.}  On the other hand,  condition~\ref{cond:b} states that a restricted accepting run is basically an accepting run defined in Section~\ref{sec:ltl} that is further defined over a pair of initial and accepting vertices. In Section~\ref{sec:app}, we extract smaller sub-NBAs from the NBA $\autop$ for each pair of initial and accepting vertices, which helps reduce complexity of the problem.

 Conditions~\ref{cond:c}-\ref{cond:f} require that  the completion of a subtask in a restricted accepting run   automatically activates the subtasks that immediately follow it; see Example~\ref{exmp:1}. This ensures that robots are given adequate time to undertake subsequent subtasks after completing the current subtask. Accepting runs that do not satisfy conditions \ref{cond:c}-\ref{cond:d} are disregarded. In fact, in Section~\ref{sec:prune} we prune vertices and edges in the NBA that violate these conditions, further reducing the size of the NBA.  Finally, the implication $\gamma(\vertex{prior}, \vertex{accept}) \simplies  \gamma(\vertex{accept}, \vertex{next})$ in condition~\ref{cond:f}  requires that the robot locations enabling the last edge in the prefix part of a restricted accepting run  also enable the first edge in the suffix part. As a result, we can find the prefix and suffix parts of a restricted accepting run separately. Otherwise, the progress in the NBA $\autop$  may get stuck since these two edge labels need to be satisfied at two consecutive time instants, similar to conditions~\ref{cond:d} and \ref{cond:e}. Also, as the suffix part of a restricted accepting run  is a loop, robots need to return to their initial locations in the suffix part after executing the suffix part once. The relation $\gamma(\vertex{prior}, \vertex{accept}) \simplies  \gamma(\vertex{prior}', \vertex{accept})$ requires  that the  initial locations in the suffix part of a restricted accepting run  enable the edge  $(\vertex{prior}', \vertex{accept})$, which ensures that the robots can travel back to the initial locations in the suffix part and, as a result, activate  the transition in  $\autop$ back to the vertex $\vertex{accept}$ that allows to repeat the suffix part $\rho^{\text{suf}}$. Finally, we make the following assumption on the structure of the NBA $\ccalA_\phi$.

 \begin{asmp}[Existence of restricted accepting runs]\label{asmp:run}
   The set of restricted accepting runs in the NBA $\ccalA_\phi$ is non-empty.
 \end{asmp}

We note that the sets of restricted accepting runs for tasks~\hyperref[task:i]{(i)} and~\hyperref[task:ii]{(ii)} satisfy Assumption~\ref{asmp:run}. Common robotic tasks, such as sequencing and coverage, have NBAs that contain restricted accepting runs. However, there is also a small subclass of LTL where the ``next"  operator  directly precedes an atomic proposition that violates this assumption. For instance, $\lozenge (\pi_{1,1}^{2,1} \wedge \bigcirc \pi_{1,2}^{3,1})$ requires a second  robot to visit region $\ell_3$ immediately after the first robot reaches $\ell_2$, which does not allow for any physical time between the completion of the two consecutive subtasks. On the other hand, the LTL formula $\lozenge (\pi_{1,1}^{2,1} \wedge \bigcirc (\pi_{1,1}^{2,1} \mathcal{U} \pi_{2,1}^3))$ satisfies the assumption.


 \subsubsection{Robot paths}
       The definition of restricted  accepting runs is based entirely on the structure of the  NBA and logical implication relations. However, Definition~\ref{defn:run} does not describe how to characterize robot paths that induce restricted accepting runs. In what follows, we discuss conditions under which robot paths satisfy restricted accepting runs. We call such paths {\em satisfying paths} and we assume that such satisfying paths exist.

 \begin{defn}[Satisfying paths of restricted accepting runs]\label{defn:same}
   Given a team of $n$ robots and a valid specification $\phi \in \textit{LTL}^\chi$, a robot path $\tau$ is a satisfying path that induces a restricted accepting run, if the following conditions hold:
   \begin{noindlist}
   \item \label{asmp:a} If a vertex label is satisfied by the path $\tau$, it is always satisfied by the same clause that is always satisfied by the same fleet of robots;
   \item  \label{asmp:b} If a clause in an edge label is satisfied by the path $\tau$, then a clause in the end vertex label is also satisfied. Moreover, the fleet of robots satisfying the positive subformula of the clause in the end vertex label is the same as the fleet of robots satisfying the positive  subformula of the clause in the corresponding edge label;
     \item \label{asmp:c} Robot locations  enabling the edges $(\vertex{accept}, \vertex{next})$ and  $(\vertex{prior}', \vertex{accept})$ in the suffix part of a restricted accepting run  are identical to robot locations enabling the  edge $(\vertex{prior}, \vertex{accept})$ in the prefix part.
   \end{noindlist}
 \end{defn}

 Definition~\ref{defn:same} is closely related to the definition of a restricted accepting run. Specifically, condition~\ref{asmp:a} in Definition~\ref{defn:same} requires that once a fleet of robots satisfies a vertex label in a restricted accepting run, then these robots remain idle during the next time instant so that the same clause in this vertex label is still satisfied. This satisfies condition~\ref{cond:a} in Definition~\ref{defn:run}. Furthermore, condition~\hyperref[asmp:b]{(b)} in Definition~\ref{defn:same} requires that once a fleet of robots satisfies an edge label in a restricted accepting run, then these robots remain idle during the next time instant so that the clause in the end vertex label that is implied by the clause that is satisfied in the edge label is also satisfied. This satisfies condition~\ref{cond:d} in Definition~\ref{defn:run}.

 Finally, condition~\hyperref[asmp:c]{(c)}  in Definition~\ref{defn:same} requires that the robot locations  enabling the edge $(\vertex{prior}, \vertex{accept})$ in the prefix part of a restricted accepting run coincide with the initial locations of the robots that enable the edge  $(\vertex{prior}', \vertex{accept})$ in the suffix part of the restricted accepting run, as per condition~\ref{cond:f} in Definition~\ref{defn:run}. Therefore, condition~\hyperref[asmp:c]{(c)} in Definition~\ref{defn:same} requires that the robots travel along a loop
  so that the suffix part of the restricted accepting run is executed indefinitely. In what follows, we make the following assumption.

 \begin{asmp}[Existence of satisfying paths]\label{asmp:same}
   There exist robot paths that satisfy the restricted accepting runs in the NBA $\ccalA_\phi$.
 \end{asmp}

 \subsection{Outline of the proposed method}
          An overview of our proposed method is shown in Alg.~\ref{alg:ltlmrta}, which first finds prefix paths and then suffix paths.  The process of finding prefix or suffix paths consists of relaxation and correction stages. 
             Specifically, during the relaxation stage, we ignore the negative literals in the NBA $\ccalA_\phi$ and formulate a MILP to allocate subtasks to robots and determine time-stamped robot waypoints  that satisfy the task assignment. To this end, we first prune the NBA $\autop$ by deleting infeasible transitions and then relax it by removing negative subformulas  so that transitions in the relaxed NBA are solely satisfied by robots that meet at certain regions [line~\ref{ltl:relax}]; see Section~\ref{sec:prune}. The idea to temporarily remove negative literals from the NBA is motivated by ``lazy collision checking'' methods in robotics and allows to simplify the planning problem as the constraints are not considered during planning and are only checked during execution, when needed. {Then,  since by condition~\ref{cond:b} in Definition~\ref{defn:run}, restricted accepting runs contain  only one initial vertex and one accepting vertex, for every sorted pair of initial and accepting vertices by length in the relaxed NBA, we extract a sub-NBA of smaller size [line~\ref{ltl:prefix}];} see Section~\ref{sec:pregraph}, where $\auto{subtask}(\vertex{init}, \vertex{accept})$ is the sub-NBA including only initial vertex $\vertex{init}$, accept vertex $\vertex{accept}$ and other intermediate vertices. The sub-NBAs are used to extract subtasks and temporal orders between them captured by a set of posets ([lines~\ref{ltl:pruneprefix}-\ref{ltl:inferprefix}], see Section~\ref{sec:poset}), and construct routing graphs, one for each poset,  that capture the regions that the robots need to visit and the temporal order of the visits so that the subtasks extracted from the sub-NBAs are satisfied; see Section~\ref{sec:graph}. Finally, given the routing graph corresponding to each poset we formulate a MILP inspired by the vehicle routing problem to obtain a high-level task allocation plan along with time-stamped waypoints that the robots need to visit to satisfy the task assignment [lines~\ref{ltl:prefixgraph}-\ref{ltl:prefixmilp}]; see Section~\ref{sec:path}. During the correction stage, we introduce the negative literals back into the NBA and formulate a collection of generalized multi-robot path planning problems, one for each poset, to design low-level executable robot paths that satisfy the original specification ([line~\ref{ltl:prefixgmrpp}], see Section~\ref{sec:lowlevel}). Viewing the final states of the prefix paths as the initial states, a similar process is conducted for the sub-NBA $\auto{subtask}(\vertex{accept}, \vertex{accept})$ to find the suffix paths. Alg.~\ref{alg:ltlmrta} can terminate after a specific number of paths is found or all possible alternatives are explored.  Under the mild assumptions discussed in Section~\ref{sec:asmp}, completeness of our proposed method is shown in Theorem~\ref{thm:completeness} in Section~\ref{sec:correctness}.

\begin{algorithm}[!t]
       \caption{Algorithm for LTL-MRTA}
       \LinesNumbered
       \label{alg:ltlmrta}
       
       Prune and relax the NBA \label{ltl:relax}\;
         \ForEach{\textup{sorted sub-NBA} $\auto{subtask}(\vertex{init}, \vertex{accept})$ \label{ltl:prefix}}{
                \Comment*[r]{Compute the prefix path}
         Prune the sub-NBA $\auto{subtask}(\vertex{init}, \vertex{accept})$\label{ltl:pruneprefix}\;
         Infer the set of posets $\{P_{\textup{pre}}\}$\label{ltl:inferprefix}\;
         \ForEach{\textup{sorted poset} $P_{\textup{pre}}$}{
            Build the routing graph \label{ltl:prefixgraph}\;
            Formulate MILP to get prefix high-level plans \label{ltl:prefixmilp}\;
            Formulate generalized multi-robot path planning to get prefix low-level paths \label{ltl:prefixgmrpp}\;
                            \Comment*[r]{Compute the suffix path}
            Prune the sub-NBA $\auto{subtask}(\vertex{accept}, \vertex{accept})$\label{ltl:prunesuffix}\;
                     Infer the set of posets $\{P_{\textup{suf}}\}$\label{ltl:infersuffix}\;
         \ForEach{\textup{sorted poset} $P_{\textup{suf}}$}{
            Build the routing graph\label{ltl:suffixgraph}\;
            Formulate MILP to get suffix high-level plans\label{ltl:suffixmilp}\;
            Formulate generalized multi-robot path planning to get suffix low-level paths\label{ltl:suffixgmrpp}\;
         }
         }}   
\end{algorithm}  


 \begin{rem}
 We note that Assumption~\ref{asmp:same} on the existence of satisfying paths is only a sufficient condition  that needs to be satisfied to show completeness of our proposed method, as shown in the theoretical analysis of Section~\ref{sec:correctness}. The robot path returned by our method may not be a satisfying path, although  it still satisfies the specification $\phi$.
 \end{rem}

 \section{Extraction of  Subtasks from the NBA and Inferring their  Temporal Order}\label{sec:app}
 In this section, we first prune and relax the NBA $\ccalA_\phi$ by removing infeasible transitions and negative literals. As discussed before, this step is motivated by "lazy collision checking" methods in robotics and allows to simplify the planning problem by checking constraint satisfaction during the execution of the plans rather than their synthesis. Then, we extract sub-NBAs from the relaxed NBA and use these sub-NBAs to obtain sequences of subtasks and a temporal order between them that need to be satisfied so that the global specification is satisfied. 
  \subsection{{Pruning and relaxation of the NBA}}\label{sec:prune}
 To prune infeasible transitions from the NBA $\ccalA_\phi$, we first delete  all  edges labeled with $\bot$, as they cannot be enabled. We also delete vertices and edges in $\ccalA_\phi$ that do not belong to restricted accepting runs, as defined in Definition~\ref{defn:run}. Specifically, we delete all  vertices without self-loops except for the initial and accepting vertices, as per condition \ref{cond:c} in Definition~\ref{defn:run}. Furthermore, for every  vertex other than the accepting vertex, we delete all its incoming edges with edge labels that do  not strongly imply its vertex label, as per  condition~\ref{cond:d} in Definition~\ref{defn:run}. Finally, we delete every vertex, except for the initial vertex, that cannot be reached by other vertices.  We note that  these pruning steps do  not compromise any feasible solution to Problem~\ref{prob:1} that induces a restricted accepting run in $\autop$, as shown in~Lemma~\ref{prop:prune} in Appendix~\ref{app:correctness}.

 We denote by $\autop^-$ the resulting pruned  NBA.  Given the pruned NBA $\autop^-$, we further relax it by replacing each negative literal in vertex or edge labels with $\top$. Let $\auto{relax}$ denote the relaxed NBA. Note that, when the specification $\phi$ does not involve negative atomic propositions, we have  $\auto{relax} = \autop^-$.  Furthermore, Lemma~\ref{prop:inclusion} in Appendix~\ref{app:correctness} states that the language accepted by $\autop^-$ is included in the language accepted by $\auto{relax}$, so this relaxation step does not remove feasible solutions to Problem~\ref{prob:1}. In other words, $\auto{relax}$ is an over-approximation of $\autop^-$. However, a solution to Problem~\ref{prob:1} based on $\auto{relax}$ may not satisfy the specification $\phi$. Note that $\autop^-$ and $\auto{relax}$  are sub-NBAs  of $\autop$ in terms of vertices and edges. Thus, labels and runs in $\autop^-$ and $\auto{relax}$ can be mapped to labels and runs in  $\autop$. For instance, for an edge label $\gamma$ in $\auto{relax}$, we denote by $\gamma_{\phi}$ the corresponding label in $\autop$ (including  negative literals).

 \begin{figure}
   \centering
   \subfigure[$\auto{relax}$ for task {(i)}]{
     \includegraphics[width=0.5\linewidth]{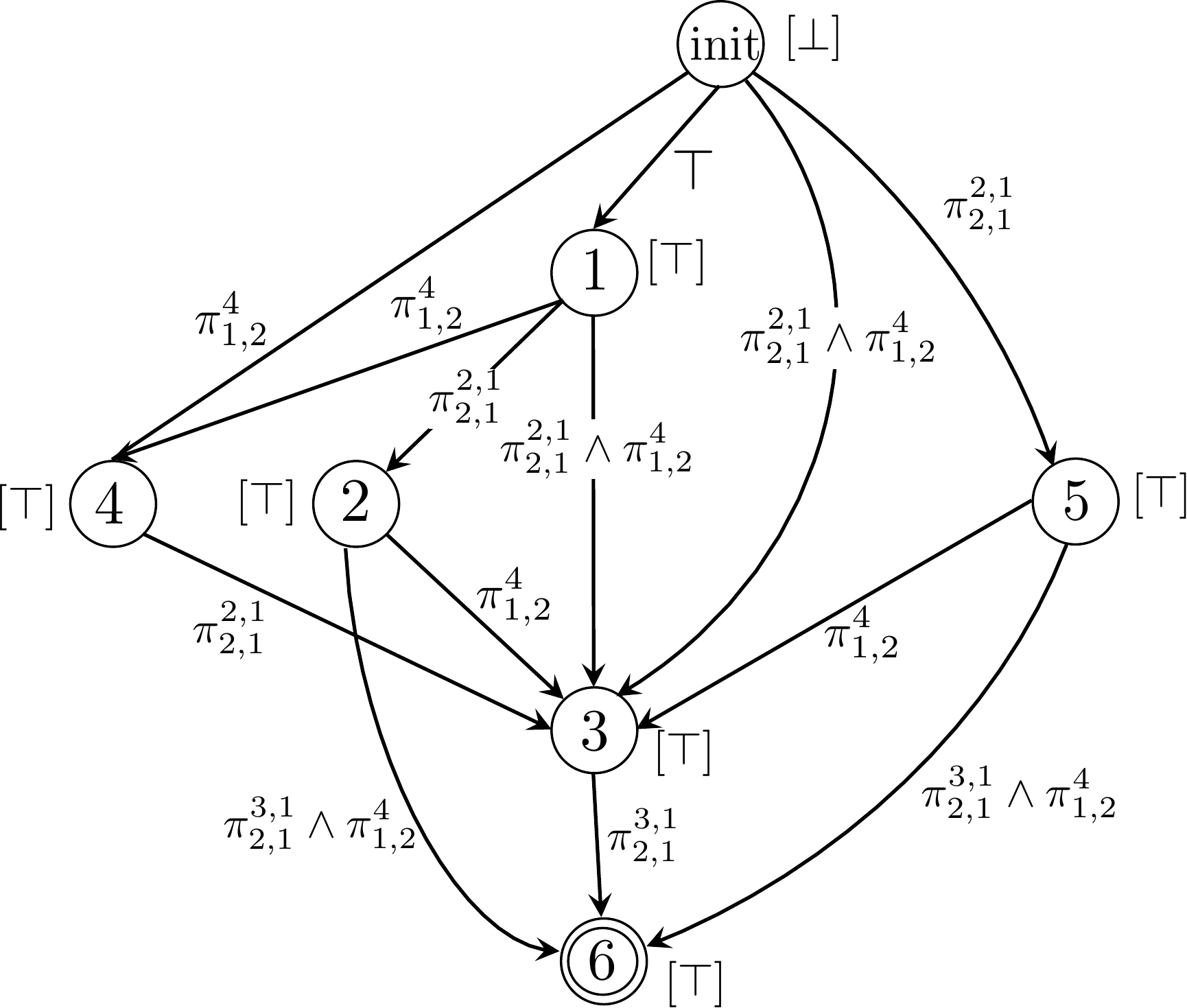}\label{fig:nba_i_relax}}
    \subfigure[$\auto{relax}$ for task {\it (ii)}]{
      \includegraphics[width=0.35\linewidth]{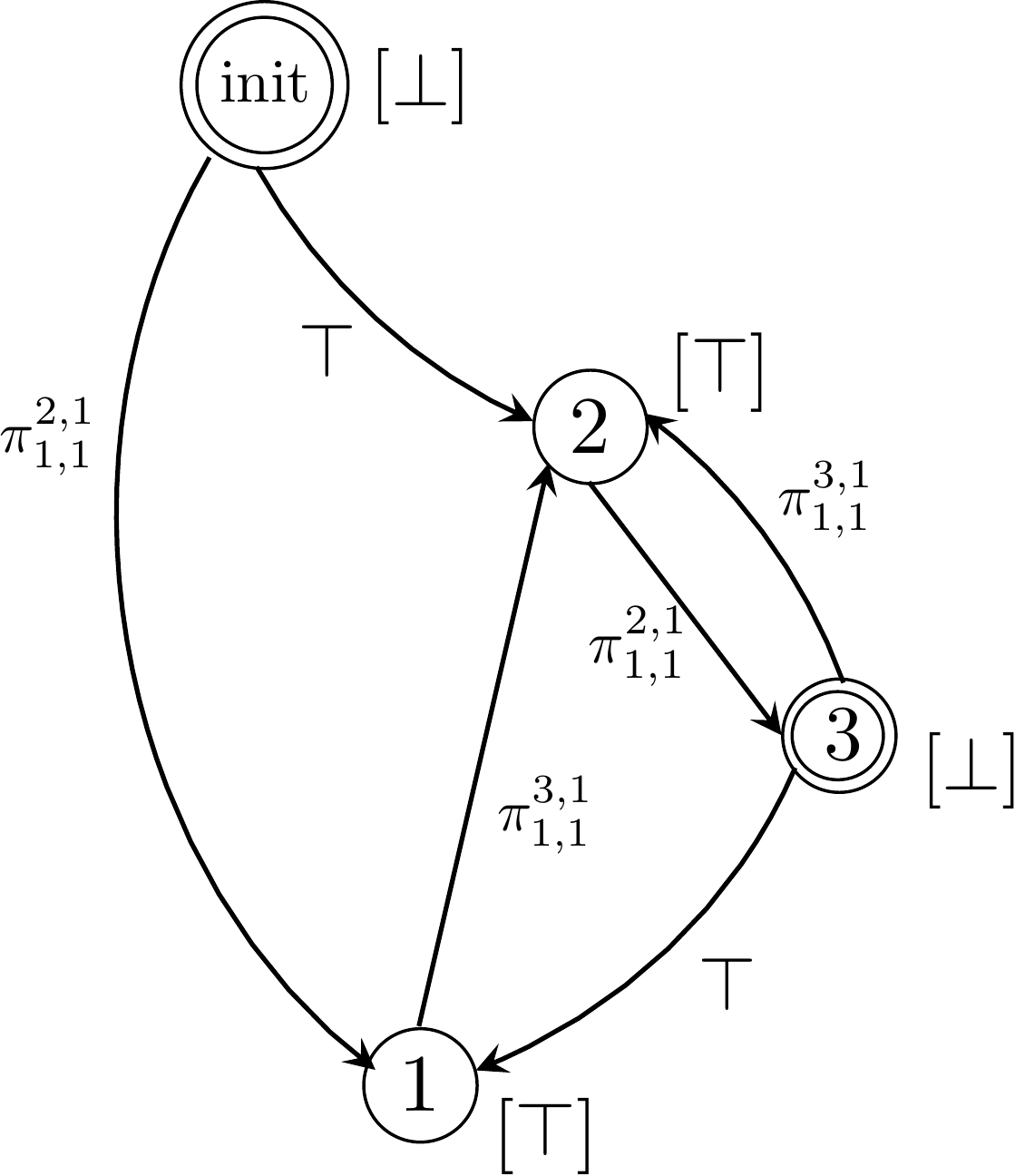}\label{fig:nba_ii_prune}}
    \caption{The relaxed NBA $\auto{relax}$ for tasks~\hyperref[task:i]{(i)} and~\hyperref[task:ii]{(ii)}.}
    \label{fig:relax}
 \end{figure}

 \begin{cexmp}{exmp:1}{Pruning and relaxation of the NBA $\autop$}
    The pruned NBA $\autop^-$ for the task \hyperref[task:i]{(i)} is the same as the original NBA in Fig.~\ref{fig:nba_i}. The relaxed NBA $\auto{relax}$ is shown in Fig.~\ref{fig:nba_i_relax}. The pruned NBA $\autop^-$ for the task \hyperref[task:ii]{(ii)} is the same as the relaxed NBA $\auto{relax}$ which is shown in Fig.~\ref{fig:nba_ii_prune}. Particularly, $\autop^-$  is obtained from $\autop$ in Fig.~\ref{fig:nba_ii} by removing edges $(v_1, \vertex{init}), (v_2, \vertex{init}),  (v_3, \vertex{init})$ and replacing $\gamma(\vertex{init})$ with $\bot$.
 \end{cexmp}

  \subsection{Extraction of sub-NBA \upshape  $\auto{subtask}$  from $\auto{relax}$}\label{sec:pregraph}
  In this section, we extract multiple sub-NBAs from the relaxed NBA $\auto{relax}$, one for every pair of initial and accepting vertices in $\auto{relax}$. Then, in Section~\ref{sec:poset}, we determine the temporal order among subtasks in every sub-NBA.

  \subsubsection{Sorting the pairs of initial and accepting vertices by path length}\label{sec:sort}  As required by condition \ref{cond:b} in Definition~\ref{defn:run}, every  restricted accepting run  in $\auto{relax}$ contains   one pair of initial and accepting vertices. In what follows, we sort all pairs of initial and accepting vertices in $\auto{relax}$ in an ascending order so that the pair of initial and accepting vertices connected by a restricted accepting run with the shortest length appears first. Then in Section~\ref{sub-NBA:1}, we extract a sub-NBA from $\auto{relax}$ for each pair in this ascending order. Intuitively, the sub-NBAs corresponding to  restricted accepting runs of shorter length generally will contain fewer subtasks to be completed.

  \paragraph{Computation of the shortest simple prefix path}{Given a pair of an initial vertex $v_0$ and an accepting vertex $\vertex{accept}$ in $\auto{relax}$, we first compute  the shortest simple path from  $v_0$ to $\vertex{accept}$ in terms of the number of edges/subtasks,  where a simple path does not contain any  repeating vertices, as per  condition~\ref{cond:a} in Definition~\ref{defn:run} that excludes cycles from restricted accepting runs. This step corresponds to the prefix part of a restricted accepting run.
   To this end, we first remove all other initial vertices and accepting vertices from $\auto{relax}$. This will not affect the restricted  accepting runs in $\auto{relax}$ associated with the pair  $v_0$ and $\vertex{accept}$ due to condition \ref{cond:b}  in Definition~\ref{defn:run}.  Then, depending on whether   the initial vertex $v_0$ has a self-loop, we proceed as follows.}\label{para:path}

 \mysubparagraph{If $v_0$ does not have a self-loop, i.e., $\gamma(v_0)=\bot$}{We remove all outgoing edges  $v_0$ in $\auto{relax}$ with label $\gamma$,   if the initial robot locations do not satisfy the corresponding edge label $\gamma_{\phi}$ (including the negative literals) in $\autop$. We emphasize that we need to check satisfaction of $\gamma_\phi$ in the NBA $\ccalA_\phi$ instead of satisfaction of $\gamma$ in the relaxed NBA $\auto{relax}$, since if initial robot locations cannot enable an edge starting from $v_0$ in $\autop$, there is no reason to consider  this edge in any NBA.}\label{sec:initial}

 \mysubparagraph{If $v_0$ has a self-loop, i.e., $\gamma(v_0)\neq\bot$}{We check whether the initial robot locations satisfy $\gamma_{\phi}(v_0)$ in the NBA $\ccalA_\phi$. If yes, we do nothing; otherwise, we proceed as in case \ref{sec:initial} in this part and  remove the self-loop of $v_0$ as well as all its outgoing edges in  $\auto{relax}$ if the initial robot locations do not satisfy the corresponding edge label $\gamma_{\phi}$ in $\autop$.}

 Next, the shortest simple path connecting $v_0$ and $\vertex{accept}$ can be found using Dijkstra's algorithm.  Note that if a vertex is both an initial and accepting vertex, we treat it once  as the initial vertex and once as the accepting vertex, although it appears twice in the shortest simple path.

 \paragraph{Computation of the shortest simple suffix  cycle} {Next, we compute the shortest simple cycle around $v_\text{accept}$ in $\auto{relax}$, where repeating vertices only appear at the beginning and at the end of the simple cycle.} \label{sec:shortestcycle} This step corresponds to the suffix part of a restricted accepting run, which is conducted in the original NBA  $\auto{relax}$.  If $\vertex{accept}$ in $\auto{relax}$ has a self-loop, then the length of the shortest simple cycle is 0. Otherwise, similar to steps in Section~\ref{para:path}  used to find  the shortest simple prefix path, we first remove all other accepting vertices from~$\auto{relax}$ and then remove all initial vertices (including $v_0$) if they do no have self-loops.  In this way, the only vertex that does not have a self-loop is the accepting vertex $\vertex{accept}$.  This will not affect those restricted accepting runs that are related to  $v_0$ and $\vertex{accept}$  due to conditions \ref{cond:b} and \ref{cond:c} in Definition~\ref{defn:run}.


 Finally, the length associated with the pair $v_0$ and $\vertex{accept}$ is equal to the total length of the shortest simple prefix path and the shortest simple suffix cycle connecting these vertices in $\auto{relax}$. By default, if no simple path or cycle exists for the pair $v_0$ and $\vertex{accept}$, the length is infinite, which means there is no restricted accepting run for this pair. We repeat this process for all pairs of initial and accepting vertices in $\auto{relax}$ and sort them in ascending order in terms of the total length. As discussed before, we plan first for pairs with shorter length since they contain  fewer subtasks to be completed.

 \subsubsection{Extraction of the sub-NBA \upshape $\auto{subtask}$}\label{sub-NBA:1}
 For every  pair of vertices  $v_0$ and $v_\text{accept}$ in $\auto{relax}$ connected by a simple path of finite total length in the above ascending order, our goal is to determine time-stamped task allocation plans for all robots that induce the simple prefix path and simple suffix cycle in $\auto{relax}$ connecting $v_0$ and $\vertex{accept}$. To do this,  we extract one sub-NBA from the NBA $\auto{relax}$ that we can use to construct the prefix part of the plan and one that we can use to construct the suffix part of the plan, respectively.  Here, we discuss the sub-NBA for the prefix part. The  sub-NBA for the suffix part is similar and is discussed in Appendix~\ref{sec:suf}.

 Given the pair of vertices $v_0$ and $\vertex{accept}$, we construct a prefix sub-NBA $\auto{subtask}$ by the following three steps. {First, we follow exactly the same steps in Section~\ref{para:path} that computes the shortest simple prefix path to prune the NBA $\auto{relax}$.}
   Next, we remove all outgoing edges from $\vertex{accept}$ if $\vertex{accept} \not=v_0$, because we focus on the prefix part. Finally, let $\ccalV_{\text{s}}$ denote the set that contains all remaining vertices in $ \auto{relax}$ that belong to some path  connecting $v_0$ and $\vertex{accept}$. Then, we construct a sub-NBA $\auto{subtask}=(\ccalV_{\text{s}}, \ccalE_\text{s})$ from $\auto{relax}$ that includes all edges that connect the vertices in $\ccalV_{\text{s}}$. The sub-NBA $\auto{subtask}$ contains prefix parts of all restricted accepting runs associated with the pair $v_0$ and $\vertex{accept}$.

  \begin{figure}
        \centering
        \subfigure[$\auto{subtask}$ for task~{(i)}]{
          \includegraphics[width=0.45\linewidth]{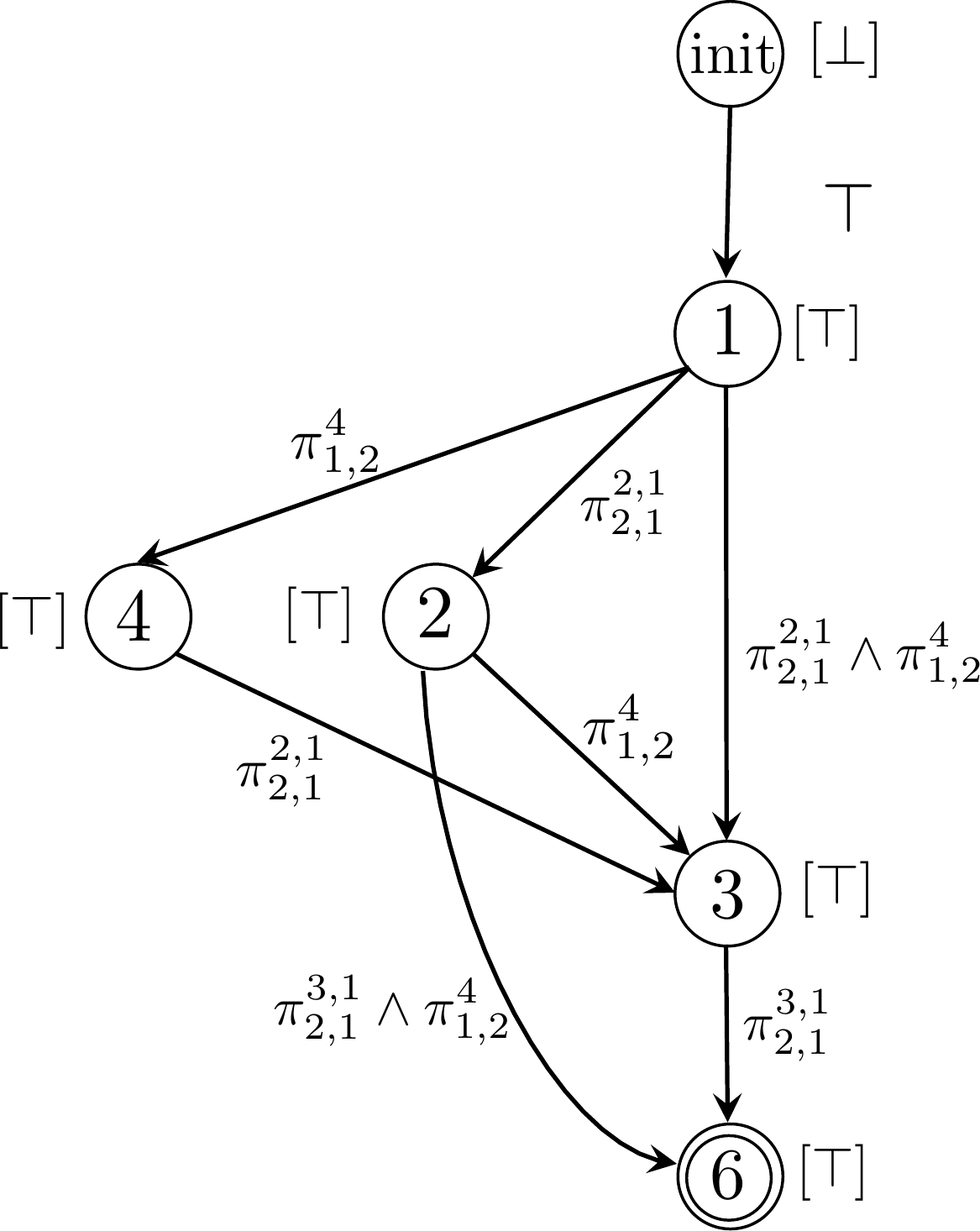}\label{fig:nba_i_subtask}}
        \subfigure[$\auto{subtask}$ for task~{(ii)}]{
          \includegraphics[width=0.3\linewidth]{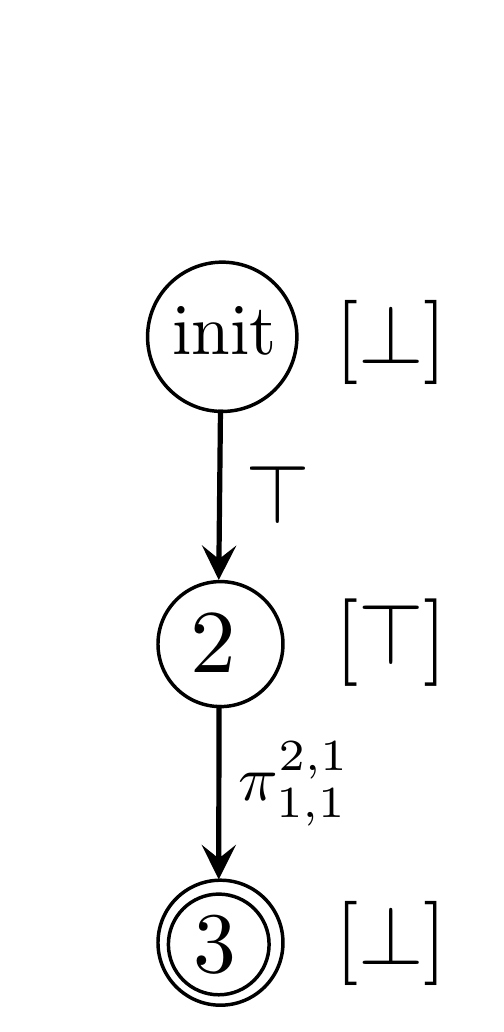}\label{fig:nba_ii_subtask}}
        \caption{Sub-NBA $\auto{subtask}$ for the prefix part of tasks~\hyperref[task:i]{(i)} and~\hyperref[task:i]{(ii)} in Example~\ref{exmp:1}, obtained from the NBA $\auto{relax}$ in Fig.~\ref{fig:relax}.}
        \label{fig:auto_subtask}
      \end{figure}

     \begin{cexmp}{exmp:1}{Sub-NBA $\auto{subtask}$}
        The sub-NBA $\auto{subtask}$ for the prefix parts of plans associated with tasks~\hyperref[task:i]{(i)} and~\hyperref[task:i]{\it (ii)} are shown in~Fig.~\ref{fig:auto_subtask}. For task~\hyperref[task:i]{(i)}, given the pair $\vertex{init}$ and $v_6$ in the relaxed NBA $\auto{relax}$ in Fig.~\ref{fig:nba_i_relax}, the total length is $3+0=3$ (edges $(\vertex{init}, v_3), (\vertex{init}, v_4), (\vertex{init}, v_5)$ were removed since $\vertex{init}$ does not have a self-loop and  all  robots initially located inside region $\ell_1$  do not satisfy their labels; see Fig.~\ref{fig:workspace}). The NBA $\auto{subtask}$, shown in  Fig.~\ref{fig:nba_i_subtask} is obtained by removing edges $(\vertex{init}, v_3), (\vertex{init}, v_4), (\vertex{init}, v_5), (v_5, v_3),  (v_5, v_6)$ and vertex $v_5$ from $\auto{relax}$. For task~\hyperref[task:i]{\it (ii)}, given the pair $\vertex{init}$ and $\vertex{init}$,  there is no cycle leading back to $\vertex{init}$, so the total length is infinite and there is no corresponding sub-NBA $\auto{relax}$. The total length for the pair $\vertex{init}$ and $v_3$ is $2+2=4$. The NBA $\auto{subtask}$ is shown in  Fig.~\ref{fig:nba_ii_subtask}, where edges $(\vertex{init}, v_1)$ and $(v_1, v_2)$ are removed since $\vertex{init}$ does not have a self-loop and initial robot locations do not satisfy their labels.
     \end{cexmp}

 \begin{cexmp}{exmp:1}{Subtasks in $\auto{subtask}$}
   The sub-NBA  $\auto{subtask}$ is composed of subtasks that need to be satisfied in specific orders to reach the accepting vertex. For instance, the path $\vertex{init}, v_1, v_4, v_3, v_6$ in Fig.~\ref{fig:nba_i_relax} requires that first  $\langle1,2\rangle$ visits the control room $\ell_4$, then $\langle2,1\rangle$ visit the office building $\ell_2$ and finally the same two robots of type 1 drop off the mail at the delivery site  $\ell_3$. By definition of task \hyperref[task:i]{(i)}, the temporal order between these subtasks specifies that the time when $\langle1,2\rangle$ visits the control room $\ell_4$ is independent from the time when $\langle2,1\rangle$ pick up the mail at the building $\ell_2$, and that $\langle1,2\rangle$ visiting  $\ell_4$ and $\langle2,1\rangle$  visiting $\ell_2$ should occur prior to $\langle2,1\rangle$ visiting the delivery site $\ell_3$.
 \end{cexmp}

   \subsubsection{Pruning the sub-NBA \upshape $\auto{subtask}$} \label{sub-NBA:2}
   Observe that the sub-NBA $\auto{subtask}$ in Fig.~\ref{fig:nba_i_subtask} still constitutes a large portion of  $\auto{relax}$ in Fig.~\ref{fig:nba_i_relax}, {which is common in practice, since there are typically many more edges than  vertices in $\auto{relax}$}. However, some edges/subtasks are ``redundant'' in that they can be decomposed into more elementary edges/subtasks.  Therefore, in what follows, we further prune the NBA $\auto{subtask}$  by removing such redundant edges.

 Recall  Definition~\ref{defn:subtask} where subtasks are defined by their edge labels and starting vertex labels.  Next we define  the notion of equivalent subtasks.

 \begin{defn}[Equivalent subtasks]\label{defn:eq}
  Subtasks $(v_1, v_2)$ and $(v'_1, v'_2)$ in an NBA $\auto{}$ are equivalent, denoted by $(v_1, v_2) \sim (v'_1, v'_2)$, if $\gamma(v_1) = \gamma(v'_1)$, $\gamma(v_1, v_2)=\gamma(v_1', v'_2)$ and they  are not in the same path that connects the same  pair of initial and accepting vertices.
 \end{defn}

\begin{figure}
   \centering
   \subfigure[ID property]{
     \includegraphics[width=0.5\linewidth]{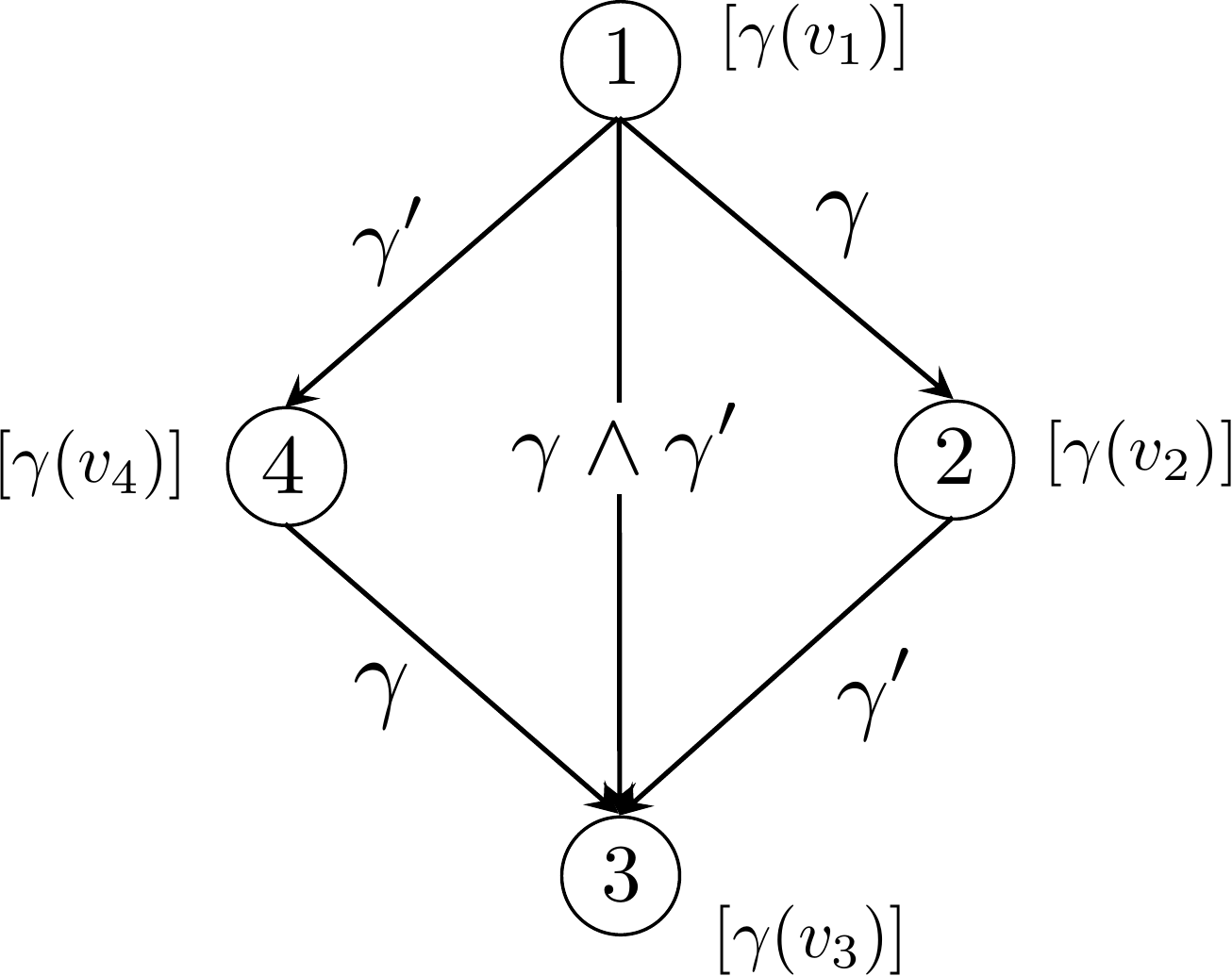}\label{fig:id}}
   \hspace{1em}
   \subfigure[ST property]{
     \includegraphics[width=0.3\linewidth]{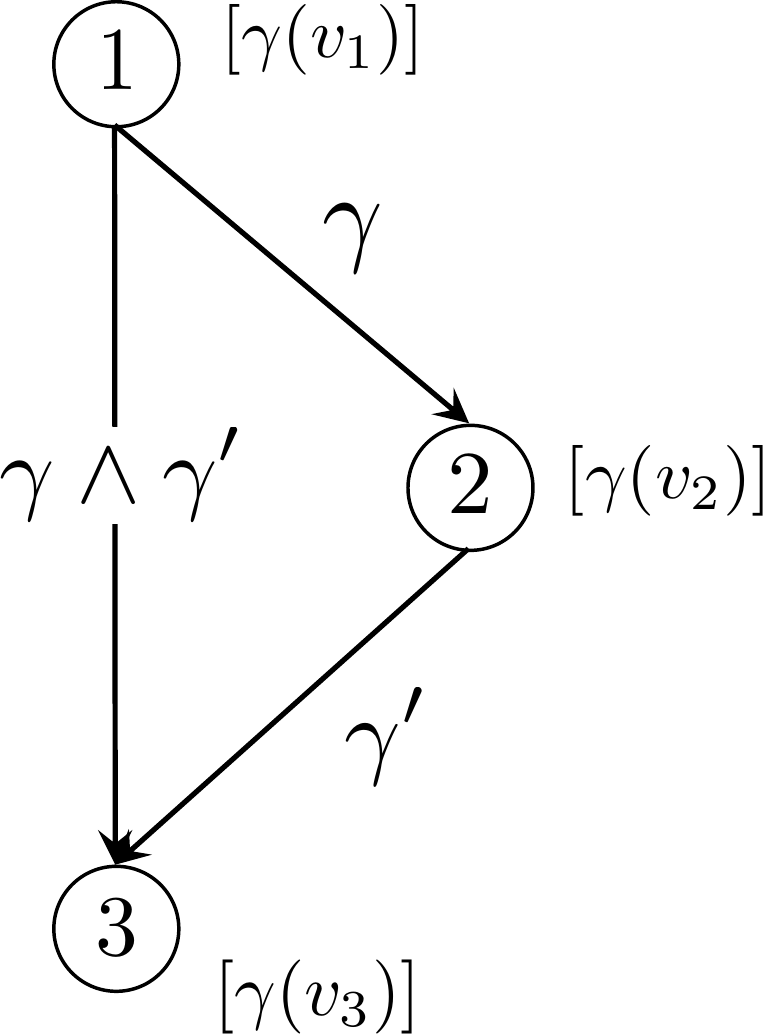}\label{fig:st}}
   \caption{Independent diamond and sequential triangle properties.}\label{fig:property}
 \end{figure}

 The last condition in Definition~\ref{defn:eq} is necessary since two subtasks in the same path mark different progress towards completing a task, even if they have identical labels. {Recall that in task~\hyperref[task:i]{(i)} in Example~\ref{exmp:1}, certain regions can be visited in parallel. To capture the parallel visits, we define the following two properties over vertices in $\auto{subtask}$, namely, the independent diamond (ID) property adapted from~\cite{stefanescu2006automatic}  and the sequential triangle (ST) property over vertices; see also Fig.~\ref{fig:property}.}
 \begin{defn}[Independent diamond property]\label{defn:id}
   Given four different vertices $v_1, v_2, v_3, v_4$ in the NBA $\auto{subtask}$, we say that these four vertices satisfy the ID property if
   \begin{noindlist}
     \setlength\itemsep{0em}
   \item \label{id:a} $\gamma(v_1) = \gamma(v_2) = \gamma(v_4)$;
   \item  \label{id:b} $v_1 \xrightarrow{\gamma}_B v_2 \xrightarrow{\gamma'}_B v_3$;
   \item \label{id:c} $v_1 \xrightarrow{\gamma'}_B v_4 \xrightarrow{\gamma}_B v_3$;
   \item \label{id:d} $v_1 \xrightarrow{\gamma \wedge \gamma'}_B v_3$;
   \item \label{id:e} $\gamma_{\phi}(v_3) = \top$ if $v_3 = \vertex{accept}$.
   \end{noindlist}
 \end{defn}
 Intuitively, if vertices $v_1, v_2, v_3$, and $v_4$ in $\auto{subtask}$ satisfy the ID property (see Fig.~\ref{fig:id}), then conditions  \ref{id:a}-\ref{id:c} in Definition~\ref{defn:id} imply that the subtasks $(v_1, v_2) \sim (v_4, v_3)$ and $(v_1, v_4) \sim (v_2, v_3)$ in $\auto{subtask}$ are equivalent, while conditions \ref{id:b}-\ref{id:d} in Definition~\ref{defn:id} state that their order is arbitrary, i.e., one can proceed the other or they can occur simultaneously.  We refer to $(v_1, v_3)$ as the {\it composite} subtask and $(v_1, v_2)$, $(v_1, v_4)$ as the {\it elementary} subtasks.  Although both can lead to vertex $v_3$,  composite subtasks are ``redundant'',  since elementary subtasks can be executed independently and, therefore, their labels are easier to satisfy, compared to composite tasks that need to be executed simultaneously and, therefore, more conditions need to hold so that their labels are satisfied.   Note that we conduct the $\top$-check in condition \ref{id:e} in Definition~\ref{defn:id} on the NBA $\autop$ so that condition~\ref{cond:f} in Definition~\ref{defn:run} is satisfied which means that the set of restricted accepting runs is not affected if the edge $(v_1, v_3)$ is removed. This result is formally shown  in Lemma~\ref{prop:sub-NBA} in Appendix~\ref{app:correctness}. In words, if $\gamma_{\phi}(v_3) \ne \top$ with $v_3 = \vertex{accept}$, and  if a restricted accepting run traverses edges $(v_1, \vertex{accept})$ and $(\vertex{accept}, \vertex{next})$ where $\vertex{accept}\neq \vertex{next}$, then condition~\ref{cond:f} in Definition~\ref{defn:run} states  that $\gamma_{\phi}(v_1, \vertex{accept}) \simplies \gamma_{\phi}(\vertex{accept}, \vertex{next})$. However $\gamma_{\phi}(v_2, \vertex{accept})$ and $\gamma_{\phi}(v_4, \vertex{accept})$ may not  imply $\gamma_{\phi}(\vertex{accept}, \vertex{next})$ since they are subformulas of $\gamma_{\phi}(v_1, \vertex{accept})$. Therefore, removing the composite edge $(v_1, \vertex{accept})$ risks emptying the  set of restricted accepting runs.

 \begin{defn}[Sequential triangle property]\label{defn:st}
   Given three different vertices $v_1, v_2, v_3$ in the NBA $\auto{subtask}$, we say that these three vertices $v_1, v_2, v_3$ satisfy  the ST property if 
   \begin{noindlist}
     \setlength\itemsep{0em}
   \item \label{st:a}  $v_1 \xrightarrow{\gamma}_B v_2 \xrightarrow{\gamma'}_B v_3$;
   \item \label{st:b} $v_1 \xrightarrow{\gamma \wedge \gamma'}_B v_3$;
     \item \label{st:c} $\gamma_{\phi}(v_3) = \top$ if $v_3 = \vertex{accept}$.
   \end{noindlist}
 \end{defn}
 If vertices $v_1, v_2$, and $v_3$ in $\auto{subtask}$ satisfy  the ST  property (see Fig.~\ref{fig:st}), then conditions \ref{st:a} and \ref{st:b} in Definition~\ref{defn:st} state that subtask $(v_1, v_2)$ should be satisfied no later than $(v_2, v_3)$. Note that if vertices $v_1, v_2, v_3, v_4$ satisfy  the ID property, then $v_1, v_2, v_3$ and $v_1, v_4, v_3$ satisfy  the ST property. Using these two properties, we remove all edges from $\auto{subtask}$ associated with composite subtasks  and denote by $\auto{subtask}^-$ the resulting pruned $\auto{subtask}$. A composite subtask can be an elementary subtask of another composite subtask at a higher layer. Thus, removing composite subtasks  is vital for reducing the size of $\auto{subtask}$. Similar to  pruning  $\autop$ to get $\autop^-$, the feasibility of Problem~\ref{prob:1} is not compromised by pruning composite subtasks from the NBA $\auto{subtask}$, as shown in Lemma~\ref{prop:sub-NBA}. 
 \begin{figure}[t]
     \centering
     \subfigure[$\auto{subtask}^-$ for task {(i)}]{
       \label{fig:sub-NBA}
       \includegraphics[width=0.29\linewidth]{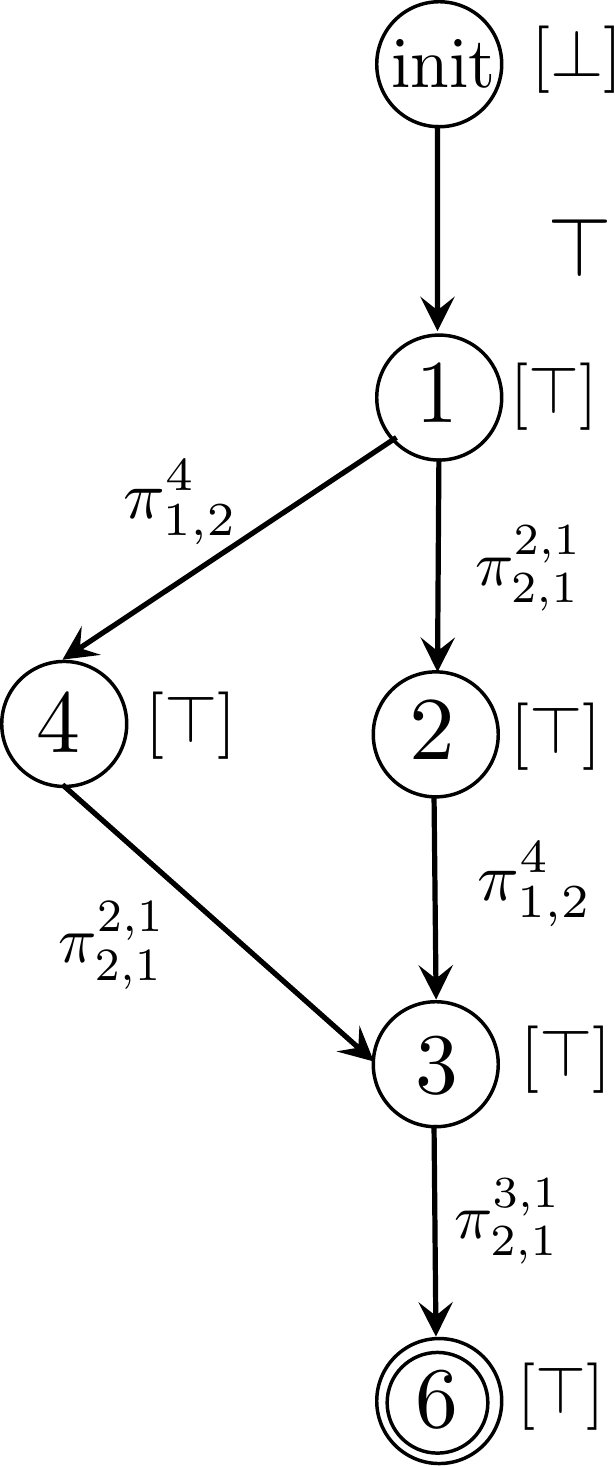}}
     \hspace{1em}
     \subfigure[Subtasks and poset]
     {\label{fig:subtask}
       \includegraphics[width=0.61\linewidth]{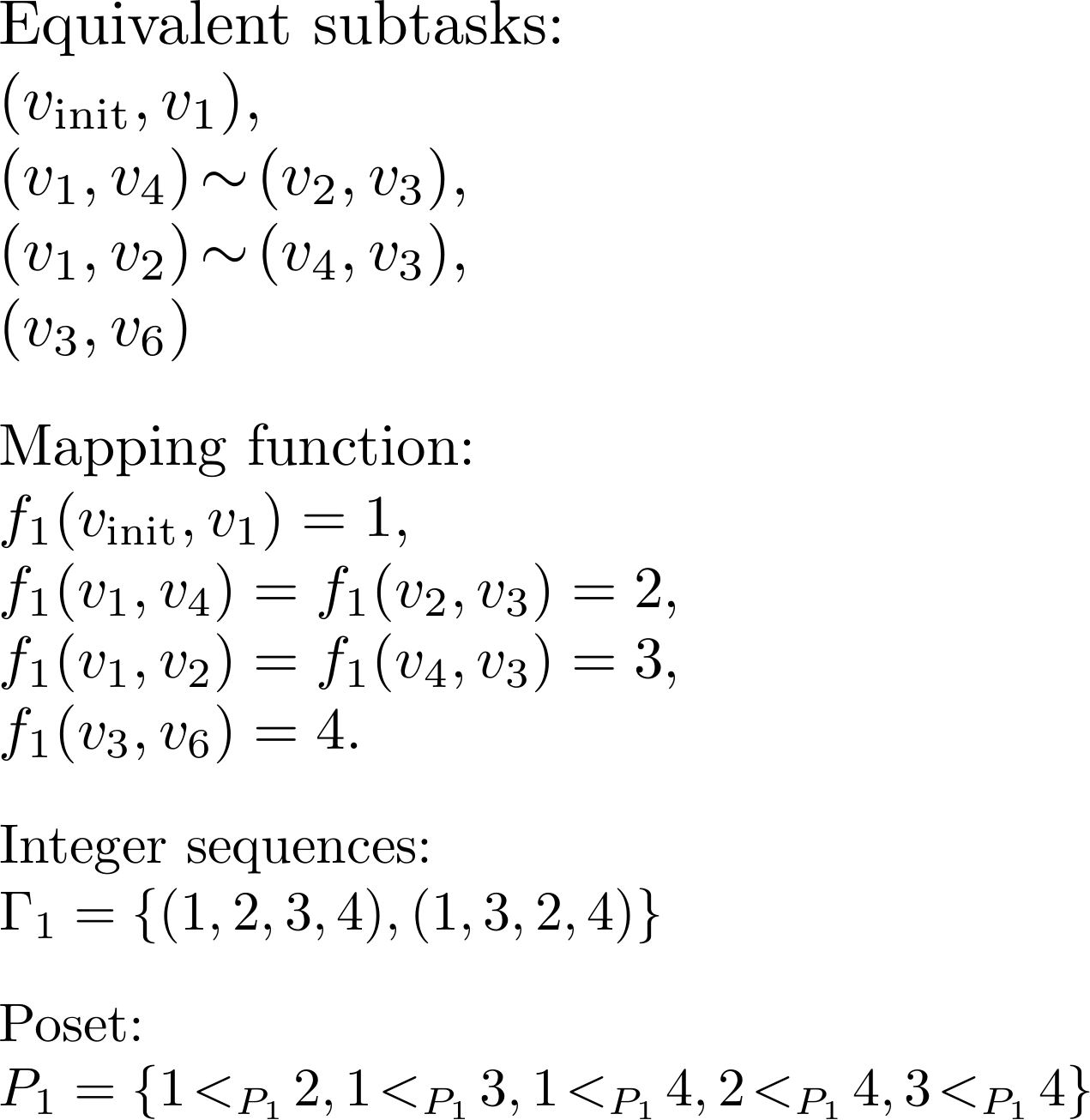}}
         \caption{The NBA $\auto{subtask}^-$ and corresponding subtasks.} \label{fig:subtask_prune}
 \end{figure}
 \begin{cexmp}{exmp:1}{ID and ST properties and the resulting NBA $\auto{subtask}^-$}
   In the NBA $\auto{subtask}$ for task~\hyperref[task:i]{(i)}, shown in Fig.~\ref{fig:nba_i_subtask}, the vertices $v_1, v_2, v_3, v_4$  satisfy the ID property and the  vertices $v_2, v_3, v_6$ ($\gamma_\phi(v_6)=\top$ in Fig.~\ref{fig:nba_i}) satisfy the ST property. Thus we delete $(v_1, v_3)$ and $(v_2, v_6)$. The resulting  $\auto{subtask}^-$ is shown in Fig.~\ref{fig:sub-NBA}. The NBA $\auto{subtask}^-$ of task~\hyperref[task:ii]{(ii)} is the same as $\auto{subtask}$ since there are no composite subtasks.
 \end{cexmp}

 \subsection{Inferring the temporal   order between subtasks in  $\auto{subtask}^-$}\label{sec:poset}
 In this section,  we infer the temporal relation between subtasks in the pruned NBA $\auto{subtask}^- = (\ccalV_\text{s}, \ccalE_\text{s})$.  For this,  we rely on  partially ordered sets introduced in Section~\ref{sec:partial}. Specifically,  let $\Theta$ denote the set that collects all simple paths connecting $v_0$ and $\vertex{accept}$ in $\auto{subtask}^-$. We focus on simple paths since condition~\ref{cond:a} in Definition~\ref{defn:run} excludes cycles. 
 Given a simple path $\theta \in\Theta$, let $\ccalT(\theta)$ denote the set of subtasks in $\theta$. We say that two simple paths $\theta_1$ and $\theta_2$ have the same set of subtasks if $\ccalT(\theta_1) =\ccalT(\theta_2)$. Then we partition $\Theta$ into subsets of  simple paths that contain the same set of subtasks, that is, $\Theta = \cup_e \Theta_e$ where $\ccalT(\theta_1) = \ccalT(\theta_2)$ for all $\theta_{1}, \theta_{2} \in \Theta_e $ and {$\theta_1 \not=\theta_2$}. The reason for this partition is that we want to map simple paths in $\auto{subtask}^-$ to posets, and the set of linear extensions  generated by a poset has the same set of elements.

 Given a subset $\Theta_e$ of simple paths in the partition, with a slight abuse of notation, let $\ccalT(\Theta_e)$ denote the set of corresponding  subtasks. Let the function $f_e:\ccalT(\Theta_e)\to [|\ccalT(\Theta_e)|]$ map each subtask to a distinct positive integer. {Note that two different subtasks in two different subsets $\Theta_e$ and $\Theta_{e'}$ may be mapped to the same integer; however, we treat these two subsets separately.} Using $f_e$, we can map every path in $\Theta_e$ to a sequence of integers, denoted by $S_e$. Let $\Gamma_e$ collect all sequences of integers for all paths in $\Theta_e$, so $|\Theta_e| =  |\Gamma_e|$. Moreover, all sequences of integers in $\Gamma_e$ are permutations of each other  and we denote this  base set by $X_e = [|\ccalT(\Theta_e)|]$. For every sequence $S_e \in \Gamma_e$, let $S_e[i]$ denote its $i$-th entry. We define a linear order $L_{X_e} = (X_e, <_L)$ such that  $S_e[i]  <_L  S_e[j] $ if $i  <  j$. In other words, the subtask $S_e[i]$ should be completed prior to  $S_e[j]$. Then, let $\Xi_e$ collect all linear orders over $X_e$ that can be defined from sequences in $\Gamma_e$. A poset $P_e  = (X_e, <_{P_e})$  containing the maximum number of linear orders in $\Xi_e$ can be found using the algorithm proposed in~\cite{heath2013poset} for the partial cover problem, where the order represents the precedence relation. Note that $\Xi_e$ may not be identical to $\ccalL(P_e)$, the set of all linear extensions of $P_e$. Thus, after obtaining poset $P_e$, each of the remaining linear orders in $\Xi_e$ that are not covered by $P_e$ are treated as separate totally ordered sets, that are posets  as well. In this way, we do not discard any posets.

 Finally, given a partition $\{\Theta_e\}$ and a corresponding set of posets $\{P_e\}$, we sort $\{P_e\}$ lexicographically first in descending order in terms of the width of posets and then in ascending order in terms of the height. Recall that the width of a poset is the cardinality of its maximal antichain, and its height is the cardinality of its maximal chain; see Section~\ref{sec:partial}. Intuitively, the wider a poset is, the more temporally independent subtasks it contains. The shorter a poset is, the fewer  subtasks  it has. We consider first wider posets since they impose less restrictions on the high-level plans compared to shorter posets. Every linear extension of subtasks in a poset produces a simple path  connecting $v_0$ and $\vertex{accept}$ in $\auto{subtask}^-$. 
 \begin{cexmp}{exmp:1}{Temporal constraints}
   For task~\hyperref[task:i]{(i)}, there are two simple paths in $\auto{subtask}^-$ leading to $v_6$ and all have the same set of four edges, thus,  $\Theta_1 = \{\vertex{init},v_1, v_4,v_3,v_6$; $\vertex{init}, v_1, v_2,v_3, v_6$\}; see~Fig~\ref{fig:sub-NBA}.  The design of equivalent subtasks, mapping function, integer sequence and the poset are shown in Fig.~\ref{fig:subtask}. The temporal relation implies that subtasks $(v_1, v_4)$ and $(v_1, v_2)$ are independent, which agrees with our observation. For task~\hyperref[task:ii]{(ii)}, the NBA $\auto{subtask}^-$ in Fig.~\ref{fig:nba_ii_subtask} only has one path of two subtasks that generates a totally ordered set where  every two subtasks are comparable.
 \end{cexmp}

 \begin{rem}
   If the size of sub-NBA $\auto{subtask}^-$ is still large, leading to large number of simple paths, we can select a fixed number of simple paths, similar to finding a fixed number of runs in~\textup{\cite{kloetzer2020path}}. This will not severely compromise the diversity of the selected simple paths since a lot of simple paths are combinations of the same set of elementary subtasks.
 \end{rem}

 \section{Design of High-Level Task Allocation Plans and Low-Level Executable Paths}\label{sec:solution}
 {In this section, we synthesize plans that satisfy  the LTL specification $\phi$ by first generating a time-stamped task allocation plan that respects the temporal order between subtasks that need to be satisfied in order to satisfy the specification,  and then obtaining a low-level executable path that also satisfies the negative literals that we removed from $\auto{relax}$ in Section~\ref{sec:prune}. In what follows, we discuss the synthesis of a  prefix path; a similar process is used to synthesize  the suffix path in Appendix~\ref{sec:suf}. Specifically, to synthesize high-level prefix plans, we  iterate over the sorted set of posets $\{P_\text{pre}\}$, where $P_\text{pre}$ is a poset corresponding to a simple prefix path in $\auto{subtask}^-$ and, for every poset in  $\{P_{\text{pre}}\}$ we formulate a MILP to assign robots to tasks and determine a high-level plan, i.e., a sequence of time-stamped waypoints, that the robots need to visit to satisfy the subtasks in the corresponding simple path in $\auto{subtask}^-$. Note that, given  a poset $P\in \{P_\text{pre}\}$, every element in the corresponding base set $X_{P}$  is an integer  associated with  an edge/subtask in the  NBA $\auto{subtask}^-$. Since a solution to the proposed  MILP is effectively a linear extension of the poset $P$, the corresponding plan sequentially satisfies the vertex and edge labels of all subtasks in $\auto{subtask}^-$ associated with the elements in $X_P$.  Therefore,  this plan produces  a simple path in $\auto{subtask}^-$ that connects $v_0$ and $\vertex{accept}$. To  obtain  the  low-level executable path, for every subtask in this simple path, we formulate a generalized multi-path robot planning problem that considers the negative literals that were removed from $\auto{relax}$ in Section~\ref{sec:prune}.}
 
 The proposed MILP  is inspired by the vehicle routing problem (VRP) with temporal constraints~\citep{bredstrom2008combined}. In the VRP, a fleet of vehicles traverses a given set of customers such that all vehicles depart from and return to the same depot, and each customer is visited by exactly one vehicle. 
{Compared to the VRP with temporal constraints~\citep{bredstrom2008combined}, the LTL-MRTA problem is significantly more complicated.}  First, robots are not required to return to their initial locations. Instead, there may exist robots that need to execute the task forever corresponding to the  ``always'' LTL operator. Second, there may exist labeled regions that do not need to be visited at all and others that need to be visited exactly once,  more than once, or infinitely many times. {Finally, visits of regions and visiting times are subject to logical constraints induced by the NBA $\auto{subtask}^-$.} 

\subsection{Construction of the prefix routing graph}\label{sec:graph}
 We first construct  the vertex set and then the edge set of the routing graph  $\ccalG$. Both constructions consist of  four layers that iterate over the edges, then the labels, then the clauses, and finally, the literals in $\auto{subtask}^-$. The outline of the algorithm is shown in Alg.~\ref{alg:milpgraph}.  
 An illustrative graph for task~\hyperref[task:i]{(i)} is  shown in Fig.~\ref{fig:milp}.
 \subsubsection{Construction of the vertex set}\label{sec:vertex} The vertex set $\ccalV_\ccalG$ consists of three types of vertices, namely, location vertices  related to initial robot locations, literal vertices related to edge labels in the sub-NBA $\auto{subtask}^-$, and literal vertices related to vertex labels in the sub-NBA $\auto{subtask}^-$. Specifically, we construct the location vertices as follows.

\begin{figure}[t]
   \centering
   \includegraphics[width=0.7\linewidth]{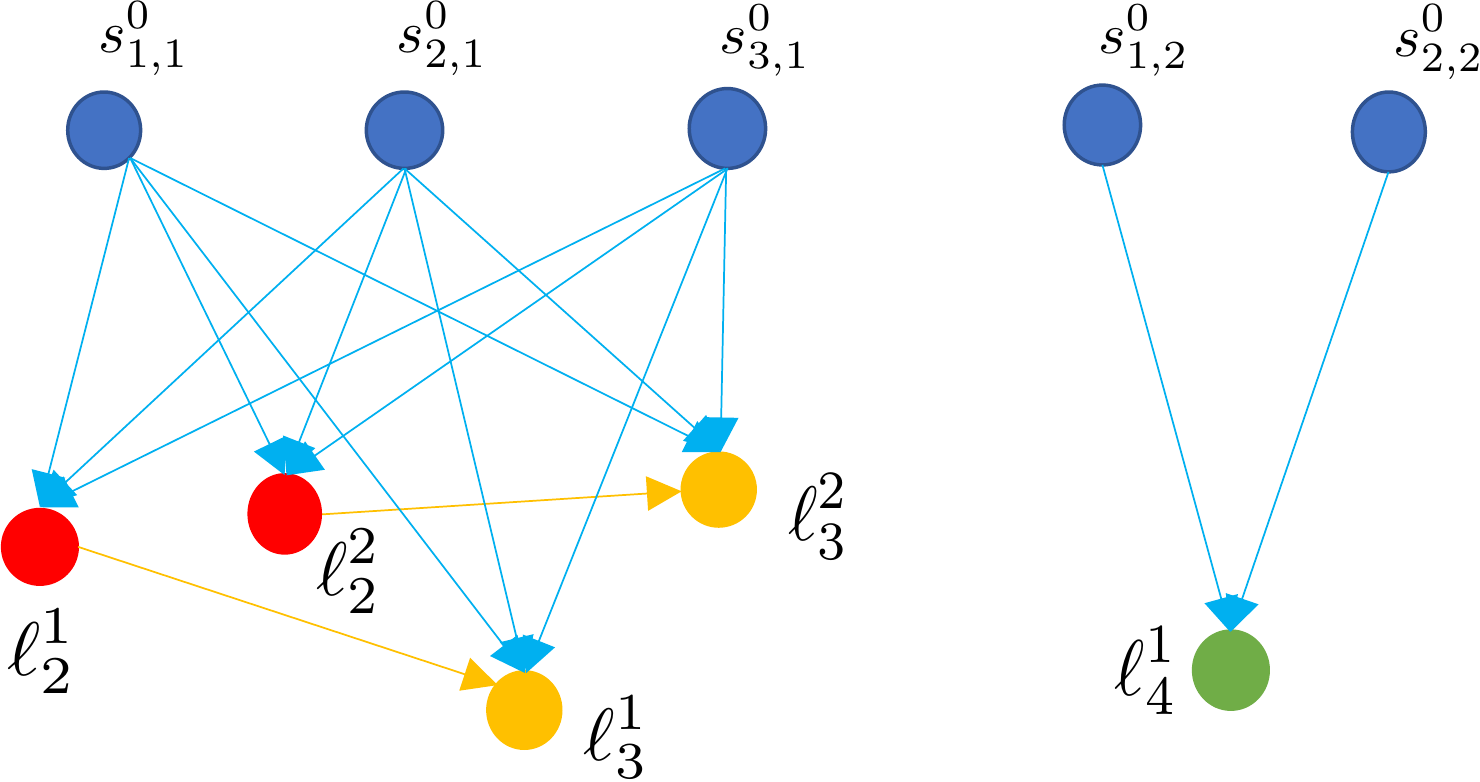}
   \caption{Routing graph $\ccalG$ for task~\hyperref[task:i]{(i)}. $s^0_{1,1}$, $s^0_{2,1}$ and $s^0_{3,1}$ are initial locations of three robots of type 1 and $s^0_{1,2}$ and $s^0_{2,2}$ are initial locations of two robots of type 2 (see case~\ref{vertex:initial}). Red dots $\ell_{2}^{1}$ and $\ell_{2}^{2}$ correspond to the edge label $\ap{2}{1}{2,1}$ of element 3, i.e., edge $(v_1, v_2)$ in $X_P$; see Fig.~\ref{fig:subtask}. Yellow  dots $\ell_{3}^{1}, \ell_{3}^{2}$ correspond to the edge label $\ap{2}{1}{3,1}$ of element 4, and green dot $\ell_{4}^{1}$ corresponds to the edge label $\ap{1}{2}{4}$ of element 2 (see Appendix~\ref{vertex:edge}). No dots correspond to vertex labels since all vertex labels are either $\top$ or $\bot$.  The edges from $\ell_2^{1}$ to $\ell_3^{1}$ and from $\ell_2^{2}$  to $\ell_3^{2}$ are due to $3 <_{P} 4$. }
     \label{fig:milp}
 \end{figure}

 \paragraph{Location vertices associated with initial robot locations}{First we create $n$ vertices, collected in the set $\ccalV_{\text{init}}\subseteq \ccalV_{\ccalG}$ such that each vertex points to the initial location $s^0_{r,j}$ of robot $[r,j]\in\ccalK_j, \forall j\in [m]$ [line~\ref{milp:init}, Alg.~\ref{alg:milpgraph}] (see blue dots in Fig.~\ref{fig:milp}).\label{vertex:initial}}

 To obtain the set of literal vertices in $\ccalV_\ccalG$, we iterate over subtasks in $X_{P}$. Given a subtask $e = (v_1, v_2) \in X_{P}$, we construct vertices for the  edge label $\gamma(v_1, v_2)$ and the starting vertex label $\gamma(v_1)$, if they are neither $\top$ nor $\bot$. Specifically, we take the following steps.

 \paragraph{Literal vertices associated with edge labels}{If $\gamma(v_1, v_2) \not= \top$, we operate on $\gamma(v_1, v_2) =  \bigvee_{p\in \ccalP} \bigwedge_{q \in \ccalQ_{p}} \ap{i}{j}{k,\chi}$ starting by iterating over the clauses $\ccalC_p^{\gamma} \in \clause{\gamma}$ in the label, and then over the literals in each clause $\ccalC_p^\gamma$ [lines~\ref{milp:for}-\ref{milp:i}, Alg.~\ref{alg:milpgraph}]. The literal $\ap{i}{j}{k,\chi}\in $ $\mathsf{lits}^+(\ccalC_p^{\gamma})$ implies that at least $\ag{i}{j}$, i.e., $i$ robots of type $j$,  should visit the target region $\ell_k$ simultaneously. Hence, we create $i$ vertices in $\ccalV_\ccalG$  all associated with region $\ell_k$. If $\ag{i}{j}$ visit these $i$ vertices simultaneously, one robot per  vertex, then $\ap{i}{j}{k,\chi}$ is true. Note that if $\chi\not=0$, the robots visiting these $i$ vertices should be the same as those visiting another $i$ vertices  associated with another literal with the same nonzero connector, which is ensured by the MILP formulation; see the red, yellow, and green dots in Fig.~\ref{fig:milp}.}\label{vertex:edge}

 \begin{algorithm}[!t]
       \caption{Construct the routing graph}
       \LinesNumbered
       \label{alg:milpgraph}
       \KwIn {Poset $P$}
       \Comment*[r]{Create the vertex set}
       Create the vertex set $\ccalV_{\text{init}}$ for initial locations \label{milp:init}\;
       \Comment*[r]{vertices for  labels}
       \For{$e = (v_1, v_2) \in X_P$ \label{milp:for}}{
         \If{$\gamma(v_1, v_2) \not= \top$ \label{milp:nottrue}}{
           \For{$\ccalC_p^\gamma \in \clause{\gamma} $\label{milp:clause}}{
             \For{$\ap{i}{j}{k,\chi} \in \mathsf{lits}^+(\ccalC_p^{\gamma})$\label{milp:ap}}{
               Create $i$ vertices \label{milp:i}\;
             }
           }
         }
           \If{$\gamma(v_1)\not= \top, \bot$ \label{milp:nottrue2}}{
             Create vertices by following lines~\ref{milp:clause}-\ref{milp:i} \label{milp:repeat}\;
           }

       }
       \Comment*[r]{Create the edge set}
       \For{$e = (v_1, v_2) \in X_P$ \label{milp:for2}}{
         \If{$\gamma(v_1, v_2) \not= \top$ \label{milp:e}}{
             \For{$\ccalC_p^\gamma \in \clause{\gamma} $\label{milp:clause2}}{
               \For{$\ap{i}{j}{k,\chi} \in \mathsf{lits}^+(\ccalC_p^{\gamma})$\label{milp:lits}}{
                 $(i)$ Vertices of  initial robot locations \label{milp:i2}\;
                 $(ii)$ Vertices of prior subtasks \label{milp:ii}\;
                 $(iii)$ Vertices associated with  $\gamma(v_1)$ \label{milp:iii}\;
               }
             }
         }
         \If{$\gamma(v_1) \not= \top,\bot$ \label{milp:nottrue3}}{

           \If{$S_2^e= \emptyset$}{
             Create edges by following lines~\ref{milp:clause2}-\ref{milp:i2} \label{milp:repeati}\;
           }
           \ElseIf{\upshape $S_2^e\ne \emptyset$}{
             Create edges from vertices associated with subtasks in $S_2^e$  \label{milp:repeatii}\;
             \If{$X_{\prec_P}^e = \emptyset$ {\bf and} $X_{\|_P}^e \neq \emptyset$ }{
               Create edges from vertices associated with initial robot locations \label{milp:repeatiii}\;}
          }
         }
       }
 \end{algorithm}

 \paragraph{Literal vertices associated with starting vertex labels}{After vertices in $\ccalV_\ccalG$ associated with  the edge label $\gamma(v_1, v_2)$ of subtask $e$ have been constructed, vertices in $\ccalV_\ccalG$ associated with the starting  vertex label $\gamma(v_1)$ can be constructed in the same manner if $\gamma(v_1)$ is neither $\top$ nor $\bot$ [lines~\ref{milp:nottrue2}-\ref{milp:repeat}, Alg.~\ref{alg:milpgraph}]. \label{vertex:vertex}}
 Repeating steps in Appendices~\ref{vertex:edge} and \ref{vertex:vertex} for all subtasks in $X_{P}$ completes the construction of the vertex set $\ccalV_\ccalG$. Note that each vertex in $\ccalV_\ccalG\setminus \ccalV_{\text{init}}$ is associated with a literal of a certain subtask in $X_P$. Also, each literal of a certain subtask in $X_P$ is associated with one or more vertices in $\ccalV_\ccalG\setminus \ccalV_{\text{init}}$, and the  literal specifies the region and the robot type associated with these vertices.
 To capture this correspondence, let $\ccalM^\ccalV_{e} : \ccalV_{\ccalG}\setminus \ccalV_{\text{init}} \to  X_{P} $ and $\ccalM^\ccalV_{\mathsf{lits}}  : \ccalV_\ccalG\setminus \ccalV_\text{init}  \to  \prod_{\mathsf{lits}} $  map a vertex in $\ccalV_\ccalG \setminus \ccalV_{\text{init}}$ to its associated subtask and literal, respectively, where $\prod_{\mathsf{lits}}$ is the cartesian product $ X_{P} \times\{0, 1\} \times \ccalP  \times \ccalQ_p$, and 0, 1 represent the label type, 0 for vertex label and 1 for edge label. Furthermore, let $\ccalM^\mathsf{lits}_{\ccalV}:   \prod_{\mathsf{lits}}  \to  2^{\ccalV_\ccalG}$ and $\ccalM^\mathsf{cls}_{\ccalV} :  \prod_{\mathsf{cls}} \to  2^{\ccalV_\ccalG}$ map a literal and clause to the associated vertices in $\ccalG$, respectively, where $\prod_{\mathsf{cls}}$ is the cartesian product $X_{P}\times \{0, 1\} \times \ccalP$. We also define $\ccalM^{\ccalV}_\ccalL :  \ccalV_\ccalG  \to  \ccalL$ and $ \ccalM^{\ccalV}_\ccalK :  {\ccalV_\ccalG}  \to  \{\ccalK_j\}$ that map a vertex in $\ccalV_\ccalG$ to its associated  region and robot type. Finally, if $\chi\not=0$, we define  $\ccalM_{\gamma}^\chi: \mathbb{N}^+ \to 2^{X_{P}\times \{0,1\}}$ to map $\chi$ to all labels in $X_P$ that have literals with the same connector $\chi$, {which will be used in the MILP problem in Appendix~\ref{sec:samegroup}  to encode the constraint that some regions are visited by the same $i$ robots of type $j$}.

 \begin{cexmp}{exmp:1}{Mappings for task {(i)}}
     The mappings in Fig.~\ref{fig:milp} associated with the vertex $\ell_2^{1}$ are:
     $\mathcal{M}_e^{\mathcal{V}}(\ell_{2}^{1})  = (v_1,v_2) = 3$ and $\mathcal{M}_{\mathsf{lits}}^{\mathcal{V}}(\ell_{2}^{1})  = ((v_1,v_2), 1, 1, 1)$ since the vertex $\ell_2^{1}$ corresponds to the first literal $\ap{2}{1}{2,1}$ of the first clause of the edge label of subtask $(v_1, v_2)$ in $X_P$; see also Fig.~\ref{fig:subtask_prune}.  $\mathcal{M}_\mathcal{L}^{\mathcal{V}}(\ell_{2}^{1})  = \ell_2$  and $\mathcal{M}_\mathcal{K}^{\mathcal{V}}(\ell_{2}^{1})  = \mathcal{K}_1$ since the literal $\ap{2}{1}{2,1}$ requires two robots of type 1 to visit region $\ell_2$.

     Furthermore,  the literal/clause-to-vertex mappings are:
     $\mathcal{M}^{\mathsf{lits}}_{\mathcal{V}}(((v_1,v_2), 1, 1, 1)) = \mathcal{M}^{\mathsf{cls}}_{\mathcal{V}}(((v_1,v_2), 1, 1))  = \{\ell_2^{1}, \ell_2^{2}\}$; $\mathcal{M}^{\mathsf{lits}}_{\mathcal{V}}(((v_1,v_4), 1, 1, 1)) = \mathcal{M}^{\mathsf{cls}}_{\mathcal{V}}(((v_1,v_4), 1, 1))  = \{\ell_4^{1}\}$ since the literal $\ap{1}{2}{4}$, the first literal of the first clause of the edge label of subtask $(v_1, v_4)$, requires one robot to visit region $\ell_4$. Finally, the connector-to-label mapping is: $\ccalM^{\chi}_{\gamma} (1)  = \{((v_1, v_2), 1), ((v_3, v_6), 1)\}$ since the connector 1 appears in the edge label of subtask $(v_1 ,v_2)$ and the edge label of subtask $(v_3, v_6)$.
    \end{cexmp}

 \subsubsection{Construction of  the edge set}\label{sec:edge_set} The edges in $\ccalG$ respect the partial order among subtasks captured by the poset ${P}$. We  construct the edge set $\ccalE_\ccalG$ by following a similar procedure as that used to construct the vertex set $\ccalV_\ccalG$. Specifically, we iterate over the elements in $X_P$.  For every subtask $e = (v_1, v_2) \in X_{P}$, if $\gamma(v_1, v_2) \not=\top$, we first operate on the edge label $\gamma(v_1, v_2) =  \bigvee_{p\in \ccalP} \bigwedge_{q \in \ccalQ_{p}} \ap{i}{j}{k,\chi}$ starting by iterating over the clauses $\ccalC_p^{\gamma} \in \clause{\gamma}$, and then over the literals in each clause $\ccalC_p^\gamma$ [lines~\ref{milp:for2}-\ref{milp:iii}, Alg.~\ref{alg:milpgraph}]. Specifically, recall from Appendix~\ref{sec:vertex} that the literal $\ap{i}{j}{k,\chi} \in \mathsf{lits}^+(\ccalC_p^{\gamma})$ corresponds to  $i$ vertices in $\ccalV_\ccalG$ that are associated with region $\ell_k$ that should be visited by $i$ robots. In what follows, we identify three types of {\it leaving vertices} in $\ccalV_\ccalG$ from where $i$ robots  can depart to reach  these $i$ vertices that satisfy literal $\ap{i}{j}{k,\chi}$.

 \paragraph{Location vertices}{The location vertices in $\ccalV_{\text{init}}$  associated with  robots of type $j$ are leaving vertices. We add an edge from all initial vertices to every vertex associated with literal $\ap{i}{j}{k,\chi}$ (see blue edges  in~Fig.~\ref{fig:milp}). Intuitively, robots depart from initial locations to undertake certain subtasks. These edges are associated  with a weight $T^*$ that is equal to the shortest travel time from the initial location to region $\ell_k$ and another weight  $d$ that is equal to the smallest traveling cost between the initial location and $\ell_k$, {which will be used in the MILP problem in Appendices~\ref{app:scheduling_constraints} and~\ref{sec:objective} to encode the scheduling constraints and the objective.} }\label{sec:a}

 \paragraph{Leaving vertices associated with prior subtasks}{Let $X^e_{<_{P}}$, $X^e_{\prec_{P}}$ and $X^e_{\|_{P}}$ denote the sets that collect subtasks in $X_{P}$ that are  smaller than, covered by, and incomparable to subtask $e$, respectively (see Section~\ref{sec:partial}). In words, $X^e_{<_{P}}$ contains subtasks in $X_P$ that should be completed prior to $e$, $X^e_{\prec_{P}} \subseteq X^e_{<_{P}}$ contains subtasks in $X^e_{<_{P}}$ that can be completed right before $e$, and $X^e_{\|_{P}}$ contains subtasks independent from $e$. To find leaving vertices, we iterate over $S_1^e = X^e_{<_{P}} \cup X^e_{\|_{P}}$ that includes all subtasks that can be completed prior to $e$, respecting the partial order between subtasks.}\label{sec:b}
 Given a subtask $e' = (v'_1, v'_2) \in S_1^e$, if its edge label $\gamma'(v'_1, v'_2) \not=\top$, we iterate over all clauses in $\gamma'$ and then over all literals in each clause. Specially, given a clause $\ccalC^{\gamma'}_{p'} \in \clause{\gamma'}$, for any literal $\ap{i'}{j'}{k',\chi'} \in\mathsf{lits}^+(\ccalC^{\gamma'}_{p'})$, if $j'=j$, then  literal vertices in $\ccalV_\ccalG$ associated with this literal are leaving vertices. If further $i'=i$, we randomly create $i$ one-to-one edges starting from these $i$ vertices and ending at the $i$ vertices associated with $\ap{i}{j}{k,\chi}$ (see the orange edges in Fig.~\ref{fig:milp}). Because there are exactly $i$ robots of type $j$, it suffices to build $i$ one-to-one edges. Furthermore, if $\chi=\chi'\neq0$, then literals $\ap{i}{j}{k,\chi}$ and $\ap{i'}{j'}{k',\chi'}$ must have the same number of vertices. Building $i$ one-to-one edges can guarantee that  the same  $i$ robots of type $j$ satisfy these two literals. Otherwise, if $i'\neq i$, we add $i\times i'$ edges to $\ccalE_\ccalG$ by creating an edge from any vertex associated with $\ap{i'}{j'}{k',\chi'}$ to any vertex of $\ap{i}{j}{k,\chi}$. Finally, since each region may span multiple cells, the weights $T^*$ and $d$ of these edges are set as the shortest travel time and lowest traveling cost from  $\ell_{k'}$ to $\ell_k$. After creating edges  associated with the edge label  $\gamma'(v'_1,v'_2)$ of $e'$, we identify  leaving vertices  among literal vertices in $\ccalV_\ccalG$ associated with the starting vertex label $\gamma(v'_1)$ of $e'$ and build edges in the same manner. 

 \paragraph{Leaving vertices associated with $\gamma(v_1)$ of $e$}{When the iteration over $S_1^e$ is completed, we identify leaving vertices  among literal vertices associated with the starting vertex label  $\gamma(v_1)$ of the current subtask $e$ by following the  procedure in Appendix~\ref{sec:b} for the prior subtasks. This is because $\gamma(v_1)$ becomes true before $\gamma(v_1, v_2)$.}\label{sec:c}

 So far we have constructed three types of leaving vertices corresponding to the literal $\ap{i}{j}{k,\chi}$ in $\mathsf{lits}^+(\ccalC_p^{\gamma})$ of the edge label $\gamma(v_1, v_2)$ [lines~\ref{milp:i2}-\ref{milp:iii}, Alg.~\ref{alg:milpgraph}]. We continue  constructing leaving vertices for all other literals in~$\mathsf{lits}^+(\ccalC_p^{\gamma})$ [line~\ref{milp:lits},  Alg.~\ref{alg:milpgraph}] and clauses in $\clause{\gamma}$ [line~\ref{milp:clause2},  Alg.~\ref{alg:milpgraph}].
 After constructing all edges pointing to vertices associated with literals in the edge label $\gamma(v_1, v_2)$ of the current subtask $e$ [line~\ref{milp:e},  Alg.~\ref{alg:milpgraph}],  we construct  edges pointing to vertices associated with literals in the starting vertex label $\gamma(v_1)$, by identifying leaving vertices among location vertices  and literal vertices associated with prior subtasks. Specifically, let $S_2^e = X_{\prec_P}^e \cup X_{\|_P}^e$ be the set that collects all subtasks that can occur immediately prior to subtask $e$. The satisfaction of edge labels  of subtasks in $S_2^e$  can directly lead to the starting vertex $v_1$ of $e$. We consider the following cases.

 \mysubparagraph{$S_2^e = \emptyset$}{In this case, no subtask can be completed before subtask $e$, i.e., the subtask $e$ should be the first one among all in $X_P$ to be completed. Thus, $v_1$ is identical to the initial vertex $v_0$. In this case,  we only identify location vertices as leaving vertices, as in Appendix~\ref{sec:a} [lines~\ref{milp:repeati},  Alg.~\ref{alg:milpgraph}].}\label{edge:vertex1}

 \mysubparagraph{$S_2^e \neq \emptyset$}{We  identify leaving vertices associated with prior subtasks in $S_2^e$. Given a subtask $e' = (v_1', v_2') \in S^e_2$, we find all clauses $\ccalC_{p'}^{\gamma'} \in  \clause{\gamma'}$ in the edge label $\gamma'$ of $e'$ such that, for the considered clause $\ccalC_p^\gamma \in \clause{\gamma} $ in the starting vertex label of subtask $e$, its corresponding clause $(\ccalC_p^\gamma)_{\phi}$ in $\autop$ is the subformula of   their corresponding clauses $(\ccalC_{p'}^{\gamma'})_{\phi}$ in $\autop$. Next, for each literal $\ap{i}{j}{k,\chi} \in \mathsf{lits}^+(\ccalC_p^\gamma)$ we create $i$ one-to-one edges, starting from those $i$ vertices associated with the counterpart  of literal  $\ap{i}{j}{k,\chi}$ in the found clause  $\ccalC_{p'}^{\gamma'} \in \clause{\gamma'}$  and ending at the $i$ vertices associated with $\ap{i}{j}{k,\chi}$ [lines~\ref{milp:repeatii},  Alg.~\ref{alg:milpgraph}]. We create such one-to-one edges based on condition~\ref{cond:d}  in Definition~\ref{defn:run} and condition~\hyperref[asmp:b]{(b)} in Definition~\ref{defn:same}. That is,  the edge label strongly implies its end
   vertex label, the satisfied clause in the edge label  implies the satisfied clause in the end vertex label, and the fleet of robots satisfying the clause in the vertex label belongs to the fleet of robots satisfying the clause in the incoming edge label. This is also the reason why we consider prior subtasks in $S_2^e$ rather than $S_1^e$ as in Appendix~\ref{sec:b}.}\label{edge:vertex2}

 \mysubparagraph{$X_{\prec_P}^e = \emptyset$ and $X_{\|_P}^e \neq \emptyset$}{In this case, the subtask $e$ can be the first one among all to be completed. If so,  its starting vertex label $\gamma(v_1)$ should be satisfied at the beginning. However, robots cannot depart from leaving vertices that are literal vertices  (see case~\ref{edge:vertex2} in Appendix~\ref{sec:c}), because these edges are enabled after subtask $e$. Therefore, for the vertex label $\gamma(v_1)$, we additionally identify leaving vertices pointing to initial robot locations, as in Appendix~\ref{sec:a} [lines~\ref{milp:repeatiii},  Alg.~\ref{alg:milpgraph}]. 
   Note that, if $X_{\prec_P }^e \neq \emptyset$,  there are no leaving vertices  associated with initial locations since there  exists a subtask that should be completed before $e$ and, therefore, subtask $e$ can not be the first one.}\label{edge:vertex3}
 When the iteration over all subtasks in $X_{P}$ is over, we finish the construction of  the edge set $\ccalE_\ccalG$ [line~\ref{milp:for2},  Alg.~\ref{alg:milpgraph}].

 \begin{rem}[Relaxation of strong implication in condition~\texorpdfstring{\ref{cond:d}}\, in  Definition~\ref{defn:run}]
  Condition~\ref{cond:d} in Definition~\ref{defn:run} requires that an edge label \textup{strongly implies} its end vertex label. This condition ensures  both that the satisfaction of an edge label leads to the satisfaction of its end vertex label and that when constructing the routing graph $\ccalG$, robots satisfying the  positive subformula in an end vertex label belong to robots satisfying the corresponding edge label (see step~\ref{edge:vertex2} in Appendix~\ref{sec:c}). This condition can be relaxed to requiring that an edge label \textup{implies} its end vertex label (see Definition~\ref{defn:implication}), which can still ensure that the satisfaction of an edge label implies   the satisfaction of its end vertex label, so that the previous instance of GMRPP still activates the immediately following instance of GMRPP. The only change  needed in this case is  in the pre-processing steps in Section~\ref{sec:nba} where we need to remove  all clauses  in an end vertex label that are  not a subformula of  clauses in the corresponding edge label. This way, the edge label strongly implies the remaining clauses in its end vertex label.
 \end{rem}

 \subsection{Construction of the robot prefix plans }\label{sec:path}
Given the routing graph constructed in Section~\ref{sec:graph}, the proposed  MILP contains  five types of constraints including  routing constraints, scheduling constraints, logical constraints, temporal constraints, and transition constraints; see Appendix~\ref{app:milpformulation}.   The feasibility of the  MILP and the properties of the resulting solutions are analyzed in Lemmas~\ref{prop:feasibility} and~\ref{prop:run}. Given the solution to the  MILP, we first define a time axis that includes the sorted completion times  of all subtasks in $X_P$. This time axis produces a linear extension of the poset $P$ and the plan generated by this linear extension satisfies the vertex and edge labels in a given  simple path in $\auto{subtask}^-$. Next, we extract a time-stamped task allocation plan, augmented with completion time of each subtask, for each robot that can  be used to generate low-level paths satisfying the specification $\phi$.

 \subsubsection{Time axis}\label{sec:timeaxis} The progress made in $\auto{subtask}^-$ is directly linked to the satisfaction of edge labels which, by condition~\ref{cond:d} in Definition~\ref{defn:run}, implies the satisfaction of their end vertex labels, excluding $\vertex{accept}$. Therefore, we collect the completion times of all subtasks in $X_P$ (the time when edges are enabled)
 and  sort them in an ascending order to form a single increasing time axis, denoted by $\vec{t}$. {We note that there are no identical time instants in the time axis since, by construction, the solution to the MILP is a simple path in $\auto{subtask}^-$ and subtasks in any simple path are completed at different times.}

 \subsubsection{High-level robot plans} Next we extract a high-level plan for each robot, which is a sequence of waypoints that the robots need to visit to complete the  subtasks in $X_P$ along with the time instants of these visits. Specifically, for each robot $[r,j]$, let $p_{r,j}$ denote its corresponding  high-level plan and let $t_{r,j}$ denote its  timeline. Consider also a vertex $v^*_0 \in \ccalV_{\text{init}}$ in the routing graph $\ccalG$ that is associated with the initial location of  robot $[r,j]$  and let $v^*_1$ be the  vertex that robot $r$ traverses to. Note that robot $r$ can only travel along one outgoing edge of $v_0^*$. Note also that each vertex in the routing graph $\ccalG$ is associated with a label captured in the mapping $\ccalM_{\mathsf{lits}}^{\ccalV}$. If the label associated with  $v_1^*$ is a vertex label, then we proceed to the next vertex $v_2^*$ that robot $r$ reaches from $v_1^*$, until a vertex $v^* \in \ccalV_\ccalG$ associated with an edge label is found. Then, the  region associated with this vertex $v^*$, captured by the mapping  $\ccalM^\ccalV_{\ccalL}(v^*)$, constitutes   the first waypoint robot $r$ needs to visit  to complete a subtask. We add this region $\ccalM^\ccalV_{\ccalL}(v^*)$ to the plan $p_{r,j}$. Next, the corresponding visit time indicates   the completion time of the associated subtask that is  captured by the mapping  $\ccalM_{e}^{\ccalV}(v^*)$. We add this time instance to timeline $t_{r,j}$. Since  each time instant on the time axis $\vec{t}$ corresponds to the completion of one  subtask, this visit time in $t_{r,j}$ corresponds to the time instant on $\vec{t}$ that the subtask $\ccalM_e^\ccalV(v^*)$ is completed.  {Continuing this process, we can construct for robot $[r,j]$ a sequence of waypoints and the corresponding timeline whose time instants appear on the time axis $\vec{t}$.} Given this high-level plan $\{p_{r,j}\}$, we can design low-level executable paths that reconsider the negative literals that were originally removed from the NBA $\auto{relax}$.

 \begin{cexmp}{exmp:1}{Time-stamped task allocation plan}
  {After solving the MILP for the workspace in Fig.~\ref{fig:workspace}, the high-level plans and the associated timelines for robots are as follows: $p_{2,1} = p_{3,1} = \{\ell_2, \ell_3\}, t_{2,1} = t_{3,1} = \{6, 16\},    p_{2,2} = \{\ell_4\}$, $t_{2,2} = \{10\}$.    That is, robots $[2,1]$ and $[3,1]$ visit the office building $\ell_2$ at time instant 6, then robot $[2,2]$ visits the control room $\ell_4$ at time instant 10, and finally robots $[2,1]$ and $[3,1]$ visit the delivery site  $\ell_3$ at time instant 16. The remaining  robots remain idle. Observe that the lengths of the plans differ since every robot may undertake different number of subtasks. The induced simple path in $\auto{subtask}^-$ in Fig.~\ref{fig:sub-NBA} is $\vertex{init}, v_1, v_2, v_3, v_6$. The associated time axis  is $\vec{t} = \{0, 6, 10, 16\}$, one time instant per subtask. In words, the subtask $(\vertex{init}, v_1)$ is completed at time instant 0 and the subtask $(v_1, v_2)$ is completed at time instant 6, which corresponds to the event that robots $[2,1]$ and $[3,1]$ visit the office building $\ell_2$. }
 \end{cexmp}

 \subsection{Design of low-level prefix paths}\label{sec:lowlevel}
   In this section we discuss the correction stage that re-introduces the negative literals to the NBA and corrects the high-level plans designed in Section~\ref{sec:path} (if needed) so that they satisfy  the specification $\phi$.
   To this end, we  first find the simple path in the NBA $\auto{subtask}^-$ connecting $v_0$ and $\vertex{accept}$ using  the time axis and the time-stamped task allocation plan. To satisfy the specification $\phi$, for every subtask in the simple path, we formulate a  generalized multi-robot path planning (GMRPP) problem. Each    GMRPP is essentially a generalization of the multi-robot point-to-point  navigation problem, whose goal  is to determine a collection of executable paths that allow the robots to complete the current subtask (by enabling the edge label at the end while respecting the starting vertex en route) and automatically activate the next subtask,   since the satisfaction of the edge label leads to the satisfaction of  the starting vertex of the next subtask. The details can be found in Appendix~\ref{sec:solution2mrta}, that also discusses different implementations of the proposed GMRPP (see Appendix \ref{sec:extension_gmrpp}) that depend on whether all or a subset of robots are allowed to move during the execution of the current subtask, since not all robots are responsible for the completion of this subtask, and whether the completion times of subtasks are disjoint or partially overlapping. Finally, the feasibility of the proposed  GMRPP is analyzed Lemma~\ref{prop:valid} in Appendix~\ref{app:correctness}.
   \subsection{Obtaining the best prefix-suffix  path}
   After obtaining the prefix path corresponding to a poset $P \in \{P_{\text{pre}}\}$ for the given pair $v_0$ and $\vertex{accept}$, next we find the suffix path around $\vertex{accept}$. For this, we can follow a similar process as this described in Sections~\ref{sec:graph}$\sim$\ref{sec:lowlevel} to find the prefix path for poset $P\in \{P_\text{pre}\}$, with the difference that now we treat the accepting vertex $\vertex{accept}$ as both the initial vertex $v_0$ and the accepting vertex $\vertex{accept}$. This is because the suffix path is essentially a loop, i.e., the final locations in the suffix path are identical to the initial locations in the suffix path, which are also the final locations in the prefix path; see Appendix~\ref{sec:suf} for more details.

   Specifically, given  the  pair $v_0$ and $\vertex{accept}$, we solve one MILP for each poset $P' \in \{P_{\text{suf}}\}$ to obtain a corresponding  suffix path; these MILPs can be infeasible if there are no feasible paths that induce simple paths corresponding to the poset $P'$. Then, among all suffix paths for all posets $P'\in \{P_{\text{suf}}\}$  we select the one with  the lowest cost. This best suffix path  corresponds to the  prefix path generated from a poset $P\in\{P_\text{pre}\}$ for the  given pair $v_0$ and $\vertex{accept}$. Combining this  suffix path with the corresponding  prefix path we obtain the best total path associated with the poset $P$ for the given  pair $v_0$ and $\vertex{accept}$. Then, using cost function~\eqref{eq:cost}, we select the best total path over all posets in $\{P_{\text{pre}}\}$ for the given pair $v_0$ and $\vertex{accept}$.  Finally, by iterating over all pairs of initial and accepting vertices with finite total length, we can obtain the best total path. We highlight that our method can terminate anytime once a feasible path is found, but running the algorithm longer can lead to more optimal feasible paths. Note also that by iterating over the pairs $v_0$ and $\vertex{accept}$ and the corresponding posets in the ascending order discussed in Section \ref{sec:poset}, it is more likely that the first solutions we obtain have low cost since they involve fewer subtasks that need to be accomplished. This observation is also validated numerically in Section~\ref{sec:sim}.

 \begin{figure}
   \centering
   \subfigure[$t=6$]{
     \includegraphics[width=0.29\linewidth]{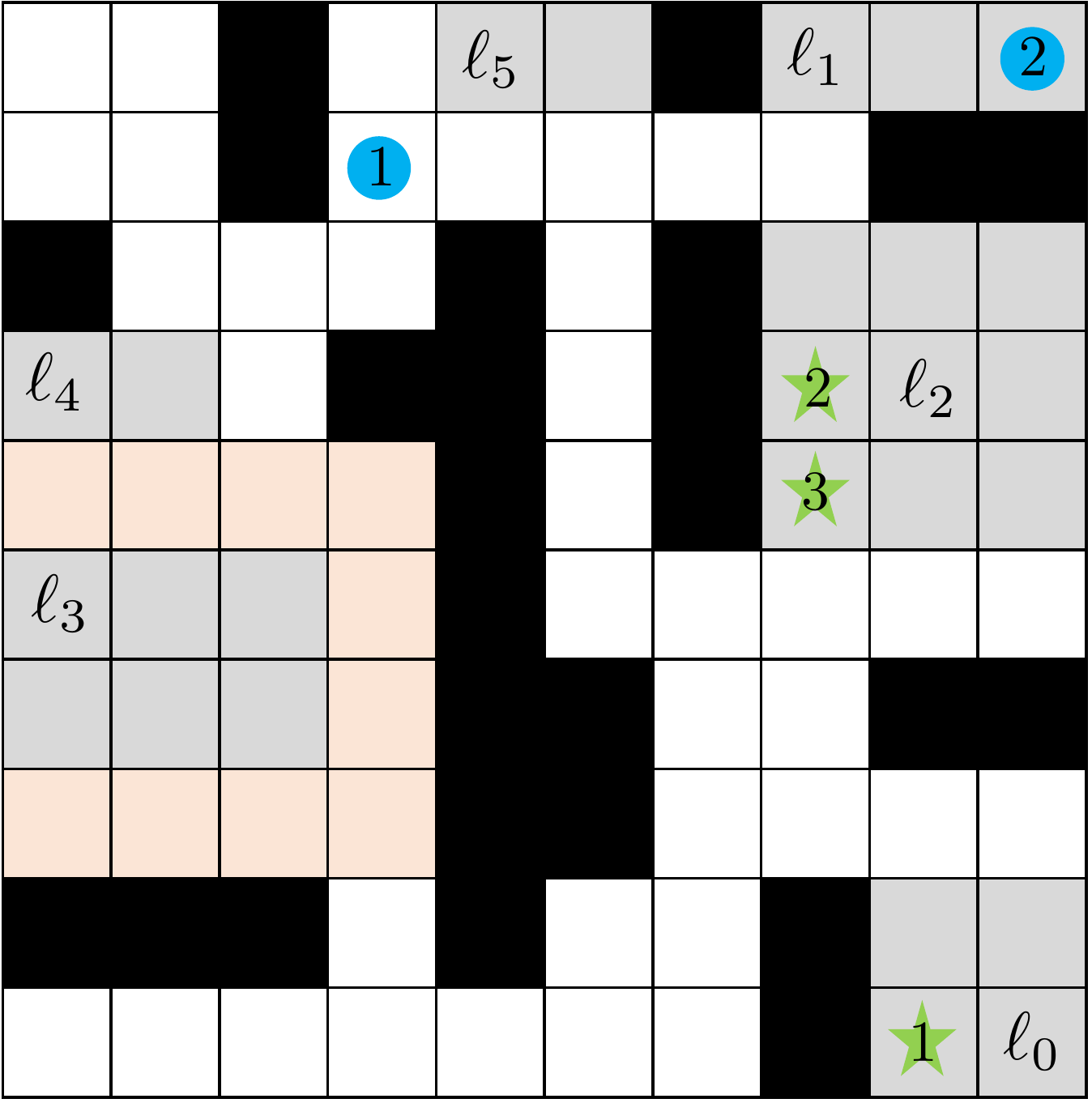}
     \label{fig:frame4}
   }
   \subfigure[$t=10$]{
     \includegraphics[width=0.29\linewidth]{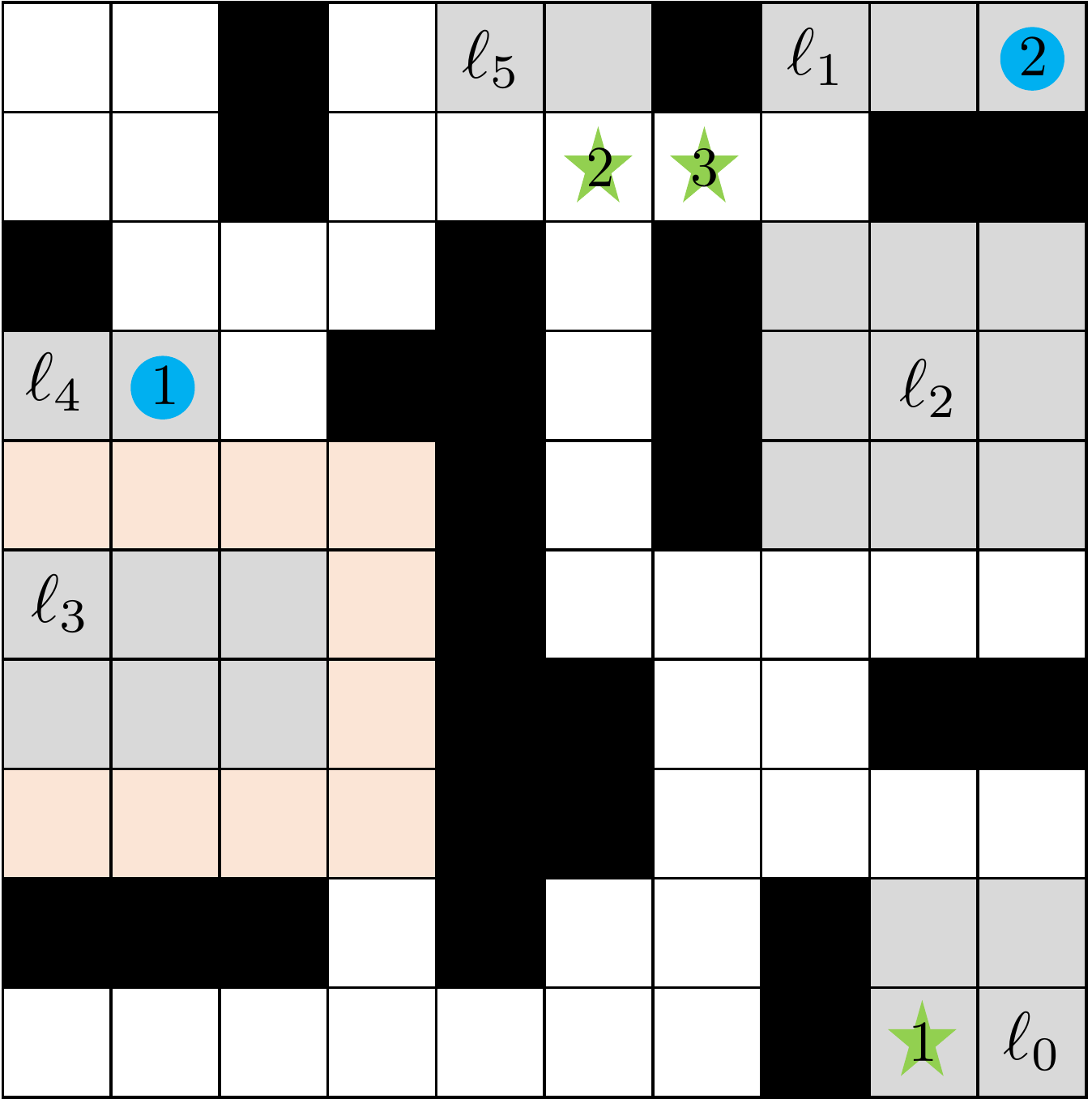}
     \label{fig:frame8}
   }
   \subfigure[$t=18$]{
     \includegraphics[width=0.29\linewidth]{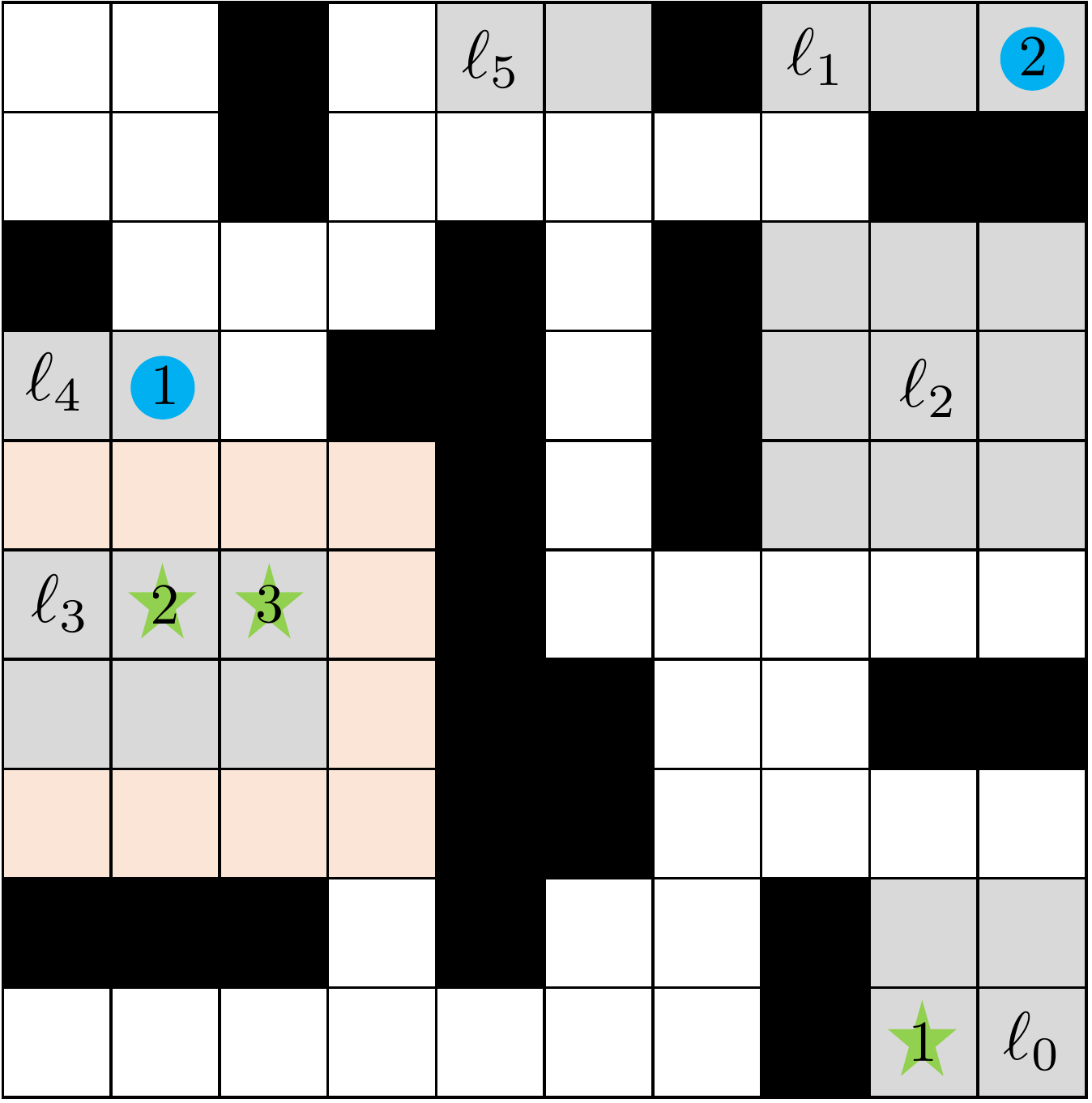}
     \label{fig:frame15}
   }
   \caption{Key frames demonstrating the execution of low-level paths that satisfy task~\hyperref[task:i]{{(i)}}. The initial configuration is shown in~Fig.~\ref{fig:workspace}. Fig.~\ref{fig:frame4} shows that at time instant 6, robots $[2,1]$ and $[3,1]$  reach the office building $\ell_2$, while robot $[1,2]$ is on the way to the control room $\ell_4$. Fig.~\ref{fig:frame8} shows at time instant 10, robot $[1,2]$ reaches the control room $\ell_4$ while robots $[2,1]$ and $[3,1]$ head towards the delivery site $\ell_3$. Finally, they reach $\ell_3$ in Fig.~\ref{fig:frame15} at time instant 18. Robots $[1,1]$ and $[2,2]$ remain idle throughout the process.}
   \label{fig:frames}
   \end{figure}
 \begin{cexmp}{exmp:1}{Low-level paths}
 When generating low-level paths for task~\hyperref[task:i]{\it {(i)}}, we also consider collision avoidance. Fig.~\ref{fig:frames} shows an array of three key frames where different subtasks are completed. Observe that task~\hyperref[task:i]{\it {(i)}} is completed at time  15,  longer than 12 given by the time-stamped task allocation plan since  the high-level plan uses the shortest travel time between regions and does not consider collision avoidance.
 \end{cexmp}

 \section{Theoretical Analysis}\label{sec:correctness}
 In this section, we analyze the completeness and soundness of our method. First we show that, with mild assumptions, our method is complete for \ltlz specifications.
 \begin{thm}[Completeness]\label{thm:completeness}
   Consider  a discrete workspace  satisfying Assumption~\ref{asmp:env}, a team of $n$ robots of $m$ types and a valid specification $\phi\in \textit{LTL}^0$. Assume also that   there exists a path $\tau = \tau^\textup{pre} [\tau^\textup{suf}]^\omega$ that induces a restricted accepting run $\rho = \rho^\textup{pre} [\rho^\textup{suf}]^\omega =  v_0, \ldots, \vertex{prior}, \vertex{accept}$ $ [\vertex{next}, \ldots, \vertex{prior}', \vertex{accept}]^\omega$ in the pre-processed NBA $\autop$ and satisfies  Assumption~\ref{asmp:same}. Then, the proposed synthesis  method can find a  robot path $\tilde{\tau}=\tilde{\tau}^{\textup{pre}} [\tilde{\tau}^{\textup{suf}}]^\omega$ that satisfies the specification $\phi$.
 \end{thm}

 The key idea in the proof of Theorem~\ref{thm:completeness} is to first show that feasible paths still exist in $\auto{subtask}^-$ and then use this fact to  show  feasibility of the MILP and GMRPP problems. The detailed proof can be found in~Appendix \ref{app:correctness}. We emphasize that the completeness result in Theorem~\ref{thm:completeness} is ensured for \ltlz rather than \ltlx formulas. This is because  task allocations captured by  induced atomic propositions in the prefix part may not lead to feasible allocations  in the suffix part. However, when the \ltlx specification can be satisfied by finite-length paths, such as co-safe LTL~\textup{\citep{kupferman2001model}} or LTL$_f$ \textup{\citep{de2013linear}}, then
 our method is complete also for \ltlx specifications; This is shown in Proposition~\ref{thm:prefix} in Appendix~\ref{app:completeness} as part of the proof of Theorem~\ref{thm:completeness}.

 \begin{rem}
   We note that the path $\tilde{\tau}$ constructed by our approach may not satisfy Assumption~\ref{asmp:same} that requires that robots close  their suffix loops at the same time the NBA $\ccalA_\phi$ transitions    to $\vertex{accept}$. However, in our  method, when the NBA $\autop$ transitions to $\vertex{accept}$, only those robots involved in the completion of the last subtask in the prefix part return to regions corresponding to their initial locations. Thereafter,  trajectories are closed.
 \end{rem}

 The following statement shows the soundness of our method, which is   a direct consequence of Theorem~\ref{thm:completeness}.
 \begin{cor}[Soundness]\label{thm:soundness}
   Consider   a discrete workspace, a team of $n$ robots of $m$ types and a valid specification $\phi\in \textit{LTL}^\chi$.  Then, the path returned by the GMRPP satisfies the specification $\phi$. Also, the specific implementation of the GMRPP is not important.
 \end{cor}

 \section{Numerical Experiments}\label{sec:sim}
 In this section we present  three case studies, implemented in Python 3.6.3 on a computer with 2.3 GHz Intel Core i5 and 8G RAM, that illustrate the correctness and scalability of our method. The MILP is solved using Gurobi~\citep{gurobi} with big-M $M_{\text{max}}=10^5$. First, we compare with the optimal solution to  examine the suboptimality of our proposed method when the NBA can be captured by one poset (thus, only one solution). Second, we generate multiple solutions for specifications with multiple posets, and compare the cost of the first solution corresponding to the widest poset to that of the subsequent solutions. We observe that the quality of the first solution obtained for the widest poset is generally very good.  Finally, we compare our method to the approach proposed in~\cite{sahin2019multi} for large workspaces and numbers  of robots and show that our method outperforms the approach in~\cite{sahin2019multi} in terms of optimality and scalability. We emphasize that the sets of restricted accepting runs of  all specifications $\phi_1-\phi_{10}$ considered in the following simulations, are nonempty, which shows that this assumption is not restrictive in practice.


 \subsection{Case study \RNum{1}: Suboptimality}
 In this case study, we examine the quality of the paths constructed for the two tasks in the Example~\ref{exmp:1}. Observe that in Fig.~\ref{fig:sub-NBA}, a unique poset corresponds to  the sub-NBA $\auto{subtask}^-$ for task~\hyperref[task:i]{(i)}. A similar  observation can be made for the sub-NBA $\auto{subtask}^-$ in Fig.~\ref{fig:nba_ii_subtask} for task~\hyperref[task:ii]{(ii)}. In the workspace shown in Fig.~\ref{fig:frames},  we randomly generate the initial locations of all robots inside label-free cells. {To measure the suboptmality of our solution in terms of  path length (travelled distance), we use brute-force search  to find the optimal cost.}
 
\begin{table}[!t]
  \caption{Statistics on the optimal cost and solutions} \label{tab:optimal}
  \begin{center}
   {\resizebox{0.8\linewidth}{!}{\begin{tabular}{cccccc}
    \toprule
    \multirow{2}{*}{task} &  \multirow{2}{*}{$J^*$} & \multicolumn{2}{c}{NoCol$+$Seq} &\multicolumn{2}{c}{Col$+$Sim} \\
     \cmidrule(r){3-4} \cmidrule(r){5-6}
     &  & cost & horizon &  cost & horizon \\
     \midrule
         \hyperref[task:i]{(i)} & 36.0$\pm$5.1  & 36.4$\pm$5.3 (35) & 23.9$\pm$4.1 & 38.8$\pm$5.5 & 19.6$\pm$2.8 \\
         \hyperref[task:ii]{(ii)} & 25.4$\pm$2.8 & 28.6$\pm$3.1 (8) & 29.2$\pm$2.9 & 28.6$\pm$3.1 & 29.2$\pm$2.9 \\
         \bottomrule
  \end{tabular}}}\\
  \end{center}
  \justify
  \scriptsize{Case Study \RNum{1}: Column ``NoCol$+$Seq'' represents the case where collision avoidance is ignored and robots move sequentially, and  column ``Col$+$Sim'' incorporates collision avoidance and simultaneous execution. The notation $J^*$ denotes the optimal cost without considering collision avoidance. The number of trials out of 50 trials where the cost cost is equivalent to the optimal cost $J^*$ are shown inside the parentheses.}
\end{table}


 Next, given the same randomly generated initial robot locations,  we implement our proposed method in the following two different ways. First, we implement GMRPP without collision avoidance and with sequential execution (see Appendix~\ref{sec:gmmpp1}). Using sequential execution, only robots participating in the subtask under consideration are assigned  target regions and the rest of the robots just move out of their way,  whereas in the case of the simultaneous execution (see Appendix~\ref{sec:extension_essential}), multiple  subtasks can be undertaken at the same time, and robots that do not participate in the current  subtasks simultaneously  move towards  their  target points for subsequent subtasks. Second, we implement GMRPP with collision avoidance (see Appendix~\ref{sec:extension_collision}) and with simultaneous execution.  Both implementations employ the full execution (see Appendix~\ref{sec:gmmpp1}), in which all robots are allowed to  move. Note that in the partial execution  (see Appendix~\ref{sec:extension_partial}), only necessary robots participating in the current   subtask are allowed to move and the remaining robots are treated as obstacles.  Table~\ref{tab:optimal} shows statistical results on  the path costs and path time horizons (number of time stamps), averaged over 50 trials. For task~\hyperref[task:i]{(i)}, the MILP for the high-level plan includes 105 variables and 183 constraints; for task~\hyperref[task:ii]{(ii)}, it includes 33 variables and 61 constraints to find the prefix plan and 94 variables and 162 constraints to find the suffix plan.

 Without considering collision avoidance, the cost returned by our method is  close to the optimal cost, especially for task~\hyperref[task:i]{(i)} that only requires paths of finite length. In 35 out of 50 trials, our method can identify the exact optimal solutions. For task~\hyperref[task:ii]{(ii)}, the additional cost  arises from planning separately for the prefix and suffix parts.  In the prefix part, the robot can visit the cell in region $\ell_2$ that is the closest to its initial location, however, it may incur additional cost to return to this cell in the suffix part. The costs when considering  collision avoidance are also close to the optimal cost,  indicating that often robots follow the shortest path. As for the path  horizon, observe that, for task~\hyperref[task:i]{(i)}, simultaneous execution results in shorter horizons since one robot of type 2 can move towards region $\ell_4$ while two robots of type 1 leave from their initial locations for region $\ell_2$. Nonetheless, for task~\hyperref[task:ii]{(ii)}, the horizon remains almost the same,  since the corresponding subtasks cannot be executed in parallel by the same robot.

\begin{table*}[!t]
  \caption{Results for  specifications $\phi_3-\phi_8$}
  \label{tab:quality}
  \begin{center}
  \resizebox{0.8\linewidth}{!}{\begin{tabular}{c|ccccc|cc|cccccc}
    \toprule
    \multirow{2}{*}{Task} &  \multirow{2}{*}{$N_{\text{pair}}$} &  \multirow{2}{*}{$|\auto{}|$} & \multirow{2}{*}{$|\autop|$}  & \multirow{2}{*}{$|\auto{subtask}^{-,\text{pre}}|$}   & \multirow{2}{*}{$|\auto{subtask}^{-, \text{suf}}|$} &  \multirow{2}{*}{MILP$^\text{pre}$} &  \multirow{2}{*}{MILP$^\text{suf}$} & \multicolumn{2}{c}{$N_{\text{sol}}=1$} & \multicolumn{2}{c}{$N_{\text{sol}}=5$} & \multicolumn{2}{c}{$N_{\text{sol}}=10$}\\
  \cmidrule(r){9-10}  \cmidrule(r){11-12}     \cmidrule(r){13-14}
    & & & & & & &&  cost & time(sec) & cost & time(sec) & cost & time(sec)\\
    \midrule
    $\phi_3$ & 8 & (20, 142) & (20, 49) & (3, 2) & (5, 5) & (45, 78)& (276, 397)& 66.4$\pm$4.7 & 1.7$\pm$0.2 &  ---  &  ---  &   ---  &   ---\\
    $\phi_4$ & 4 &  (10, 57) & (10, 31) & (3, 2) & (3, 3) & (45, 78)& (130, 210) & 61.4$\pm$4.8 & 1.4$\pm$0.2 &  ---  &  ---  &  ---  &  --- \\
    $\phi_5$ &2 &  (11, 31) & (11, 25) & (10, 19) & (3, 3) & (78, 141)& (76, 142) & 17.9$\pm$5.7 & 0.5$\pm$0.1 & 17.9$\pm$5.7 & 2.2$\pm$0.7  & --- & --- \\
    $\phi_6$ &1  &(4, 9) & (4, 8) & (4, 5) & (4, 5) & (117, 203)& (204, 308) & 33.3$\pm$7.1 & 1.2$\pm$0.5  &  30.8$\pm$5.7 & 3.2$\pm$1.1 &  30.8$\pm$5.7 & 4.5$\pm$1.5 \\
    $\phi_7$ &3 & (24, 140) & (24, 104) & (22, 57) & (9, 18) & (124, 194)& (93, 164) &  45.4$\pm$7.1 & 2.5$\pm$0.3  &  45.4$\pm$7.1 & 2.9$\pm$0.3 &  45.4$\pm$7.1 & 3.9$\pm$0.3 \\
    $\phi_8$ &4 & (15, 83) & (15, 41) & (8, 13) & (8, 14) & (120, 210)& (201, 325) & 74.0$\pm$6.2  & 1.6$\pm$0.2 & 74.0$\pm$6.2  & 8.7$\pm$0.5  & 74.0$\pm$6.2  & 22.4$\pm$1.2  \\
    \bottomrule
  \end{tabular}} 
  \end{center}
  \justify \footnotesize{Case Study \RNum{2}: $N_{\text{pair}}$ is the number of pairs of initial and accepting vertices, $|\ccalA|$, $|\autop|$, $|\auto{subtask}^{-,\text{pre}}|$ and $|\auto{subtask}^{-,\text{suf}}|$ are the size of the NBA before and after pre-processing, for the prefix and suffix parts from which the first solutions are obtained, respectively.  MILP$^{\text{pre}}$ and MILP$^{\text{suf}}$ are the size of MILP of the first solution. The symbol ``---'' means that only one solution found for $\phi_3$ and $\phi_4$, and less than or equal to 5 solutions found for $\phi_5$.}
\end{table*}

 \subsection{Case study \RNum{2}: Quality of the first solution} Common to the two specifications in the first case study is that the sub-NBA $\auto{subtask}^-$ for the prefix and suffix parts can be concisely captured by one poset, which may not be the case for most specifications. Here, we consider various specifications that can produce many posets  and examine the quality of the first solutions obtained for the widest poset (see Section~\ref{sec:poset}) by comparing to subsequent solutions obtained for subsequent posets. We use the same workspace and robot team as in Example~\ref{exmp:1}. The considered specifications are as follows:
 \begingroup
 \allowdisplaybreaks
 \begin{align*}
   \phi_3  = & \, \square \lozenge (\ap{2}{1}{2,1} \wedge \lozenge (\ap{2}{1}{3,1} \wedge \lozenge (\ap{2}{1}{4,1} \wedge \lozenge \ap{2}{1}{5,1}  )   )  ), \\
   \phi_4 = & \square \lozenge (\ap{2}{1}{2,1} \wedge \lozenge \ap{2}{1}{3,1}) \wedge \square (\ap{1}{1}{5,2} \simplies \bigcirc (\ap{1}{1}{5,2} \;\ccalU \, \ap{1}{2}{4})) \\
   &  \wedge \square \neg \ap{2}{1}{4}\\
   \phi_5  = & \, \lozenge (\ap{1}{2}{4,1} \wedge \bigcirc (\ap{1}{2}{4,1} \;\ccalU \, \ap{2}{1}{3})) \wedge \square \lozenge ( \ap{1}{2}{4,1} \wedge \lozenge \ap{1}{2}{3,1}), \\
   \phi_6  = & \,  \square \lozenge (\ap{1}{1}{3,1} \,\vee \, \ap{1}{1}{5,1}) \wedge \square \lozenge \ap{1}{1}{2,1}  \wedge \square \lozenge (\ap{2}{2}{3} \,\vee\, \ap{2}{2}{5})  \\
   & \wedge \square \neg \ap{2}{1}{4} \wedge \square \neg \ap{2}{2
   }{4}, \\
   \phi_7  = & \,  \square \lozenge (\ap{1}{2}{4} \wedge \bigcirc (\lozenge \neg \ap{1}{2}{4})) \wedge  \square \lozenge (\ap{1}{1}{5} \wedge \bigcirc (\lozenge \neg \ap{1}{1}{5} )) \\
   & \wedge  \lozenge  (\ap{3}{1}{3} \wedge \ap{2}{2}{3}), \\
   \phi_8 = & \, \square \lozenge  (\ap{2}{2}{4,1} \wedge \lozenge (\ap{2}{2}{2,1} \wedge \lozenge \ap{2}{2}{5,1}))  \wedge  \neg \ap{1}{2}{2} \;\ccalU \, \ap{2}{2}{4,1} \\
   & \wedge \neg \ap{1}{2}{5} \; \ccalU \, \ap{2}{2}{4,1} \wedge (\square \lozenge \ap{2}{1}{5}\, \vee \,  \square \lozenge \ap{2}{1}{3}),
 \end{align*}
 \endgroup
 where (a) $\phi_3$ requires that the same two robots of type 1 meet first at regions $\ell_2$, then  $\ell_3$, next  $\ell_4$, and finally at $\ell_5$, repeating this process infinitely often;
 (b) $\phi_4$ requires that the same two robots of type 1 meet at region $\ell_2$ and then $\ell_3$ infinitely many times. Also, every time one  robot of type 1  visits region $\ell_5$, it should stay there until one robot of type 2 visits region $\ell_4$. Finally, at most one robot of type 1 should be  at region $\ell_4$ at any time;
 (c) $\phi_5$ requires that one robot of type 2 visits region $\ell_4$ and stays there until two robots of type 1 reach region $\ell_3$. This type 2 robot should visit regions $\ell_4$ and then $\ell_3$ infinitely many times;
 (d) $\phi_6$ requires that the same  robot of type 1 visits regions $\ell_3$ or $\ell_5$ infinitely many times and this robot visits region $\ell_2$ infinitely many times, while two robots of type 2 meet at regions $\ell_3$ or $\ell_5$ infinitely many times.  Finally, at most one robot of any type can be  present at region $\ell_4$;
 (e) $\phi_7$ requires that one robot of type 2 periodically visits region $\ell_4$ while one robot of type 1 periodically visits region $\ell_5$. All robots must eventually  meet at region $\ell_3$;
 (f) $\phi_8$ requires that two robots of type 2 meet at region $\ell_4$, then $\ell_2$ and next $\ell_5$, repeating this infinitely many times  with the restriction that no robots of type 2 reach regions $\ell_2$ and $\ell_5$ before two robots of type 2 meet at region $\ell_4$ for the first time. Ultimately, two robots of type 1 should meet at region $\ell_5$ or $\ell_3$ infinitely often.

 These  specifications  involve various operators and are representative of commonly used complex tasks in robotics applications. For example, $\phi_3$ can capture surveillance and data gathering tasks~\citep{smith2011optimal,guo2017distributed}, and the subformula $\square \lozenge (\ap{1}{1}{3,1} \,\vee\, \ap{1}{1}{5,1})$ in $\phi_6$ can specify intermittent connectivity tasks where robots are required to meet at communication regions infinitely often~\citep{kantaros2018distributed,kantaros2019temporal,khodayi2019distributed}. Furthermore, subformula $\square \neg \ap{2}{2}{4}$ in $\phi_5$ can be used to represent collision avoidance among robots and $\neg \ap{1}{2}{2} \;\ccalU \, \ap{2}{2}{4,1}$ in $\phi_8$ can  prioritize certain subtasks to others.

 We executed our method 20 times  for each specification. In each trial, we randomly generated initial robot locations inside the label-free cells such that  no two robots occupy the same cell. We considered collision avoidance, as well as full  and simultaneous execution. In Table~\ref{tab:quality}, we report  the number of pairs of initial and accepting vertices in the NBA $\autop$ before pre-processing,  the size (number of vertices and edges) of the NBA $\autop$ before and after pre-processing, and the size of the sub-NBA $\auto{subtask}^-$ for the prefix and suffix parts from which the first solutions are obtained. {The size (number of variables and constraints) of the MILP for the prefix and suffix part of the first solution is also displayed.}\footnote{{The size of the MILP differs for different solutions since they may be generated from different posets of subtasks. We only report the results for the first solution since we aim to examine the quality of the first solution.}} We terminate our method until all solutions or the first 10 solutions are generated, whichever comes first. We record the smallest cost   achieved by the first solution, after the first 5  and  10 solutions along with the runtimes.

 In Table~\ref{tab:quality}, we observe that  the size of sub-NBA $\auto{subtask}^-$ is dramatically reduced compared to the size of NBA before pre-processing, especially for specifications $\phi_3$, $\phi_7$ and $\phi_8$, considerably reducing the computation times. It takes about 20 seconds to get 10 solutions for specification $\phi_8$. Except for specification $\phi_6$, the first solution returned by our method is also the lowest cost solution. For specification $\phi_6$, the best solution corresponds to one of the first 5 posets. This is  because  our optimization-based  method sorts the set of  posets in part according to their height so that posets with smaller numbers of subtasks are considered first  (see Section~\ref{sec:poset}). Therefore, we can terminate our method only after a few solutions  have been obtained, which is especially important when the complexity of the planing problem increases,  as in the next case study.

 \subsection{Case study \RNum{3}: Scalability}

 In this case study, we examine the scalability of our proposed method with respect to the size of the workspace and the number of robots. Specifically, we first  compare our method to the Bounded-Model-Checking-based (BMC) method in~\cite{sahin2019multi} and then, we examine the effect of full or partial  execution on the performance.
  \begin{figure}[!t]
     \centering
     \includegraphics[width=0.65\linewidth]{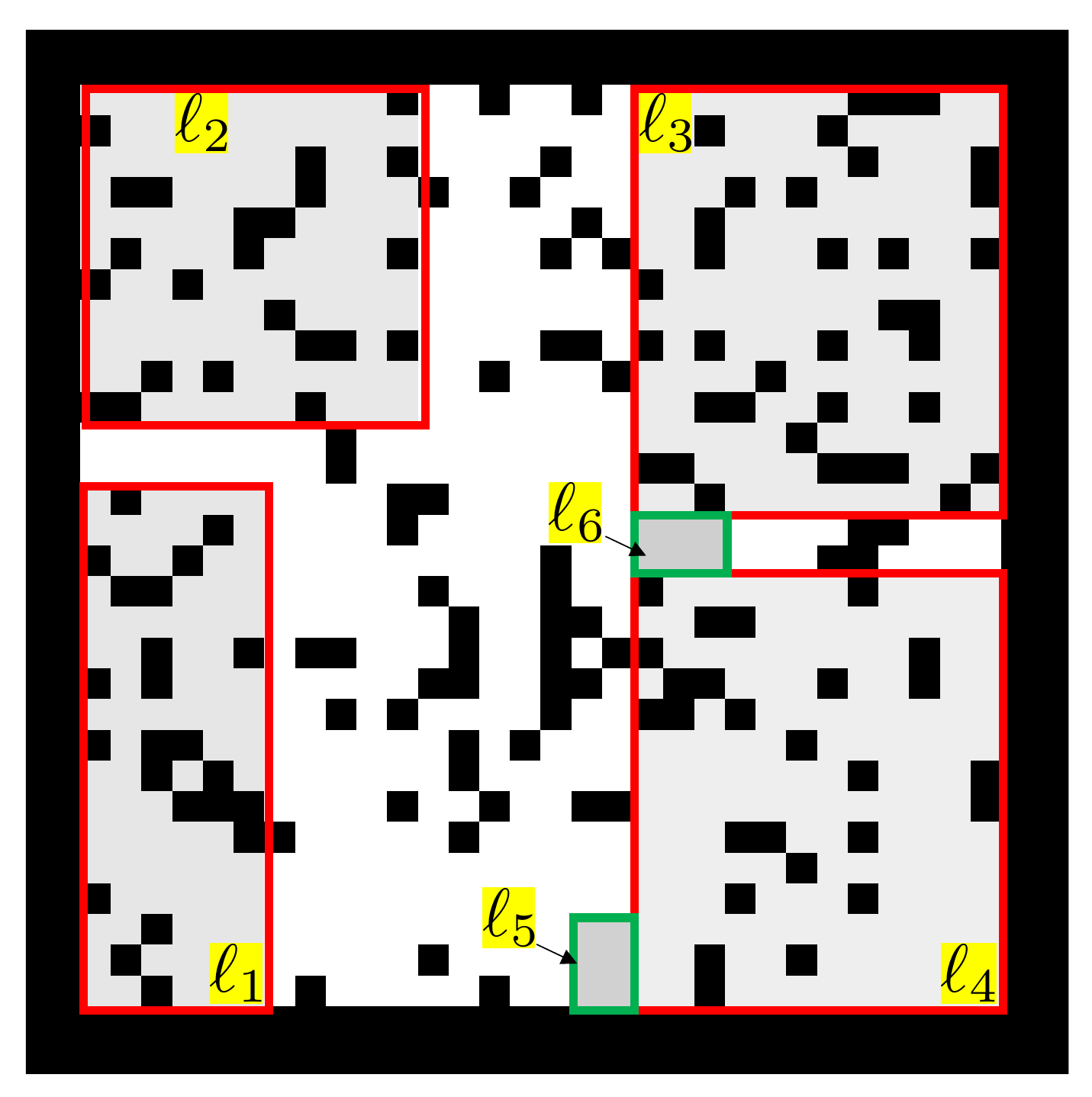}
     \caption{Grid world from~\cite{sahin2019multi}}\label{fig:scalability}
  \end{figure}

 \subsubsection{Comparison with the BMC method}
Similar to our method,~\cite{sahin2019multi} also adopts a hierarchical framework, which improves the scalability of methods in~\citep{sahin2017synchronous,sahin2019multirobot} that address feasible control synthesis over LTL$^0$. 
 For the purpose of comparison,  we borrow the workspace used in~\cite{sahin2019multi}, a 30-by-30 grid world containing 6 regions $\ell_i, i=1,\ldots,6$; shown in Fig.~\ref{fig:scalability}. At each trial, $20\%$ of cells are randomly selected as obstacles. We consider a team of $n$  robots of the same type whose initial locations are randomly sampled inside region $\ell_1$.  The specification we consider is given by~\cite{sahin2019multi}:
 \begin{align*}
   \phi_9 = \,   \square \lozenge \ap{n}{1}{2} \wedge   \square \lozenge \ap{n/2}{1}{3} &\,\wedge \square \lozenge \ap{n/2}{1}{4} \\
   & \wedge \neg \ap{1}{1}{4} \,\ccalU\, (\ap{1}{1}{5} \wedge \ap{1}{1}{6}),
 \end{align*}
 which requires (a) all robots to  meet at region $\ell_2$ infinitely often, (b) at least half of the robots to meet at regions $\ell_3$ and $\ell_4$, respectively, infinitely often, and (c) robots should not visit region $\ell_4$ until at least one robot is inside region $\ell_5$ and one robot is inside region $\ell_6$ at the same time. We vary the number  of robots $n$ from 4 to 30, which produces  a product transition system that has up to $(30\times30)^{30}  \approx 10^{90}$ states.

 The size of the NBA is independent from the number of robots. The NBA $\autop$ has one pair of initial and accepting vertices, 5 vertices and 10 edges (excluding self-loops). The sub-NBA $\auto{subtask}^-$ for the prefix part has 5 vertices and 5 edges and for the suffix part has 4 vertices and 5 edges. In the implementation of our method, we employ the full and simultaneous execution. We record runtimes  and cost of the first feasible solutions, where the cost is the sum of the cost of the prefix and suffix parts. 
 Both methods consider collision avoidance. The horizon increases by 10 when no solution exists for the GMRPP, until the considered horizon exceeds  the initial horizon by 100. {The source code for~\cite{sahin2019multi} can address robots of the same type, and is available at~\cite{sahincltl}.}
 The statistical results  averaged over 10 trials are shown in Table~\ref{tab:scalability}. For $n=30$ robots, the MILP to find the prefix plan includes 89179 variables and 91244 constraints and the MILP to find the suffix plan includes 112519 variables and 115482 constraints.
 \begin{table}[!t]
   \caption{Results with respect to the number of robots}\label{tab:scalability}
   \centering
   \resizebox{\linewidth}{!}{\begin{tabular}{ccccc}
       \toprule
       \multirow{2}{*}{$n$} & \multicolumn{2}{c}{Our method} & \multicolumn{2}{c}{BMC method} \\
       \cmidrule(r){2-3}  \cmidrule(r){4-5}
       & cost & time(sec) & cost & time(sec) \\
       \midrule
       4 & 270.6$\pm$4.4 & 62.4$\pm$1.4 & 944.4$\pm$21.2 & 76.5$\pm$13.8\\
       8 & 513.0$\pm$30.2 & 124.9$\pm$9.2 & 1819.0$\pm$149.9 & 334.9$\pm$153.9 \\
       12 & 794.6$\pm$11.1 & 187.4$\pm$9.1 &  2217.0$\pm$163.8 &  704.3$\pm$178.0\\
       16 & 1080.2$\pm$14.7 & 502.0$\pm$225.4 & 2725.8$\pm$149.2 & 1135.8$\pm$123.7\\
       30 & 2509.4$\pm$168.9 & 4072.1$\pm$985.4 & --- &--- \\
       \bottomrule
   \end{tabular}}
 \end{table}

 Observe in  Table~\ref{tab:scalability} that our method outperforms the BMC method both in terms of runtimes and optimality of the solutions. Specifically, as the number of robots increases, the runtime of our method is about half the runtime of the BMC method but the cost returned by our method is about 1/3 of the cost of the solutions obtained using the BMC method. The reason is that we optimize the cost at both the high level and the low level, while the BMC method only considers feasibility. For $n=30$ robots, the BMC method did not produce a solution within 2 hours.  Furthermore, the efficiency  of the low-level path planner has significant impact on the runtime. In our method, the number of times that the path planner is invoked is the same  or smaller than the number of subtasks in the simple path extracted from  the high-level plan (see Appendix~\ref{sec:run}). {On the other hand, the BMC method abstracts the given environment by aggregating states with the same observation, where transitions between abstract states are defined by whether they share the same boundary.}  Then, each transition in the high-level plan obtained by the BMC method is converted into one instance of multi-robot path planning problem. Obviously, the number of transitions in the BMC method  is larger than the number of subtasks in our method, since each subtask may take multiple transitions.

 \subsubsection{Full vs. partial GMRPP execution} We use the same workspace as in Fig.~\ref{fig:scalability}  and consider a team of $n$ homogeneous robots that are subject to the specification:
 \begin{align*}
   \phi_{10} = \, \lozenge (\ap{3}{1}{5} \,\vee \, \ap{3}{1}{6}) &\,\wedge   \square \lozenge (\ap{n/2}{1}{2,1} \wedge \lozenge \ap{n/2}{1}{4,1}) \\
   & \wedge  \square \lozenge \ap{n/4}{1}{3} \wedge \square \neg \ap{4}{1}{6},
 \end{align*}
 which requires that (a) at least 3 robots eventually meet at either region $\ell_5$ or $\ell_6$, (b) a fleet of  at least half robots meet at region $\ell_2$ and then the same robots meet at region  $\ell_4$, infinitely often, (c) at least a quarter robots meet at region $\ell_3$ infinitely often, and (d) always no more than 3 robots can be present at region $\ell_6$ at the same time.

 Before pre-processing, there are two pairs of initial and accepting vertices in the NBA $\autop$ that contains  8 vertices and 27 edges. After pre-processing,  the NBA $\autop$  has 8 vertices and 20 edges. For the first pair of initial and accepting vertices, the sub-NBA $\auto{subtask}^-$ associated with  the prefix part has 7 vertices and 10 edges, and the sub-NBA associated with suffix part has 5 vertices and 7 edges. We compare the performance of our method for the full and partial execution in the GMRPP problem and for an increasing number of robots up to 32.  The results averaged over 10 trials are shown in Table~\ref{tab:scalability2}. For $n=32$ robots, the MILP to find the prefix plan includes 41428 variables and 42533 constraints and the MILP to find the suffix plan includes 78625 variables and 81200 constraints. It can be seen  that our method with  partial execution in the GMRPP problem  takes less time than with full execution. This advantage becomes more significant as the number of robots increases since in this case, a larger number of robots that do not participate in the current subtask can remain idle and can be treated as obstacles in the GMRPP. For example, the subtask requiring that at least 3 robots meet at region $\ell_5$ or $\ell_6$, only involves 3 robots no matter how large the robot team is. On the other hand, the full execution of the GMRPP problem  results in slightly larger cost which suggests that even though all robots are allowed to move, those robots that do not participate in the specific subtask rarely move because our method optimizes the cost. Observe that for the partial execution of the GMRPP problem and for 32 robots,  no solutions are generated in 3 out of 10 trials. This is  due to the fact that robots treated as obstacles affect the obstacle-free workspace and, therefore, may  make the GMRPP infeasible. Thus, the partial execution of the GMRPP problem can be more effective in large workspaces with few robots, where a few idle robots do not significantly alter the obstacle-free environment.

 \begin{table}[!t]
   \caption{Results with respect to the number of robots.}\label{tab:scalability2}
   \centering
   \resizebox{\linewidth}{!}{\begin{tabular}{ccccc}
       \toprule
       \multirow{2}{*}{$n$} & \multicolumn{2}{c}{Full execution} & \multicolumn{2}{c}{Partial execution} \\
       \cmidrule(r){2-3}  \cmidrule(r){4-5}
       & cost & time(sec) & cost & time(sec) \\
       \midrule
       4 & 181.4$\pm$17.7 & 89.5$\pm$5.0 & 180.4$\pm$20.1 & 65.8$\pm$10.1\\
       8 & 356.6$\pm$16.0 & 198.9$\pm$12.3 & 354.2$\pm$15.2 & 129.3$\pm$4.9 \\
       12 & 573.5$\pm$63.3 & 350.7$\pm$25.4 &  554.3$\pm$49.4 &  192.5$\pm$10.4\\
       16 & 774.2$\pm$59.0 & 561.0$\pm$44.4 & 763.0$\pm$50.7 & 278.9$\pm$8.9\\
       32 & 1560.4$\pm$160.7 & 1886.8$\pm$696.0 & 1524.6$\pm$30.6$^*$ &  778.1$\pm$134.9 \\
       \bottomrule
   \end{tabular}}
   \begin{tablenotes}
     \scriptsize
   \item $^*$ 3 out of 10 trials failed.
   \end{tablenotes}
 \end{table}

 \section{Conclusion}\label{sec:conclusion}
 In this work, we consider the problem of allocating tasks, expressed as global LTL specifications,  to teams of heterogeneous mobile robots. This  problem cannot be solved using  existing model checkers since all possible allocations of robots to tasks can result in LTL formulas that are prohibitively long. We proposed a hierarchical approach to solve this problem  that first solves an MILP to obtain a high-level  time-stamped allocation of robots to tasks  and then formulates a sequence  of multi-robot path planning problems to obtain the low-level executable paths. We proved that, with mild assumptions, the proposed method is complete and we provided extensive simulations that showed that our method outperforms the state-of-the-art  BMC method in terms of optimality and scalability. Scalability of our method is primarily due to a clever relaxation of the NBA that captures the LTL specification, that involves removing the negative literals. This relaxation is motivated by ``lazy collision checking'' methods for point-to-point navigation, and significantly simplifies the high-level planning problem as constraint violation is not considered during planning and instead it is only checked during execution when needed. To the best of our knowledge, this is the first time that ``lazy collision checking'' methods are used and shown to be effective for high-level planning tasks.


\bibliographystyle{IEEEtran}
\bibliography{xl_bib}

 \begin{appendices}

 \section{Time-Stamped Task Allocation}\label{app:milpformulation}
 {In this section, we first formulate the MILP to obtain the time-stamped task allocation plan for the prefix part. Next, we present a similar process for the suffix part. Finally, we discuss extensions of the MILP to address problem-specific requirements.}

 \subsection{Construction of the prefix MILP}\label{app:appendix_prefix_milp}
  To formulate the proposed MILP, we define two types of variables: the routing variables $x_{uvr} \in\{0, 1\}$ and the scheduling variables $t^-_{vr}, t^+_{vr} \in \mathbb{N}$, where $x_{uvr}=1$  if robot  $r\in \ccalM^\ccalV_\ccalK(v)$ traverses the edge $(u,v) \in \ccalE_\ccalG$, and $t^-_{vr}, t^+_{vr}$ are times when robot $r$ should arrive at and is allowed to leave from vertex $v \in \mathcal{V}_\ccalG$. We assume that robot $r$ is still at vertex $v$ at departure time $t^+_{vr}$. Since the satisfaction of the edge label is instantaneous, if $v \in \ccalV_\ccalG$ is associated with an edge label, we have
   $t_{vr}^- = t_{vr}^+$, which means that the robot is allowed to leave at the next time instant. As for the vertex label, we have $t_{vr}^-  \leq t_{vr}^+$, which means that the robot should stay where it is to wait for the satisfaction of the corresponding edge label.

 \subsubsection{Routing constraints}\label{app:routing_constraints} These constraints are associated with vertices and restrict the flow of robots  between connected vertices in $\ccalV_\ccalG$. Specifically, $\forall\, v \in \ccalV_\ccalG\setminus\ccalV_{\text{init}}$, the constraint that  $v$ is visited by at most one robot of type $\ccalM^\ccalV_{\ccalK}(v)$ can be written as
 \begin{align}
    & \sum_{u:(u,v)\in \ccalE_\ccalG} \sum_{r\in \ccalM^\ccalV_{\ccalK}(v)} x_{uvr}  \leq 1,  \label{eq:1}
 \end{align}
 which is not a strict equality since the clause that vertex $v$ is associated with can be false. In this case, there is no need to visit this vertex. Moreover, the  constraint that the inflow is no less than the outflow at any vertex $\,v \in \ccalV_\ccalG\setminus\ccalV_{\text{init}}$ can be written as
 \begin{align}
 & \sum_{w:(v,w)\in \ccalE_\ccalG} x_{vwr}   \leq  \sum_{u:(u,v)\in \ccalE_\ccalG} x_{uvr}, \;\forall\, r \in \ccalM^\ccalV_{\ccalK}(v), \label{eq:2}
 \end{align}
 which states that  robots can remain idle if they are not assigned a subtask. Then, $\forall \,v \in \ccalV_{\text{init}}$, the initial conditions associated with constraint~\eqref{eq:2} are
 \begin{subequations}\label{eq:2.5}
   \begin{align}
     & \sum_{w:(v,w)\in \ccalE_\ccalG} x_{vwr}  \leq 1, \;\text{if}\; r = r_v \in \ccalM^\ccalV_{\ccalK}(v), \label{eq:2.5a}\\
     & \sum_{w:(v,w)\in \ccalE_\ccalG} x_{vwr}  = 0, \;\forall r \in \ccalM^\ccalV_{\ccalK}(v)\setminus \{ r_v\}, \label{eq:2.5b}
   \end{align}
 \end{subequations}
 where $r_v$ refers to the specific robot at the initial location $\ccalM^\ccalV_{\ccalL}(v) = s^0$ if $v \in \ccalV_{\text{init}}$.
  \subsubsection{Scheduling constraints}\label{app:scheduling_constraints} These constraints are also associated with vertices and capture  the temporal relation on a vertex or between visits of two connected vertices. First, we require positivity of scheduling variables, $\forall \, v \in \ccalV_\ccalG$, i.e.,
 \begin{align} \label{eq:3}
   0 \leq t_{vr}^{-}, t_{vr}^+ \leq M_{\text{max}} \sum_{u:(u,v)\in \ccalE_\ccalG} x_{uvr}& , \forall r\in \ccalM^\ccalV_{\ccalK}(v),
 \end{align}
 where $M_{\text{max}}$ is a large positive integer. The constraint~\eqref{eq:3} implies that $t_{vr}^- = t_{vr}^+ = 0$ if vertex $v$ is not visited by robot $r$. The initial condition associated with constraint~\eqref{eq:3} is
 \begin{align}\label{eq:3.5}
    t_{vr}^- = t_{vr}^+ = 0,  \quad   \;\forall r \in \ccalM^\ccalV_{\ccalK}(v), \;\forall\, v \in \ccalV_{\text{init}}.
 \end{align}
 {The scheduling constraints between visiting times of two connected vertices considering the travel time,} $\forall\, r\in \ccalM^\ccalV_{\ccalK}(v),\, \forall \,(u,v)\in \ccalE_\ccalG$, are
 \begin{subequations}\label{eq:4}
   \begin{align}
     & \hspace{-0.6em}t_{ur}^+  + (T^*_{uv} + 1)  x_{uvr}  \leq t_{vr}^- + M_{\text{max}} (1 - x_{uvr}),  \nonumber \\
     & \pushright{\text{if}\; u\|_{P} v,} \label{eq:4a} \\
     & \hspace{-0.6em} t_{ur}^+ +  T^*_{uv} x_{uvr} \leq  \,t_{vr}^- + M_{\text{max}} (1 - x_{uvr}),\; \text{otherwise}.  \label{eq:4b}
   \end{align}
 \end{subequations}
 {where $T^*_{uv}$ is the shortest travel time between regions that vertices $u$ and $v$ correspond to and $u \|_{P} v$ means that the subtask $\ccalM^\ccalV_{{e}}(u)$ is incomparable to $\ccalM^\ccalV_e(v)$, corresponding to the cases in Appendices~\ref{sec:b} and \ref{sec:c}.}
 When $x_{uvr}=1$, constraint~\eqref{eq:4a} becomes $t_{ur}^+  + T^*_{uv} + 1  \leq t_{vr}^-$ and constraint~\eqref{eq:4b} becomes $t_{ur}^+  + T^*_{uv}  \leq t_{vr}^-$. Because $T^*_{uv} \geq 0$, constraints~\eqref{eq:4} ensure that $t_{vr}^-$ should be no less than $t_{ur}^+$ if $x_{uvr}=1$. Note that a cycle in $\ccalG$ must include a pair of incomparable vertices, since all comparable vertices constitute a chain. Constraint~\eqref{eq:4a} prevents cycles in $\ccalG$ where all vertices correspond to the same region. For instance, consider such a cycle $u_1,u_2,\ldots, u_c, u_1$. Without  constraint~\eqref{eq:4a}, a solution with zero travel time satisfies constraints~\eqref{eq:1},~\eqref{eq:2} and~\eqref{eq:4b}, resulting in $x_{u_1 u_2 r} = ,\ldots,=x_{u_c u_{1} r}= 1$ for a robot $r$, without this robot actually visiting any vertex from its initial location.  Constraint~\eqref{eq:4a} is functionally similar to the subtour elimination constraint in vehicle routing problems which prevents any solution that consists of a disconnected tour. We leverage a  term 1 to ensure that time increases along the edge that connects incomparable vertices, thus preventing visiting of a cycle.

   \subsubsection{Logical constraints}\label{sec:labelconstraints} These constraints associate vertices with subtasks and  encode the logical relation between labels, clauses and literals, and the realization of literals.
                 {Given a subtask $e\in X_P$, every vertex or edge label  $\gamma  = \bigvee_{p\in \ccalP} \bigwedge_{q\in \ccalQ_p} \ap{i^q}{j^q}{k^q,\chi^q}$ (neither $\top$ nor $\bot$) is true as long as one of its clauses is true.  To this end, we associate each clause  $\ccalC_{p}^{\gamma} \in \clause{\gamma}$ with a binary variable $b_p$ such that $b_p=1$ if the $p$-th clause  $\ccalC_{p}^{\gamma}$ is true. Hence, the label $\gamma$ being true can be encoded as
 \begin{align}\label{eq:c}
    \sum_{p  \in \ccalP} b_p = 1.
 \end{align}
 That is, one and only one clause is true, which is justified by condition~\hyperref[asmp:a]{(a)} in  Definition~\ref{defn:same} which states that it is the same clause in a vertex label that is satisfied. The logical relation, between a clause and its literals, that the satisfaction of the clause is equivalent to the satisfaction of all its literals, is written as
 \begingroup\makeatletter\def\f@size{9}\check@mathfonts
 \def\maketag@@@#1{\hbox{\m@th\normalsize\normalfont#1}}%
 \begin{align}
  \!\!\!  \left. \left[ \sum_{q\in \ccalQ_p} \sum_{v \in \ccalM^\mathsf{lits}_{\ccalV}(e,0|1,p,q)} \sum_{u: (u,v) \in \ccalE_\ccalG} \sum_{r\in \ccalM^\ccalV_{\ccalK}(v)} x_{uvr} \right] \middle/ {\sum_{q\in \ccalQ_p} i^q} \right.= b_p, \label{eq:6}
 \end{align}
 \endgroup
 which  connects  the routing variables $x_{uvr}$ with the logical variables $b_p$. In words, if $b_p=1$, then every vertex associated with the $p$-th clause should be visited by one robot.
 {Let $$z_{q} = \sum_{v \in \ccalM^\mathsf{lits}_{\ccalV}(e,0|1,p,q)} \sum_{u: (u,v) \in \ccalE_\ccalG} \sum_{r\in \ccalM^\ccalV_{\ccalK}(v)} x_{uvr}$$ be the inner summation in~\eqref{eq:6}. If $b_p=1$, then all literals in $\mathsf{lits}^+(\ccalC_p^{\gamma})$ are true. In this case, for the $q$-th literal $\ap{i^q}{j^q}{k^q,\chi^q} \in \mathsf{lits}^+(\ccalC_p^\gamma)$, all $i^q$ vertices in ${\ccalM^\mathsf{lits}_{\ccalV}(e, 0|1, p, q)}$ should be visited, so $\sum_{u: (u,v) \in \ccalE_\ccalG} \sum_{r\in \ccalM^\ccalV_{\ccalK}(v)} x_{uvr} = 1$ for each vertex $v \in {\ccalM^\mathsf{lits}_{\ccalV}(e,0|1,p,q)}$, and therefore $z_{q}= i^q$, and the left side of constraint~\eqref{eq:6} becomes
   \begin{align*}
     \left. \sum_{q\in \ccalQ_p} z_q \middle/ {\sum_{q\in \ccalQ_p} i^q} = {\sum_{q\in \ccalQ_p} i^q}\middle/{\sum_{q\in \ccalQ_p} i^q} \right. = 1 = b_p.
   \end{align*}
 If $b_p=0$, all $x_{uvr}$ in constraint~\eqref{eq:6} equal 0, which implies that no vertices need to be visited for false clauses. Combining constraints~\eqref{eq:3} and~\eqref{eq:6}, $\sum_{r\in \ccalM^\ccalV_{\ccalK}(v)} t^-_{vr}$ equals 0 if the  clause that $v$ is associated with is false. That is, a robot remains idle if it is not responsible for the satisfaction of any clause. 

     Note that the logical relation in constraint~\eqref{eq:6} only requires  that some vertices should be visited at some point in time  to satisfy all literals. Next, we formulate the synchronization constraint requiring that, if $\gamma$ is an edge label and  the $p$-th clause $\ccalC_p^\gamma$ is true, all vertices in~$\ccalM^\mathsf{cls}_\ccalV(e, 1,p)$ should be visited at the same time since the satisfaction of edge labels is instantaneous.
       We define the pairwise vertex set induced from the clause $\ccalC_p^\gamma$ as  $\ccalV_{\gamma, p}^{\text{sync}} = \{(u,v)\,|\, u, v\in \ccalM^\mathsf{cls}_\ccalV(e,1,p), u\not= v \}$. If $b_p=1$, visiting any pair in $\ccalV_{\gamma, p}^{\text{sync}}$ simultaneously is written as
 \begin{align}
   & \quad\quad  \sum_{r\in \ccalM^\ccalV_{\ccalK}(u)} t_{ur}^-  = \sum_{r\in \ccalM^\ccalV_{\ccalK}(v)} t_{vr}^- , \quad\forall (u,v) \in \ccalV_{\gamma, p}^{\text{sync}}.\label{eq:7}
 \end{align}
 When $b_p=0$, constraint~\eqref{eq:7} also holds since both sides equal 0.

 \subsubsection{Temporal constraints}\label{sec:temporal} These constraints capture the temporal orders between subtasks. We first introduce the notions of the activation and completion time of a subtask. Then, given a subtask $e$, there are three types of temporal constraints associated with the activation and completion times (see Definition~\ref{defn:time} below),  for the subtask $e$ or between subtasks.
 \begin{defn}[Activation and completion time of a subtask or its starting vertex label]\label{defn:time}
 {Given a subtask $e = (v_1, v_2)$, we define its  activation time  (equivalently, the activation time  of its starting  vertex label) as the time instant when  its vertex label $\gamma(v_1)$ becomes  true. Similarly, we define  the completion time  of a subtask (equivalently, the completion time of its starting vertex label) as the time instant when  its edge label $\gamma(v_1, v_2)$ becomes true (or the last time its starting vertex label $\gamma(v_1)$ is true). The span of a subtask (or its starting vertex label) is the time interval stating  from the activation time and ending  at the completion time.}
 \end{defn}

 \paragraph{Temporal constraints associated with one subtask} {These constraints capture the relation that the completion time of a subtask should lie in the span of its starting vertex label, or exactly one time step after  the completion of its starting vertex label. Intuitively, the ``avoid'' part of a subtask  should be maintained until the ``reach'' part is realized.} \label{sec:onesubtask}

 For this, we define the auxiliary variable  $t_e$ to denote the completion time of the subtask $e$, i.e., time when its edge label  becomes true. We have
 \begin{align}\label{eq:edgetime}
   t_e = \sum_{p\in \ccalP} \sum_{r\in \ccalM^\ccalV_{\ccalK}(v_p)}t_{v_p r}^-,
 \end{align}
 where $v_p$ is randomly selected from $\ccalM^\mathsf{cls}_\ccalV(e,1,p)$ due to constraints~\eqref{eq:c} and~\eqref{eq:7} that require that  only one clause of an edge label is true and all associated vertices are visited at the same time.

 When the starting vertex $v_1$ has a self-loop, and its label $\gamma(v_1)$ is not  $\top$ ({if this is not the case, there are no vertices in $\ccalG$ associated with $\gamma(v_1)$}), the temporal relation, $\forall\, \ccalC_p^{\gamma} \in \clause{\gamma(v_1)}$ and $\forall v \in \ccalM^\mathsf{cls}_\ccalV(e,0,p)$, can be written as
   \begin{align}
  \sum_{r\in \ccalM^\ccalV_{\ccalK}(v)} t_{vr}^-   \leq  t_e \leq
     \sum_{r\in \ccalM^\ccalV_{\ccalK}(v)} t_{vr}^+ + 1  + M_{\text{max}} (1 - b_{p}). \label{eq:17}
   \end{align}
   If $b_p=0$, then by constraint~\eqref{eq:6} no robot visits vertex $v$, so~$ \sum_{r\in \ccalM^\ccalV_{\ccalK}(v)} t_{vr}^- = 0 \leq t_e$, i.e., the left inequality in~\eqref{eq:17} holds. The right inequality in~\eqref{eq:17} holds trivially. Only when $b_p=1$, i.e., when the $p$-th clause in the vertex label is true, does  constraint~\eqref{eq:17} become active. {Note that constraint~\eqref{eq:17} implies that the span of a subtask is not necessarily equal to the span of its starting vertex label.}  On the other hand, when the starting vertex $v_1$ does not have a self-loop,  $v_1$ is identical to the initial vertex $v_0$. Recall that in Section~\ref{sec:prune} we remove all vertices without self-loops except for the  initial and accepting vertices. Hence, in $\auto{subtask}^-$, only $v_0$ and $\vertex{accept}$ are allowed not to have self-loops but $\vertex{accept}$ cannot be the starting vertex. Therefore, $v_1 = v_0$. If $\gamma(v_0) = \bot$,  {constraint~\eqref{eq:17} implies  that the edge label of subtask $e$ should be satisfied at time instant 0}, i.e.,
   \begin{align}
  t_e = 0, \quad \text{if}\; \gamma ({v_0}) = \bot. \label{eq:tis0}
   \end{align}

 \paragraph{Temporal constraints associated with the completion of two sequential subtasks}{
 These constraints impose the precedence relation that subsequent subtasks should be completed after prior subtasks are completed.   Given the current subtask $e$, we collect its prior subtasks in the set $X^e_{\prec_{P}}$  rather than a larger set $X^e_{<_{P}}$~{(defined in Appendix~\ref{sec:b})}. That is, we consider  subtasks that are covered by $e$  due to the transitivity property of the partial order.} \label{sec:constraintonedge}
 If $X^e_{\prec_{P}}$ is nonempty, we iterate over subtasks in it.  Given $e' \in X^e_{\prec_{P}}$,  we can capture the requirement that the subtask $e'$ is completed before the current subtask $e$ by the constraint
 \begin{align}\label{eq:12}
   t_{e'} +  1  \leq t_e, \; \forall \, e' \in X_{\prec P}^e,
 \end{align}
 where the term 1 excludes the case where two edge labels become true simultaneously, violating the precedence relation.

 \paragraph{Temporal constraints associated with   the  completion of the current subtask and the activation  of the  subtask immediately following it}{}
 These constraints capture the precedence relation that the   current subtask $e$ should be completed at most one time step  before  the subtask immediately following it is activated. Otherwise, progress in the sub-NBA induced from the poset $P$ will be trapped at subtask  $e$ if there is no  subtask  immediately after it is activated. To capture this requirement, we define $|X_P|\cdot |X_P -1|$ auxiliary binary variables $b_{ee'}$ for any two different subtasks  $e, e' \in X_P$,  such that $b_{ee'}=1$  if subtask $e'$  occurs immediately after subtask $e$. Furthermore, we define the set   $S_3^e = X^e_{\succ_{P}} \cup X^e_{\|_{P}}$ that collects all subtasks whose activation can immediately follow the completion of subtask $e$.  In what follows, we proceed based on whether $ X^e_{\succ_{P}} \neq \emptyset$.

 \mysubparagraph{$ X^e_{\succ_{P}} \neq \emptyset$}{In this case, there must exist a subtask that occurs after $e$. Then, the  constraint that there exists a subtask in $S_3^e$ that occurs immediately after $e$ can be written as}\label{activation:a}
 \begin{align}\label{eq:bafter}
  \sum_{e'\in  S_3^e} b_{ee'} = 1.
 \end{align}
 If the subtask $e'$ indeed occurs immediately  after subtask $e$, then it should be completed after subtask $e$, that is,
 \begin{align}\label{eq:after}
   t_e + 1 \leq  t_{e'} + M_{\text{max}} (1 - b_{ee'}),\; \forall\, e' \in S_3^e.
 \end{align}
 To establish the transition between subtasks, the subtask $e'$ that occurs immediately  after subtask $e$ should be activated at most one time step after the completion of $e$. That is, $\forall\, e' = (v_1', v_2') \in S_3^e, \forall \, \ccalC_p^{\gamma(v'_1)} \in \clause{\gamma(v'_1)} , \forall \,v \in \ccalM^\mathsf{cls}_\ccalV(e',0,p)$, we have
 \begin{align}\label{eq:20}
   & \sum_{r\in \ccalM^\ccalV_{\ccalK}(v)} t_{vr}^-   \leq t_{e}  + 1 + M_{\text{max}} (1 - b_{ee'}).
 \end{align}
 If the $p$-th clause in the vertex label $\gamma(v_1')$ is true, constraint~\eqref{eq:20} requires that the associated vertices are visited at most one time step after the completion of $e$. Otherwise if the $p$-th clause is false, the left side of~\eqref{eq:20} becomes 0 and the constraint~\eqref{eq:20} holds trivially. {If $\gamma(v_1')=\top$, the  subtask $e'$ can be viewed as being activated at time instant 0.} Thus, constraint~\eqref{eq:20} is satisfied trivially.

 \mysubparagraph{$ X^e_{\succ_{P}} = \emptyset$}{In this case, if subtask $e$ is completed after the  subtasks in  $X_{\|_P}^e$, then it is the last  subtask  to be completed in  $X_P$. Thus, there is no subtask to be activated any more. Otherwise, if subtask $e$ is not the last subtask, then there exists a subtask that occurs after $e$, same as in case~\ref{activation:a}. To determine whether subtask $e$ is the last subtask, we define $|X_P|\cdot |X_P -1|$ auxiliary binary variables $b_{e}^{e'}$ for any two different subtasks  $e, e' \in X_P$,  such that $b_{e}^{e'}=1$  if and only if  $t_{e} > t_{e'}$, i.e., if and only if subtask $e$ is completed after $e'$. This implication can be written as, $\forall\, e, e' \in X_P $ and $e\neq e'$,}\label{activation:b}
 \begin{subequations}\label{eq:diff}
   \begin{align}
     b_e^{e'} + b_{e'}^e &  = 1, \label{eq:diff_a}\\
   M_{\text{max}} (b_{e}^{e'} - 1) \leq t_e -  t_{e'} & \leq M_{\text{max}} b_{e}^{e'} - 1. \label{eq:diff_b}
 \end{align}
 \end{subequations}
 Constraints~\eqref{eq:diff} require  that  no two subtasks are completed at the same time and that $b_e^{e'}=1$ if and only if $t_e > t_{e'}$. Assume $t_e = t_{e'}$. From constraint~\eqref{eq:diff_b}, we get $b_e^{e'}=b_{e'}^{e}=1$, which violates constraint~\eqref{eq:diff_a}. {Although independent subtasks can occur simultaneously, constraint~\eqref{eq:diff} requires that they occur serially so that the solution to the MILP gives rise to a simple path in $\auto{subtask}^-$ that is a linear extension of the poset $P$.} When $t_e > t_{e'}$, the right side of constraint~\eqref{eq:diff_b} implies $b_e^{e'}=1$; when $t_e < t_{e'}$, the left side of constraint~\eqref{eq:diff_b} implies $b_{e}^{e'}=0$.


  Furthermore, we define $z = |X^e_{\|_{P}}|$. Observe that, for $e' \in X^e_{\|_{P}} $, the term $z -  \sum_{e' \in X^e_{\|_{P}}  } b_{e}^{e'} = 0$ if $e$ is the last completed task; otherwise it is   positive. If subtask $e$ is not the last subtask, there should be a subtask in $X^e_{\|_{P}}$ that occurs immediately after $e$. This requirement can be written as
 \begin{subequations}\label{eq:afterparallel}
   \begin{align}
   \sum_{e' \in X_{\|_P}^e} b_{ee'} & \le 1,\label{eq:afterparallel_a}\\
   z - \sum_{e' \in X^e_{\|_{P}}  } b_{e}^{e'} - M_{\text{max}}  \sum_{e' \in X_{\|_P}^e} b_{ee'} &  \leq 0, \label{eq:afterparallel_b} \\
   \sum_{e' \in X_{\|_P}^e} b_{ee'} - M_{\text{max}}(z -  \sum_{e' \in X^e_{\|_{P}}  } b_{e}^{e'}) & \leq0 . \label{eq:afterparallel_c}
 \end{align}
 \end{subequations}
 {If subtask $e$ is completed after all subtasks in  $X_{\|_P}^e$}, then $z - \sum_{e' \in X^e_{\|_{P}}  } b_{e}^{e'} = 0$, and constraint~\eqref{eq:afterparallel_c} gives $ \sum_{e' \in X_{\|_P}^e} b_{ee'}=0$, i.e., there is no subtask that follows  $e$ immediately. Constraint~\eqref{eq:afterparallel_a} becomes $0\le 1$ and~\eqref{eq:afterparallel_b} becomes $0\le 0$. Both hold trivially.
 Otherwise, if subtask is not the last subtask, i.e., if $z - \sum_{e' \in X^e_{\|_{P}}  } b_{e}^{e'} > 0$, then constraints~\eqref{eq:afterparallel_a} and~\eqref{eq:afterparallel_b} give $ \sum_{e' \in X_{\|_P}^e} b_{ee'}=1$. Constraint~\eqref{eq:afterparallel_c} holds trivially. Finally, after determining the subtask $e'$ that occurs immediately after $e$, we impose the same constraints as~\eqref{eq:after} and~\eqref{eq:20}.

 Note that constraints~\eqref{eq:bafter}-\eqref{eq:afterparallel} in \ref{activation:a} and \ref{activation:b} ensure that, for a subtask in $X_P$, except for the last one,  there exists another subtask that immediately follows it. However, it is possible that two different subtasks are followed by the same subtask, which cannot be excluded by constraint~\eqref{eq:bafter}. To avoid this situation, next we impose the constraint that except for the  first subtask  to be completed in $X_P$, each subtask  can only immediately follow one subtask. Combined with constraints~\eqref{eq:bafter}-\eqref{eq:afterparallel}, we guarantee the one-to-one correspondence between any two consecutive subtasks in a linear extension. Recall that $S_2^e = X^e_{\prec_{P}} \cup X^e_{\|_{P}}$. We proceed based on whether $X^e_{\prec_{P}} = \emptyset$ or not.

 \mysubparagraph{$X^e_{\prec_{P}} \neq \emptyset$}{In this case,  subtask $e$ cannot be the first subtask to be completed, that is, it has to immediately follow one subtask in $S_2^e$. This requirement is captured by the constraint }
   \begin{align}\label{eq:follow1}
  \sum_{e' \in S_2^e} b_{e'e} = 1.
   \end{align}

  \mysubparagraph{$X^e_{\prec_{P}}= \emptyset$}{In this case, $X^e_{<_{P}} = \emptyset$,  so there is no subtask prior to $e$. Recall that the binary variable $b_e^{e'}=1$ if subtask $e$ is completed after $e'$ and no two subtasks are completed at the same time. Therefore, $b_e^{e'}=0$ if $e$ is completed prior to $e'$, and further  the term $\sum_{e' \in X_{\|_P}^e } {b}_{e}^{e'}=0$ if $e$ is the first completed task; otherwise it is   positive. Then, the constraint that each subtask in $X_{\|_P}^e$, except the first one,  immediately follows another subtask can be written as}
 \begin{subequations}\label{eq:follow}
   \begin{align}
   \sum_{e' \in X_{\|_P}^e} b_{e'e} & \le 1,\label{eq:follow_a}\\
  \sum_{e' \in X^e_{\|_{P}}  } b_{e}^{e'} - M_{\text{max}}  \sum_{e' \in X_{\|_P}^e} b_{e'e} &  \leq 0, \label{eq:follow_b} \\
   \sum_{e' \in X_{\|_P}^e} b_{e'e} - M_{\text{max}} \sum_{e' \in X^e_{\|_{P}}  } b_{e}^{e'} & \leq0 . \label{eq:follow_c}
 \end{align}
 \end{subequations}
 When $e$ is the first subtask, i.e., when  $\sum_{e' \in X^e_{\|_{P}}  } b_{e}^{e'}=0$, then constraints~\eqref{eq:follow_b} and~\eqref{eq:follow_c} give  $\sum_{e' \in X_{\|_P}^e} b_{e'e} = 0$. Otherwise, if $e$ is not the first subtask, i.e., if $\sum_{e' \in X^e_{\|_{P}}  } b_{e}^{e'} > 0$, then constraints~\eqref{eq:follow_a} and~\eqref{eq:follow_b} give $\sum_{e' \in X_{\|_P}^e} b_{e'e} = 1$; same as~\eqref{eq:follow1}.

 \paragraph{Temporal constraints associated with the activation of the first subtask}{}
 We analyzed the temporal constraints on the completion of the current subtask and the activation of subsequent subtasks above. However, if the subtask $e$ is the first subtask to be completed, there is no subtask whose completion activates $e$, which should be activated at the beginning. To determine the first subtask in $X_P$,  let $P_{\text{max}}$ be the set that collects subtasks that can be the first ones to be completed, which are referred to as the maximal elements in a poset $P$. An element in a poset $P$ is a maximal element if there is no larger element in $P$ than itself. That is, for  any subtask  $e \in P_{\text{max}}$, we have $X_{\prec_P}^e = \emptyset$. If the first completed subtask $e = (v_1 ,v_2)$ has a self-loop and the vertex label is not $\top$ (it is activated at the beginning if $\gamma(v_1)=\top$), we require that the vertex label $\gamma(v_1)$  be activated at time 0, which implies that the associated vertices in $\ccalG$ should be visited at time 0,   i.e., $\forall\, e \in P_{\text{max}}, \forall\, \ccalC_p^{\gamma(v_1)} \in \clause{\gamma(v_1)}, \forall \,v \in \ccalM^\mathsf{cls}_\ccalV(e,0,p)$,
 \begin{align}\label{eq:zeroactivation}
   & \sum_{r\in \ccalM^\ccalV_{\ccalK}(v)} t_{vr}^-  \leq M_{\text{max}} (\sum_{e' \in P_{\text{max}}\setminus \{e\}  } {b}_{e}^{e'} + 1- b_p).
 \end{align}
 Only when $e$ is the first subtask to be completed, i.e, when $\sum_{e' \in P_{\text{max}}\setminus \{e\}  } {b}_{e}^{e'}=0$ and when the associated clause is true, i.e., when $b_p=1$, {should the vertices associated with the $p$-th clause} be visited by robots at time 0, i.e., $\sum_{r\in \ccalM^\ccalV_{\ccalK}(v)} t_{vr}^-  \leq 0$. When $|P_{\text{max}}|=1$, constraint~\eqref{eq:zeroactivation} will be reduced to  $ \sum_{r\in \ccalM^\ccalV_{\ccalK}(v)} t_{vr}^-  \leq M_{\text{max}} (1- b_p).$ Note that if the vertex label $\gamma(v_1)$ has no self-loop, then $v_1$ is identical to $v_0$. We have discussed  this case in constraint~\eqref{eq:tis0}.

 Recall that in cases~\ref{edge:vertex2} and~\ref{edge:vertex3} in Appendix~\ref{sec:c} when constructing the edges for vertex labels of subtasks in $P_{\text{max}}$ (a subtask $e$ is in $P_{\text{max}}$ if $X_{\prec_P}^e=\emptyset$), their leaving vertices fall into two categories,  location vertices and literal vertices associated with immediately preceding subtasks. To satisfy condition~\hyperref[asmp:b]{(b)}  in Definition~\ref{defn:same} that the satisfied clause in the edge label implies the satisfied clause in the end vertex label and the same fleet of robots satisfy these two clauses, we require that the starting vertex label of the first completed  subtask in $P_{\text{max}}$ should be satisfied by robots coming from location vertices in $\ccalV_{\text{init}}$, and the starting vertex label of the remaining subtasks in $P_{\text{max}}$ should be satisfied by robots coming from literal  vertices  associated with edge labels of immediately prior subtasks. To this end, we first define an auxiliary binary variable $b_{e}^{\prec}$ such that $b_{e}^{\prec}= 1$ if and only if subtask $e$ is the first subtask in $P_{\text{max}}$. Then, we define the following constraints
 \begin{subequations}\label{eq:first_subtask}
   \begin{align}
    \sum_{e' \in P_{\text{max}}\setminus \{e\}} {b}_e^{e'}  - M_{\text{max}} (1 - b_{e}^{\prec}) & \leq 0\\
     1 - b_{e}^{\prec} -  M_{\text{max}}  \sum_{e' \in P_{\text{max}}\setminus \{e\}} {b}_e^{e'}    & \leq 0.
   \end{align}
 \end{subequations}
 Only when $e$ is the first subtask, i.e., when $ \sum_{e' \in P_{\text{max}}\setminus \{e\}} {b}_e^{e'}=0$, does  constraint~\eqref{eq:first_subtask} give $b_e^{\prec}=1$. {Then, for any clause in the starting vertex label of $e$, the constraints specifying which categories of leaving vertices robots should come from can be written as,} $\forall \, \ccalC_p^\gamma \in \clause{\gamma} $,
 \begin{subequations}\label{eq:routingforactivation}
   \begin{align}
    & \sum_{q\in \ccalQ_p} \sum_{v \in \ccalM^\mathsf{lits}_{\ccalV}(e,0,p,q)} \sum_{\substack{u: (u,v) \in \ccalE_\ccalG \\ u \not\in \ccalV_{\text{init}}}}   \sum_{r\in \ccalM^\ccalV_{\ccalK}(v)} x_{uvr} \leq M_{\text{max}} (1 - b_e^{\prec}),  \label{eq:routingforactivation_a}\\
    & \sum_{q\in \ccalQ_p} \sum_{v \in \ccalM^\mathsf{lits}_{\ccalV}(e,0,p,q)} \sum_{\substack{u: (u,v) \in \ccalE_\ccalG \\ u \in \ccalV_{\text{init}}}}   \sum_{r\in \ccalM^\ccalV_{\ccalK}(v)} x_{uvr} \leq M_{\text{max}}  b_e^{\prec}. \label{eq:routingforactivation_b}
   \end{align}
 \end{subequations}
 When subtask $e$ is the first one to be completed in $P_{\text{max}}$, i.e., when $b_e^{\prec}=1$, then constraint~\eqref{eq:routingforactivation_a}, combined with constraint~\eqref{eq:6}, states that robots should come from location vertices in~$\ccalV_{\text{init}}$. However, when subtask $e$ is not  the first subtask to be completed in $P_{\text{max}}$, i.e., when $b_e^\prec=0$, then constraint~\eqref{eq:routingforactivation_b} requires that robots should come from  literal vertices associated with immediately prior subtasks.

 \subsubsection{Same-$\ag{i}{j}$ constraints}\label{sec:samegroup}
 Next, we encode the constraint that some subtasks are executed by the same $i$ robots of type $j$, which are indicated by the same nonzero connector $\chi$. Given a nonzero connector $\chi$, we can identify all vertex or edge labels that have literals with the same connector $\chi$  by the mapping  $\ccalM_\mathsf{\gamma}^\chi(\chi)$. In an edge label $(e, 1)  \in  \ccalM_\mathsf{\gamma}^\chi(\chi)$ or a vertex label $(e, 0)  \in  \ccalM_\mathsf{\gamma}^\chi(\chi)$, each clause has at most one literal $\aap{i}{j}{k}{\chi}$ with connector $\chi$ and it is associated with $i$ vertices in $\ccalG$. We enumerate these $i$ vertices and denote the $b$-th vertex by $v_k^b$. Then for any two labels $\gamma, \gamma'\in\ccalM_\mathsf{\gamma}^a(\chi)$ and any two clauses $\ccalC_p^\gamma \in \clause{\gamma}$ and $\ccalC_{p'}^{\gamma'} \in \clause{{\gamma'}}$ that have  literals $\aap{i}{j}{k}{\chi}$ and $\aap{i}{j}{k'}{\chi}$, respectively, {the constraint} that the corresponding literals are satisfied by the same $i$ robots of type $j$, $\forall \,b \in [i]$ and  $\forall \, r \in \ccalK_j$, can be written as
 \begingroup
 \allowdisplaybreaks
 \begin{subequations}\label{eq:same}
   \begin{align}
     & \sum_{u:(u, v_k^b) \in \ccalE_\ccalG} x_{u v_k^b r} + M_{\text{max}} (b_p - 1) \nonumber \\& \quad\quad\quad \leq  \sum_{u:(u, v_{k'}^b) \in \ccalE_\ccalG} x_{u v_{k'}^b r} + M_{\text{max}}(1 - b_{p'}),\\
     & \sum_{u:(u, v_{k'}^b) \in \ccalE_\ccalG} x_{u v_{k'}^b r} + M_{\text{max}} (b_{p'} - 1) \nonumber \\& \quad\quad\quad \leq \sum_{u:(u, v_{k}^b) \in \ccalE_\ccalG} x_{u v_{k}^b r} + M_{\text{max}}(1 - b_{p}),
   \end{align}
 \end{subequations}
 \endgroup
 where $v_{k'}^b$ is the $b$-th vertex associated with $\aap{i}{j}{k'}{\chi}$. Only when $b_p = b_{p'}=1$, does~\eqref{eq:same} become active. Then, $\sum\nolimits_{u:(u, v_k^b) \in \ccalE_\ccalG} x_{u v_k^b r} = \sum\nolimits_{u:(u, v_{k'}^b) \in \ccalE_\ccalG} x_{u v_{k'}^b r} $, i.e., two $b$-th vertices $v_k^b$ and $v_{k'}^b$ are visited by the same robot $r$.

 \subsubsection{Constraints associated with the transition between the prefix and suffix parts}\label{sec:transition} Since we synthesize plans for the  prefix and suffix parts separately, to ensure that the final locations of the prefix part seamlessly transition to the suffix part, we impose constraints on the final locations of the prefix part, which are determined by the satisfied clause in the edge label of the  subtask that is the last one to be completed.

 To this end,  we first find the set of subtasks, denoted by $P_{\text{min}}$, in the poset $P$ that can be the last ones to be completed, which are referred to as the minimal elements in a poset. An element in a poset $P$  is a minimal element if there is no smaller element in $P$ than itself. Then, we iterate over subtasks in $P_{\text{min}}$ when formulating the MILP each time selecting a different  subtask $e\in P_{\text{min}}$ to be the last one, which can be written as
 \begin{align}\label{eq:lastsubtask0}
   b_e^{e'} = 1, \;  \forall\, e' \in X_P \setminus\{e\}.
 \end{align}
 After selecting the last subtask to be completed, we next select one clause in its edge label $\gamma$ that needs to be satisfied. We iterate over all clauses in the edge label of the last subtask each time selecting a different clause $\ccalC_p^\gamma \in \clause{\gamma}$ to be true, i.e.,
 \begin{align}\label{eq:lastclause}
   b_p = 1.
 \end{align}
 If a path for Problem~\ref{prob:1} cannot be detected when  $e$ is the last subtask in the prefix part  and its $p$-th clause is set to be true,  then we continue iterating  over clauses in the edge label of $e$. If a path for Problem~\ref{prob:1} cannot be detected after iterating over clauses in the selected edge label, we select another subtask in $P_{\text{min}}$ to be the last one and repeat the same process.

 \begin{rem}
   {Assuming there is a feasible path  $\tau$ to Problem~\ref{prob:1}, the  constraints~\eqref{eq:lastsubtask0}-\eqref{eq:lastclause} on the final locations of the prefix part allow us to identify the same clause satisfied by the final locations of the prefix part as that satisfied by the assumed feasible path $\tau$. This ensures the feasibility of  the suffix part and the completeness of our method; see also Theorem~\ref{thm:completeness}.} Note that the constraints~\eqref{eq:lastsubtask0}-\eqref{eq:lastclause} are necessary for establishing the completeness of our proposed method. However, there may exist multiple solutions to Problem~\ref{prob:1}, and it will be computationally inefficient to try all possibilities for the last subtasks and the corresponding clauses.  We found that often in practice, omitting constraints~\eqref{eq:lastsubtask0}-\eqref{eq:lastclause}  did not make Problem~\ref{prob:1} infeasible. Therefore, constraints~\eqref{eq:lastsubtask0}-\eqref{eq:lastclause} can be initially omitted from the formulation of the MILP.
 \end{rem}

 \subsubsection{MILP objective}\label{sec:objective} The objective is to minimize the weighted sum of the travel cost and travel time, i.e.,
 \begin{align}\label{equ:obj}
   \min \; \;&  \alpha \sum_{(u,v)\in \ccalE_\ccalG}  \sum_{r \in \ccalM^\ccalV_{\ccalK}(v)} d_{uv} x_{uvr} +  (1 - \alpha) \sum_{e\in X_P} t_e,
 \end{align}
 where $\alpha$ is a user-specified parameter and  $d_{uv}$ is the travel cost, e.g., travel distance.
  Compared to Problem~\ref{prob:1},  objective~\eqref{equ:obj} involves  optimization of time. In practice, we observed that without optimizing  time, some scheduling variables can take large values, which impacts the generation of the low-level path. Note that travel cost, e.g., travel distance, and travel time are typically non-conflicting objectives.
                 }}

 \subsection{Construction of the robot suffix path}\label{sec:suf}
 In this section, we construct the robot path for the suffix part. We assume that the high-level plan found in Section~\ref{sec:path} for the prefix part has been used to generate low-level paths as in Appendix~\ref{sec:solution2mrta}, which induces a run in $\autop$ connecting $v_0$ and $\vertex{accept}$. Thus, the final robot locations of the prefix part are known. In what follows, we proceed depending on whether the vertex $\vertex{accept}$ has a self-loop or not.
 If the  vertex $\vertex{accept}$ has a self-loop, we first examine whether the final locations of the prefix part satisfy its label $\gamma_\phi(\vertex{accept})$. {If yes, we conclude that the prefix path we have found also  satisfies the specification $\phi$.} Otherwise, we remove this self-loop since it does not contribute to the identification of the suffix paths. By treating the final locations of the prefix paths as the initial robot locations of the suffix paths, the suffix paths aim to drive the progress in~$\autop$ back to vertex $\vertex{accept}$ and send robots to the initial locations of the suffix part to close the trajectories.
 The basic idea is to  view the simple cycle around $\vertex{accept}$ as a simple path, by treating the accepting vertex at the beginning of this simple path as the initial vertex $v_0$ and the other accepting vertex at the end as the goal to be reached. Then, starting from the NBA $\autop^-$ in Section~\ref{sec:prune}, we can follow a procedure similar to the prefix part to obtain paths for the suffix part.

 \begin{figure}[!t]
     \centering
     \includegraphics[width=0.4\linewidth]{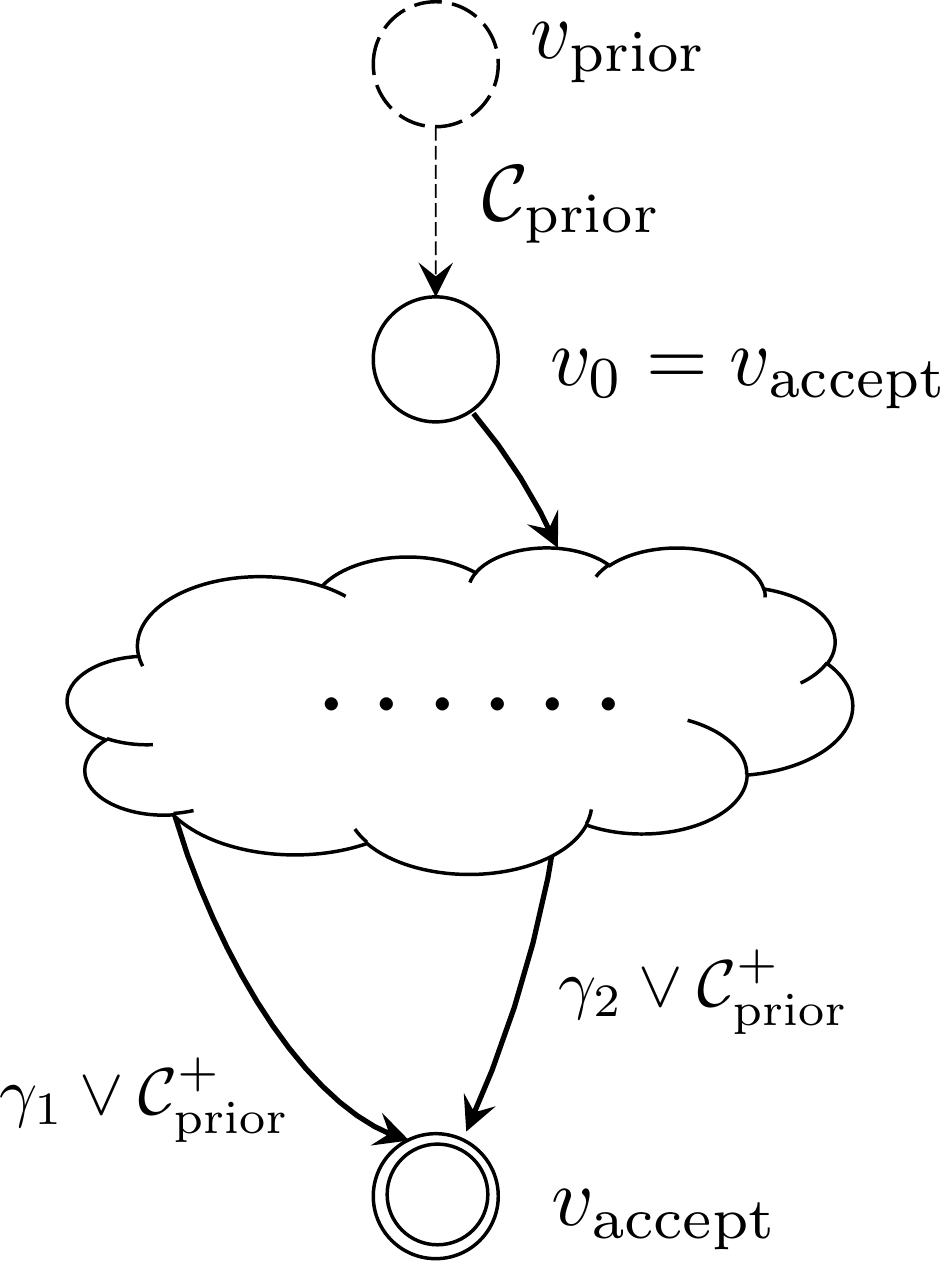}
     \caption{$\auto{subtask}^-$ for the suffix part when $\vertex{accept}$ does not have a self-loop, where $\gamma_1$, $\gamma_2$ and $\gamma_3$ are edge labels and $\ccalC_{\text{prior}}^+$ is positive subformula that is satisfied by final robot locations of the prefix part; see Appendix~\ref{sec:suf_milp}.}
     \label{fig:suffix}
 \end{figure}

  \begin{figure}
     \centering
     \includegraphics[width=0.5\linewidth]{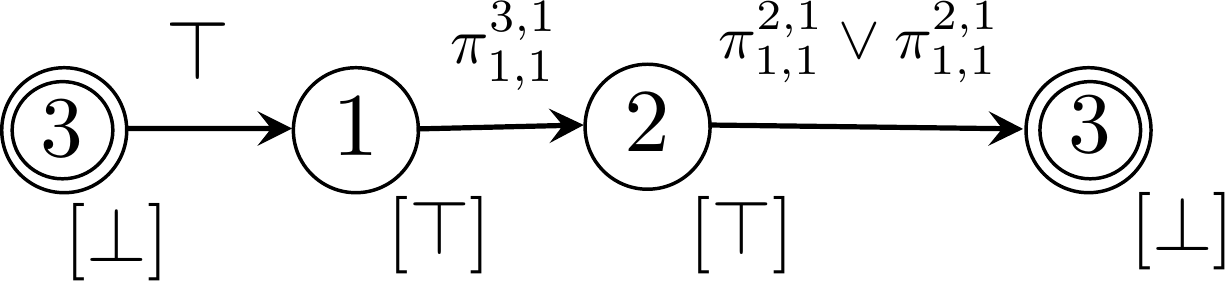}
     \caption{$\auto{subtask}^-$ for task (ii) obtained from Fig.~\ref{fig:nba_ii_prune}. A clause $\ap{1}{1}{2,1}$ is added to the edge label $\gamma(v_2, v_3)$; see Appendix~\ref{sec:suf_milp}.} \label{fig:suffix_ii}
   \end{figure}

  \subsubsection{Extracting subtasks and inferring the temporal order from the NBA}\label{sec:suf_prune}~\\
   \paragraph{Extraction of sub-NBA $\auto{subtask}$ from $\auto{relax}$}{}\label{sec:suf_extract} First, based on the NBA $\autop^-$ we obtain the relaxed NBA $\auto{relax}$, as in Section~\ref{sec:prune}. Then, to obtain the sub-NBA $\auto{subtask}$, similar to finding the shortest simple cycle around $\vertex{accept}$ in Section \ref{sec:shortestcycle}, we remove all other accepting vertices from $\auto{relax}$ and all initial vertices if they do not have self-loops. Let $\gamma_{\phi}(\vertex{prior}, \vertex{accept})$ denote the edge label corresponding to the last completed subtask in the prefix part. After generating the low-level paths for the prefix part, the final robot locations of the prefix part satisfy $\gamma_{\phi}(\vertex{prior}, \vertex{accept})$ (see Fig.~\ref{fig:lasso}).
   Next, since $\vertex{accept}$ does not have a self-loop,  we remove all outgoing edges from $\vertex{accept}$ (acting as $v_0$) from $\auto{relax}$ if $\gamma_{\phi}(\vertex{prior}, \vertex{accept})$ does not imply its edge label in $\autop$. Also, we remove all incoming edges to $\vertex{accept}$ (acting as $\vertex{accept}$) from $\auto{relax}$ if the corresponding edge label in $\autop$ is not implied by $\gamma_\phi(\vertex{prior}, \vertex{accept})$. By condition \ref{cond:f} in Definition~\ref{defn:run}, these edges do not appear in any restricted accepting run if the corresponding prefix part traverses the edge $(v_{\text{prior}}, \vertex{accept})$. Also, if the set of restricted accepting runs is nonempty, there exist accepting vertices in $\auto{relax}$ that have  outgoing edges and incoming edges for which this implication holds.  Note that the implication check is   conducted in $\autop$.
 Finally, we  follow a similar process as that described in Section~\ref{sub-NBA:1} to  extract a sub-NBA $\auto{subtask}$ from $\auto{relax}$ for the pair $\vertex{accept}$ (acting as $v_0$) and $\vertex{accept}$. The structure of $\auto{subtask}$ is shown in Fig.~\ref{fig:suffix} where $v_0$ does not have a self-loop. We also depict the vertex $\vertex{prior}$ in the prefix part for better understanding.  Then, we prune $\auto{subtask}$ to obtain the NBA $\auto{subtask}^-$, as in Section~\ref{sub-NBA:2}.

   \begin{cexmp}{exmp:1}{$\auto{subtask}^-$ for the suffix part}
 The suffix part for task~\hyperref[task:i]{(i)} only consists of the accepting vertex $\vertex{accept}$. The sub-NBA $\auto{subtask}^-$ associated with the suffix part for task~\hyperref[task:ii]{(ii)} is a cycle $v_3, v_1, v_2, v_3$; see Fig.~\ref{fig:suffix_ii}.
   \end{cexmp}
   \paragraph{Inferring temporal order between subtasks in $\auto{subtask}^-$}{} We collect all simple cycles around $\vertex{accept}$ in the set $\Theta$. Because the initial robot locations for the suffix part satisfy $\gamma_{\phi}(\vertex{prior}, \vertex{accept})$, they also satisfy  the label of the last edge in every simple cycle $\theta\in \Theta$ since  edges in $\auto{relax}$ whose labels are not implied by $\gamma_\phi(\vertex{prior}, \vertex{accept})$ are removed from $\auto{relax}$  when constructing $\auto{subtask}$.
   By following such a simple cycle $\theta$, not only the transition is driven back to $\vertex{accept}$, but further  robots are able to return to their initial locations to close the trajectories. 
  Finally, we infer a set of posets $\{P_\text{suf}\}$ from simple cycles in $\Theta$ and sort them as Section~\ref{sec:poset}.

  \subsubsection{Finding the suffix path  on $\auto{subtask}^-$}\label{sec:suf_milp} Similar to the prefix part, we find the suffix path that, combined with the prefix path, satisfies the specification by iterating over the set of posets $\{P_\text{suf}\}$. By condition~\hyperref[asmp:c]{(c)} in Definition~\ref{defn:same}, robots need to return to their initial locations to close the suffix loop and  drive the transition in $\autop$ back to $\vertex{accept}$. To ensure that our method is complete,  we achieve these two goals separately. First, those robots participating in the satisfaction of the positive subformula in $\gamma_\phi(\vertex{prior}, \vertex{next})$ return to regions corresponding to their initial locations of the suffix part (not necessarily returning to their initial locations), and at the same time drive the transition in $\ccalA_{\phi}$ back to $\vertex{accept}$. Then, all robots return to their initial locations of the suffix part while not violating the specification $\phi$. We also discuss how to achieve these two goals above at the same time in Appendix~\ref{app:loop}.

 To achieve the first step, we  find the satisfied clause, denoted by $\ccalC_{\text{prior}}$, in the edge label $\gamma_\phi(\vertex{prior}, \vertex{next})$; we denote by  $\ccalC_{\text{prior}}^+$ and $\ccalC_{\text{prior}}^-$ the positive and negative subformulas in $\ccalC_{\text{prior}}$, respectively.  Next, we find the set $P_{\text{min}}$ of subtasks in $X_P$ that can be the last one to be completed. For each subtask $e = (v_1, v_2) \in P_{\text{min}}$, we augment its edge label $\gamma(v_1, v_2)$ with the clause $\ccalC_{\text{prior}}^+$, i.e., $\gamma(v_1, v_2) = \gamma(v_1, v_2) \vee \ccalC_{\text{prior}}^+$; see also Fig.~\ref{fig:suffix}.  If $e \in P_{\text{min}}$ is the last subtask to be completed, we require that the clause $\ccalC_{\text{prior}}^+$ is satisfied. We say that $\ccalC_{\text{prior}}^+$ is satisfied if those robots involved in satisfying $\ccalC_{\text{prior}}^+$ in the prefix part return to regions including their initial locations. For instance, in Fig.~\ref{fig:suffix_ii} that shows $\auto{subtask}^-$ for task~\hyperref[task:ii]{(ii)} in Example~\ref{exmp:1}, we augment the label of the last edge with the clause $\ccalC_{\text{prior}}^+ = \ap{1}{1}{2,1}$, which is the clause in the last edge label of the prefix part; see Fig.~\ref{fig:nba_ii_subtask}. Combined with the negative subformula $\ccalC_{\text{prior}}^-$, if $\ccalC_{\text{prior}}^+ \wedge \ccalC_{\text{prior}}^-$ is satisfied, the original edge label $\gamma_\phi(v_1, v_2)$  will also be satisfied since, by condition~\ref{cond:f} in Definition~\ref{defn:run}, the label $\gamma_\phi(\vertex{prior}, \vertex{accept})$ implies the original edge label $\gamma_\phi(v_1, v_2)$ (acting as $\gamma_\phi(\vertex{prior}', \vertex{accept})$; see Fig.~\ref{fig:lasso}). In this way, robots return to regions  that contain  their initial locations and at the same time drive the transition in $\auto{subtask}^-$ back to $\vertex{accept}$. In what follows, we first construct the routing graph and  formulate the MILP for the first step, and then design low-level paths so that the robots can return to their initial locations while satisfying the specification $\phi$.

   \paragraph{Construction of the routing graph}{}\label{app:suffix_graph} Given the poset $P$, we build a routing graph $\ccalG = (\ccalV_G, \ccalE_G)$ following almost the same steps as in Appendix~\ref{sec:graph}. The only differences are related to the augmented clause $\ccalC_{\text{prior}}^+$.   When building the vertex set $\ccalV_\ccalG$, each time we encounter a literal $\ap{i}{j}{k,\chi}$ in $\mathsf{lits}^+(\ccalC_{\text{prior}}^+)$, we create $i$ vertices and let each vertex point to the region $\ell_k$.  Also, we build a one-to-one correspondence between  vertices and robots satisfying this literal in the prefix part since these robots need to return to their initial regions. We emphasize that each such vertex is associated with a single robot instead of a type of robots.  The mappings are created  as in Appendix~\ref{sec:vertex}.

   When  building the edge set $\ccalE_\ccalG$, there are no outgoing edges from  vertices associated with the literals in $\ccalC_{\text{prior}}^+$ since these are the vertices where the robots will be located at the final moment. To construct the  incoming edges to these vertices we treat  $\ccalC_{\text{prior}}^+$ as a regular clause  in an edge label but with one exception.  Recall  in Appendix~\ref{sec:b} that identifies  the leaving vertices  associated with prior subtasks. In this case, when the number of leaving vertices is the same as  the number of end vertices, we randomly create   one-to-one edges between leaving vertices and end vertices  since all robots visiting these leaving vertices belong to the same type. However, here we create edges from all leaving vertices to each vertex associated with literal $\ap{i}{j}{k,\chi}$ in $\ccalC_{\text{prior}}^+$ since each vertex associated with this literal  is only allowed to be visited by a specific robot of a specific type.

 \paragraph{Formulation of the MILP problem}{}\label{app:appendix_suffix_milp}  To find a high-level plan, we formulate a MILP  based on the routing graph $\ccalG$
 by following a similar process as in the case of  the prefix part, but with the exception that at the end, some robots need to return to their initial regions. The MILP formulation for the suffix part results in the same  constraints as~\eqref{eq:1}-\eqref{eq:same} in Appendix~\ref{app:appendix_prefix_milp} with the following two exceptions.

 \mysubparagraph{Returning to initial regions}{The first exception results from the requirement that while driving transition back to $v_{\text{accept}}$, robots need to return to  regions corresponding to their initial locations. To this end, we define binary variables $b_e$ for each subtask in $P_{\text{min}}$ that can be the last one to be completed, such that $b_e$ equaling 1 implies the satisfaction of the augmented clause $\ccalC_{\text{prior}}^+$; see Fig.~\ref{fig:suffix}. First, we require that one  and only one $b_e$ can be true, i.e.,}
 \begin{align}\label{eq:one_suffix}
   \sum_{e \in P_{\text{min}}} b_e = 1.
 \end{align}
 {If $b_e=0$ for a subtask $e$ in $P_{\text{min}}$, then one of the remaining clauses in the edge label of $e$ must be satisfied according to constraint $\eqref{eq:c}$, which is reduced to the case where $\ccalC_{\text{prior}}^+$  does not exist.}

 Next, we encode the constraint that when completing the last subtask in $P_{\text{min}}$,  robots must return to their respective regions. Recall that in Appendix~\ref{app:appendix_prefix_milp} we defined the binary variable $b_{e}^{e'}=1$ which equals 1 if subtask $e$ is completed after  $e'$, i.e., $t_e >  t_{e'}$. To determine this last satisfied subtask in $P_{\text{min}}$,
   we define $z = |P_{\text{min}}| -1$. Then, for any $e \in P_{\text{min}}$, the term $\sum_{e' \in  P_{\text{min}}\setminus \{e \} } b_{e}^{e'} - z  = 0$ if $e$ is the last subtask to be completed. Thus, the requirement that some robots return  to their respective regions to complete the last subtask  can be written as
 \begin{align}\label{eq:lastsubtask}
   1 + M_{\text{max}} ( &  \sum_{e' \in  P_{\text{min}}\setminus \{e \} } b_{e}^{e'} - z)  \leq  b_e \nonumber \\
   & \leq 1 + M_{\text{max}} (z - \sum_{e' \in  P_{\text{min}}\setminus \{e \} } b_{e}^{e'}),
 \end{align}
 for any subtask $e \in P_{\text{min}}$. Only when $e$ is the last subtask in $P_{\text{min}}$, does $b_e = 1$ come into effect.

 Given a subtask $e = (v_1, v_2) \in P_{\text{min}}$, similar to constraint~\eqref{eq:6}, the following constraint states that when $b_e=1$, i.e., when subtask $e$ is  the last one to be completed,  each vertex in $\ccalG$  associated with clause $\ccalC_{\text{prior}}^+$ of $\gamma(v_1, v_2)$ will be visited by a specific robot, for the $q$-th literal $\ap{i^q}{j^q}{k^q,\chi^q}$ in $\ccalC_{\text{prior}}^+$,
 \begin{align}\label{eq:return_suffix}
   \left.   \left[ \sum_{v \in \ccalM_\ccalV^\mathsf{lits}(e, 1, p_e, q)}   \sum_{u: (u,v) \in \ccalE_\ccalG}     \sum_{r_v = \ccalM_{\ccalK}^\ccalV(v) }   x_{uvr} \right] \middle/ {i^q} \right. = b_e,
  \end{align}
 where $p_e$ is the index of the clause $\ccalC_{\text{prior}}^+$ in $\gamma(v_1, v_2)$ and robot $r_v$ is the specific robot that should visit vertex $v$.

 \mysubparagraph{same-$\ag{i}{j}$ constraints}{{The second exception relates to  the same-$\ag{i}{j}$ constraints in Appendix~\ref{sec:samegroup}. The goal is to ensure that  the same  $i$ robots of type $j$ satisfy those atomic propositions with the same non-zero connectors that appear both in the prefix and suffix NBA.} Specifically, after solving the MILP for the prefix part, for any connector $\chi\not=0$ that appears in the specification $\phi$, we check whether any literal that includes this connector was involved in the prefix part. If yes, then these literals with the same nonzero connector should be satisfied by the same $\ag{i}{j}$. We denote by $\ccalK^\chi \subseteq \ccalK_j$ the set of $\ag{i}{j}$ that make these literals true and  by  $r^b$ the $b$-th robot in the enumerated set $\ccalK^\chi$. When dealing with the suffix part, for any label $\gamma \in\ccalM_\mathsf{\gamma}^\chi(\chi)$ where literals with connector $\chi$ appear and any clause $\ccalC_p^\gamma \in \clause{\gamma}$ that has  literal $\aap{i}{j}{k}{\chi}$, the constraint that this literal is satisfied by the same $\ag{i}{j}$  can be written as}
 \begin{align}\label{eq:same_suffix}
     \sum_{u:(u, v_k^b) \in \ccalE_\ccalG} x_{u v_k^b r^b} =  b_p, \;\forall\, r^b \in \ccalK^\chi,
 \end{align}
 where $v_k^b$ is the $b$-th vertex in the set of vertices in the routing graph $\ccalG$  that are associated with literal $\ap{i}{j}{k,\chi}$. If the clause $\ccalC_p^\gamma$ is true, i.e., if $b_p=1$, then the $b$-th vertex $v_k^b$ is visited by the $b$-th robot $r^b \in \ccalK^\chi$. On the other hand, if $\ccalK^\chi = \emptyset$, that is, if literals that share this $\chi$ do not appear in the prefix part, we turn to constraint~\eqref{eq:same} to impose the same-$\ag{i}{j}$ constraints.

 By condition \ref{cond:f} in Definition~\ref{defn:run}, there is a clause  $\ccalC'$ in the edge label $\gamma(\vertex{prior}', \vertex{accept})$ that is a subformula of $\ccalC_{\text{prior}}$. Therefore, some robots returning to regions corresponding to their initial locations  will enable the positive subformula in $\ccalC'$ of $\gamma(\vertex{prior}', \vertex{accept})$ and, at the same time ensure that any positive literal in the clause $\ccalC'$ with nonzero connector uses the same group of robots as the literal in the clause $\ccalC_{\text{prior}}^+$ of $\gamma(\vertex{prior}', \vertex{accept})$ with  the same connector.  In other words, when robots head back to their initial locations, the same-$\ag{i}{j}$ constraints over the last completed subtask are satisfied automatically. Robots can safely return to their initial locations  without  violating the same-$\ag{i}{j}$ constraints.

 \paragraph{Closing the suffix loops}{}\label{sec:closing}
 After solving the MILP for the first step for the suffix part, we utilize the method in  Appendix~\ref{sec:solution2mrta} to obtain low-level paths that drive the transition in $\autop$ back to $\vertex{accept}$, and at the same time ensure that robots involved in $\ccalC_{\text{prior}}^+$ return to regions corresponding to their initial locations of the suffix part. Note that to generate   the low-level paths for the last completed subtask $(\vertex{prior}', \vertex{accept})$, we need to satisfy  the augmented  clause $\ccalC_{\text{prior}}$  in the  edge label $\gamma_\phi(\vertex{prior}', \vertex{accept})$. By conditions~\ref{cond:d} and~\ref{cond:f} in Definition~\ref{defn:run}, we have $\gamma_\phi(\vertex{prior}, \vertex{accept}) \simplies \gamma_\phi(\vertex{accept}, \vertex{next})$ and $\gamma_\phi(\vertex{accept}, \vertex{next}) \simplies_s \gamma_\phi(\vertex{next})$, thus the initial locations of the suffix part satisfy the edge label $\gamma_\phi(\vertex{accept}, \vertex{next})$ and the  vertex label $\gamma_\phi(\vertex{next})$; so do the final locations in the low-level paths that enable $\gamma_\phi(\vertex{prior}', \vertex{accept})$ since they satisfy $\ccalC_{\text{prior}}$. Next, to close the suffix loop, robots return to their initial locations starting from the final locations in the low-level paths, while satisfying the clause $\ccalC_{\text{prior}}$ en route, thus satisfying $\gamma_{\phi}(\vertex{next})$. Because those robots involved in $\ccalC_{\text{prior}}^+$ have returned to their initial regions, and  by Assumption~\ref{asmp:env} each region spans consecutive cells,  they can return to their initial locations by traveling inside these  regions. In this way, the NBA $\autop$ remains at vertex $\vertex{next}$, thus the specification $\phi$ is not violated. The problem of finding  the path that travels inside the regions can be formulated as a generalized multi-robot path planning problem; see Appendix~\ref{sec:mapp}.

  \begin{rem}\label{rem:one-two-step}
    Closing the suffix loops in two steps is important to show the completeness of our proposed method. In a  single step approach where robots return to their initial locations in the suffix part at the same time that the NBA $\autop$ transitions to $\vertex{accept}$, the robots return to their initial locations to satisfy the last subtask $(\vertex{prior}', \vertex{accept})$. However, it is  possible  that the initial locations violate the vertex label $\gamma_\phi(\vertex{prior}')$ of the last subtask. {In this case, once the robots  reach regions corresponding to their initial locations (not necessarily reaching initial locations),  the edge label of the last subtask is satisfied and the last subtask  has to be completed since its vertex label is violated.} Nonetheless, it is possible that, at this moment,  robots have not reached  their initial locations inside these regions if some regions cover multiple cells. Therefore, this single-step approach may fail in this case. In practice such scenario rarely occurs. In fact, a single step approach generally works well in practice. Nevertheless, the proposed two-step method allows to guarantee completeness of our approach.
  \end{rem}


  \paragraph{Returning to initial locations in one step}{}\label{app:loop}
 To  ensure that the robots returning to their initial locations and progressing towards  the accepting vertex $\vertex{accept}$ in the NBA $\autop$ is made in one step, we first define a positive atomic proposition $\pi_{\text{init}}$ which is true if all robots return to their initial locations at the end of the suffix paths. Then, we replace $\ccalC_{\text{prior}}^+$ on the edge label of the last subtasks with $\pi_{\text{init}}$; see Fig.~\ref{fig:suffix}. If $\pi_{\text{init}}$ is satisfied, the original edge label $ \gamma_\phi(\vertex{prior}', \vertex{accept})$ in $\autop$ will also be satisfied since the initial robot locations satisfy $\ccalC_{\text{prior}}$ in $\gamma_\phi(\vertex{prior}, \vertex{accept})$ and $\gamma_\phi(\vertex{prior}, \vertex{accept}) \simplies \gamma_\phi(\vertex{prior}', \vertex{accept})$.  We adopt the first step in Appendix~\ref{sec:suf_milp} with all exceptions related to the difference between $\pi_{\text{init}}$ and $\ccalC_{\text{prior}}^+$. {That is, $\pi_{\text{init}}$ requires all robots to return to their initial locations, whereas $\ccalC_{\text{prior}}^+$ in the suffix part requires only those robots participating in the satisfaction of $\ccalC_{\text{prior}}^+$ in the prefix part to return to regions corresponding to their initial locations.}

 Recall that the vertex set $\ccalV_\text{init} \subseteq \ccalV_\ccalG$ contains vertices pointing to the initial robot locations. When building vertices in $\ccalV_\ccalG$ for the literal $\pi_{\text{init}}$, we create a copy of vertices in $\ccalV_{\text{init}}$ and associate these vertices  with the literal $\pi_{\text{init}}$, so that each vertex points to one single cell which is the initial location of a specific robot. The incoming edges of these vertices are constructed by treating $\pi_{\text{init}}$ as a regular clause (single literal with connector $\chi=0$) of an edge label. Specifically, given a subtask $e\in P_{\text{min}}$, let robot $r_v$ denote the specific robot that should visit vertex $v$ associated with its literal $\pi_{\text{init}}$. We identify all vertices in $\ccalV_\ccalG$ that are associated with robots of the same type as $r_v$ and are related to initial locations (see Appendix~\ref{sec:a}), prior subtasks of $e$ (see Appendix~\ref{sec:b}) or vertex labels of the same subtask (see Appendix~\ref{sec:c}). Then, we create an edge from each one of these vertices   to vertex $r_v$. No outgoing edges exist for these vertices.  The remaining steps to build the routing graph are the same as those in Appendix~\ref{app:suffix_graph}.

 When formulating the MILP, the only difference is in the constraint~\eqref{eq:return_suffix}, that is, $\pi_{\text{init}}$ is true if and only if each vertex in $\ccalG$ that is associated with  $\pi_{\text{init}}$ is visited by a specific robot among the whole fleet of $n$ robots, i.e.,
 \begin{align}
   \left.   \left[ \sum_{v \in \ccalM_\ccalV^\mathsf{lits}(e, 1, p_e, 1)}   \sum_{u: (u,v) \in \ccalE_\ccalG}     x_{uvr_v} \right] \middle/ {n} \right. = b_e,
  \end{align}
 where robot $r_v$ is the specific robot that should visit vertex $v$. Note that robots returning to their initial locations in one step   does not guarantee the completeness of our proposed method, as discussed in Remark~\ref{rem:one-two-step}.

 \subsection{Extensions of the MILP}\label{sec:extension}
 {One advantage of the proposed MILP for the prefix and suffix parts is that it is associated with  each subtask individually,  which allows us to impose additional constraints on certain subtasks to address problem-specific requirements.} In this section, we present possible extensions of the \ltlx formula  and  introduce variations to the MILP formulation by considering more interesting constraints.

 \subsubsection{Requiring specific robots to participate in a subtask}
 Given an atomic proposition $\ap{i}{j}{k,\chi}$ that appears in the \ltlx formula, we can require that  a specific subset of robots, denoted by  ${\ccalK}'_j \subseteq \ccalK_j$,  participates or not in the satisfaction of this formula. Suppose the set of vertices  in $\ccalG$ that are associated with this literal is $\ccalM_{\ccalV}^\mathsf{lits} (e, 0|1, p, q)$. Then for each specific robot $r \in \ccalK'_j$, we have
 \begin{align}
   \sum_{u:(u,v) \in \ccalE_\ccalG} x_{uvr} = I, \quad  \forall \, v \in \ccalM_{\ccalV}^\mathsf{lits} (e, 0|1, p, q)
 \end{align}
 where $I\in \{0, 1\}$. We set $I$ to 1 if we require every robot in $\ccalK'_j$ to participate in the satisfaction of $\ap{i}{j}{k,\chi}$ (where $|\ccalK'_j|\leq i$) and $I=0$ if no robot in $\ccalK'_j$ should  be involved. 
 \subsubsection{Managing the number of participating robots}
 {When completing the task specified by the \ltlx formula, it may be desirable to dispatch as few robots as possible to keep the whole system at a small scale. On the other hand, we may want to dispatch as many robots as possible to enhance the efficiency.} This requirement can be handled by adding another term to the MILP objective in~\eqref{equ:obj} as follows
 \begin{align}\label{eq:objective1}
   \min \; \; &  \alpha_1 \sum_{(u,v)\in \ccalE_\ccalG}  \sum_{r \in \ccalM^\ccalV_{\ccalK}(v)} d_{uv} x_{uvr} +  \alpha_2  \sum_{e\in X_P} t_e  \nonumber \\ & \pm  \alpha_3 \sum_{v\in \ccalV_{\text{init}}}   \sum_{w:(v,w)\in \ccalE_\ccalG} x_{vwr_v},
 \end{align}
 where $\alpha_1 + \alpha_2 + \alpha_3=1$, $r_v$ is the specific robot at vertex $v \in \ccalV_{\text{init}}$, and the final term  captures the number of robots that leave their initial locations,  equivalent to the  number of robots that are assigned to the desired subtasks. The positive sign in the last term in objective~\eqref{eq:objective1} corresponds to the case where fewer robots are needed while the negative sign corresponds to the case where more diverse robots are needed.

 \subsubsection{Prohibiting the  use of the same robots} In  Definition~\ref{defn:valid} of the \ltlx formula, we handled the requirement that two atomic propositions with the same nonzero connector must be satisfied by the same fleet of robots. Alternatively, we can impose the restriction that some atomic propositions involving the same robot type but different nonzero connectors must be satisfied by two disjoint fleets of robots. For instance, the formula $\lozenge \ap{1}{1}{1,1} \wedge \lozenge \ap{2}{1}{2,2}$ requires that the robots that visit region $\ell_1$ are different from those two robots that visit region $\ell_2$. Given such two atomic propositions that need to be  satisfied by different robots, suppose the sets of vertices in $\ccalG$ that are associated with these two literals are $\ccalM_{\ccalV}^\mathsf{lits} (e, 0|1, p, q)$ and $\ccalM_{\ccalV}^\mathsf{lits} (e', 0|1, p', q')$. Then, $\forall \, r \in \ccalK_j$, this requirement can be written as
 \begin{align}\label{eq:atmost1}
   \sum_{v\in \ccalM_{\ccalV}^\mathsf{lits} (e, 0|1, p, q)} & \sum_{u: (u,v) \in \ccalE_\ccalG} x_{uvr} + \nonumber \\
  & \hspace{-2em}  \sum_{v\in \ccalM_{\ccalV}^\mathsf{lits} (e', 0|1, p', q')} \sum_{u: (u,v) \in \ccalE_\ccalG}  x_{uvr} \leq 1.
 \end{align}
 Constraint~\eqref{eq:atmost1} states that robot $r$ of type $j$ can visit at most one vertex among the vertices that are associated with these two literals.

 \section{Design of Low-Level Paths that Satisfy  the Original LTL Task}\label{sec:solution2mrta}
   This section presents the correction stage  that concretizes the high-level plan obtained in Section~\ref{sec:path} to satisfy the specification $\phi$.
   We first find a simple path from the NBA $\auto{subtask}^-$ that connects $v_0$ and $\vertex{accept}$, based on the time axis and the time-stamped task allocation plan, and then find the counterpart of this simple path from the NBA $\autop$. To satisfy the specification $\phi$, while following  the high-level plan, we formulate a sequence of generalized multi-robot path planning (GMRPP) problems  to design low-level executable paths.

 \subsection{Extraction of the simple path from the sub-NBA \upshape $\auto{subtask}^-$}\label{sec:run}

 Recall that each distinct time instant on the time axis $\vec{t}$ obtained in Section~\ref{sec:path} has a one-to-one correspondence with subtasks in the set $X_P$, and the sorted time axis generates a linear extension of subtasks in $X_P$ that induces a simple path in $\auto{subtask}^-$ that connects $v_{0}$ and $\vertex{accept}$. {In this section, we proceed along the time axis $\vec{t}$ to extract this simple path using a graph-search version  of the backtracking search algorithm, which is a variant of the depth-first search~\citep{russell2002artificial}. In our graph-search method, each vertex is searched at most once.}  The outline of this algorithm is shown in Alg.~\ref{alg:extract}.

   To this end, we define $c\in \mathbb{N}$ as the {\it global counter} which keeps track of the progress made along the time axis $\vec{t}$. Specifically, $c$ is the index of the subtask that has  been completed most recently. Therefore, $\vec{t}(c+1)$ is  the  completion time of the  subtask, denoted by $e' = (v'_1, v_2')$, that is  the next one to be completed. Let $v_1$ denote the vertex in $\auto{subtask}^-$ that is the  most recently reached. The set $frontier$ is a last-in-first-out queue that stores vertices that are available for expansion and  the set $explored$ stores vertices that have been expanded.  {At each iteration, among all subtasks with  the starting vertex $v_1$,} we find the one $(v_1, v_2)$ that is  equivalent to subtask $e'$ [line~\ref{run:edgelabel}, Alg~\ref{alg:extract}]. Then, after time instant $\vec{t}(c+1)$,  vertex $v_2$ becomes the most recently reached vertex. We next increase the global counter by 1 and add it to $frontier$, a last-in-first-out queue [line~\ref{run:put}, Alg.~\ref{alg:extract}]. The iteration will terminate when the accepting vertex $\vertex{accept}$ is reached.

  \begin{algorithm}[t]
       \caption{Extract the  simple path  from $\auto{subtask}^-$ }
       \LinesNumbered
       \label{alg:extract}
       \KwIn {time axis $\vec{t}$, sub-NBA $\auto{subtask}^-$}
       $v_1 = v_0, c = 0,  frontier=\{(v_1, c)\}, explored=\emptyset$ \label{run:initialization}
       \While{\upshape $v_1 \neq \vertex{accept}$}{
         Remove $(v_1, c)$ from $frontier$ and add $v_1$ to $explored$\;
         Obtain the subtask $e'=(v'_1, v_2')$ that is associated with $\vec{t}(c+1)$\;
         \For{\upshape $(v_1, v_2) \in \auto{subtask}^-$}{
           \If{\upshape   $ \gamma(v_1, v_2) = \gamma(v'_1, v'_2)$ {\bf and} $\gamma(v_1) = \gamma(v'_1)$ \label{run:edgelabel}}{
             Determine {(1)} essential clause, {(2)} essential robots, {(3)} negative clause, and {(4)}  sequence of vertices leading to  $v_2$\label{run:info}\;
             \If{$v_2$ not in $frontier$ and $explored$}{
               Add $(v_2, c+1)$ to $frontier$\label{run:put}\;
             }
           }
         }

       }
       {\bf return} the simple path leading to $v_1$\;

  \end{algorithm}

  When a subtask $(v_1, v_2)$ in~$\auto{subtask}^-$ is matched with the subtask $e'$ that is completed at $\vec{t}(c+1)$, we keep track of the following information:  \ref{essential:clause} the exact clauses that are\ satisfied in the vertex label $\gamma(v_1)$  and edge label $\gamma(v_1, v_2)$ since only one clause in each label is true by constraint~\eqref{eq:c} in the MILP,  \ref{essential:robots} the subset of robots that participate in the satisfaction of each literal in these clauses,
  \ref{essential:negative} the negative subformula in $\autop$ that is in conjunction  with the satisfied clause found in \ref{essential:clause} but is replaced with $\top$ during the relaxation stage in Section~\ref{sec:prune}, and  {(4)} the sequence of vertices in $\auto{subtask}^-$ that have been visited up to vertex $v_2$. This information will be used to formulate the generalized multi-robot path planning problems later. {In what follows, we discuss \ref{essential:clause}-\ref{essential:negative} in further detail and omit step (4) since it is straightforward.}

  \mysubparagraph{Essential clauses}{Given the edge label or vertex label $\gamma=\bigvee_{p\in \ccalP} \bigwedge_{q\in \ccalQ_p} \ap{i}{j}{k,\chi}$ of subtask $(v_1, v_2)$ that is neither $\top$ nor $\bot$, we refer to the unique satisfied clause as the {\it essential clause} and denote it by $\gamma^+$. Recall that in Appendix~\ref{sec:labelconstraints}  we define a binary variable $b_p$ representing the truth of the $p$-th clause in a given label; see constraint~\eqref{eq:c}. Thus, we find the essential clause by locating the clause $\ccalC_{p}^{\gamma} \in \clause{\gamma}$ such that $b_p=1$. On the other hand, when the vertex or edge label is $\top$, by default, we define the essential clause as $\top$.}\label{essential:clause}

  \mysubparagraph{Essential robots}{We refer to the set of robots whose collective behavior satisfies the positive literals in the essential clause as the {\it essential robots}. Recall in Appendix~\ref{app:appendix_prefix_milp} that the binary variable $x_{uvr}$ represents  robot $r$ visiting a vertex  $v$ in the routing graph $\ccalG$.    For the $q$-th literal $\ap{i}{j}{k,\chi}$ in the essential clause $\ccalC_p^\gamma$ of a vertex or edge label of subtask $(v_1, v_2)$, we determine its essential robots by locating  the associated $x_{uvr}$ whose value is 1. That  is, for each associated vertex $v\in \ccalM_{\ccalV}^\mathsf{lits} (e, 0|1, p, q)$, we identify the robot $r\in \ccalM_\ccalK^\ccalV(v)$ such that $\exists\, (u,v)\in\ccalE_\ccalG$, making $x_{uvr}=1$.  On the other hand, if the essential clause is $\top$, there are no essential robots.}\label{essential:robots}

  \mysubparagraph{Negative clause}{The collective behavior of essential robots satisfies the essential clauses in $\auto{subtask}^-$. For an essential clause that is not $\top$, there exists a unique clause in the NBA $\autop$ that only differs from the essential clause in that it may contain the conjunction of the negative literals that were removed during the relaxation stage. We refer to this conjunction of negative literals as the {\it negative clause} and denote it by $\gamma^-$, which will be satisfied by the low-level paths. By default, we define the negative clause as $\top$, if the corresponding clause in $\autop$ does not have negative literals. Finally, the conjunction $\gamma^+ \wedge \gamma^-$ of an essential clause and its corresponding negative clause constitutes a {\it complete clause} in~$\autop$.}\label{essential:negative}

  When the vertex label $\gamma(v_1)$ or edge label $\gamma(v_1, v_2)$ of subtask $(v_1, v_2)$ in $\auto{subtask}^-$ is $\top$, the associated essential clause is also $\top$. However, the negative clause may not be $\top$, which happens when there exists a  clause in the corresponding label in $\autop$ that only includes negative literals.   Note that by condition~\hyperref[asmp:b]{(b)} in Definition~\ref{defn:same}, the complete clause of the vertex $v_1$ is implied by the complete clause of the edge that is immediately preceding the current subtask $(v_1, v_2)$. If the label is the edge label $\gamma(v_1, v_2)$, we randomly select one among the clauses that only include negative literals. Otherwise,  if the label is the vertex label $\gamma(v_1)$, and further if the current subtask is not the first one, we select one as the negative clause (acting as the complete clause), that is  implied by the complete clause in the edge that is immediately preceding the current subtask $(v_1, v_2)$. We can obtain this edge since in step (4) we keep track of the sequence of vertices that lead to vertex $v_2$. This ensures that when the edge label $(v_1, v_2)$ is enabled due to the satisfaction of its complete clause, the complete clause in its end  vertex label can be satisfied automatically.   On the other hand, if the current subtask is indeed the first one, we randomly select a negative  clause that is satisfied by the initial robot locations.

  \subsection{Generalized multi-robot path planning}\label{sec:mapp}
 Leveraging the correspondence between the NBA  $\autop$ and the sub-NBA $\auto{subtask}^-$,  we can find the counterpart in $\autop$ of the simple path in $\auto{subtask}^-$ obtained in Appendix~\ref{sec:run}. We denote by $\theta_{\phi}$ this counterpart, which corresponds to a sequence of temporally sequential subtasks. Our goal is to find a collection of executable paths  that induce the simple path $\theta_{\phi}$ in $\autop$. To achieve this, we formulate the execution of each subtask in $\theta_{\phi}$ into a {\it generalized multi-robot path planning problem (GMRPP)}. Compared to traditional  multi-robot path planning that, given an initial robot configuration, designs paths to reach the target configuration, the GMRPP imposes additional constraints on the intermediate configurations.

 Observe that given the completion time of two consecutive subtasks on the time axis $\vec{t}$, we can obtain the tightest span of the second subtask's vertex label; see Definition~\ref{defn:time}. Specifically, the activation time of the second subtask's vertex label is at most one time step after the completion time of the first subtask; see also constraint~\eqref{eq:20} in Appendix~\ref{app:appendix_prefix_milp} that captures the temporal relation between two subtasks. On the other hand, the  completion time of the vertex label of the second subtask is at most one time step before the completion time of the second subtask; see also constraint~\eqref{eq:17} in Appendix~\ref{app:appendix_prefix_milp} that captures the temporal relation for the same subtask.  To design the low-level paths,  we let each robot visit waypoints in its individual plan $p_{r,j}$ sequentially, {possibly at different time instants than those in its timeline $t_{r,j}$. This is because   the individual timeline is obtained using the shortest travel time between regions (see scheduling constraints in Appendix~\ref{app:scheduling_constraints}) and omitting collision avoidance between robots,} but the relative temporal relations with other robots  are kept. Also, we maintain the tightest span of the vertex label of the considered subtask between completion time of two consecutive subtasks.

 To this end, for each robot $[r,j] \in \ccalR$, we define a {\it local counter} $\zeta_{r,j} \in \mathbb{N}$  that  keeps track of how much progress has been made along the individual plan $p_{r,j}$. Specifically, $\zeta_{r,j} = a$ indicates that the $a$-th waypoint in the plan $p_{r,j}$ is the one  visited by  robot $[r,j]$  most recently. Furthermore, recall that the  global clock $c$ monitors the index of the most recently completed subtask  along the time axis $\vec{t}$, which also captures the execution progress along the simple path  $\theta_{\phi}$ since a one-to-one correspondence exists between time instants in $\vec{t}$ and subtasks in $\theta_\phi$. In what follows, we provide the ingredients for the construction of GMRPP.

  \subsubsection{Ingredients of GMRPP}\label{sec:gmmpp1} Consider a subtask $e = (v_1, v_2)$ generated by the simple path $\theta_{\phi}$ that is the next one to be completed. Let $\gamma_1^+$ and $\gamma_1^-$ denote the essential and negative clauses associated with the vertex label $\gamma_\phi(v_1)$, respectively. Similarly, we define $\gamma_{1,2}^{+}$ and $\gamma_{1,2}^{-}$  for the edge label $\gamma_\phi(v_1, v_2)$. The goal of a GMRPP   is to determine a collection of executable paths such that robots complete the current subtask (by satisfying the complete clause  $\gamma_{1,2}^+ \wedge \gamma_{1,2}^-$ at the end while respecting the complete clause $\gamma_1^+ \wedge \gamma_1^-$ en route) and automatically activate the next subtask after completion since the complete clause $\gamma_{1,2}^+ \wedge \gamma_{1,2}^-$ implies the complete clause associated with the starting vertex of the next subtask. We refer to $\gamma_1^-$ as the {\it running constraint} and $\gamma_{1,2}^- $  as the {\it terminal constraint}. Next, we determine three types of robots that are directly involved in the execution of the current subtask $e$.

 \mysubparagraph{Essential robots associated with constraint $\gamma_1^+$}{We collect essential robots associated with essential clauses in $\gamma_{1}^{+}$ in the set $\ccalR_{1}$,  where robots need to remain at certain target regions.}\label{sec:essential_a}

 \mysubparagraph{Essential robots associated with target $\gamma_{1,2}^{+}$}{We collect essential robots associated with the essential clause  $\gamma_{1,2}^{+}$ in the set $\ccalR_{1,2}$, where  robots need to reach certain target regions.}\label{sec:essential_b}

  \mysubparagraph{Robots associated with running and terminal constraints $\gamma_1^-$ and $\gamma_{1,2}^-$}{The robots, in this case, are different from the previous two types since they are related to negative clauses $\gamma_1^-$ or $\gamma_{1,2}^-$. These robots, unless they are involved in the first two cases, navigate without specific targets, only to satisfy the bound imposed by the negative literals on the number of certain types of robots in some regions. We collect them in the set $\ccalR^-$, which contains all robots that belong to certain types involved
  in $\gamma_1^-$ or $\gamma_{1,2}^-$, i.e., $\ccalR^- = \left\{\ccalK_j: \neg \ap{i}{j}{k} \in \mathsf{lits}^- (\gamma_1^-  \vee \gamma_{1,2}^-) \right\}$.}\label{sec:essential_c}

  Let $\ccalR_e = \ccalR_1 \cup\ccalR_{1,2}  \cup \ccalR^-$ denote the set that collects all robots directly involved in the current subtask, and $\ccalR_0 = \ccalR \setminus\ccalR_e$ collects the remaining robots. To formulate the GMRPP,  we define by $X_I$ and $X_G$ the sets of initial and target locations, respectively, such that $X_I(r,j) \in S$ and $X_G(r,j)\subseteq S$ are the initial  and target locations of robot $[r,j] \in \ccalR$. Specifically, the initial robot locations are  where the robots are at the end of the subtask immediately preceding $e$. The target region of  robot $[r,j] \in \ccalR_{1,2}$ is determined  by its associated literal in $\gamma_{1,2}^+$, which is also given by  $p_{r,j}(\zeta(r,j)+1)$. Similarly, the target region of robot $[r,j] \in \ccalR_{1}$ can also be determined by its associated literal in $\gamma_1^+$. There are no specific target locations for robots in $\ccalR^- \cup \ccalR_0$.

  Finally, let $\tau'_{r,j}$ denote the path segment of robot $[r,j] \in \ccalR$, where $\tau'_{r,j}(t)$ denotes the robot location at time $t$ for $ t = 0,  \ldots, T$, where time instants 0  and $T$ correspond to the completion time of the immediately preceding subtask and the current subtask, respectively. Next, the generalized multi-robot path planning problem, adapted from~\cite{yu2016optimal}, is defined as follows.
  \begin{defn}[Generalized multi-robot path planning]\label{defn:gmmpp}
     Given a discrete workspace $E$, a set of robots $\ccalR= \ccalR_e \cup \ccalR_0$ where $\ccalR_e = \ccalR_1 \cup\ccalR_{1,2} \cup \ccalR^-$, a set of initial locations $X_I$, a set of target regions $X_G$, the running constraint $\gamma_1^-$, the terminal constraint $\gamma_{1,2}^-$, and the horizon $T$, find a collection of path segments $\tau'_{r,j}$ for all robots $[r,j]\in \ccalR$ such that {\it (i)} every robot $[r,j] \in \ccalR_{1,2}$ starts from the initial location and arrives at the target region at time instant $T$, i.e., $ \tau'_{r,j}(0) = X_I(r,j)$ and  $ \tau'_{r,j}(T) \in X_G(r,j)$, $\forall [r,j] \in  \ccalR_{1,2}$; {\it (ii)} every robot $[r,j] \in \ccalR_1$ remains in the target region for all time except $0$ and $T$, i.e., $ \tau'_{r,j}(0) = X_I(r,j)$ and $\tau'_{r,j}(t) \in X_G(r,j)$ for all $t=1,\ldots, T-1$;
   and {\it (iii)} the paths $\{\tau'_{r,j}\}, \forall [r,j] \in \ccalR^-$,  satisfy the running constraint $\gamma^-_1$ for all times except at $0$ and $T$, and also satisfy the terminal constraint $\gamma^-_{1,2}$ at time instant $T$.
  \end{defn}

\begin{figure}[!t]
    \centering
    \includegraphics[width=0.9\linewidth]{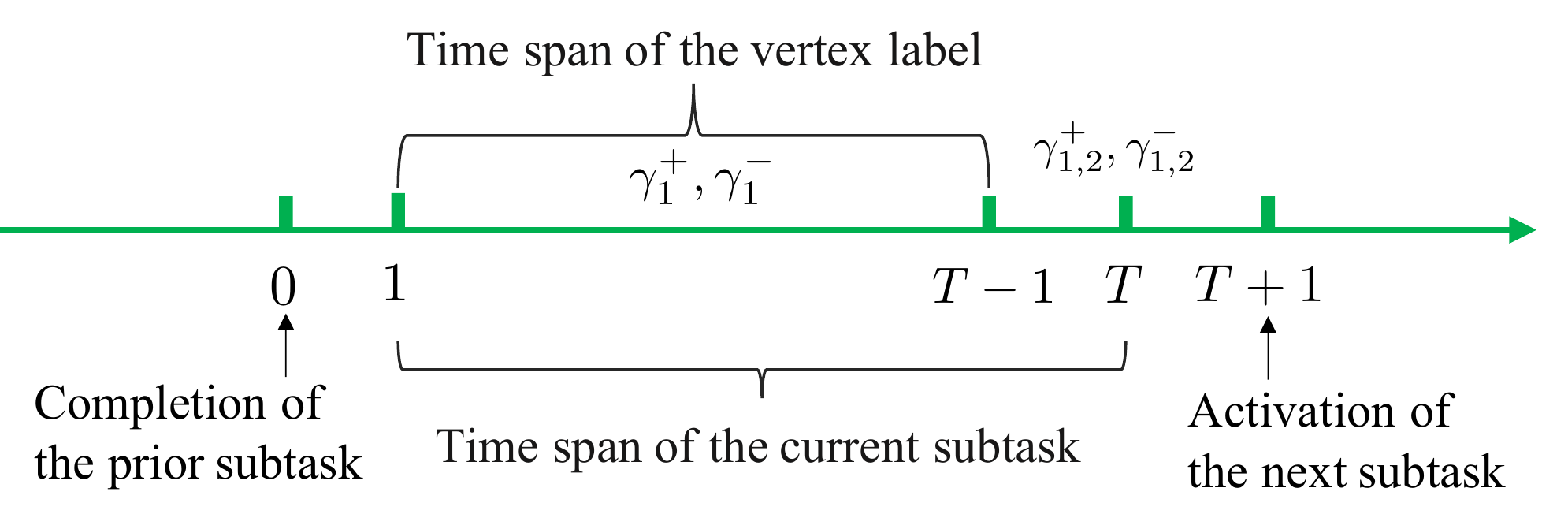}
    \caption{The time relation within one GMRPP. The essential and negative clauses are aligned with the time instants when they should be satisfied.}
    \label{fig:timeline}
  \end{figure}

 Fig.~\ref{fig:timeline} illustrates the time relation within one instance of GMRPP.  The paths do not need to satisfy  $\gamma_1^+$ and $\gamma_1^-$ at time instants 0 and  $T$ since the tightest span of the vertex label of the current subtask can be one time step after the completion of the immediately preceding subtask, which is indicated by time 0, and one time step before the completion of the current subtask, which is indicated by time $T$. In Appendix~\ref{sec:solution2mapp}, we discuss how to solve the GMMPP with horizon $T$.  The paths returned by this GMRPP complete the subtasks $(v_1, v_2)$  and meanwhile activate the vertex label of $v_2$, i.e., the next subtask. 

 \begin{rem}
   In the formulation of the GMRPP, we did not take into account  collision avoidance between robots, which will be addressed in Appendix~\ref{sec:extension_collision}.
 \end{rem}

  \begin{rem}
    Note that subtasks are executed sequentially as discussed above since we only assign target regions to those robots $\ccalR_1 \cup \ccalR_{1,2}$ directly involved in the current subtask $e$.  Thus, we refer to this as the sequential execution. However, robots that participate in subsequent subtasks can move together with the robots that participate in the current subtask $e$ by heading toward some ``intermediate'' targets,  so that after the current subtask $e$ is completed, these robots associated with subsequent subtasks have already traveled part of their routes towards finishing their respective subtasks. We will present this simultaneous execution  in Appendix~\ref{sec:extension_essential}.
  \end{rem}

 \begin{rem}
   We refer to the execution of the subtask $e$ discussed above as the full execution since it mobilizes all robots in the workspace. However, most times not all robots need to move for one specific subtask since  only a subset of robots are responsible for the satisfaction of this subtask. In Appendix~\ref{sec:extension_partial} we discuss a partial execution where only necessary robots in $\ccalR_e$, are allowed to move and the rest of the robots stay put. The partial execution shares most similarity with the full execution.
 \end{rem}

 \begin{algorithm}[!t]
     \caption{Executable multi-robot path planning}
     \algorithmendnote{The right-side time instant $\vec{t}(c+1)$ in~line~\ref{seq:axis} is the one before updates.}
       \LinesNumbered
       \label{alg:sequentialMAPP}
       \KwIn {Workspace $E$,  robot team $\ccalR$, subtask sequence $\theta_{\phi}$, waypoint sequence $\{p_{r,j}\}$, time sequence $\{t_{r,j}\}$, NBA $\autop$ and $\auto{subtask}^-$}
       \Comment*[r]{Initilization}
       $\tau_{r,j} = s_{r,j}^0 $, $\zeta_{r,j} = 0, \forall\, [r,j] \in \ccalR$, $c = 0$ \label{seq:initilization}\;
       \Comment*[r]{Sequential GMRPP solutions}
       \For{$e = (v_1, v_2) \in \theta_{\phi}$ \label{seq:terminate_1}}{
         \If{\upshape $\vec{t}(1)=0$ \label{seq:initial}}{
           $\zeta_{r,j} = \zeta_{r,j} + 1, \; \forall \, [r,j] \in \ccalR_{1,2}$ \label{seq:counter_initial}\;
           $c = c + 1$ \label{seq:c_initial}\;
         }
         \Else{
           Formulate the GMRPP  \label{seq:ingredients}\;
           Solve the GMRPP for horizon $T$ as in Appendix~\ref{sec:solution2mapp} to obtain paths $\tau'_{r,j}$, $\forall [r,j]\in \ccalR$\label{seq:mapp}\;
           \Comment*[r]{Update}
           Concatenate paths:
             $\tau_{r,j} = \tau_{r,j} \cdot \tau'_{r,j}[1:T_e], \;\forall \,[r,j] \in \ccalR $
           \label{seq:path}\;
            Update individual timeline:  $t_{r,j}(\zeta) =  t_{r,j}(\zeta)  + T_e - (\vec{t}(c+1) - \vec{t}(c))$, $\forall \,\zeta \geq  \zeta_{r,j}$, $ \forall\, [r,j] \in  \ccalR$ \label{seq:timeline} \;
           Upadte time axis: $\vec{t}(c') = \vec{t}(c') + T_e - (\vec{t}(c+1) - \vec{t}(c))$, $ \forall \, c' \geq c+1$  \label{seq:axis}\;
           Update local counter: $\zeta_{r,j} = \zeta_{r,j} + 1, \; \forall \, [r,j] \in \ccalR_{1,2}$ \label{seq:counter}\;
            Update global counter: $c = c + 1$ \label{seq:c}\;

         }
       }
 \end{algorithm}

 \subsubsection{Sequential GMRPP solutions to find low-level paths that induce the simple path $\theta_{\phi}$}
  The GMRPP algorithm to design executable paths under the  full execution is outlined in Alg.~\ref{alg:sequentialMAPP}.  We initialize all local counters and the global clock to 0 [line~\ref{seq:initilization}, Alg.~\ref{alg:sequentialMAPP}]. The algorithm  terminates when iteration over  the subtasks in the simple path $\theta_{\phi}$ is finished [line~\ref{seq:terminate_1}, Alg.~\ref{alg:sequentialMAPP}].

  We first check whether the first time instant $\vec{t}(1)$ on the time axis is 0. If $\vec{t}(1)=0$, then the first subtask in the simple path $\theta_{\phi}$, i.e., the edge label of the first subtask, is satisfied by the initial robot locations. Thus, we increase the local counters of robots that participate and the global counter by 1 [lines~\ref{seq:initial}-\ref{seq:c_initial}, Alg.~\ref{alg:sequentialMAPP}]. Otherwise, we solve the corresponding GMRPP as in Appendix~\ref{sec:solution2mapp}.  We initialize $T$ by $\vec{t}(c+1) - \vec{t}(c)$ (by default $\vec{t}(0)=0$),  which is the difference between the completion time of the immediately preceding subtask and the current one. 
  We denote by $T_e$  the final $T$  when  the GMRPP has a solution [line~\ref{seq:mapp}, Alg.~\ref{alg:sequentialMAPP}]. Given a solution to the generalized multi-robot planning problem, Alg.~\ref{alg:sequentialMAPP} proceeds with the following updates [lines~\ref{seq:path}-\ref{seq:c}, Alg.~\ref{alg:sequentialMAPP}].


   First, for each robot $[r, j] \in \ccalR$, we append the path segment $\tau'_{r,j}(t)$, for all $t=1,\ldots, T_e$, to its already-executed path $\tau_{r,j}$ [line~\ref{seq:path}, Alg.~\ref{alg:sequentialMAPP}]. 
  Note that the final waypoints will be the initial locations of the next instance of GMRPP.  Moreover, for each robot $[r,j] \in \ccalR$, we increase the time instants in $t_{r,j}$ with indices larger than or equal to  $\zeta_{r,j}$ by $T_e - (\vec{t}(c+1) - \vec{t}(c))$ [line~\ref{seq:timeline}, Alg.~\ref{alg:sequentialMAPP}], where $\vec{t}(c+1) - \vec{t}(c)$ is the time span of the current subtask  given by the high-level plan whereas $T_e$ is the actual time span given by the low-level executable path.
     Similarly, we increase the time instants in $\vec{t}$ with indices larger than or equal to  $c+1$ by $T_e - (\vec{t}(c+1)- \vec{t}(c))$.
     In this way,  the subsequent subtasks in the high-level plan  that have not been executed are shifted into the future by the same amount in order to maintain the correct temporal relation (precedence or simultaneity) between visits to waypoints in $\{p_{r,j}\}$. Next, we increase  the local counter by 1 for all robots in $\ccalR_{1,2}$, which reflects local progress towards completing their individual plans [line~\ref{seq:counter}, Alg.~\ref{alg:sequentialMAPP}]. Similarly, we increase the global counter $c$ by 1 [line~\ref{seq:c}, Alg.~\ref{alg:sequentialMAPP}].

 \subsection{Solution to the generalized multi-robot path planning problem}\label{sec:solution2mapp}

 Conventional multi-robot path planning problems  find feasible or optimal paths for groups of robots starting from a set of initial locations and ending at a set of desired target locations; see, e.g., \cite{yu2016optimal} and the references therein. To find executable paths satisfying the subtasks, we generalize the multi-robot path planning problem in several ways. First,  we extend the notion of a single target location to a target region such that reaching any cell in the target region suffices. Second, the path segments that complete the  subtask  satisfy the complete clauses in the corresponding vertex label and edge label. Third, in the partial execution,  only a subset of robots  directly involved in the current or future subtasks are allowed to move.
 \begin{figure}[t]
   \centering
   \includegraphics[width=0.8\linewidth]{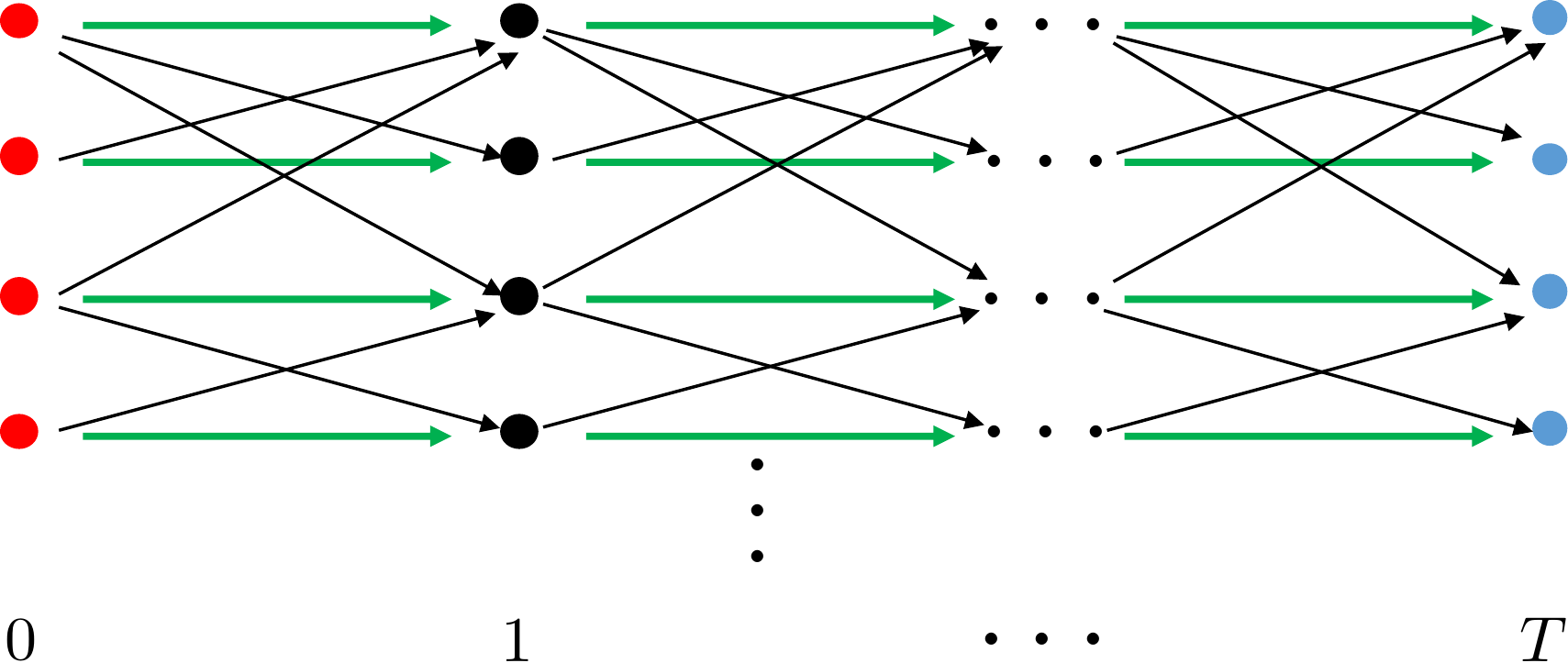}
   \caption{Time-expanded graph over horizon $T$ (modified from~\cite{yu2016optimal})}
   \label{fig:mapp}
 \end{figure}

 \begin{figure}[t]
   \centering
   \includegraphics[width=0.5\linewidth]{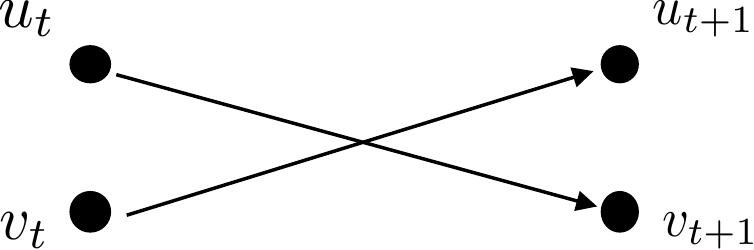}
   \caption{Merge-split gadget for avoiding head-on collision (modified from~\cite{yu2016optimal})}
   \label{fig:gadget}
 \end{figure}

   In what follows, we adapt the method proposed in~\cite{yu2016optimal} to solve the GMRPP under the full execution  with given horizon $T$.  The key idea is to construct a time-expanded graph $\ccalG_T = (\ccalV_T, \ccalE_T)$ of the workspace which contains $T$ copies of the free cells in the workspace $E$; see Fig.~\ref{fig:mapp}. We organize the vertices and edges of this time-expanded graph $\ccalG_T$ in a matrix structure, so that each row corresponds to a free cell in the workspace $E$ and each column corresponds to a time instant $t\in \{0,\ldots,T\}$. Then, a vertex $u_t \in \ccalV_T$ that appears in the $t$-th column of this matrix structure indicates whether the cell $u\in E$ is occupied by a robot at  time instant $t$. The edges in $\ccalG_T$ capture  adjacency relations between neighboring cells in $E$ and consecutive time instants in $\{0,\ldots,T\}$. Specifically, for any two adjacent cells $u$ and $v$ in $E$, an ``X''-shape structure, referred to as a {\it merge-split gadget}, is created to capture the transition from vertex $u$ at time $t$ to vertex $v$ at time $t+1$; see also Fig.~\ref{fig:gadget}. In this way, robots traveling along a given row in the matrix structure corresponding to $\ccalG_T$ effectively remain idle at their current cell, while robots switching between different rows in $\ccalG_T$ transition between adjacent cells in $E$. We say that a sequence of transitions in $\ccalG_T$ form $t=0$ to $t=T$ produces a robot path in the workspace $E$.

 Next, we formulate an Integer Linear Programming (ILP) problem to solve the GMRPP. Let $s_{uvrj} \in\{0,1\}$ be the routing variable such that $s_{u_{t}v_{t+1}rj} = 1$ if robot $[r,j] \in \ccalR$ traverses the edge $(u_t,v_{t+1}) \in \ccalE_T$. In what follows, we describe the constraints and objective of this ILP.

 \subsubsection{Routing constraints}
 The constraint that each edge can be traversed by at most one robot at a given time is given by
 \begin{align}
   \sum_{[r,j] \in \ccalR} s_{u_t v_{t+1}rj} \leq 1, \quad \forall \,(u_t,v_{t+1}) \in \ccalE_T,
 \end{align}
 for all $t=0,\ldots, T-1$. Furthermore, the flow conservation constraint is written as,
   \begin{align}\label{equ:flow}
   \sum_{u_{t-1}:(u_{t-1}, v_t) \in \ccalE_T}   s_{u_{t-1} v_t rj} &  =   \sum_{w_{t+1}:(v_t, w_{t+1}) \in \ccalE_T} s_{v_t w_{t+1} rj},
   \end{align}
 for all robots $[r,j]\in \ccalR$ and all $t=1,\ldots,T-1$. This means that every robot that arrives at a vertex in $\ccalG_T$ has to leave that vertex at the next time step. Next, the constraints at the initial time are encoded as,
 \begin{subequations}\label{equ:source}
   \begin{align}
     \sum_{v_1:(u_0, v_1) \in \ccalE_T} s_{u_0v_1 rj} &  = 1,  \\
     \sum_{v_1:(w_0, v_1) \in \ccalE_T} s_{w_0v_1 rj} & = 0, \quad \forall\, w \in E\setminus u,
   \end{align}
 \end{subequations}
 for all robots $[r,j]\in \ccalR$, where $u_0$ is the vertex associated with the cell $u = X_I(r,j)$ where robot $[r,j]$ is at the initial time. Constraints~\eqref{equ:source} state that robot $[r,j]$ has to depart from its initial location. Note that this departure is in the graph $\ccalG_T$ and is associated  with time rather than physical location.

 \subsubsection{Target constraint}
 The general constraints that robot $[r,j]$ in $\ccalR_1$ and $\ccalR_{1,2}$  arrives at a cell   in the target region $X_G(r,j)$ at certain time instant $t$ can be encoded as
   \begin{align}\label{equ:sink}
   \sum_{v: v \in X_G(r,j) } \;\sum_{u_{t-1}:(u_{t-1}, v_{t}) \in \ccalE_T} s_{u_{t-1} v_{t} rj} &  = 1.
   \end{align}
  Specifically, $t$ in constraint~\eqref{equ:sink} takes  values ranging from $1, \ldots, T-1$ when encoding the constraint that robot $[r,j] \in\ccalR_1$  stays at the target region $X_G(r,j)$ to maintain the truth of  the vertex label of the current subtask.  For the constraint that robot $[r,j]\in\ccalR_{1,2} $ arrives at a cell in $X_G(r,j)$ at the time instant $T$ to complete the current subtask, we have $t = T$ in constraint~\eqref{equ:sink}. 

  \subsubsection{Running and terminal constraints} The general running and terminal constraints that negative literals $\neg \ap{i}{j}{k}$ should be respected at certain time instant $t$ is written as
 \begin{align}\label{equ:avoid}
   \sum_{[r,j]\in \ccalK_j} \sum_{v \in \ell_k}   \sum_{u_{t-1}: (u_{t-1}, v_{t}) \in \ccalE_T} s_{u_{t-1} v_{t} r j } \leq i - 1.
 \end{align}
 {The running constraint that all negative literals $\neg\ap{i}{j}{k} \in \mathsf{lits}^-(\gamma_1^-)$ in the vertex label} of the current subtask should be respected (excluding time instants $0 $ and $T$), can be encoded by assigning to $t$ in constraint~\eqref{equ:avoid}, values ranging from $1$ to $T-1$. Similarly, we encode the terminal constraint that the negative literal  $\neg\ap{i}{j}{k}$ in $\mathsf{lits}^-(\gamma_{1,2}^-)$  should be satisfied at the time $T$ by letting $t$ in constraint~\eqref{equ:avoid} take the value $T$.
 \subsubsection{ILP objective} The ILP objective is to minimize the total travel cost and is defined as
 \begin{align}
   \text{min} \sum_{[r,j] \in \ccalR} \; \sum_{t\in \{0,\ldots, T-1\}} \sum_{(u_t, v_{t+1}) \in \ccalE_T} d_{uv} s_{u_{t} v_{t+1} r j},
 \end{align}
 where $d_{uv} \in \mathbb{N}$ is the travel cost between cells $u$ and $v$.

 When a solution does not exist for a given horizon $T$, we increment $T$ and solve the ILP again.  The solution provides  a collection of executable paths that satisfy  the current subtask   as well as activate the next subtask at time $T$.

 \subsection{Implementations of GMRPP}\label{sec:extension_gmrpp}
 In this section, we present several implementations of the GMRPP problem. We first address the collision avoidance between robots, then we  show  how essential robots of subsequent subtasks can simultaneously move with those of the current subtask, and finally show  how only necessary robots move.

 \subsubsection{Collision avoidance}\label{sec:extension_collision}
 To handle collision avoidance, we first introduce an additional   step to  pre-process the NBA $\autop$ (see Section~\ref{sec:nba}), which removes infeasible clauses due to limited size of regions:
 \setcounter{para}{5}

 \mysubparagraph{Violation of region size}{} \label{prune:violation2} For each clause $\cp{p}{\gamma} \in \clause{\gamma}$, let $\mathsf{lits}^+(k)$ denote literals in $\mathsf{lits}^+(\cp{p}{\gamma})$ that involve region $\ell_{k}$. We delete the clause $\cp{p}{\gamma}$ (replacing it with $\bot$) if the required total number of robots  visiting region $\ell_{k}$ exceeds the number of free cells it covers, i.e., if there exists $k\in[l]$ such that $ \sum_{\ap{i}{j}{k,\chi}\in\mathsf{lits}^+(k)}   i > |\ell_k|$.

 Collision avoidance is  addressed in the low-level path planning component of our algorithm    since the high-level plan generation abstracts away the workspace. In aninstance of a GMRPP, we say that the paths of  any two distinct robots $[r,j]$ and $[r', j']$ are collision-free if there does not exist  a time instant $t \in [T]$ such that $\tau'_{r,j}(t) = \tau'_{r',j'}(t)$ (meet collision, that is, two robots occupy the same cell at the same time) or  $\tau'_{r,j}(t) = \tau'_{r',j'}(t-1) \wedge \tau'_{r',j'}(t) = \tau'_{r,j}(t-1)$ (head-on collision, that is, two robots at adjacent cells switch locations with each other). Furthermore, in the case of  the partial execution  that will be introduced in Appendix~\ref{sec:extension_partial}, we treat those robots that are not allowed to move  as obstacles, giving rise to a new workspace $E'=(S', \to_{E'})$. In the case of full execution, we have $E' = E$.
 The time-expanded graph in Fig.~\ref{fig:mapp} that captures the connectivity of the workspace is constructed based on the new workspace $E'$. Finally, we add the following collision avoidance constraints to the ILP for the GMRPP.

 Avoiding meet collisions,  $\forall v \in E'$, can be captured by the constraint
 \begin{align}\label{eq:meet}
  \sum_{[r,j]\in\ccalR}  \, \sum_{u_t: (u_t, v_{t+1}) \in \ccalE_T} s_{u_t v_{t+1} r j } \leq 1,  \; \forall \,(u_t, v_{t+1}) \in \ccalE_T,
 \end{align}
 for all $t=0,\ldots,T-1$. Moreover, avoiding head-on collisions at every  gadget, $\forall u, v \in E'$ with $u\not= v$ can be captured by the constraint
 \begin{align}\label{eq:headon}
  \sum_{[r,j]\in \ccalR} \left( s_{u_t v_{t+1} r j } + s_{v_t u_{t+1} r j } \right) \leq 1, \; \forall \,(u_t, v_{t+1}) \in \ccalE_T,
 \end{align}
 for all $t=0,\ldots,T-1$.

 \subsubsection{Simultaneous execution}\label{sec:extension_essential}
 When identifying robots  that are involved in one instance of a GMRPP in Appendix~\ref{sec:gmmpp1}, we only focused on robots $\ccalR_e =\ccalR_1 \cup \ccalR_{1,2} \cup \ccalR^-$ that are directly involved in the completion of the current subtask (see~\ref{sec:essential_a}-\ref{sec:essential_c}). However, the rest of robots that are not involved in the current subtask may concurrently move to begin the execution of  subsequent subtasks of  the current subtask $e$. Specifically, these  robots can move  towards  waypoints associated with subsequent subtasks. In what follows, we find essential robots associated with these subsequent subtasks.

 \setcounter{para}{3}
  \mysubparagraph{ Essential robots associated with subsequent subtasks}{} These robots move simultaneously with the first two types of robots in \ref{sec:essential_a}-\ref{sec:essential_b}  towards  waypoints associated with subsequent subtasks of the current subtask $e$. We collect these robots in the set $\ccalR'_{1,2}$ and identify them in the following way. First, we identify the completion time  of the current subtask, which is given by $\vec{t}(c+1)$.  Next, we iterate over the remaining robots that are not in $\ccalR_1 \cup \ccalR_{1,2}$ since they have been assigned target locations. For every robot $[r,j] \in \ccalR \setminus(\ccalR_1 \cup \ccalR_{1,2})$, the time when it should visit the next waypoint based on its local counter $\zeta_{r,j}$ is given by $t_{r,j}(\zeta_{r,j}+1)$. Note that {$t_{r, j}(\zeta_{r,j}+1) > \vec{t}(c+1)$} since we proceed along the simple path $\theta_{\phi}$ and the completion time of  subtasks that have not been considered 
  will be larger than that of the current subtask. Finally, we calculate the time difference $\Delta t = t_{r, j}(\zeta_{r,j}+1) - \vec{t}(c+1)$  and then check whether the robot $[r, j]$ can arrive at the target region $p_{r,j}(\zeta_{r,j}+1)$  within time $\Delta t$ starting from  its current location by taking the shortest route. {If not, robot $[r, j]$ should move simultaneously when completing the current subtask.} In this case, the set of robots that are involved in some subtasks becomes $\ccalR_e = \ccalR_1 \cup \ccalR_{1,2} \cup \ccalR'_{1,2}\cup \ccalR^-$.

  Next, we determine the target location $X_G(r,j)$ for robot $[r,j] \in \ccalR'_{1,2}$, which is the  location from where robot $[r,j]$ can reach the region $p_{r,j}(\zeta(r,j)+1)$ within time $\Delta t$ by taking the shortest route in the new workspace $S'$ (obtained by treating robots in the partial execution  that do not move as obstacles). To avoid collision, if the selected target location of robot $[r,j]$ has already been assigned to another robot in $\ccalR'_{1,2}$, then we select another free cell on the shortest route to be this robot's target location, which is close to the previously selected occupied cell and has not been assigned. More importantly, if a negative literal $\neg \ap{i}{j}{k}$ exists in running or terminal constraints $\gamma_1^- \vee \gamma_{1,2}^-$,  the selected free cell for robot $[r,j]$  should not be inside region $\ell_k$. In the worst scenario where such a free cell is not available for robot $[r,j]\in \ccalR_{1,2}'$, then we do not assign a specific target location to it, similar to the sequential execution. After determining the target location,  the requirement in Definition~\ref{defn:gmmpp} of GMRPP on robot $[r,j] \in \ccalR'_{1,2}$ is that it should arrive at the target waypoint at time $T$, that is, we need to design the path $\tau'_{r,j}$ such that $\tau'_{r,j}(0) = X_I(r,j)$ and $\tau'_{r,j}(T)= X_G(r,j)$. Note that the target $X_G(r,j)$ for $[r,j] \in \ccalR_{1,2}'$ is a single cell, other than a region for robots in $\ccalR_{1,2}$.
  Similar to robots in $\ccalR_{1,2}$ that complete the current subtask, this requirement can be encoded by setting $t$ equal to $T$ in constraint~\eqref{equ:sink} that handles the target constraint.

  \subsubsection{Partial execution}\label{sec:extension_partial}
  In Appendix~\ref{sec:gmmpp1}, all robots are involved in the formulation of GMRPP, which leads to a large  ILP in Appendix~\ref{sec:solution2mapp} when the size of robots is large.  To reduce the complexity, we introduce the partial execution in which only necessary robots are allowed to move and the remaining are treated as obstacles.

  First, we identify robots that need to move, which include essential robots in $\ccalR_1 \cup \ccalR_{1,2}$ and $\ccalR_{1,2}'$ when simultaneous execution is adopted. These robots have target locations. In the full execution, the set $\ccalR^-$ is defined as $\ccalR^- = \left\{\ccalK_j: \neg \ap{i}{j}{k} \in \mathsf{lits}^- (\gamma_1^-  \vee \gamma_{1,2}^-) \right\}$, which contains all robots whose types are involved in the running and terminal constraints. To shrink the size of $\ccalR^-$, for every negative literal $\neg \ap{i}{j}{k} \in \gamma_1^- \vee \gamma_{1,2}^-$, we identify the number $i'$ of robots of type $j$ that is at region $\ell_k$ at time instant 0 in each instance of GMRPP. If $i' < i$, we remove robots of type $j$, i.e., $\ccalK_j$, from $\ccalR^-$ and update this literal $\neg \ap{i}{j}{k}$ to  $\neg \ap{i-i'}{j}{k}$. Note that the case  $i' \geq i$ only happens when $\neg \ap{i}{j}{k} \in \gamma_{1,2}^-$ since at each instance of GMRPP robot locations at time instant 0 satisfy the starting vertex label, thus also satisfy $\gamma_1^-$. In this case, we replace $\ccalK_j$ in $\ccalR^-$ with $i'-i+1$ robots of type $j$ that are at region $\ell_k$ and update this literal $\neg \ap{i}{j}{k}$ to  $\neg \ap{1}{j}{k}$. In this way, we  reduce the number of robots in $\ccalR^-$. Then, the robots that need to move constitute the set $\ccalR_e = \ccalR_1 \cup \ccalR_{1,2} \cup \ccalR_{1,2}' \cup \ccalR^-$, and the remaining $\ccalR_0 = \ccalR \setminus \ccalR_e$ are treated as obstacles, giving rise to a new workspace $E' = (S', \to_{E'})$.

 The formulation of ILP to solve the GMRPP remains the same except that no variables are created corresponding to unmoved robots. {After obtaining a solution to the GMRPP, we follow similar steps as in lines~\ref{seq:path}-\ref{seq:c} in Alg.~\ref{alg:sequentialMAPP} to update relevant terms such as paths and timelines.} The exception is that in the full execution, we can concatenate paths for each robot in $\ccalR$ [line~\ref{seq:path}], while in the partial execution, the GMRPP only finds paths for robots in $\ccalR_e$. For other robots $[r,j]\in \ccalR_0$, we append $T_e$ times the last waypoint of the already-executed $\tau_{r,j}$ to $\tau_{r,j}$ since they remain idle.

 \section{Proof of Theorem~\ref{thm:completeness}}\label{app:correctness}
 {To prove Theorem~\ref{thm:completeness} that shows  the completeness of our proposed method,  we first show the completeness of the construction of the prefix part and then the completeness of the whole algorithm.} For each part, we analyze the feasibility of the MILP for the time-stamped task allocation plans  and the feasibility of the GMRPP for the low-level paths. Before presenting the main results, we first provide some necessary notation.
 \subsection{Notation}
 Given a NBA $\auto{}$, e.g., $\autop, \auto{relax}$ and $\auto{subtask}$, we define by $\ccalL_E(\auto{})$  the set of words in  $\ccalL(\auto{})$ that can be realized by robot paths. Recall that we can always map a run  in $\auto{}$ to its counterpart in $\autop$. If $\auto{}=\autop$, then the counterpart of a run is the run itself. We define by $\ccalL^\phi_E(\auto{})$  the set of words in $\ccalL_E(\auto{})$ such that for any word $w \in \ccalL^\phi_E(\auto{})$ that induces an accepting run in $\auto{}$, the counterpart of this accepting run in $\autop$ is a restricted accepting run. {In words, if $\auto{} = \auto{subtask}^-$, and if a path $\tau$ generates a word $w$ in $\ccalL_E^\phi(\auto{subtask}^-) \subseteq \ccalL_E(\auto{subtask}^-)$ and  $w$ induces a run $\rho$ in $\auto{subtask}^-$ connecting a pair $v_0$ and $\vertex{accept}$, then, we can obtain the corresponding run $\rho_\phi$ in $\autop$ that is the counterpart of the run $\rho$ in $\auto{subtask}^-$. This motivates us to modify the path $\tau$ that satisfies $\auto{subtask}^-$ to get another path that can produce this run $\rho_\phi$ in $\autop$.}   Additionally, let $\tilde{\ccalL}^{\phi}_E(\auto{}) \subseteq \ccalL^{\phi}_E(\auto{})$ collect those words in $\ccalL^{\phi}_E(\auto{})$ that can be generated by paths that satisfy Assumption~\ref{asmp:same}.

 Next, we consider the prefix and suffix parts separately. Given a pair of initial and accepting vertices, $v_0$ and $v_\text{accept}$, let $\ccalL^{\phi, v_0 \scriptveryshortarrow \vertex{accept}}_E(\auto{})$ be the set that collects finite realizable words that can generate a run in $\auto{}$ connecting $v_0$ and  $\vertex{accept}$, and further the corresponding run in $\autop$ satisfies the requirements on the prefix part of a restricted accepting run (see conditions~\ref{cond:a}-\ref{cond:d} in Definition~\ref{defn:run}). Recall that when building the sub-NBA $\auto{subtask}^-$ for the suffix part, we rely on the last subtask $(\vertex{prior}, \vertex{accept})$ in $\auto{subtask}^-$ for the prefix part to extract the sub-NBA $\auto{subtask}$ for the suffix part. {That is, we remove all outgoing edges from $\vertex{accept}$ (acting as $v_0$) from $\auto{relax}$ if $\gamma_{\phi}(\vertex{prior}, \vertex{accept})$ does not imply its edge label in $\autop$. Also, we remove all incoming edges to $\vertex{accept}$ (acting as $\vertex{accept}$) from $\auto{relax}$ if the corresponding edge label in $\autop$ is not implied by $\gamma_\phi(\vertex{prior}, \vertex{accept})$; see Appendix~\ref{sec:suf_extract}.} Also, we rely on final robot locations of the prefix part, denoted by $s_{\text{prior}}$, to interpret the augmented clause $\ccalC_{\text{prior}}^+$. {That is, $\ccalC_{\text{prior}}^+$ is satisfied if those robots involved in satisfying $\ccalC_{\text{prior}}^+$ in the prefix part return to regions including their initial locations in $s_{\text{prior}}$; see Appendix~\ref{sec:suf_milp}.} Therefore, we define by $\ccalL^{\phi, \vertex{accept} \scriptveryshortarrow \vertex{accept}}_E(\auto{}; s_{\text{prior}}, \vertex{prior})$  the set that collects finite realizable words that  can generate runs in $\auto{}$ starting from $\vertex{accept}$ and ending at $\vertex{accept}$ whose corresponding runs in $\autop$ are the suffix parts of restricted accepting runs. Furthermore,  the path generating  a  word in this set starts from $s_{\text{prior}}$ and the prefix part of this restricted accepting run  visits $\vertex{prior}$ right before $\vertex{accept}$.

 {Finally when the context is clear, we refer to the suffix MILP as the MILP in which robots returning to their initial locations and progressing towards the accepting vertex in the NBA $\autop$ is made in two steps (see Appendix~\ref{sec:suf_milp}) and the extensions in Appendix~\ref{sec:extension} to account for specific needs are not considered, and refer to the GMRPP as the GMRPP in which various implementations in Appendix~\ref{sec:extension_gmrpp}, collision avoidance, simultaneous execution and partial execution, are not considered. In what follows, we  present the main results.}

 \subsection{Existence of feasible paths in the sub-NBA $\auto{subtask}^-$}
  The following proposition  states that paths exist that can induce accepting runs in the pruned sub-NBA $\auto{subtask}^-$. This result  will be used to show the feasibility of the MILP for the time-stamped task allocation plan.
 \begin{prop}[Feasible paths for the sub-NBA $\auto{subtask}^-$]\label{thm:nba}
   Given a workspace satisfying Assumption~\ref{asmp:env} and a valid specification $\phi\in \textit{LTL}^\chi$, if there exists a path $\tau = \tau^\textup{pre} [\tau^\textup{suf}]^\omega$ inducing a restricted accepting run $\rho =  \rho^\textup{pre} [\rho^\textup{suf}]^\omega=  v_0, \ldots, \vertex{prior}, \vertex{accept} [\vertex{next}, \ldots, \vertex{prior}', \vertex{accept}]^\omega$ in $\autop$ and {this path satisfies Assumption~\ref{asmp:same}}, then there exists another path $\overline{\tau} = \overline{\tau}^{\textup{pre}}[\overline{\tau}^{\textup{suf}}]^\omega$ such that $\overline{\tau}^{\textup{pre}}$ generates a word in $\tilde{\ccalL}^{\phi,v_0\scriptveryshortarrow \vertex{accept}}_E(\auto{subtask}^-) \not= \emptyset$. Furthermore, if $ \vertex{accept} \neq \vertex{next}$, then $\overline{\tau}^{\textup{suf}}$ generates a word in  $\tilde{\ccalL}^{\phi,\vertex{accept} \scriptveryshortarrow \vertex{accept}}_E(\auto{subtask}^-; s_{\textup{prior}},v_\textup{prior}) \not= \emptyset$, where $s_\textup{prior}$ are the final robot locations of the  prefix path $\tau^{\textup{pre}}$.
 \end{prop}

 To prove Proposition~\ref{thm:nba},  we recall the main steps  to obtain the sub-NBA $\auto{subtask}^-$ in Sections~\ref{sec:prune} and~\ref{sec:pregraph} and characterize the relations between the different NBAs; see Lemmas~\ref{prop:prune}-\ref{prop:sub-NBA2}. The first lemma shows that the pruning steps in Section~\ref{sec:prune} do not affect the set of restricted accepting runs  in $\autop$ that can be incuded by realizable words.
  \begin{lem}[$\autop$ and $\autop^-$]\label{prop:prune}
 The  pruning steps in Section~\ref{sec:prune} satisfy  $\ccalL_E^\phi(\autop^-) = \ccalL^\phi_E(\autop)$.
   \end{lem}
  The proof can be found in Appendix~\ref{app:prune}. Note that any word in $\ccalL_E^\phi(\autop)$ induces a restricted accepting run in $\autop$.  A direct consequence of Lemma~\ref{prop:prune} is that, any path  generating a word $w \in \ccalL_E^\phi(\autop^-)$ satisfies  $\phi$ since the word $w$ also belongs to $\ccalL^\phi_E(\autop)$. {The following lemma shows that ignoring negative literals  expands the set of realizable words accepted by $\auto{relax}$ compared to that of $\autop^-$.}
   \begin{lem}[$\autop^-$ and $\auto{relax}$]\label{prop:inclusion}
 The relaxation stage that replaces all negative literals with $\top$ in Section~\ref{sec:prune}, satisfies   $\ccalL^\phi_E(\autop^-) \subseteq \ccalL^\phi_E(\auto{relax})$.
   \end{lem}
   The proof can be found in Appendix~\ref{app:inclusion}. Lemma~\ref{prop:inclusion} implies that a word in $\ccalL^\phi_E(\auto{relax})$ may not belong to $\ccalL^\phi_E(\autop^-)$. Hence, a path generating a word in $\ccalL^\phi_E(\auto{relax})$ may not satisfy the specification $\phi$  since $\auto{relax}$ ignores the negative literals. Next, we consider the prefix part.  The following two lemmas  show that extraction and pruning of the sub-NBA $\auto{subtask}$ for the prefix part do not empty the subset of words in $\ccalL^{\phi, v_0 \scriptveryshortarrow \vertex{accept}}_E(\auto{subtask}^-)$ that can be generated by feasible paths satisfying Assumption~\ref{asmp:same}. The detailed proofs can be found in Appendices~\ref{app:nonempty}-~\ref{app:sub-NBA}.
   \begin{lem}[$\auto{relax}$ and $\auto{subtask}$]\label{prop:nonempty}
 The extraction of the sub-NBA $\auto{subtask}$ in Section~\ref{sub-NBA:1} satisfies  $\ccalL^{\phi,v_0\scriptveryshortarrow \vertex{accept}}_E(\auto{relax}) = \ccalL^{\phi,v_0\scriptveryshortarrow \vertex{accept}}_E(\auto{subtask})$.
   \end{lem}

  \begin{lem}[$\auto{subtask}$ and $\auto{subtask}^-$]\label{prop:sub-NBA}
 The pruning steps in Section~\ref{sub-NBA:2} satisfy $\ccalL^{\phi,v_0\scriptveryshortarrow \vertex{accept}}_E(\auto{subtask}^-) \subseteq \ccalL^{\phi,v_0\scriptveryshortarrow \vertex{accept}}_E(\auto{subtask})$. {Additionally, if there exists a path $\tau = \tau^\textup{pre} [\tau^\textup{suf}]^\omega$ inducing a restricted accepting run in $\autop$ and this path satisfies Assumption~\ref{asmp:same},  then there exists a path $\overline{\tau}^{\textup{pre}}$, modified from $\tau^{\textup{pre}}$,  that can generate a word in $\tilde{\ccalL}^{\phi,v_0\scriptveryshortarrow \vertex{accept}}_E(\auto{subtask}^-)$, i.e.,  $\tilde{\ccalL}^{\phi,v_0\scriptveryshortarrow \vertex{accept}}_E(\auto{subtask}^-) \not= \emptyset$.}
  \end{lem}

  {The following corollary is a direct consequence of the proof of Lemma~\ref{prop:sub-NBA}, which implies that we can construct a sub-NBA $\auto{subtask}^-$ based on $s_{\text{prior}}$ and $v_{\text{prior}}$ obtained from the sub-NBA $\auto{subtask}^-$ for the prefix part.}
  \begin{cor}[$s_{\text{prior}}$ and $v_\text{prior}$]\label{prop:path}
   The  final configuration of the path $\overline{\tau}^{\textup{pre}}$ is $s_{\textup{prior}}$, same as the final configuration of the prefix path $\tau^{\textup{pre}}$, and the induced run  visits $v_{\textup{prior}}$ right before $\vertex{accept}$, same as the induced run   $\rho^{\textup{pre}}$.
  \end{cor}
  The following proposition  draws conclusions similar to Lemma~\ref{prop:nonempty} and~\ref{prop:sub-NBA} for the suffix part; see Appendix~\ref{app:sub-NBA2} for the proof.
  \begin{lem}[$\auto{subtask}$ and $\auto{subtask}^-$]\label{prop:sub-NBA2}
 The extraction and pruning  steps in Appendix~\ref{sec:suf_prune} satisfy $\ccalL^{\phi,\vertex{accept}\scriptveryshortarrow \vertex{accept}}_E(\auto{relax}; s_{\textup{prior}},\vertex{prior}) = \ccalL^{\phi,\vertex{accept}\scriptveryshortarrow \vertex{accept}}_E(\auto{subtask}; s_{\textup{prior}}, \vertex{prior})$. {Additionally, if there exists a path $\tau = \tau^\textup{pre} [\tau^\textup{suf}]^\omega$ inducing a restricted accepting run in $\autop$ and this path satisfies Assumption~\ref{asmp:same},  then there exists a path $\overline{\tau}^{\textup{suf}}$, modified from $\tau^{\textup{suf}}$, that can generate a word in  $\tilde{\ccalL}^{\phi, \vertex{accept} \scriptveryshortarrow \vertex{accept}}_E(\auto{subtask}^-; s_{\textup{prior}}, \vertex{prior})$, i.e.,   $\tilde{\ccalL}^{\phi, \vertex{accept} \scriptveryshortarrow \vertex{accept}}_E(\auto{subtask}^-; s_{\textup{prior}}, \vertex{prior}) \not= \emptyset$.}
  \end{lem}

  Finally, Proposition~\ref{thm:nba} can be established by combining Lemmas~\ref{prop:sub-NBA} and~\ref{prop:sub-NBA2}.

 \subsection{Completeness}\label{app:completeness}
 \subsubsection{{Completeness of the prefix part synthesis}}
 The following proposition states that, with mild assumptions, we can find a path that induces a run in $\autop$ connecting  $v_0$ and $\vertex{accept}$, which ensures the completeness of our method for specifications in \ltlx that can be satisfied by finite-length paths.
 \begin{prop}[Completeness of the synthesis method for the prefix part]\label{thm:prefix}
 Assume a workspace that satisfies   Assumption~\ref{asmp:env} and a valid specification $\phi\in \textit{LTL}^\chi$. If there exists a path $\tau = \tau^\textup{pre} [\tau^\textup{suf}]^\omega$  that induces a restricted accepting run $\rho =  \rho^\textup{pre} [\rho^\textup{suf}]^\omega = v_0, \ldots, \vertex{prior}, \vertex{accept} [\vertex{next}, \ldots, \vertex{prior}', \vertex{accept}]^\omega$ in $\autop$ and this path satisfies Assumption~\ref{asmp:same}, then the proposed synthesis method  can find a robot path $\tilde{\tau}^{\textup{pre}}$ that generates a word $\tilde{w}^{\textup{pre}}$ that induces a run $\tilde{\rho}^{\textup{pre}}$ in $\autop$  connecting the pair $v_0$ and $\vertex{accept}$. 
 \end{prop}

 We first provide the following three lemmas and then combine with Proposition~\ref{thm:nba} to conclude the proof of Proposition~\ref{thm:prefix}.
 Lemma~\ref{prop:feasibility} states that, if the poset $P$ is inferred from a set of simple paths that includes  a  simple path associated with a feasible prefix path, then the MILP in Appendix~\ref{app:appendix_prefix_milp} associated with this poset $P$ is feasible; {Lemma~\ref{prop:run} states that a simple path in $\auto{subtask}^-$ can be extracted from the solution to the MILP and discusses
   the temporal properties associated with this path,} and Lemma~\ref{prop:valid} states that the  sequence of GMRPP in Section~\ref{sec:lowlevel} associated with the extracted
 simple path is feasible.
  \begin{lem}[{Feasibility of the prefix MILP}]\label{prop:feasibility}
 If there exists a path $\overline{\tau}^{\textup{pre}}$ generating a finite word $\overline{w}^{\textup{pre}} \in \tilde{\ccalL}^{\phi,v_0\scriptveryshortarrow \vertex{accept}}_E(\auto{subtask}^-)$, and the word $\overline{w}^{\textup{pre}}$ induces a simple path $\overline{\theta}^{\textup{pre}}$ in $\auto{subtask}^-$ that belongs to the set of simple paths that generate the  poset $P$,
 then  the  prefix MILP in Appendix~\ref{app:appendix_prefix_milp} associated with this poset $P$  is feasible. 
  \end{lem}
     The detailed proof can be found in Appendix~\ref{app:feasibility}. The key idea is that, the path $\overline{\tau}^{\textup{pre}}$  generating the word $\overline{w}^{\textup{pre}}$ can give rise to a high-level plan  satisfying constraints~\eqref{eq:1}-\eqref{eq:lastclause} in Appendix~\ref{app:appendix_prefix_milp}.

 \begin{lem}[Properties of the simple path]\label{prop:run}
   If the MILP for the prefix path in Appendix~\ref{app:appendix_prefix_milp} associated with the poset $P$ produces a solution, then a simple path $\tilde{\theta}$ designed in Appendix~\ref{sec:run}  that belongs to the set of simple paths that generate the poset $P$, can be extracted from the sub-NBA $\auto{subtask}^-$. Additionally, the following properties hold for subtasks in the simple path $\tilde{\theta}$:
   \begin{noindlist}
   \item \label{property:a} The first subtask in the simple path $\tilde{\theta}$ is activated at time 0;
     \item \label{property:b}  For any subtask  $e \in \tilde{\theta}$, if its starting vertex has a self-loop, then the completion time of the subtask $e$ is no earlier than the activation of its starting vertex label, and at most one time step after the completion of its starting  vertex label;
   \item \label{property:c}  For any two consecutive subtasks  $e, e'\in \tilde{\theta}$, the latter subtask $e'$ is activated at most one time step after the completion of the former subtask $e$.
   \end{noindlist}
 \end{lem}

 The detailed proof can be found in Appendix~\ref{app:run}. Property~\ref{property:a} guarantees the initialization of the sequence of subtasks in $\tilde{\theta}$, property~\ref{property:b} ensures that each subtask in $\tilde{\theta}$ is correctly executed, and property~\ref{property:c} prevents gaps when transitioning between consecutive subtasks. Combined these three properties establish that once the first subtask is activated at time 0, each subsequent subtask  is completed successfully and inter-subtasks transitions occur  seamlessly, until the completion of the last subtask. Note that the extracted simple path $\tilde{\theta}$ may not be identical to the one induced by the word $\overline{w}^{\text{pre}} \in \tilde{\ccalL}_E^{\phi, v_0 \scriptveryshortarrow \vertex{accept}} (\auto{subtask}^-)$ in Lemma~\ref{prop:feasibility}. {The following lemma states that low-level  paths can be generated from the solution to the prefix MILP in Appendix~\ref{app:appendix_prefix_milp}}; the  proof can be found in Appendix~\ref{app:valid}.
  \begin{lem}[Feasibility of the GMRPP in Appendix~\ref{sec:mapp}]\label{prop:valid}
  Assume that the workspace satisfies Assumption~\ref{asmp:env}. Then, the sequence of the GMRPP in Appendix~\ref{sec:mapp} that constructs the simple path $\tilde{\theta}$ from the solution to the MILP in Appendix~\ref{app:appendix_prefix_milp}, is feasible. That is, every GMRPP is feasible for a time horizon $T$.
  \end{lem}
  {Combining Proposition~\ref{thm:nba} with Lemmas~\ref{prop:feasibility}-\ref{prop:valid}, we conclude the proof of Proposition~\ref{thm:prefix} on the completeness of the proposed synthesis method for the prefix part.} Note that Proposition~\ref{thm:prefix} assumes the existence of a feasible path.  By  Proposition~\ref{thm:nba},  $\tilde{\ccalL}_E^{\phi, v_0 \scriptveryshortarrow \vertex{accept}}(\auto{subtask}^-) \neq \emptyset$, therefore a path $\overline{\tau}^{\text{pre}}$ exists that generates a word $\overline{w}^{\text{pre}} \in \tilde{\ccalL}_E^{\phi, v_0 \scriptveryshortarrow \vertex{accept}}(\auto{subtask}^-) $.  Because in Section~\ref{sec:sort} we iterate over all pairs of initial and accepting vertices whose total length is not infinite, we can focus on the NBA $\auto{subtask}^-$ associated with a pair that produces a feasible path, as required in Proposition~\ref{thm:prefix}. {Moreover, since in Section~\ref{sec:poset} we create posets for all subsets of equivalent simple paths connecting the pair $v_0$ and $\vertex{accept}$, by iterating over all these posets we are guaranteed to eventually formulate the MILP over the poset that includes the simple path induced by the path $\overline{\tau}^{\text{pre}}$.} According to Lemma~\ref{prop:feasibility}, this MILP has a solution. Then, by Lemma~\ref{prop:valid}, we get that a path $\tilde{\tau}^{\textup{pre}}$ can be obtained by concatenating paths from each GMRPP since the final and initial locations of consecutive GMRPPs are identical, which completes the proof.

 \subsubsection{Completeness of the overall algorithm}
 Similar to Lemma~\ref{prop:feasibility}, we show the feasibility of the suffix MILP in Appendix~\ref{app:appendix_suffix_milp}; see Appendix~\ref{app:suffix_feasibility} for the detailed proof.
 \begin{lem}[{Feasibility of the suffix MILP}]\label{prop:suffix_feasibility}
 Assume a valid specification $\phi\in \textit{LTL}^0$ and let  $s_{\textup{prior}}$ and $v_{\textup{prior}}$ denote the final configuration of the path $\tau^{\textup{pre}}$ and the vertex before $\vertex{accept}$ in the  run $\rho^{\textup{pre}}$, respectively. If there exists a path $\overline{\tau}^{\textup{suf}}$ generating a finite word $\overline{w}^{\textup{suf}} \in \tilde{\ccalL}^{\phi,\vertex{accept}\scriptveryshortarrow \vertex{accept}}_E(\auto{subtask}^-; s_{\textup{prior}}, v_{\textup{prior}})$, and the word $\overline{w}^{\textup{suf}}$ induces a simple path  $\overline{\theta}^{\textup{suf}}$ belonging to  the set of simple paths  that generate  the poset $P$,
   then the  suffix MILP  associated with this poset $P$, composed of constraints~\eqref{eq:1}-\eqref{eq:same} in Appendix~\ref{app:appendix_prefix_milp} and constraints~\eqref{eq:one_suffix}-\eqref{eq:return_suffix} in Appendix~\ref{app:appendix_suffix_milp},  is feasible.
 \end{lem}

 Note that the results in Lemmas~\ref{prop:run} and~\ref{prop:valid} developed for the prefix part can also be applied  to the suffix part. {Combined with Lemma~\ref{prop:suffix_feasibility}, we can obtain an equivalent statement of Proposition~\ref{thm:prefix} for the suffix part. Finally, combining Proposition~\ref{thm:prefix} for the prefix part with Proposition~\ref{thm:nba}, we can establish completeness of our proposed method;  see Appendix~\ref{app:completeness_}.}

 {\rem{
  Note that Lemma~\ref{prop:suffix_feasibility} focuses on \ltlz specifications. The prefix path $\tilde{\tau}^{\text{pre}}$ in Proposition~\ref{thm:prefix} is obtained by solving the prefix MILP in Appendix~\ref{app:appendix_prefix_milp}, which may allocate a different fleet of robots than $\overline{\tau}^{\text{pre}}$ to satisfy the same induced atomic propositions. Since robots also need  to return to initial locations, this may affect the existence of  a suffix path and further the satisfaction of the same-$\ag{i}{j}$ constraint~\eqref{eq:same_suffix} in Appendix~\ref{app:appendix_suffix_milp}. Therefore, we restrict the specification to the class  \ltlz in the statement.}}

 \subsection{Detailed proofs}\label{app:proofforproof}
 \subsubsection{Proof of Lemma~\ref{prop:prune}}\label{app:prune}
 The basic idea is that the pruning steps in Section~\ref{sec:prune} do not affect any restricted accepting runs. First,  removing infeasible transitions and unreachable vertices will not exclude any realizable words. Therefore, the set of realizable words $\ccalL_E(\autop)$ does not change after this operation. Second, according to condition \ref{cond:c} in Definition~\ref{defn:run}, any restricted  accepting run  does not contain vertices without self-loops except for initial and accepting vertices. Thus, removing such vertices  will not affect the  set of restricted accepting runs in $\autop$. Third, by condition~\ref{cond:d}, any edge whose label does not strongly imply its end vertex label, except for the case that the end vertex  is an accepting vertex,  can not appear in any restricted accepting run. Thus removing such edges does not affect the set of restricted accepting runs, either.  We conclude that the operations in Section~\ref{sec:prune}  do not affect the set of restricted accepting runs in $\autop$ that can be induced by the realizable words  in $\ccalL^\phi_E(\autop)$. Moreover, these restricted accepting runs are also accepting runs in $\autop^-$, which implies that $\ccalL^\phi_E(\autop^-) = \ccalL^\phi_E(\autop)$, completing the proof.

 \subsubsection{Proof of Lemma~\ref{prop:inclusion}}\label{app:inclusion}
 The inclusion is straightforward in that, given a clause $\ccalC$ in $\autop^-$, the clause $\ccalC'$ in $\auto{relax}$ obtained by replacing negative literals in $\ccalC$ with $\top$ is a subformula of the original clause $\ccalC$. In other words, the satisfaction of the original clause $\ccalC$ implies the satisfaction of $\ccalC'$, which implies that any realizable word $w$ in $\ccalL_E(\autop^-)$ belongs to $\ccalL_E(\auto{relax})$. Thus, any word $w \in \ccalL^\phi_E(\autop^-)  \subseteq  \ccalL_E(\autop^-)$ belongs to $\ccalL_E(\auto{relax})$.

 {Next, we prove that indeed any word $\tilde{w} \in \ccalL^\phi_E(\autop^-)$ belongs to  $\ccalL^\phi_E(\auto{relax})$ is also in  $ \ccalL_E(\auto{relax})$.} Because $\tilde{w} \in \ccalL^\phi_E(\autop^-)$, it can induce  a run in $\autop^-$ whose corresponding run in $\autop$ is a restricted accepting run. Also because $\tilde{w}$ is in $\ccalL_E(\auto{relax})$ and clauses in $\auto{relax}$ are the subformulas of the clauses in $\autop^-$, $\tilde{w}$ can induce the same run in $\auto{relax}$ as the run in $\autop^-$ (same sequence of vertices). Additionally, these two runs correspond to the same restricted accepting run in $\autop^-$. Therefore, $\tilde{w} \in \ccalL^\phi_E(\auto{relax})$, i.e.,  $\ccalL^\phi_E(\autop^-)\subseteq \ccalL^\phi_E(\auto{relax})$, completing the proof.

  \subsubsection{Proof of Lemma~\ref{prop:nonempty}}\label{app:nonempty}
  Given the pair of initial and accepting vertices, $v_0$ and $\vertex{accept}$, the corresponding sub-NBA $\auto{subtask}$ is composed of all vertices and edges in $\auto{relax}$ that belong to some paths that connect $v_0$ and $\vertex{accept}$ with two exceptions. The first exception is that   all other initial and accepting vertices other than $v_0$ and $\vertex{accept}$ are removed, and the second  exception is that the self-loop of $v_0$ (if exists) is removed if the initial robot locations do not satisfy its corresponding vertex label in $\autop$, and the outgoing edges of the initial vertex are also removed  if the initial robot locations do not satisfy their corresponding edge labels in $\autop$; see Section~\ref{sub-NBA:1}. 

  To show this result, note first that, according to condition \ref{cond:b} in the Definition~\ref{defn:run}, the prefix part  does not include more than one initial vertex and more than one accepting vertex. Since $\ccalL^{\phi,v_0\scriptveryshortarrow \vertex{accept}}_E(\auto{relax})$ are related to the initial and accepting vertices $v_0$ and $\vertex{accept}$, removing other initial and accepting vertices does not affect $\ccalL^{\phi,v_0\scriptveryshortarrow \vertex{accept}}_E(\auto{relax})$.

 Second, any word $w$ that at the beginning satisfies the label of the initial vertex $v_0$ whose self-loop is removed or labels of outgoing edges that are removed, cannot be generated by feasible paths since initial robot locations violate these labels. Therefore, any such word $w$ does not belong to $\ccalL^{\phi,v_0\scriptveryshortarrow \vertex{accept}}_E(\auto{relax})$,  meaning that removing the self-loops and outgoing edges  does not affect $\ccalL^{\phi,v_0\scriptveryshortarrow \vertex{accept}}_E(\auto{relax})$, completing the proof.

 \subsubsection{Proof of Lemma~\ref{prop:sub-NBA}}\label{app:sub-NBA}
 The inclusion relation is straightforward    since $\auto{subtask}^-$ is obtained by removing edges from  $\auto{subtask}$  that are decomposable according to the sequential triangle property (see Definition~\ref{defn:st}). In what follows, we focus on the non-emptiness of $\tilde{\ccalL}^{\phi,v_0\scriptveryshortarrow \vertex{accept}}_E(\auto{subtask}^-)$. We first show that the given prefix path $\tau^{\text{pre}}$  can generate a word in $\tilde{\ccalL}^{\phi,v_0\scriptveryshortarrow \vertex{accept}}_E(\auto{subtask})$, and then show that  based on $\tau^{\text{pre}}$ another prefix path $\overline{\tau}^{\text{pre}}$ can be synthesized to generate a word in $\tilde{\ccalL}^{\phi,v_0\scriptveryshortarrow \vertex{accept}}_E(\auto{subtask}^-)$.

 To show that the given prefix path $\tau^{\text{pre}}$ can generate a word in $\tilde{\ccalL}^{\phi,v_0\scriptveryshortarrow \vertex{accept}}_E(\auto{subtask})$,  we show that the  results in Lemmas~\ref{prop:prune}-\ref{prop:nonempty} can be applied to languages $\tilde{\ccalL}(\cdot)$ that satisfy   Assumption~\ref{asmp:same}. Note that  Assumption~\ref{asmp:same} describes  how a restricted accepting run is implemented by robots. Specifically, if an accepting  word belonging to two languages $\ccalL_E^\phi(\ccalA_1)$ and $\ccalL_E^\phi(\ccalA_2)$ can be generated by robot paths that satisfy Assumption~\ref{asmp:same}, then this word should also belong to the  languages $\tilde{\ccalL}_E^\phi(\ccalA_1)$ and $\tilde{\ccalL}_E^\phi(\ccalA_2)$. Therefore, the specific implementation of  an accepting word is implemented does not affect the relation between languages. We can get that
 $ \tilde{\ccalL}_E^\phi(\ccalA_\phi^-) = \tilde{\ccalL}_E^\phi(\ccalA_\phi)$ by Lemma~\ref{prop:prune},
 $\tilde{\ccalL}_E^\phi(\ccalA_\phi^-) \subseteq \tilde{\ccalL}_E^\phi(\auto{relax})$ by Lemma~\ref{prop:inclusion},
 $\tilde{\ccalL}_E^{\phi,v_0\scriptveryshortarrow \vertex{accept}}(\auto{relax}) =  \tilde{\ccalL}_E^{\phi,v_0\scriptveryshortarrow \vertex{accept}}(\auto{subtask})$ by Lemma~\ref{prop:nonempty} and  $\tilde{\ccalL}_E^{\phi,v_0\scriptveryshortarrow \vertex{accept}}(\auto{subtask}^-) \subseteq \tilde{\ccalL}_E^{\phi,v_0\scriptveryshortarrow \vertex{accept}}(\auto{subtask})$ as discussed in the beginning of this proof.

 Next, because the accepting  word $w= w^\text{pre} [w^\text{suf}]^\omega$  generated by the path $\tau= \tau^\text{pre} [\tau^\text{suf}]^\omega$ induces a restricted accepting run and this path satisfies Assumption~\ref{asmp:same}, i.e., $w \in \tilde{\ccalL}_E^\phi(\ccalA_\phi)$, we have that  $w \in \tilde{\ccalL}_E^\phi(\ccalA_\phi^-)$ and further  $w^{\text{pre}} \in \tilde{\ccalL}_E^{\phi,v_0\scriptveryshortarrow \vertex{accept}}(\auto{relax})$.
 Therefore, $w^{\text{pre}} \in \tilde{\ccalL}_E^{\phi,v_0\scriptveryshortarrow \vertex{accept}}(\auto{subtask})$, that is, $\tau^{\text{pre}}$ generates a word in $\tilde{\ccalL}_E^{\phi,v_0\scriptveryshortarrow \vertex{accept}}(\auto{subtask})$. In what follows, we synthesize another prefix path $\overline{\tau}^{\text{pre}}$ based on $\tau^{\text{pre}}$.

 First, consider 3 different vertices $v_1, v_2, v_3$ in $\auto{subtask}$ that satisfy the ST property. Assume $\auto{subtask}$ is currently at the vertex $v_1$. We show that, given robot configuration $s$ in a path that completes the subtask $(v_1, v_3)$, i.e., a path that drives  the transition to vertex $v_3$, we can  simply repeat this robot configuration one more time so that the sub-NBA $\auto{subtask}^-$ reaches $v_3$ by  traversing edges $(v_1, v_2)$ and $(v_2, v_3)$.
 Specifically, according to Definition~\ref{defn:st} of ST property, since the robot configuration $s$ satisfies the edge label $\gamma(v_1, v_3)$ and $\gamma(v_1, v_3) = \gamma(v_1, v_2) \wedge \gamma(v_2, v_3)$, it also satisfies the edge label $\gamma(v_1, v_2)$. Thus, $s$ can drive the transition to vertex $v_2$ from $v_1$. At the next time step, if robots remain idle, the edge label $\gamma(v_2, v_3)$ can be satisfied since the robot configuration $s$ satisfies $\gamma(v_1, v_3)$ and $\gamma(v_1, v_3)$ implies $\gamma(v_2, v_3)$. Therefore, by simply repeating this robot configuration, the sub-NBA $\auto{subtask}^-$ traverses edges $(v_1, v_2)$ and $(v_2, v_3)$ to reach $v_3$, without satisfying the vertex label $\gamma(v_2)$.

 Based on this observation, we continue showing the non-emptiness of $\tilde{\ccalL}^{\phi,v_0\scriptveryshortarrow \vertex{accept}}_E(\auto{subtask}^-)$. With a slight abuse of notation, let $\rho^{\text{pre}}$ denote the run in $\auto{subtask}$ induced by the word $w^{\text{pre}}$. We assume the run $\rho^{\text{pre}}$ traverses an edge $(v_1, v_3)$ in $\auto{subtask}$ corresponding to a composite subtask, which  will be removed according to ST property. Otherwise, the run $\rho^{\text{pre}}$ will persist in $\auto{subtask}^-$. When the run $\rho^{\text{pre}}$ traverses a composite edge, we locate the robot configuration $s$ in $\tau^{\text{pre}}$ that enables this composite subtask, let the robots remain idle for one time step as discussed above, and then continue along the path $\tau^{\text{pre}}$. Let $\overline{\tau}^{\text{pre}}$ denote the new path, which also satisfies conditions~\hyperref[asmp:a]{(a)}-\hyperref[asmp:b]{(b)} in Definition~\ref{defn:same} since the robots remain idle for one time step. Furthermore, the path $\overline{\tau}^{\text{pre}}$  generates a word $\overline{w}^{\text{pre}}$ that induces a run $\overline{\rho}^{\text{pre}}$ in $\auto{subtask}^-$  traversing the two elementary edges $(v_1, v_2)$ and $(v_2, v_3)$.

 Next, we prove that since $\rho^{\text{pre}}$ is a prefix part that  satisfies the  conditions~\ref{cond:a}-\ref{cond:d} in Definition~\ref{defn:run}, so does the run  $\overline{\rho}_\phi^{\text{pre}}$ that corresponds to the run $\overline{\rho}^{\text{pre}}$. Observe that $\overline{\rho}^{\text{pre}}$ differs from $\rho^{\text{pre}}$ only in that $\rho^{\text{pre}}$ traverses edge $(v_1, v_3)$ while $\overline{\rho}^{\text{pre}}$ traverses edges $(v_1, v_2)$ and $(v_2, v_3)$ consecutively. Obviously, the run $\overline{\rho}^{\text{pre}}_{\phi}$ satisfies conditions~\ref{cond:a}-\ref{cond:c} in Definition~\ref{defn:run}.  Furthermore, $\gamma_{\phi} (v_1, v_2) \simplies_s \gamma_{\phi} (v_2)$ and $\gamma_{\phi}(v_2, v_3) \simplies_s \gamma_{\phi}(v_3)$; otherwise, they would be pruned in Section~\ref{sec:prune}. Thus, the run $\overline{\rho}^{\text{pre}}_\phi$  satisfies condition \ref{cond:d}, that is, the word $\overline{w}^{\text{pre}}$ generated by $\overline{\tau}^{\text{pre}}$ belongs to $\tilde{\ccalL}^{\phi,v_0\scriptveryshortarrow \vertex{accept}}_E(\auto{subtask}^-)$, completing the proof.

 \subsubsection{Proof of Lemma~\ref{prop:sub-NBA2}}\label{app:sub-NBA2}
 This proof is similar to the proofs of Lemmas~\ref{prop:nonempty} and~\ref{prop:sub-NBA}. Recall from Appendix~\ref{sec:suf_prune} that  to obtain the sub-NBA $\auto{subtask}$, we remove  all other accepting vertices from $\auto{relax}$, all initial vertices without self-loops, all outgoing edges from $\vertex{accept}$ if the corresponding label is not implied by the label $\gamma_\phi(\vertex{prior}, \vertex{accept})$, which is the  edge label in the NBA $\autop$ that corresponds to the last completed subtask in the prefix part, and  all incoming edges to $\vertex{accept}$ if the corresponding label is not implied by the label $\gamma_\phi(\vertex{prior}, \vertex{accept})$. According to conditions \ref{cond:b} and~\ref{cond:f} in Definition~\ref{defn:run}, the suffix part does not traverse these removed vertices and edges. Therefore, removing them does not affect $\ccalL^{\phi,\vertex{accept} \scriptveryshortarrow \vertex{accept}}_E(\auto{relax}; s_{\textup{prior}}, \vertex{prior})$, i.e., $\ccalL^{\phi,\vertex{accept} \scriptveryshortarrow \vertex{accept}}_E(\auto{relax};  s_{\textup{prior}}, \vertex{prior}) = \ccalL^{\phi,\vertex{accept} \scriptveryshortarrow \vertex{accept}}_E(\auto{subtask}; s_{\textup{prior}}, \vertex{prior})$.

 To prove that $\tilde{\ccalL}^{\phi,\vertex{accept} \scriptveryshortarrow \vertex{accept}}_E(\auto{subtask}^-; s_{\textup{prior}},\vertex{prior}) \neq \emptyset$, we follow similar steps as those in the proof of   Lemma~\ref{prop:sub-NBA}. First, the suffix path $\tau^{\text{suf}}$ can generate a word in $\tilde{\ccalL}^{\phi,\vertex{accept} \scriptveryshortarrow \vertex{accept}}_E(\auto{subtask}; s_{\textup{prior}},\vertex{prior})$, same as  $\tau^{\text{pre}}$ can generate a word in $\tilde{\ccalL}^{\phi,v_0 \scriptveryshortarrow \vertex{accept}}_E(\auto{subtask})$. {Second, the path $\overline{\tau}^{\text{suf}}$ can be obtained from $\tau^{\text{suf}}$ by  repeating the robot configuration  that completes a composite subtask one more time, so that the resulting run traverses two elementary edges successively. This drives the transition to the same vertex as that reached by traversing a composite edge.}
 Since $\vertex{accept}$ does not have a self-loop,  when $v_3 = \vertex{accept}$, we do not remove the composite edge $(v_1, v_3)$ from $\auto{subtask}^-$ for the  suffix part (see Definition~\ref{defn:st}). Thus, we can reuse the  robot configuration that enables the edge $(v_1, \vertex{accept})$, which satisfies condition~\hyperref[asmp:c]{(c)} in Definition~\ref{defn:same} that robots return to $s_{\text{prior}}$, completing the proof.

 \subsubsection{Proof of Lemma~\ref{prop:feasibility}}\label{app:feasibility}
 {Consider a path $\overline{\tau}^\text{pre}$ that generates a finite word $\overline{w}^\text{pre} \in \tilde{\ccalL}^{\phi,v_0\scriptveryshortarrow \vertex{accept}}_E(\auto{subtask}^-)$ inducing a simple path $\overline{\theta}^{\text{pre}}$ in $\auto{subtask}^-$. Then, for two consecutive subtasks in the simple path $\overline{\theta}^{\text{pre}}$, it is possible that their edge labels are satisfied by two consecutive symbols in the word $\overline{w}^\text{pre}$. In this case, the starting  vertex label of the second subtask is not satisfied by the path $\overline{\tau}^\text{pre}$.  We first show that $\overline{\tau}^{\text{pre}}$ can be used to construct  a new path such that this new path  can also induce the simple path $\overline{\theta}^{\text{pre}}$ in $\auto{subtask}^-$, and the starting  vertex label, if exists, of each subtask in the simple path $\overline{\theta}^{\text{pre}}$ is satisfied by the new path  at least once. This is because constraint~\eqref{eq:c} states that  any vertex or edge label of any subtask in a simple path must be satisfied. Since the new path is similar to the original path $\overline{\tau}^\text{pre}$, with a slight abuse of notation, we still use $\overline{\tau}^{\text{pre}}$ to denote the new path. Given the new path $\overline{\tau}^\text{pre}$,  our goal is to show that $\overline{\tau}^\text{pre}$ can generate  a time-stamped task allocation plan that can also be generated by a solution to the prefix MILP in Appendix~\ref{app:appendix_prefix_milp}.} To this end, we first obtain an {\it essential word} $w^*$ based on the word $\overline{w}^\text{pre}$  generated by the path $\overline{\tau}^\text{pre}$, such that the path $\overline{\tau}^\text{pre}$ can also generate the essential word and the essential word $w^*$ is the tightest word that can induce the same run $\overline{\rho}^\text{pre}$ as $\overline{w}^\text{pre}$ does; see Appendix~\ref{app:word}. Then, we show that the essential word $w^*$ can produce a graph that is a subgraph of the routing graph $\ccalG$ built in Appendix~\ref{sec:graph}; see Appendix~\ref{app:graph}. Finally, we show that this subgraph  can be viewed as a graphical solution to the MILP; see Appendix~\ref{app:milp}.

 The construction of new path is straightforward.  According to condition~\ref{cond:d} in Definition~\ref{defn:run}, for any subtask $e$ in the simple path $\overline{\theta}^{\text{pre}}$ that is not the first one to be completed, its starting vertex label  is  strongly implied by the edge label of the subtask $e'$ immediately preceding $e$. Therefore, when the  edge label of subtask $e'$ is enabled, robots can remain idle for one time step to satisfy the starting vertex label of subtask $e$. Also, the satisfied clause in the edge label of $e'$ implies the satisfied clause in the starting vertex label of subtask $e$. On the other hand, if subtask $e$ is the first subtask in the simple path $\overline{\theta}^{\text{pre}}$, and also the vertex $v_0$ has a self-loop, then the initial robot locations satisfy the label of $v_0$. Similarly robots can remain idle to make the label of $v_0$ true at least once. We still use $\overline{\tau}^\text{pre}$ to denote the new path as the only change is the idleness of robots. The new path $\overline{\tau}^\text{pre}$ still generates a word belonging to $\tilde{\ccalL}^{\phi,v_0\scriptveryshortarrow \vertex{accept}}_E(\auto{subtask}^-)$ and induces the same simple path $\overline{\theta}^{\text{pre}}$ as the original path. {Note that $\overline{\theta}^{\text{pre}}$ is a linear extension of the poset $P$ based on which, the prefix MILP in Appendix~\ref{app:appendix_prefix_milp} is formulated.} No two subtasks are completed at the same time in the path $\overline{\tau}^\text{pre}$.

 \paragraph{Construction of the essential word}{Given the path $\overline{\tau}^\text{pre}$ that induces the simple path $\overline{\theta}^{\text{pre}}$, let $\overline{w}^\text{pre} = \sigma_0 \sigma_1 \sigma_2\ldots \sigma_k$ denote the generated finite  word in $\tilde{\ccalL}^{\phi,v_0\scriptveryshortarrow \vertex{accept}}_E(\auto{subtask}^-)$, and $\overline{\rho}^\text{pre}=v_0 v_1 v_2\ldots \vertex{accept}$  denote the induced run in $\auto{subtask}^-$. Next we obtain  an {\it essential word}, denoted by $w^* =  \sigma^*_0 \sigma^*_1 \sigma^*_2\ldots \sigma^*_k$, such that  $\sigma^*_i \subseteq \sigma_i$ is the tightest subset of atomic propositions that enables a clause of label $\gamma(v_i, v_{i+1})$, where $v_i$ and $ v_{i+1}$ are consecutive vertices in the run $\overline{\rho}^\text{pre}$, so that  removing any atomic proposition from $\sigma^*_i$ violates this clause. We identify the satisfied clause in the label $\gamma(v_i, v_{i+1})$  and add all positive literals in this clause  to $\sigma^*_i$. If two sets of atomic propositions $\sigma_i$ and $\sigma_j$ correspond to  the  same vertex label, then $\sigma^*_i = \sigma^*_j$ since by  condition~\hyperref[asmp:a]{(a)} in Definition~\ref{defn:same}, it is always the same clause that is satisfied in a vertex label. Furthermore, if $\sigma_i$ corresponds to an edge label and $\sigma_{i+1}$ corresponds to the immediate following  vertex label, then $\sigma^*_{i+1} \subseteq \sigma^*_i$ since by condition~\hyperref[asmp:b]{(b)}, the satisfied clauses in the  edge labels implies the satisfied clauses in the immediately following vertex labels. By default, $\sigma_i^* = \{\top\}$ if $\gamma(v_i, v_{i+1})=\top$.  In this way, we have that $w^* \in \tilde{\ccalL}^{\phi,v_0\scriptveryshortarrow \vertex{accept}}_E(\auto{subtask}^-)$
 since it induces the same run $\overline{\rho}^\text{pre}$ as $\overline{w}^\text{pre}$ does, and that the path $\overline{\tau}^\text{pre}$ generating the word $\overline{w}^\text{pre}$ can generate the word $w^*$.}\label{app:word}

\begin{figure}[!t]
   \centering
   \includegraphics[width=\linewidth]{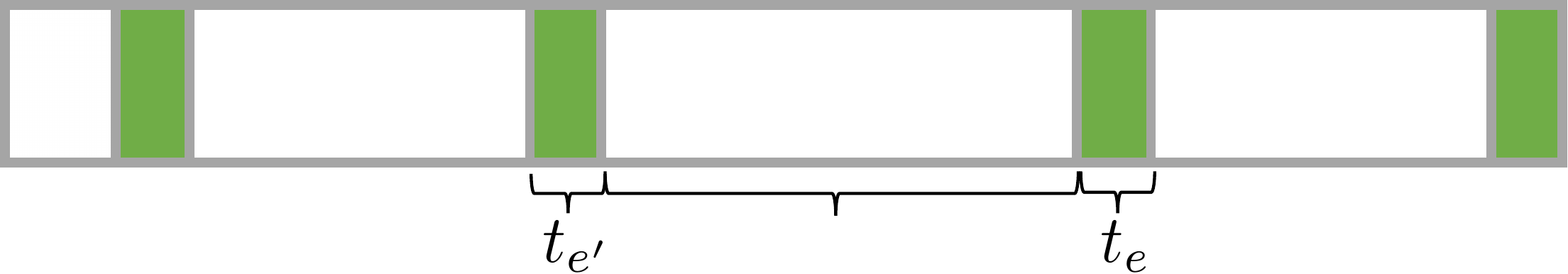}
   \caption{The divided essential word $w^*$.}
   \label{fig:word}
 \end{figure}

 \paragraph{Construction of a subgraph of the routing graph $\ccalG$}{}\label{app:graph}
 In this part, we construct a routing graph $\ccalG_{w^*}$ based on the essential word $w^*$ which is a subgraph of the routing graph $\ccalG$ built in Appendix~\ref{sec:graph}.
 Given the essential word $w^*$, we can divide it into parts by locating the components where edges in the induced run $\overline{\rho}^\text{pre}$ in $\auto{subtask}^-$ are enabled. Fig.~\ref{fig:word} demonstrates such a partition where green columns represent single time instants when edges  are enabled, i.e., subtasks are completed, e.g., time instants $t_{e'}$ and $t_{e}$ where $e'$ is the subtask that is completed immediately preceding $e$, and {the white areas between any two green columns represent the time intervals, e.g., $[t_{e'}+1, t_{e}-1]$}, when the vertex labels are satisfied. Note that $t_{e}-1 \geq t_{e'}+1$ since we adjusted the path $\overline{\tau}^{\text{pre}}$ so that each starting vertex label is satisfied at least one. In this way,  the time interval $[t_{e'}+1, t_{e}-1]$ and the time instant $t_e$ make up the time span of the  subtask $e$ in the simple path $\overline{\theta}^{\text{pre}}$. Thus, given the path $\overline{\tau}^\text{pre}$, we can obtain an array of time spans of subtasks in $\overline{\theta}^{\text{pre}}$ such that  consecutive time spans are disjoint  with  others and subtasks are completed sequentially. In what follows, we build a graph $\ccalG_{w^*}$ based on the essential word $w^*$. We begin with the vertex set.

 \mysubparagraph{Construction of the vertex set}{}\label{app:vertex}

  \subphase{Location vertices associated with initial robot locations}{First, we create the vertex set $\ccalV_{\text{init}}$ that corresponds to initial robot locations, as in Appendix~\ref{vertex:initial} for $\ccalG$. We assign visit time $t_{vr}^- = t_{vr}^+ = 0$ to each vertex $v \in \ccalV_{\text{init}}$,  where robot $r$ is the specific robot that is associated with $v$. In what follows, we also create vertices associated with clauses in edge or vertex labels that are satisfied by the path $\overline{\tau}^\text{pre}$.}\label{app:initial}

   \subphase{Literal vertices associated with edge labels}{Consider a time instant $t_e$ when the edge $e = (v_1, v_2)$ in  $\overline{\theta}^{\text{pre}}$ is enabled. The set $\sigma_{t_e}^*$ of atomic propositions  contains all literals appearing in the single clause satisfied in the edge label $\gamma(v_1, v_2)$. If $\gamma(v_1, v_2) = \top$, then $\sigma_{t_e}^* = \{\top\}$ and we do not create any vertices, as in Appendix~\ref{vertex:edge}. Otherwise, for each atomic proposition $\ap{i}{j}{k,\chi}$ in $\sigma_{t_e}^*$, we know that there are $i$ robots of type $j$ at region $\ell_k$ in the $t_e$-th configuration of the path $\overline{\tau}^\text{pre}$, and we also know which these $i$ robots are. Similar to  the construction of  the routing graph $\ccalG$ in Appendix~\ref{vertex:edge}, we construct $i$ vertices pointing to region $\ell_k$. Recall that we associated all robots of type $j$ with each of these $i$ vertices in $\ccalG$. However, for $\ccalG_{w^*}$, we know which specific $i$ robots of type $j$ visit region $\ell_k$ at time $t_e$ by checking the path $\overline{\tau}^\text{pre}$. We create a one-to-one correspondence between these $i$ robots with these $i$ vertices. In this way, each vertex is visited by one specific robot. These robots are referred to as the essential robots in Appendix~\ref{sec:run}. Furthermore, the time a specific  robot $r$ visits its assigned vertex $v$ is $t_e$, which is denoted by $t_{vr}^- = t_{vr}^+ = t_e$. Continuing this way, we create vertices for other atomic propositions in $\sigma_{t_e}^*$, which only correspond to a single clause  satisfied in $\gamma(v_1, v_2)$. Recall that when building the vertex set of $\ccalG$ in Appendix~\ref{vertex:edge}, we build such vertices for each clause in the given edge label. Therefore, the set of vertices in $\ccalG_{w^*}$ corresponding to the edge label satisfied at $t_e$ is a subset of the vertex set in $\ccalG$ for the same edge label.}\label{app:edge}

   \subphase{Literal vertices associated with vertex labels}{Following the same logic, we build the vertex set for the satisfied clause in the starting vertex label of $e$. We proceed depending on whether $e$ is the first completed subtask. If $e$ is not the first completed subtask in the simple path $\overline{\theta}^{\text{pre}}$,  according to Definition~\ref{defn:same}, the clauses satisfied in this vertex label remain the same, that is, $\sigma^*_t$'s remain the same for all $t= t_{e'}+1, \ldots, t_{e}-1$ where $e'$ is the subtask immediately preceding $e$. Also, it is the same fleet of robots that satisfy  this clause. Likewise, we can associate each vertex with one single robot, and the visit time interval is set as $[t_{e'}+1, t_{e}-1]$. That is, the robot $r$ remains at its assigned vertex $v$  within this time interval,
  denoted by $t_{vr}^- = t_{e'}+1$ and $t_{vr}^+ = t_{e}-1$. {This vertex set also exists in $\ccalG$} since vertices are created for any starting vertex label in Appendix~\ref{vertex:vertex}. Otherwise, if  $e$ is the first completed subtask in the path $\overline{\tau}^\text{pre}$ and its starting vertex has a self-loop, then,  we create vertices for the satisfied clause as usual and associate them with time interval $[0, t_e-1]$. That is, $t_{vr}^-=0$ and $t_{vr}^+ = t_e-1$. Recall also that, the routing graph $\ccalG$ contains  vertices for all clauses in all starting  vertex labels.
 Furthermore, we do not create vertices for $\top$ or $\bot$ labels, similar to the case in Appendix~\ref{vertex:vertex}. Therefore, we can conclude that the vertex set of $\ccalG_{w^*}$ is a subset of that of $\ccalG$.}\label{app:vertex}

 \mysubparagraph{Construction of the edge set}{}\label{app:edge_set} Next, we prove that the edge set in $\ccalG_{w^*}$ is also a subset of the edge set in $\ccalG$. Consider the edge label that is satisfied at the time instant $t_e$. For a vertex $v$ among those associated with this edge label, we already know the robot $r$ that visits  $v$. Our goal is to determine the unique vertex in $\ccalG_{w^*}$ from which robot $r$ comes. Let $\overline{\tau}_{r,j}$ denote  the path of robot $r$ of type $j$. Going backward from the $(t_{e}-1)$-th waypoint in $\overline{\tau}_{r,j}$ (included), we identify the most recent time instant $t\leq t_e -1$  when robot $r$ takes part in the satisfaction of a literal $\ap{i}{j}{k,\chi}$ that appears in the set $\sigma^*_{t}$ of atomic propositions, that is, participate in a certain subtask.

 \subphase{Time instant $t$ does not exist}{In this case, subtask $e$ is the first subtask that robot $r$ participates in, and we can create an edge starting from the vertex $u$ that is associated with the initial location of robot $r$ and ending at vertex $v$. We assign the travel time $T_{uv}=t_e$ to the edge $(u,v)$, which is obtained by  $T_{uv} = t_{vr}^- - t_{ur}^+ = t_e - 0$.
 In Appendix~\ref{sec:a}, the edge $(u,v)$ is  also  created in $\ccalG$.}

 \subphase{Time instant $t$ exists}{In this case, let $e'$ denote the subtask that the literal $\ap{i}{j}{k,\chi}$ corresponds to. If $e\neq e'$,   $e'$ occurs before $e$ in the given path $\overline{\tau}^\text{pre}$ since the time spans of subtasks are disjoint and $t < t_e$. Thus, $e' \in X_{<_P}^{e} \cup X_{\|_P}^e$. We identify the vertex $u$ in $\ccalG_{w^*}$ that is associated with this literal $\ap{i}{j}{k,\chi}$ and is visited by robot $r$, then create an edge starting from $u$ and ending at $v$. Furthermore, we assign the weight $t_{vr}^- - t_{ur}^+$ to this edge (where $t_{vr}^-=t_e$), which is the travel time of robot $r$  between these two consecutive subtasks. We emphasize that the edge $(u, v)$ also exists in $\ccalG$ since vertices $u$ and $v$ are associated with the same  robot type, and $u$ is associated with a prior subtask $e'$ of $e$. In Appendix~\ref{sec:b} that discusses  leaving vertices from prior subtasks, the edge $(u,v)$ exists in $\ccalG$.}

 \subphase{$e'=e$}{In this case, the vertex $u$ that robot $r$ visits is associated with the starting vertex label of the same subtask  $e$. We create the edge $(u,v)$ and assign the travel time  $T_{uv} = t_{vr}^- - t_{ur}^+$. This edge is also created in $\ccalG$ in Appendix~\ref{sec:c}. Therefore, all edges in $\ccalG_{w^*}$ with end vertices  associated with edge labels also exist in $\ccalG$.}

 Following the same logic, we create edges associated with the starting vertex label of subtask $e$.  Given a vertex $v$ in $\ccalG_{w^*}$ that is associated with the vertex label $\gamma(v_1)$ of subtask $e$,  we find  the associated specific  robot $r$ of type $j$.

 \subphase{Subtask $e$ is the first completed subtask in the path $\overline{\tau}^\textup{pre}$} {If the  starting vertex $v_0$ of subtask $e$ has a self-loop, then the initial locations should satisfy the starting vertex label of $e$. We locate the vertex $u$ associated with the initial location of robot $r$ of type $j$, create an edge starting from vertex $u$ and ending at $v$, and assign the travel time $T_{uv}=0$. This edge also exists in cases~\ref{edge:vertex1} or~\ref{edge:vertex2} in Appendix~\ref{sec:c}.}

 \subphase{Subtask $e$ is not the first completed subtask}{{We move backwards   from the $t_{e'}$-th waypoint (the subtask $e'$ immediately precedes  $e$) in the path $\overline{\tau}_{r,j}$ to find the most recent time instant $t$ that this robot has participated in another subtask preceding $e$. Condition~\hyperref[asmp:b]{(b)} in Definition~\ref{defn:same} states that all the robots satisfying the starting vertex label of a given subtask  belong to the robots that satisfy the edge label of the subtask immediately preceding the given subtask, which implies that $t$ should be identical to $t_{e'}$ since the path $\overline{\tau}^\text{pre}$ satisfies condition~\hyperref[asmp:b]{(b)}.}  We locate the vertex $u$ associated with the edge label of subtask $e'$ that robot $r$ visits, create an edge between $u$ and $v$, and assign the travel time $T_{uv} = t_{vr}^- - t_{ur}^+$ to the edge. This edge is also created in case~\ref{edge:vertex2} in Appendix~\ref{sec:c}.  Thus, the edge set of  $\ccalG_{w^*}$ is a subset of the edge set of $\ccalG$. Finally, we  conclude that the graph $\ccalG_{w^*}$ constructed from  the essential word $w^*$ is a subgraph of the routing graph $\ccalG$ used to formulate  the prefix MILP in Appendix~\ref{app:appendix_prefix_milp}.}\label{app:strongly}

 The graph $\ccalG_{w^*}$ has the property that there are no cycles and any two paths in $\ccalG_{w^*}$, starting from vertices pointing to initial robot locations  and ending at vertices without outgoing edges, do not share the same vertex since each path is associated with a specific robot. Therefore, every vertex except for the starting and end vertices in one path has indegree 1 (number of incoming edges) and outdegree 1 (number of outgoing edges). Moreover, vertices in $\ccalG_{w^*}$ are assigned  the tightest visit time intervals for the specific robot. Consequently, starting from the vertex corresponding to the initial location of robot $r$ of type $j$, we can extract a high-level plan $p_{r,j}$ for this robot by traversing along edges, which is a concise description of the low-level path $\overline{\tau}_{r,j}$. Observe that, given a feasible solution to the prefix MILP in Appendix~\ref{app:appendix_prefix_milp}, we can build a subgraph of $\ccalG$ by removing any vertices and edges that are not visited by any robots and assigning robots and visit times to the remaining vertices. In this sense, such a subgraph of $\ccalG$ can be viewed as the graphical depiction of the solution to the MILP. In what follows, we show that the graph $\ccalG_{w^*}$ is such a graph. That is, it gives rise to a feasible solution that satisfies constraints~\eqref{eq:1}-\eqref{eq:lastclause} in Appendix~\ref{app:appendix_prefix_milp}.

 \paragraph{Satisfaction of the prefix MILP constraints in Appendix~\ref{app:appendix_prefix_milp}}{}\label{app:milp}\hfill

 \setcounter{para}{0}
 \mysubparagraph{Routing constraints}{} Any vertex in $\ccalG_{w^*}$ is visited by a single robot of certain type, thus, constraint~\eqref{eq:1} that each vertex is visited by at most one robot of certain type is satisfied as follows. Given a vertex $v\in \ccalG_{w^*}$, its associated robot $r$ and unique vertex $u$ that is connected to $v$, we set $x_{uvr}=1$  and $x_{uvr'}=0$ for other robots $r'$ of the same type as $r$. In what follows, we omit the detailed assignment when it is clear to recognize. Furthermore, each vertex not in $\ccalV_{\text{init}}$  is either a sink vertex (indegree is 1, outdegree is 0) or a vertex with indegree equal to its  outdegree. Therefore, the flow constraint~\eqref{eq:2} is satisfied. For each vertex in $\ccalV_\text{init}$ of $\ccalG_{w^*}$, its outdegree is either 0 or 1, thus, constraint~\eqref{eq:2.5a} is satisfied. Each vertex in $\ccalV_\text{init}$ is associated with a unique robot, which satisfies constraint~\eqref{eq:2.5b}.

 \mysubparagraph{ Scheduling constraints}{} Since the visit time of each vertex  is non-negative and the visit time associated with the vertices in $\ccalV_{\text{init}}$ is set to $t_{vr}^- = t_{vr}^+ = 0$ (see case~\ref{app:initial} in Appendix~\ref{app:graph}), constraints~\eqref{eq:3} and~\eqref{eq:3.5} are trivially satisfied. When creating edges in $\ccalG_{w^*}$, we denote the travel time $T_{uv}$  between connected vertices $u$ and $v$ in $\ccalG_{w^*}$ by $t_{vr}^- - t_{ur}^+$, which is the actual time robot $r$ needs to travel between regions associated with $u$ and $v$. Obviously, $T_{uv}$ is no less than the shortest travel time $T^*_{uv}$ between these two regions, i.e., $T_{uv}^* \leq T_{uv}$. When $u \in \ccalV_\text{init}$ or $u <_P v$ or $(u,v)\in X_P$ and  when robot $r$ travels along the edge $(u, v)$, i.e., $x_{uvr}=1$, constraint~\eqref{eq:4b} holds since $t^+_{ur} + T^*_{uv} \leq t^+_{ur} + T_{uv} = t^-_{vr}$. Next, we show that, when $u \|_P v$, constraint~\eqref{eq:4a} can be satisfied if all robots remain idle for one time step within the time interval $[t_{ur}^+, t_{vr}^-]$. More importantly, the elongated path can still generate a word belonging to $\tilde{\ccalL}^{\phi,v_0\scriptveryshortarrow \vertex{accept}}_E(\auto{subtask}^-)$. This analysis proceeds depending on the types of NBA vertices that vertices $u$ and $v$ are associated with.

 \subphase{Starting vertex $u$ in $\ccalG_{w^*}$ is associated with a vertex label in $\auto{subtask}^-$}{Recall that we assign $t_e-1$ to $t_{ur}^+$ when constructing the vertex set which is the time instant right before the subtask is completed (see case~\ref{app:vertex} in Appendix~\ref{app:graph}). Thus, at the time instant $t_{ur}^+$, the run $\overline{\rho}^{\text{pre}}$ has not left the NBA vertex in $\auto{subtask}^-$ that vertex $u$ is associated with. We can repeat one more time the locations of all robots in the path $\overline{\tau}^\text{pre}$ at the time instant $t_{ur}^+$  so that the run visits the same NBA vertex  one more time. In this way, the travel time assigned to  the edge $(u,v)$ becomes $T_{uv}+1$, where $T_{uv}$ is the time robot $r$ takes in the path $\overline{\tau}^\text{pre}$ and 1 is the extra time it takes when all robots remain idle for one time step. Therefore, constraint~\eqref{eq:4a} is satisfied.}

 \subphase{End vertex $v$ is associated with a vertex label in $\auto{subtask}^-$}{Recall that we assign $t_{e'}+1$ to $t_{vr}^-$ when constructing the vertex set which is the time instant right after a subtask is completed (see case~\ref{app:vertex} in Appendix~\ref{app:graph}). The vertex label is satisfied at $t_{vr}^-$ since the  label of each vertex in the simple path $\overline{\theta}^{\text{pre}}$ is satisfied at least once according to the construction of the path $\overline{\tau}^\text{pre}$ at the beginning of Appendix~\ref{app:feasibility}. Thus, we can repeat one more  time the locations of all robots in the path $\overline{\tau}^\text{pre}$ at the time instant $t_{vr}^-$  so that the run visits the same NBA vertex  one  more time. Same as before, the travel time assigned to  the edge $(u,v)$ becomes $T_{uv}+1$.}

 \subphase{Both $u$ and $v$ in $\ccalG_{w^*}$ are associated with an edge label in $\auto{subtask}^-$}{These two vertices must correspond to two different subtasks. Furthermore, there must be an NBA vertex with a  self-loop between these two subtasks in the simple path $\overline{\theta}^{\text{pre}}$. This is because  according to condition~\ref{cond:c} in Definition~\ref{defn:run}, only initial and accepting vertices are allowed not to have self-loops but these vertices cannot be between two edges in the path $\overline{\theta}^{\text{pre}}$ if they do not have self-loops. In this case, robots can remain idle for one more time  when this vertex label is true. Same as before, the travel time assigned to  the edge $(u,v)$ becomes $T_{uv}+1$.}

 Therefore, constraint~\eqref{eq:4a} is satisfied  in these three cases. We emphasize that the path $\overline{\tau}^\text{pre}$ after modification still produces the prefix part of a restricted accepting run since the only result of idleness is  that a vertex in the run $\overline{\rho}^{\text{pre}}$ is visited for  one more time step.  Thus, requiring that all  robots remain  idle for a period of time does not affect the satisfaction of other constraints. In what follows, we still focus on the path $\overline{\tau}^\text{pre}$ since if it satisfies the others constraints, so does the modified path.

 \mysubparagraph{Logical constraints}{} Each set $\sigma^*_i$ of atomic propositions in the essential word $w^*$ collects all literals inside one clause, and all $\sigma^*_i$'s that are associated with the same vertex label collect literals of the same clause. Therefore, constraint~\eqref{eq:c} that one and only one clause is true is satisfied. Although the path $\overline{\tau}^\text{pre}$ can simultaneously satisfy more than one clauses in a label, we construct the essential word $w^*$ by selecting only one clause, and build the graph $\ccalG_{w^*}$ based on $w^*$. In this sense, we can state that only one clause is true on the graph $\ccalG_{w^*}$. Moreover, because every vertex in $\ccalG_{w^*}$ associated with the same clause is visited by one robot, constraint~\eqref{eq:6} is satisfied.  When constructing the vertex set of $\ccalG_{w^*}$ associated with edge labels (see case~\ref{app:edge} in Appendix~\ref{app:graph}), we associate each vertex $v$ corresponding to the same edge label with time $t_{vr}^- = t_{vr}^+ = t_e$. Therefore, the simultaneous visit constraint~\eqref{eq:7} is satisfied.

 \mysubparagraph{Temporal constraints}{}

 \subphase{Temporal constraints on one subtask}{} As discussed before, each vertex $v$ in $\ccalG_{w^*}$ associated with the same edge label is assigned time $t_{vr}^- = t_{vr}^+ = t_e$, and only one clause is true. Therefore, constraint~\eqref{eq:edgetime} specifying the completion time is satisfied. In case~\ref{app:vertex} in Appendix~\ref{app:graph}, we associate each vertex $v$ corresponding to the same vertex label (neither $\bot$ nor $\top$) with the same arriving time $t_{vr}^- = t_{e'}+1$ and the same leaving time $t_{vr}^+ = t_{e}-1$, where $t_e$ is the completion time of the subtask that the vertex label corresponds to and $t_{e'}$ is the completion time of the subtask immediately preceding $e$. Since subtasks in  the simple path $\overline{\theta}^{\text{pre}}$ are sequentially completed, $t_{vr}^- = t_{e'}+1\leq t_e$, thus the left side of constraint~\eqref{eq:17} is satisfied. The right side is satisfied trivially since $t_e = t_e-1+1 = t_{vr}^+ +1$. Next, if the initial vertex does not have a self-loop, and if the outgoing edge in the simple path $\overline{\theta}^{\text{pre}}$ is labeled with $\top$, there is no vertex in $\ccalG_{w^*}$ that corresponds to this edge. In this case, we can define the completion time of this edge label as 0, as stated by~\eqref{eq:tis0}. Otherwise, if the outgoing edge label is not $\top$,  the initial robot locations  satisfy this edge label, that is, the set $\sigma^*_0$ of atomic propositions include literals that appear in the satisfied clause in the edge label.
 Therefore, vertices are created in $\ccalG_{w^*}$ for these literals and the assigned visit time corresponds to the index of $\sigma_0^*$, i.e., 0. Thus, in this case~\eqref{eq:tis0} also holds.

 \subphase{Temporal constraints on the completion of two sequential subtasks}{}
 Since the simple path $\overline{\theta}^{\text{pre}}$ in $\auto{subtask}^-$, induced by the path $\overline{\tau}^\text{pre}$, is a linear extension of the poset $P$, we have that the temporal  order of subtasks in $\overline{\theta}^{\text{pre}}$ respects the partial order in the poset $P$. Thus, given a subtask $e$ in the simple path $\overline{\theta}^{\text{pre}}$, any subtask $e'\in \overline{\theta}^{\text{pre}}$ with $e' \prec_P e$ is completed before $e$ in the path $\overline{\tau}^\text{pre}$. Therefore, $t_{e'} +1 \leq t_e$, which satisfies constraint~\eqref{eq:12}.

 \subphase{Temporal constraints on the completion of the
 current subtask and the activation of subsequent
 subtasks}{}
 For each subtask $e$ except for the last one in the simple path $\overline{\theta}^{\text{pre}}$, the subtask $e'$ immediately following it belongs to the set $X_{\succ_P}^e \cup X_{\|_P}^e$. If $e' \in X_{\succ_P}^e \neq \emptyset$ as in case~\ref{activation:a} in Appendix~\ref{sec:temporal}, since the only subtask in the simple path $\overline{\theta}^{\text{pre}}$ that immediately follows $e$ is subtask $e'$, constraint~\eqref{eq:bafter} holds. The completion time  of subtasks $e'$ immediately following  $e$ in the simple path is larger than the completion time  of $e$ by at least 1. Therefore, constraint~\eqref{eq:after} is satisfied. 
 Furthermore, we associate  each vertex $v$ associated with the edge label of subtask $e'$  with the arrival time $t_{vr}^- = t_{e}+1$, which satisfies constraint~\eqref{eq:20}. Since the simple path $\overline{\theta}^{\text{pre}}$ is a linear extension of the poset $P$, no subtasks are completed at the same time, which satisfies constraint~\eqref{eq:diff}. If $X_{\succ_P}^e = \emptyset$ as in case \ref{activation:b} in Appendix~\ref{sec:temporal}, then $e' \in X_{\|_P}^e$. Because subtask $e'$ follows $e$, i.e., $b_e^{e'}=0$, and  subtask $e$ is not the last completed one, constraints~\eqref{eq:afterparallel_a}-\eqref{eq:afterparallel_c} are satisfied by setting $b_{ee'}=1$.  If $e$ is the last subtask in the simple path, $b_{ee'}=0$ and $b_{e}^{e'}=1$ in~constraints~\eqref{eq:afterparallel_a}-\eqref{eq:afterparallel_c}.

 On the other hand,  any subtask $e$ in the simple path $\overline{\theta}^{\text{pre}}$ except for the first one,  immediately follows a subtask $e'$, which should be in $X_{\prec_P}^e \cup X_{\|_P}^e$. If $e'\in  X_{\prec_P}^e \neq \emptyset$, since the only subtask in the simple path $\overline{\theta}^{\text{pre}}$ that immediately precedes $e$ is subtask $e'$, constraint~\eqref{eq:follow1} holds. Otherwise, if $X_{\prec_P}^e = \emptyset$, then $e \in X_{\|_P}^e$. As subtask $e'$ precedes $e$, i.e., $b_e^{e'}=1$, and subtask $e$ is not the first completed one, constraints~\eqref{eq:follow_a}-\eqref{eq:follow_c} are satisfied by setting $b_{e'e}=1$.  If $e$ is the first subtask in the simple path, $b_{e'e}=0$ and $b_e^{e'}=0$ in~constraints~\eqref{eq:follow_a}-\eqref{eq:follow_c}.

 \subphase{Temporal constraints on the activation of the first subtask}{}
 Consider the first completed subtask $e$ in the path $\overline{\tau}^\text{pre}$. If its starting vertex has a self-loop, and also the vertex label is not $\top$,
 then this vertex label is satisfied at least once by the construction of the path $\overline{\tau}^\text{pre}$, and  there are vertices in $\ccalG_{w^*}$ associated with the satisfied clause in this vertex label. Since the assigned time $t_{vr}^-$ to these vertices is 0 (see case~\ref{app:vertex} in Appendix~\ref{app:graph}), constraint~\eqref{eq:zeroactivation} is satisfied. For the first subtask $e$, since there is no subtask before it, the robots  visiting vertices associated with the starting vertex label of $e$ come from  vertices in $\ccalV_{\text{init}}$. Thus, constraint~\eqref{eq:routingforactivation_a} is satisfied for subtask $e$. For subtasks in $P_{\text{max}}$ other than $e$, the vertices associated with their vertex labels are connected to vertices associated with the edge labels of subtasks that are completed immediately before them (see case~\ref{app:strongly} in Appendix~\ref{app:graph}). Therefore, constraint~\eqref{eq:routingforactivation_b} is satisfied.

 \mysubparagraph{Same-$\ag{i}{j}$ constraints}{}
 Any two vertex subsets in $\ccalG_{w^*}$ that are associated with two literals that share the same nonzero connector  are visited by the same fleet of robots since the essential word $w^*$ belongs to $\title{\ccalL}_E^{v_0 \scriptveryshortarrow \vertex{accept}}(\auto{subtask}^-)$. We enumerate these two vertex subsets such that there is a one-to-one correspondence between vertices in these two subsets and the matched pair of vertices are visited by the same robot. In this way, constraint~\eqref{eq:same} is satisfied.

 \mysubparagraph{Constraints on the transition between the prefix and suffix parts}{}
 Since  we iterate over all subtasks that can be the last  to be completed and also iterate over all clauses in the edge label of the selected last subtask, we can formulate a prefix MILP in Appendix~\ref{app:appendix_prefix_milp} in which the selected last subtask and the clause are  the same as that induced by the feasible path $\overline{\tau}^\text{pre}$. Therefore, constraints~\eqref{eq:lastsubtask0} and~\eqref{eq:lastclause} are satisfied, which completes the proof.

 \subsubsection{Proof of Lemma~\ref{prop:run}}\label{app:run}
 To prove that a simple path $\tilde{\theta}$ can be extracted from the sub-NBA $\auto{subtask}^-$,  we  show that the solution to the MILP in Appendix~\ref{app:appendix_prefix_milp} gives rise to a simple path in the set of simple paths $\Theta$ from which the poset $P$ is inferred. {Then, since  the graph-search version of the depth-first search on finite graphs is complete and since the  backtracking search is a form of a depth-first-search~\citep{russell2002artificial},  such a simple path can  be found in $\auto{subtask}^-$.} In the prefix MILP in Appendix~\ref{app:appendix_prefix_milp}, we define a variable for each subtask in $X_P$ which indicates its completion time (see Appendix~\ref{sec:temporal}) and require that the completion times of two subtasks are different (see constraint~\eqref{eq:diff}).  Thus, we can sort the subtasks in $X_P$  in an ascending order with respect to their completion time. The sorted subtasks respect the partial order in $P$ since precedence relations among subtasks are captured by constraint~\eqref{eq:12}, which means that the sequence  of sorted subtasks is a linear extension of the poset $P$. Furthermore, the set of simple paths $\Theta$ is equivalent to the set of  linear extensions of the poset $P$, which is ensured in Section~\ref{sec:poset}. Therefore, the sequence of sorted subtasks corresponds to a simple path in $\Theta$, which is in $\auto{subtask}^-$.

 In what follows, we prove the three properties of this simple path $\tilde{\theta}$ as stated in Lemma~\ref{prop:run}: First, if the initial vertex $v_0$ of the initial subtask in the simple path $\tilde{\theta}$  does not have a self-loop,  according to constraint~\eqref{eq:tis0}, it must be completed at time 0. Thus, the activation time is also 0. Otherwise, if $v_0$ has a self-loop and also the vertex label in $\auto{subtask}^-$ is $\top$, then it can  be activated at anytime, including 0; else if $v_0$ has a self-loop for which the label is not $\top$, according to constraint~\eqref{eq:zeroactivation}, the activation of the vertex label is 0. Therefore, property~\ref{property:a} in Lemma~\ref{prop:run} holds.

 Second, for any subtask $e$ in the simple path $\tilde{\theta}$, if its starting vertex label has a self-loop and the vertex label is $\top$,  property~\ref{property:b} in Lemma~\ref{prop:run} holds trivially. Otherwise,  if the vertex label is not $\top$,  constraint~\eqref{eq:17} ensures that  property~\ref{property:b} in Lemma~\ref{prop:run} holds.

 Finally,  for  any two consecutive subtasks $e$ and $e'$ in the simple path $\tilde{\theta}$, we  prove that it is exactly $e' \in S_3^e = X^e_{\succ_{P}} \cup X^e_{\|_{P}}$ that makes $b_{ee'}=1$ in constraint~\eqref{eq:bafter}. If so, according to constraint~\eqref{eq:20}, subtask $e'$ is activated at most one time step after the completion of $e$. Therefore, property~\ref{property:c} in Lemma~\ref{prop:run} holds. To see this, we use  induction.

 Consider $e_0$ and $e_1$ to be the first two subtasks in the simple path $\tilde{\theta}$. Because $e_1$ is completed immediately after $e_0$, we have that $e_1\in X^{e_0}_{\succ_{P}} \cup X^{e_0}_{\|_{P}} $. Thus, ${e_0} \in X^{e_1}_{\prec_{P}} \cup X^{e_1}_{\|_{P}}$. Since subtask $e_1$ is not the first one in $\tilde{\theta}$, it must immediately follow a subtask. By constraints~\eqref{eq:follow1}-\eqref{eq:follow}, there must exist  a subtask $e \in X^{e_1}_{\prec_{P}} \cup X^{e_1}_{\|_{P}} $ such that $b_{e e_1}=1$.
 Assume that $e \neq e_0$. By constraint~\eqref{eq:after}, subtask $e$ must be completed before $e_1$, but it is only subtask $e_0$ that occurs before $e_1$ in the simple path $\tilde{\theta}$, a contradiction. Therefore, $e  = e_0$, i.e., it is exactly subtask $e_1$ that makes $b_{e_0 e_1}=1$. Next,  assume that for any two consecutive subtasks $e_{i-1}$ and $e_{i}$ in the simple path $\tilde{\theta}$,  it holds that $b_{e_{i-1} e_i} = 1$. Given the next two subtasks $e_{i}$ and $e_{i+1}$, assume that $e_{i+1}$ immediately follows subtask $e'$ but $e' \neq e_i$. By constraint~\eqref{eq:after}, $e'$ is completed before $e_{i+1}$. However, $e'$ cannot be any subtask in $e_0, \ldots, e_{i-1}$, since this will contradict constraint~\eqref{eq:bafter} that the immediately following subtask of any subtask in $e_0, \ldots, e_{i-1}$ is unique. Therefore, $e' = e_i$, completing the proof.

 \subsubsection{Proof of Lemma~\ref{prop:valid}}\label{app:valid}
 Given a subtask $e = (v_1, v_2)$ in the simple path $\tilde{\theta}$, the goal of the   GMRPP in Appendix~\ref{sec:mapp} is to design paths for the robots to reach locations that satisfy the complete clause $\gamma_{1,2}^+ \wedge \gamma_{1,2}^-$ in the edge label $\gamma(v_1, v_2)$ so as to complete the current subtask and activate the next subtask, while respecting the complete clause $\gamma_1^+ \wedge \gamma_1^-$ in the starting vertex label $\gamma(v_1)$ en route. Starting from the first subtask in the simple path $\tilde{\theta}$, we proceed along the simple path to prove that each GMRPP instance with initial locations generated by the previous instance is feasible.

 Consider the first subtask $(v_1, v_2)$ with $v_1 = v_0$. We first discuss the case where  the initial vertex $v_0$ has a self-loop in the sub-NBA $\auto{subtask}^-$, which implies that the initial robot locations satisfy the label $\gamma_{\phi}(v_0)$ in the NBA $\autop$, otherwise we  remove its self-loop (see Sections~\ref{sec:sort} and \ref{sub-NBA:1}). We continue based on whether the sets of essential robots $\ccalR_1$ and $\ccalR_{1,2}$ are disjoint.

 \paragraph{$\ccalR_1 \cap \ccalR_{1,2} = \emptyset$}{According to property~\ref{property:b} in Lemma~\ref{prop:run}, the complete clause $\gamma_1^+ \wedge \gamma_1^-$ can only become false when $\gamma_{1,2}^+ \wedge \gamma_{1,2}^-$ becomes true. At the initial time 0, according  to property~\ref{property:a} in Lemma~\ref{prop:run}, the robot locations satisfy $\gamma_1^+ \wedge \gamma_1^-$, including those robots in $\ccalR_{1,2}$.  {Thus, all robots can move around safely  within their respective regions without violating $\gamma_1^-$.} By Assumption~\ref{asmp:env}, there is a label-free path between any two regions, and between any label-free cells and any regions. Thus, the robots in $\ccalR_{1,2}$ can move to label-free cells without passing through  other regions, and, therefore, they can travel along label-free paths to reach label-free cells that are adjacent to their target regions. This process does not  violate  the negative clause $\gamma_1^-$. Also, $\gamma_1^+$ is satisfied due to $\ccalR_1 \cap \ccalR_{1,2} = \emptyset$. At this point, the essential clause $\gamma_{1,2}^+$ can be satisfied in one  time step. If, at this time, there are robots in $\ccalR^-\setminus \ccalR_{1,2}$ that violate  $\gamma_{1,2}^-$ (robots in $\ccalR_{1,2}$ stay at label-free cells now), then without passing through   other regions, these robots  move to locations within their respective regions from where they can reach the label-free cells in one time step. This process also respects the complete clause $\gamma_1^+ \wedge \gamma_1^-$. Finally, at the same time, the robots  in $\ccalR_{1,2}$ move to their target regions and the robots in $\ccalR^-\setminus \ccalR_{1,2}$ that violate $\gamma_{1,2}^-$   move from their regions to label-free cells. As a result, the complete label $\gamma_{1,2}^+ \wedge \gamma_{1,2}^-$ is satisfied. Note that robots moving to target regions to satisfy $\gamma_{1,2}^+$  do not violate the negative clause $\gamma_{1,2}^-$ since infeasible clauses are removed during the  pre-processing step~\hyperref[prune:exclusion2]{(4)} in Section~\ref{sec:nba}.}\label{app:gmrpp_a}

 \paragraph{$\ccalR_1 \cap \ccalR_{1,2} \neq \emptyset$}{In this case, for a robot $r \in  \ccalR_1 \cap \ccalR_{1,2}$, the shortest travel time between its source region and target region is less than or equal to 1 since by constraint~\eqref{eq:17} in Appendix~\ref{app:appendix_prefix_milp}, the completion time of a subtask is at most one time step after  the completion time of its starting vertex label. This implies that its source  region and target region are identical or adjacent. Same as in Appendix~\ref{app:gmrpp_a} where $\ccalR_1 \cap \ccalR_{1,2} = \emptyset$, the robots in $\ccalR_{1,2} \setminus \ccalR_1$ move to label-free cells from where they can reach  their target regions  in one time step, while the robots in $\ccalR_{1,2} \cap \ccalR_1$ move to locations within their respective source regions from where they can reach the target regions and leave the source regions in one time step. Without pass through other regions, the complete clause $\gamma_1^+ \wedge \gamma_1^-$ remains satisfied. Next, similar to Appendix~\ref{app:gmrpp_a} where $\ccalR_1 \cap \ccalR_{1,2} = \emptyset$, the robots in $\ccalR^-\setminus \ccalR_{1,2}$ move to locations within their respective regions from where they can reach the label-free cells in one time step. In this way, $\gamma_{1,2}^+ \wedge \gamma_{1,2}^-$ can be satisfied at the next time step.}\label{app:gmrpp_b}

 We have shown the feasibility of GMRPP when the initial vertex $v_0$ in $\auto{subtask}^-$ has a self-loop. In the case where $v_0$ does not have a self-loop in $\auto{subtask}^-$, the initial robot locations satisfy the complete clause  $\gamma_{1,2}^+ \wedge \gamma_{1,2}^-$ in the  edge label; otherwise, the edge is  removed (see Sections~\ref{sec:sort} and \ref{sub-NBA:1}). We do not formulate the GMRPP in this case; see line~\ref{seq:terminate_1} in Algorithm~\ref{alg:sequentialMAPP}.

 Whether or not $v_0$ has a self-loop, the complete clause $\gamma_{1,2}^+ \wedge \gamma_{1,2}^-$ is satisfied at last. By condition~\ref{cond:d} in  Definition~\ref{defn:run} and condition~\hyperref[asmp:b]{(b)} in Definition~\ref{defn:same}, the complete clause in the end vertex label $\gamma_\phi(v_2)$ can be satisfied automatically, which activates the next subtask, as per property~\ref{property:c} in Lemma~\ref{prop:run} states that   the next subtask is activated at most one time step after the current one.  We can apply the same logic in Appendices~\ref{app:gmrpp_a} where $\ccalR_1 \cap \ccalR_{1,2} = \emptyset$ and \ref{app:gmrpp_b} where $\ccalR_1 \cap \ccalR_{1,2} \neq \emptyset$ to the remaining subtasks in the simple path $\tilde{\theta}$ since each subtask being activated by the previous subtask is similar to the first subtask being activated by initial robot locations, completing the proof.

 \subsubsection{Proof of Lemma~\ref{prop:suffix_feasibility}}\label{app:suffix_feasibility}
 The MILP in Appendix~\ref{app:appendix_suffix_milp} for the suffix part shares most constraints with that for the prefix part,  so we can follow the same procedure as  in Appendix~\ref{app:feasibility}.

 First, we construct an essential word $w^*$ based on the given path $\overline{\tau}^{\text{suf}}$, as in Appendix~\ref{app:word}.  Note that the essential word is constructed with respect to the sub-NBA $\auto{subtask}^-$ in Fig.~\ref{fig:suffix} where we add a  clause $\ccalC_{\text{prior}}^+$ to each edge label of subtasks that can be the last to be completed. By condition~\hyperref[asmp:c]{(c)} in Definition~\ref{defn:same}, all robots return to  $s_{\text{prior}}$ at last while driving the transition back to $\vertex{accept}$. This implies that those robots involved in the clause $\ccalC^+_{\text{prior}}$ of $\gamma(\vertex{prior}, \vertex{accept})$ return to regions corresponding to their initial locations. By Assumption~\ref{asmp:env}, any region spans consecutive cells. Therefore, the last waypoint in $\overline{\tau}^{\text{suf}}$ must satisfy the clause $\ccalC_{\text{prior}}^+$, and further the last set of atomic propositions in $w^*$ is $\ccalC_{\text{prior}}^+$. Following steps similar to Appendix~\ref{app:graph}, we can construct a graph $\ccalG_{w^*}$ that is also a subgraph of the routing graph $\ccalG$.

 Next, since the MILP in Appendix~\ref{app:appendix_suffix_milp} for the suffix part includes constraints~\eqref{eq:1}-\eqref{eq:same} in Appendix~\ref{app:appendix_prefix_milp} for prefix MILP, our analysis for these constraints is the same as that in Appendix~\ref{app:milp}. Thus, we focus on constraints~\eqref{eq:one_suffix}-\eqref{eq:same_suffix} from Appendix~\ref{app:appendix_suffix_milp}. First, no two subtasks are satisfied at the same time in the given path $\overline{\tau}^{\text{suf}}$. Thus, constraints~\eqref{eq:one_suffix} and~\eqref{eq:lastsubtask} are  satisfied. Each vertex in $\ccalG_{w^*}$  associated with  clause $\ccalC_{\text{prior}}^+$ of the last subtask is visited by a specific  robot. Therefore, constraint~\eqref{eq:return_suffix} is satisfied. 

 \subsubsection{Proof of Theorem~\ref{thm:completeness}}\label{app:completeness_}
 We emphasize that we discuss the class  \ltlz of formulas in this proof. Because \ltlz$\subset$ {LTL}$^\chi$, Proposition~\ref{thm:prefix} and Lemma~\ref{prop:suffix_feasibility} apply also to the class {LTL}$^0$. Proposition~\ref{thm:prefix} ensures  that we can find   a feasible prefix part $\tilde{\tau}^{\text{pre}}$ that induces a run $\tilde{\rho}^{\text{pre}}$ connecting $v_0$ and $\vertex{accept}$. Therefore, our goal is to prove that a corresponding suffix part $\tilde{\tau}^{\text{suf}}$ exists.

 We divide the proof into two cases depending on whether the suffix part $\rho^{\text{suf}}$ of the run induced by the assumed path $\tau=\tau^{\text{pre}} [\tau^{\text{suf}}]^w$ in Theorem~\ref{thm:completeness} is a single vertex or not. When $\rho^{\text{suf}}$ only consists of the accepting vertex $\vertex{accept}$, we next show that condition~\ref{cond:e} in Definition~\ref{defn:run} can be  satisfied. If $\gamma_{\phi}(\vertex{accept})=\top$, this condition is satisfied automatically. Otherwise, if $\gamma_\phi(\vertex{accept})\neq \top$, because in condition {(c)} in Definition~\ref{defn:st} we do not remove any composite edges leading to $\vertex{accept}$ and in constraints~\eqref{eq:lastsubtask0} and~\eqref{eq:lastclause} in Appendix~\ref{app:appendix_prefix_milp} we iterate over  subtasks that can be the last  to be completed, eventually we can formulate a prefix MILP  where the edge label of the  last subtask implies the label of vertex $\vertex{accept}$. Therefore, condition~\ref{cond:e} in Appendix~\ref{app:appendix_prefix_milp} is  met, which means that the final locations of the prefix path $\tilde{\tau}^{\text{pre}}$ satisfy the vertex label $\gamma_\phi(\vertex{accept})$. Combined with Proposition~\ref{thm:prefix}, we can find a path that induces a run in $\autop$  connecting $v_0$ and $\vertex{accept}$ and ensures that the NBA $\autop$ remains at $\vertex{accept}$ forever, satisfying the specification $\phi$.

 Next, we discuss the case where the suffix part $\rho^{\text{suf}}$ includes more than one  vertices. Recall that ${s}_{\text{prior}}$ and ${v}_{\text{prior}}$ are the last waypoint in the prefix path $\tau^{\text{pre}}$  and the last vertex before $\vertex{accept}$ in the induced run $\rho^{\text{pre}}$. We denote by   $\tilde{s}_{\text{prior}}$ and $\tilde{v}_{\text{prior}}$  the last waypoint in the found prefix path $\tilde{\tau}^{\text{pre}}$  and the last vertex before $\tilde{v}_\text{accept}$ in its induced run, respectively. By Proposition~\ref{thm:nba}, there exists a suffix path $\overline{\tau}^{\text{suf}}$ generating a word in $\tilde{\ccalL}_E^{\phi, \vertex{accept}\scriptveryshortarrow \vertex{accept}} (\auto{subtask}^-;  s_{\text{prior}}, v_{\text{prior}})$. However, this word may not belong to $\tilde{\ccalL}_E^{\phi, \vertex{accept}\scriptveryshortarrow \vertex{accept}} (\auto{subtask}^-;  \tilde{s}_{\text{prior}}, \tilde{v}_{\text{prior}}) $, since the pair ${s}_{\text{prior}}$ and $\vertex{prior}$ may not be same as the pair $\tilde{s}_{\text{prior}}$ and $\tilde{v}_{\text{prior}}$. In what follows  we show that a feasible suffix path $\doverline{\tau}^{\text{suf}}$, modified from $\overline{\tau}^{\text{suf}}$, exists that generates a finite word $\doverline{w}^{\text{suf}} \in \ccalL_E^{\phi, \vertex{accept}\scriptveryshortarrow \vertex{accept}} (\auto{subtask}^-;  \tilde{s}_{\text{prior}},  \tilde{v}_{\text{prior}})$ and satisfies conditions~\hyperref[asmp:a]{(a)} and~\hyperref[asmp:b]{(b)} in Definition~\ref{defn:same}. Then, we rely on Lemma~\ref{prop:suffix_feasibility}   to prove the final result. {Note that the word $\doverline{w}^{\text{suf}}$  belongs to language $\ccalL_E^{\phi, \vertex{accept}\scriptveryshortarrow \vertex{accept}}$ instead of $\tilde{\ccalL}_E^{\phi, \vertex{accept}\scriptveryshortarrow \vertex{accept}}$ since the path $\doverline{\tau}^{\text{suf}}$ does not satisfy condition \hyperref[asmp:c]{(c)} in Definition~\ref{defn:same}.}

 First, we show that eventually we have $\tilde{v}_{\text{prior}} = {v}_{\text{prior}}$ and $\tilde{s}_{\text{prior}}$ and ${s}_{\text{prior}}$ satisfy the same clause $\ccalC_{\text{prior}}$. As we iterate over subtasks that can be the last  to be completed in the prefix part (see constraint~\eqref{eq:lastsubtask0} in Appendix~\ref{sec:transition}), we can eventually formulate a prefix MILP  whose solution gives rise to a run with $\tilde{v}_{\text{prior}} = {v}_{\text{prior}}$. Furthermore, as we iterate over clauses in the selected last subtask (see constraint~\eqref{eq:lastclause}), we can obtain the final configuration $\tilde{s}_{\text{prior}}$ such that $\tilde{s}_{\text{prior}}$ and ${s}_{\text{prior}}$ satisfy the same clause $\ccalC_{\text{prior}}$ in the edge label $\gamma_\phi(\vertex{prior}, \vertex{accept})$. Note however that it is possible that different robots in $\tilde{s}_{\text{prior}}$ and $s_{\text{prior}}$ satisfy the positive subformula $\ccalC_\text{prior}^+$. Based on $\tilde{s}_{\text{prior}}$ and $\vertex{prior}$, we can obtain a sub-NBA $\auto{subtask}^-$ for the suffix part (see Appendix~\ref{sec:suf_prune}), which  differs from the sub-NBA obtained based on  $s_{\text{prior}}$ and $\vertex{prior}$ only in the interpretation of the  clause $\ccalC_\text{prior}^+$, that is,  on those robots that  return to their respective regions; these regions are identical for $\tilde{s}_{\text{prior}}$ and $s_{\text{prior}}$ since they satisfy the same clause $\ccalC_{\text{prior}}^+$. In other words, these two sub-NBAs are graphically equivalent. We  denote by $\auto{subtask}^-( \tilde{s}_{\text{prior}},  {v}_{\text{prior}})$ the sub-NBA based on $\tilde{s}_{\text{prior}} $ and $  {v}_{\text{prior}}$.

 Next, based on the fact that $\overline{\tau}^{\text{suf}}$ generates a word $\overline{w}^{\text{suf}} \in \ccalL_E^{\phi, \vertex{accept}\scriptveryshortarrow \vertex{accept}} (\auto{subtask}^-;  {s}_{\text{prior}},  {v}_{\text{prior}})$, we construct another feasible path $\doverline{\tau}^{\text{suf}}$, modified from $\overline{\tau}^{\text{suf}}$, that generates a word $\doverline{w}^{\text{suf}} \in \ccalL_E^{\phi, \vertex{accept}\scriptveryshortarrow \vertex{accept}} (\auto{subtask}^-;  \tilde{s}_{\text{prior}},  {v}_{\text{prior}})$. The path $\doverline{\tau}^{\text{suf}}$ begins with the final locations $\tilde{s}_{\text{prior}} $ of the found prefix part $\tilde{\tau}^{\text{pre}}$ in Proposition~\ref{thm:prefix}. By condition~\ref{cond:f} in Definition~\ref{defn:run}, $\gamma_\phi(\vertex{prior}, \vertex{accept}) \simplies \gamma_\phi(\vertex{accept}, \vertex{next}) $, therefore,  $\tilde{s}_{\text{prior}}$ and ${s}_{\text{prior}}$ satisfy the same clause in label  $\gamma_\phi(\vertex{accept}, \vertex{next})$ since they satisfy the same clause $\ccalC_{\text{prior}}$ in $\gamma_\phi(\vertex{prior}, \vertex{accept})$. Moreover,  by conditions~\ref{cond:d} and~\ref{cond:f} in Definition~\ref{defn:run}, $\tilde{s}_{\text{prior}}$ and ${s}_{\text{prior}}$ satisfy the same clauses in the edge label $\gamma_\phi(\vertex{accept}, \vertex{next})$ and vertex label $\gamma_\phi(\vertex{next})$, respectively. Therefore, $\tilde{s}_{\text{prior}}$ and ${s}_{\text{prior}}$ satisfy the same clause in $\gamma(\vertex{next})$ in the sub-NBA $\auto{subtask}^-( \tilde{s}_{\text{prior}},  {v}_{\text{prior}})$. Note that, at the time  when the sub-NBA $\auto{subtask}^-( \tilde{s}_{\text{prior}},  {v}_{\text{prior}})$ remains at the vertex $\vertex{next}$,  robots can start from $\tilde{s}_{\text{prior}}$ and reach a configuration  $\overline{s}_{\text{prior}}$ that is almost identical to $s_{\text{prior}}$ except for the specific robots at the specific cells. {In other words, if we consider robots in the same type to be indistinguishable, $\overline{s}_{\text{prior}}$ is identical to $s_{\text{prior}}$.}  To show this,  we construct a one-to-one correspondence between robots in $\tilde{s}_{\text{prior}}$ and robots in ${s}_{\text{prior}}$.

 {Specifically, for a literal $\ap{i}{j}{k}$ in the satisfied clause of $\gamma(\vertex{next})$ in $\auto{subtask}^-$, we identify $i$ robots of type $j$ in configuration $\tilde{s}_{\text{prior}}$ that satisfy this literal and another  $i$ robots of type $j$ in ${s}_{\text{prior}}$. Then we construct a random one-to-one correspondence between these two sets of robots, i.e., $i$ pairs of robots, such that every robot from the $i$ robots associated with $\tilde{s}_{\text{prior}}$ starts  from its location in $\tilde{s}_{\text{prior}}$, travels inside region $\ell_k$ and reaches the location in $s_{\text{prior}}$ where its paired robot is.} This  maintains the satisfaction of $\gamma(\vertex{next})$. This point-to-point navigation is feasible since those $i$ robots associated with $\tilde{s}_{\text{prior}}$ and their corresponding robots associated with $s_{\text{prior}}$ are all in region $\ell_k$ and, according to Assumption~\ref{asmp:env} every region spans consecutive cells. Enumerating  other robots of type $j$ in $\tilde{s}_{\text{prior}}$ that do not participate in the satisfaction of $\gamma(\vertex{next})$, we can construct another one-to-one correspondence between them and those of type $j$ in $s_{\text{prior}}$ that do not participate in the satisfaction of $\gamma(\vertex{next})$. {Such robots in $\tilde{s}_{\text{prior}}$, by Assumption~\ref{asmp:env},  can leave their regions corresponding to their locations in $\tilde{s}_\text{prior}$ to go to label-free cells without passing through other regions, then travel along  label-free paths to the regions where their paired robots are located in $s_{\text{prior}}$ and finally reach the specific cells inside these regions.} Robots traveling  inside  regions  do not violate the label $\gamma(\vertex{next})$ since $\tilde{s}_{\text{prior}}$ and $s_{\text{prior}}$ satisfy $\gamma(\vertex{next})$. In this way, robots reach a configuration $\overline{s}_{\text{prior}}$ while the NBA $\auto{subtask}^-( \tilde{s}_{\text{prior}},  {v}_{\text{prior}})$ remains at the vertex $\vertex{next}$.

 {When robots reach the configuration $\overline{s}_{\text{prior}}$ from $\tilde{s}_{\text{prior}}$, conditions~\hyperref[asmp:a]{(a)} and \hyperref[asmp:b]{(b)} in Definition~\ref{defn:same} are not violated since these robots in $\tilde{s}_{\text{prior}}$ that participate in the satisfaction of a clause in $\gamma(\vertex{next})$ do not leave their respective regions.} We append this path segment from $\tilde{s}_{\text{prior}}$ to  $\overline{s}_{\text{prior}}$ to the current $\doverline{\tau}^{\text{suf}}$. {Note that $\overline{s}_{\text{prior}}$ and $s_\text{prior}$ are identical if robots that belong   to the same type are indistinguishable.
  After reaching the configuration $\overline{s}_{\text{prior}}$, every robot travels along  the suffix path in $\overline{\tau}^{\text{suf}}$ (both beginning with and ending at  $s_{\text{prior}}$) that its paired robot does. Appending this path to the currrent $\doverline{\tau}^{\text{suf}}$ concludes the construction of $\doverline{\tau}^{\text{suf}}$.} At last, the transition in the sub-NBA $\auto{subtask}^-( \tilde{s}_{\text{prior}},  {v}_{\text{prior}})$ is driven back to $\vertex{accept}$. Note that the last configuration in $\doverline{\tau}^{\text{suf}}$ is not identical to $\tilde{s}_{\text{prior}}$, that is, robot trajectories are not closed yet. Thus, condition~\hyperref[asmp:c]{(c)} in Definition~\ref{defn:same} is not met. However, in the last configuration of $\doverline{\tau}^{\text{suf}}$ those robots participating in the satisfaction of $\ccalC_{\text{prior}}^+$ return to their respective regions since $\overline{\tau}^{\text{suf}}$ at last returns to $s_{\text{prior}}$, and $s_{\text{prior}}$ and $\tilde{s}_{\text{prior}}$ satisfy the same positive subformula  $\ccalC_{\text{prior}}^+$. Therefore, we can construct a path $\doverline{\tau}^{\text{suf}}$ that satisfies conditions~\hyperref[asmp:a]{(a)} and \hyperref[asmp:b]{(b)} in Definition~\ref{defn:same} and generates a word $\doverline{w}^{\text{suf}} \in \ccalL_E^{\phi, \vertex{accept}\scriptveryshortarrow \vertex{accept}} (\auto{subtask}^-;  \tilde{s}_{\text{prior}},  {v}_{\text{prior}})$.  More importantly, those robots participating in the satisfaction of $\ccalC_{\text{prior}}^+$ return to  regions corresponding to their initial locations.

 Subsequently, from Lemma~\ref{prop:suffix_feasibility} we conclude that we can obtain a low-level path and we denote it  by  $\tilde{\tau}^{\text{suf},1}$. We note that Lemma~\ref{prop:suffix_feasibility} assumes that a path exists  satisfying condition~\hyperref[asmp:c]{(c)} in Definition~\ref{defn:same} which requires robots to return to their initial locations, while in $\doverline{\tau}^{\text{suf}}$ only those robots participating in satisfying $\ccalC_{\text{prior}}^+$  return to their respective regions. Even so, it suffices to establish the feasibility of the MILP (excluding constraint~\eqref{eq:same_suffix}) in Appendix~\ref{app:appendix_suffix_milp} for the suffix part since in the MILP, the clause needed to be satisfied in the last completed subtask is $\ccalC_{\text{prior}}^+$ (see Fig.~\ref{fig:suffix}), i.e., robots are not required to return to their initial locations. After obtaining the path $\tilde{\tau}^{\text{suf},1}$,  the run in $\autop$ induced by $\tilde{\tau}^{\text{suf},1}$ is a cycle around the accepting vertex $\vertex{accept}$.

 Finally, we prove that  closing the trajectories in Appendix~\ref{sec:closing} is feasible. The last configuration in the low-level path $\tilde{\tau}^{\text{suf},1}$ satisfies the clause $\ccalC_{\text{prior}}$, and so do the initial locations $\tilde{s}_{\text{prior}}$. Also, the robots in the last configuration of $\tilde{\tau}^{\text{suf},1}$ participating in the satisfaction of $\ccalC_{\text{prior}}^+$ are identical to those in $\tilde{s}_{\text{prior}}$. Therefore, they  can return to their initial locations in $\tilde{s}_{\text{prior}}$ inside the same regions, while maintaining the truth of $\ccalC_{\text{prior}}^+$. {The rest of the robots can return to their initial locations by leaving their regions in the last configuration to go to label-free cells,} then traveling along the label-free paths to the regions where their initial locations in $\tilde{s}_{\text{prior}}$ are located, and finally returning to initial locations inside these regions. This  respects the negative subformula $\ccalC_{\text{prior}}^-$ since both the last configuration in $\tilde{\tau}^{\text{suf},1}$ and $\tilde{s}_{\text{prior}}$ satisfy $\ccalC_{\text{prior}}$. We denote by $\tilde{\tau}^{\text{suf,2}}$ the path segment from the last configuration of $\tilde{\tau}^{\text{suf,1}}$ to $\tilde{s}_{\text{prior}}$, which satisfies the clause $\ccalC_{\text{prior}}$, and further satisfies the label $\gamma_\phi(\vertex{prior}, \vertex{next})$ and $\gamma_\phi(\vertex{next})$ according to conditions~\ref{cond:d} and \ref{cond:f} in Definition~\ref{defn:run}.
 Therefore, the NBA $\autop$ can remain at vertex $\vertex{next}$ while robots execute  the path segment $\tilde{\tau}^{\text{suf,2}}$. In a nutshell, we leverage the vertex $\vertex{next}$ to reach $\overline{s}_{\text{prior}}$ from $\tilde{s}_{\text{prior}}$ in order to reuse the suffix path $\overline{\tau}^{\text{suf}}$, and similarly {we leverage the vertex $\vertex{next}$ to deviate from the path $\overline{\tau}^{\text{suf}}$ in order to return to $\tilde{s}_{\text{prior}}$.} Finally, we can obtain the suffix path by concatenating $\tilde{\tau}^{\text{suf},1}$ with $\tilde{\tau}^{\text{suf},2}$, i.e., $\tilde{\tau}^{\text{suf}} = \tilde{\tau}^{\text{suf},1} \tilde{\tau}^{\text{suf},2}$, which gives rises to a path $\tilde{\tau} = \tilde{\tau}^{\text{pre}} [\tilde{\tau}^{\text{suf}}]^\omega$ that satisfies the specification $\phi$, completing the proof.

 \end{appendices}

\end{document}